\documentclass[lettersize,journal]{IEEEtran}
\pdfoutput=1
\usepackage{graphicx}  %插入图片的宏包
\usepackage{float}  %设置图片浮动位置的宏包
\usepackage{amsmath,amsfonts}
\usepackage{array}
\usepackage{subfig}
\usepackage{textcomp}
\usepackage{stfloats}
\usepackage{url}
\usepackage{verbatim}
\usepackage{graphicx}
\usepackage{cite}
\usepackage{xspace}
\usepackage{algorithmic}
\usepackage{algorithm}
\usepackage{caption}
\usepackage{enumerate}
\usepackage{enumitem}
\usepackage{hyperref}
\usepackage{multirow, makecell}
\usepackage{nccmath}
\usepackage{ragged2e}
\usepackage{threeparttable} 
\usepackage{tabularx}
\usepackage{xspace}
\usepackage{color}

\newcommand{\ie}{\emph{i.e.},\xspace}

\newcommand{\eg}{\emph{e.g.},\xspace}
\newcommand{\etc}{\emph{etc.}\xspace}
\newcommand{\etal}{\emph{et al.}\xspace}

\ifodd 1

\newcommand\rev[1]{\textcolor{black}{#1}}
\newcommand\rrev[1]{\textcolor{black}{#1}}

\else

\newcommand\rev[1]{#1}

\fi

\begin{document}

\title{Enabling Resource-efficient AIoT System with Cross-level Optimization: A survey}

% \title{Cross-level System Software for Resource-efficient AIoT Applications: A survey}

% % \thanks{This paper was produced by the IEEE Publication Technology Group. They are in Piscataway, NJ.}% <-this % stops a space
% \thanks{Manuscript received April 15, 2023.}
% }
%\markboth{Journal of \LaTeX\ Class Files,~Vol.~00, No.~0, April~2023}{}

\author{Sicong~Liu,~\IEEEmembership{Member,~IEEE,} 
        Bin~Guo,~\IEEEmembership{Senior Member,~IEEE, } 
        \textup{Cheng~Fang, }
        \textup{Ziqi~Wang, } 
        \textup{Shiyan~Luo, } 
        
        Zimu~Zhou,~\IEEEmembership{Member,~IEEE,} 
        Zhiwen~Yu,~\IEEEmembership{Senior Member,~IEEE}
        % <-this % stops a space
\IEEEcompsocitemizethanks{\IEEEcompsocthanksitem Sicong Liu, 
Bin Guo, Cheng Fang, Ziqi Wang, Shiyan Luo, and Zhiwen Yu were with the Department of Computer Science, Northwestern Polytechnical University, Xi'an,
China. 
Zhiwen Yu was also with the Harbin Engineering University, Harbin,
China.
% \protect 
% Email: guobin.keio@gmail.com
% \IEEEcompsocthanksitem 
Zimu Zhou was with the School of Data Science, City University of Hong Kong. 
Corresponding authors: Bin Guo (guobin.keio@gmail.com) and Zhiwen Yu (zhiwenyu@nwpu.edu.cn).
}
}

\maketitle

\begin{abstract}
The emerging field of artificial intelligence of things (AIoT, \textit{AI+IoT}) is driven by the widespread use of intelligent infrastructures and the impressive success of deep learning (DL). 
With the deployment of DL on various intelligent infrastructures featuring rich sensors and weak DL computing capabilities, a diverse range of AIoT applications has become possible. 
However, DL models are notoriously resource-intensive.
Existing research strives to realize near-/realtime inference of AIoT live data and low-cost training using AIoT datasets on resource-scare infrastructures.
Accordingly, the accuracy and responsiveness of DL models are bounded by resource availability.
To this end, the algorithm-system co-design that jointly optimizes the \textit{resource-friendly DL models} and \textit{model-adaptive system scheduling} improves the runtime resource availability and thus pushes the performance boundary set by the standalone level.
Unlike previous surveys on resource-friendly DL models or hand-crafted DL compilers/frameworks with partially fine-tuned components, this survey aims to provide a broader optimization space for more free resource-performance tradeoffs.
The cross-level optimization landscape involves various granularity, including the DL model, computation graph, operator, memory schedule, and hardware instructor in both on-device and distributed paradigms.
Furthermore, due to the dynamic nature of AIoT context, which includes heterogeneous hardware, agnostic sensing data, varying user-specified performance demands, and resource constraints, this survey explores the context-aware inter-/intra-device controllers for automatic cross-level adaptation. 
Additionally, we identify some potential directions for resource-efficient AIoT systems. 
By consolidating problems and techniques scattered over diverse levels, we aim to help readers understand their connections and stimulate further discussions.
\end{abstract}


\section{Introduction}
\label{sec:intro}

% \IEEEPARstart{A}{} number of applications have been developed for consumer, enterprise, and societal AIoT, including predictive maintenance, intelligent healthcare, smart cities, housing, \etal.

\IEEEPARstart{T}he Artificial Internet of Things (AIoT), also known as \textit{AI+IoT}, was coined in 2017 and quickly gained widespread attention~\cite{ghosh2018artificial}.
On the one hand, the rapid development of deep learning (DL) has led to the emergence of numerous intelligent services.
% , such as autonomous driving~\cite{nguyen2018deep} and security monitoring~\cite{li2021deep}. 
% On the other hand, the richer sensors and enhanced DL computing capabilities of intelligent infrastructures have given rise to a new category of devices known as \textit{AIoT devices}, which are distinct from traditional IoT sensor nodes in terms of a broader range of sensing capabilities, performing complex computations directly on the device, and greater connectivity capabilities.
\rev{On the other hand, the richer sensors and enhanced DL computing capabilities of intelligent infrastructures have given rise to a new category of devices known as \textit{AIoT devices}. 
AIoT devices are distinct from traditional IoT sensor nodes due to their broader range of sensing capabilities, ability to perform complex computations directly on the device, and greater connectivity capabilities.
}

Moreover, there is a growing trend to integrate DL-powered intelligence into tiny embedded AIoT devices for two primary reasons. 
\textit{First}, the proliferation of AIoT devices has resulted in a massive increase in distributed sensing data captured in various modalities~\cite{cisco}. 
By executing DL inference and training tasks on resource-scarce AIoT devices, rather than transmitting data to remote centers, \eg cloud, like traditional IoT, we can save bandwidth and latency while guaranteeing recognition accuracy.
% from multiple viewpoints.
% with diverse physical properties. 
%
\textit{Second}, AIoT applications such as medical assistance and security monitoring collect sensitive user information, posing well-known privacy risks~\cite{hassan2019privacy}. 
It is preferable to process data locally or at trusted nearby devices. 
However, modern DL models are notoriously resource-intensive, making it challenging to achieve real-time inference and low-cost training on resource-scarce AIoT devices.

\begin{figure}[t]
  \centering
  \includegraphics[width=.48\textwidth]{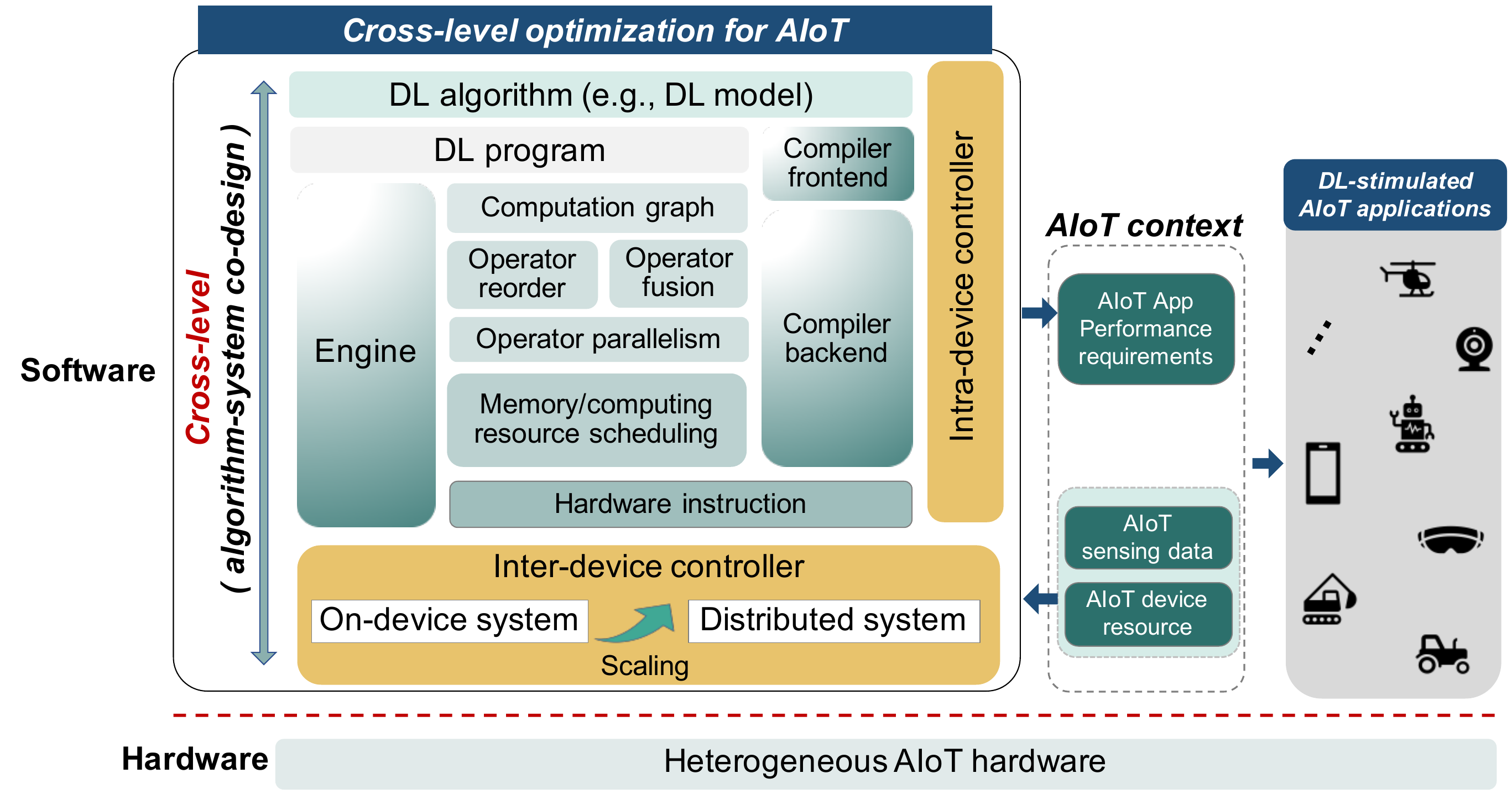}
  \caption{Illustration of cross-level optimization for the resource-efficient AIoT system, spanning the resource-friendly algorithm, model-adaptive system scheduling, to context-aware intra-/inter-device controllers.} 
  \label{fig:sys_plus}
\end{figure}

Given these challenges, previous research has explored \textit{resource-friendly DL models} and \textit{resource scheduling} techniques, either individually or in combination.
\textit{First}, \textit{resource-friendly DL model compression} have been widely investigated to reduce the resource demands of DL models at the algorithm level.
They include standalone model compression techniques~\cite{howard2017mobilenets, ray2021review}, and automated neural architecture search (NAS) frameworks~\cite{guo2021mistify, liu2019edge, liu2021adaspring}.
However, despite extensive research on model compression, compressed DL models typically compromise accuracy to reduce resource demands.
% (\eg accuracy, latency, memory usage, computation cost, and energy efficiency).
\textit{Second}, some \textit{hand-crafted DL compilers/frameworks} fine-tune diverse system levels to reuse the input data and reduce runtime overhead.
For example, TVM~\cite{chen2018tvm} optimizes DL operators at the computation graph level, Tensorflow Runtime (TFRT)~\cite{tensorflowruntime} implements efficient execution of the computing kernel, and TensorflowXLA~\cite{tensorflowxla} designs a linear algebra compiler engine for the TensorFlow framework.
\textit{Third}, algorithm-system co-design brings better performance.
For example, MCUNet~\cite{lin2020mcunet} co-designs the TinyNAS, for DL model design, and TinyEngine, for code compilation and memory scheduling.
However, algorithm-system co-design is difficult for non-experts.
And existing techniques only manually optimize partial system levels to provide a feasible solution.

To this end, providing a relatively complete picture of the cross-level optimization space is necessary. 
This can assist researchers in finding suitable technique combinations for their specific requirements more freely and help automated frameworks build a finer-grained search space to push the boundary of performance-resource tradeoffs. 
We summarize the cross-level optimization space as follows:
\begin{itemize}
\item \textit{Resource-friendly algorithm level} is concerned with specifying DL models to balance performance and resource constraints. 
However, the accuracy of lightweight models is degraded and bounded by resource budgets. 
\item \textit{Model-adaptive system scheduling level} comprises various fine-grained granularities, \eg computation graph, operator, memory, compiler, engine, and instructor.
This level aims at utilizing hardware resources to their fullest capacity without compromising model accuracy.
% Diverse enabling techniques over these levels can increase data reuse, avoid memory fragmentation, and reduce data movement to different degrees.
\end{itemize}
Joint optimizing across these levels with bi-directional feedback can improve runtime resource availability and performance-resource tradeoff for AIoT.
However, no existing surveys have extensively covered all of these levels.
The most related previous surveys include \cite{benmeziane2021hardware, joshi2022enabling, wang2020convergence,chen2019deep,li2020deep}.
\textit{First}, \rev{many mobile and embedded DL surveys focus on resource-friendly model compression~\cite{benmeziane2021hardware, zhang2021compacting} from the algorithm level}. 
\textit{Second}, surveys on edge computing~\cite{wang2020convergence,chen2019deep} spanning networking and computation offloading~\cite{chen2019deep, joshi2022enabling}, pay little attention to the distributed cross-level optimization, \eg data movement in partitioned computation graphs and memory allocation across devices at runtime.
\textit{Third}, \cite{li2020deep} maps the DL computation to hardware. 
However, it targets DL-oriented compiler optimization, which does not co-design the DL models with compilers.
This paper comprehensively analyzes the resource-efficient AIoT systems, covering all of these levels. 
As depicted in Figure \ref{fig:sys_plus}, \rev{it not only encompasses optimization techniques used in stand-alone fields like on-device DL~\cite{lecun2015deep}, distributed DL~\cite{tang2022computational, zhou2019edge}, and tiny systems~\cite{nauth2009embedded} but also expands the optimization possibility in the AIoT context.}

In particular, the dynamic nature of AIoT context poses a significant challenge: how to adaptively optimize cross-level DL systems to meet varying application performance requirements while satisfying resource constraints.
The \textit{dynamic context} includes a range of factors, including heterogeneous AIoT resources, agnostic live data, varying user-specified performance demands, and device-imposed resource constraints.
To address these challenges, the AIoT system should also contain context-aware controllers across these levels:
\begin{itemize}
\item \textit{Intra-device cross-level controller} establishes a control flow across different system levels to automate the dynamic context awareness, optimization technique combination, and adaptive co-design loop.
\item \textit{Inter-device cross-level controller} judges the complementarity of distributed AIoT sensing data and resources, scaling cross-level systems between on-device and distributed schemes. 
\end{itemize}

\rev{Therefore, we identify essential enabling technologies for diverse tasks. 
Each of them encompasses cross-layer optimization, but with different constraints.
}
\rev{
\begin{itemize}
    \item Cross-level optimization for On-device DL inference ($\S$ \ref{subsec:inference_1}). 
    It aims to achieve better real-time performance and higher accuracy by minimizing DL model redundancy and maximizing on-device resource capability.
    \item Cross-level optimization for Distributed DL inference ($\S$ \ref{subsec:inference_m}).
    By aggregating more computing resources and sensing sources, it can further optimize latency and  accuracy than the on-device scheme. 
    It operates with a similar cross-level spectrum but utilizes distributed scheduling.
    \item Cross-level optimization for on-device DL training ($\S$ \ref{subsec:train_1}).
    DL training is more complex than inference.
    It aims to reduce training costs and maintain accuracy.
    \item Cross-level optimization for distributed DL training ($\S$ \ref{subsec:train_m}).
    It further coordinates data fusion and resource aggregation to optimize DL training efficacy and efficiency.
    % across swarm.
    \item Resource-efficient AIoT applications ($\S$ \ref{sec:system}).
    The aforementioned techniques and systems stimulate a wide range of AIoT applications with flexibility and adaptivity.
\end{itemize}
}

In summary, the key contributions of this work can be summarized as follow:

\begin{itemize}
    \item To the best of our knowledge, this is the first to describe the characteristics and architectures of the resource-efficient AIoT system exactly.
    It provides a \textit{cross-level spectrum} of the system optimization space for AIoT.% 
    \item We propose a novel taxonomy of existing techniques, summarizing how state-of-the-art address issues across different levels of the resource-efficient AIoT system.
    Additionally, we demonstrate how context-aware controllers can automatically select cross-level techniques for AIoT.
    \item 
    We discuss open issues in resource-efficient AIoT systems and suggest potential future research directions.
\end{itemize}

This section introduces the background and leads to the motivation for this survey. 
\rev{
In the rest of this paper, we present fundamentals of the resource-efficient AIoT system in $\S$ \ref{sec:overview}, introduce the enabling techniques across diverse levels for DL inference and training tasks in $\S$ \ref{sec:inference} and $\S$ \ref{sec:train}, respectively.
And then, we list related AIoT systems and applications in $\S$ \ref{sec:system}.
Finally, we discuss the open issues in $\S$ \ref{sec:issue} and conclude this paper in $\S$ \ref{sec:conclu}.
}
\section{Fundamentals of Resource-efficient AIoT system}
\label{sec:overview}

\begin{figure*}[t]
	\centering 
	\subfloat[Development of related areas]{\label{fig:concept_1}
		\includegraphics[height=0.125\linewidth]{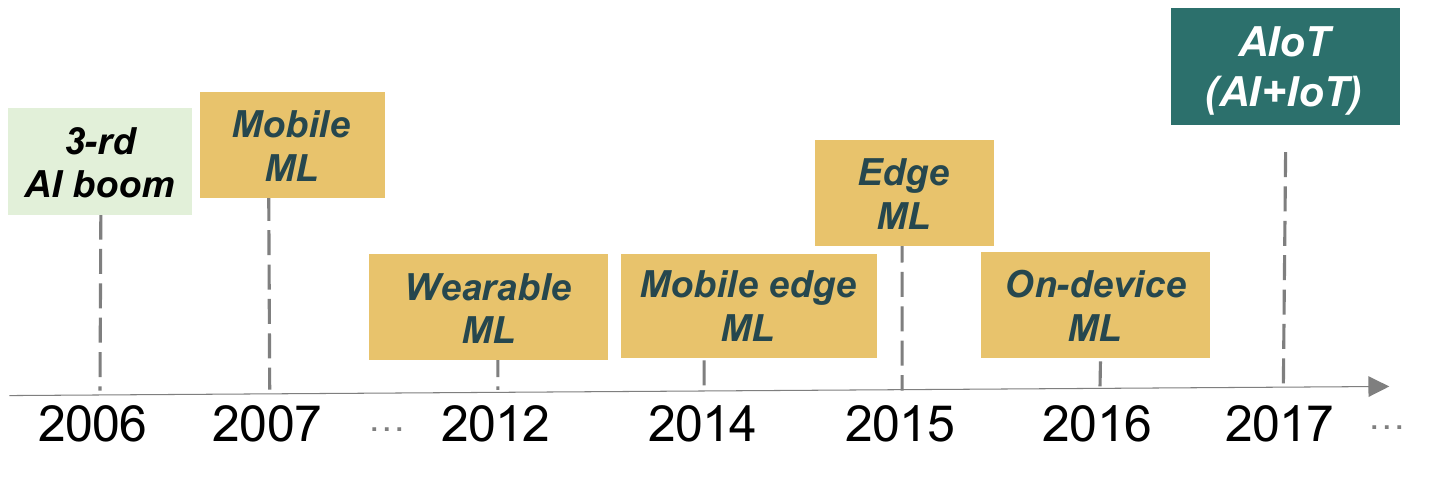}}
    \hspace{2mm}
    \subfloat[Development of DL models (a showcase of vision fields) and frameworks]{\label{fig:concept_2}
		\includegraphics[height=0.15\linewidth]{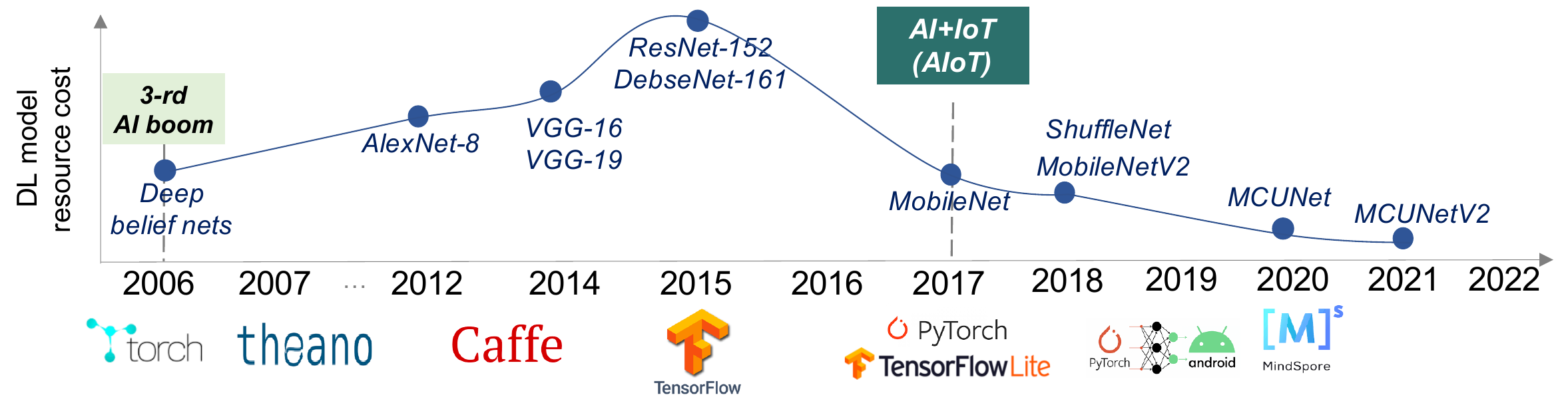}}
\caption{Illustration of various related areas and frameworks.}
\label{fig:concept}
\end{figure*}

This section presents an overview of the resource-efficient AIoT system, departing from existing related areas.

\begin{figure}[t]
  \centering
  \includegraphics[width=.48\textwidth]{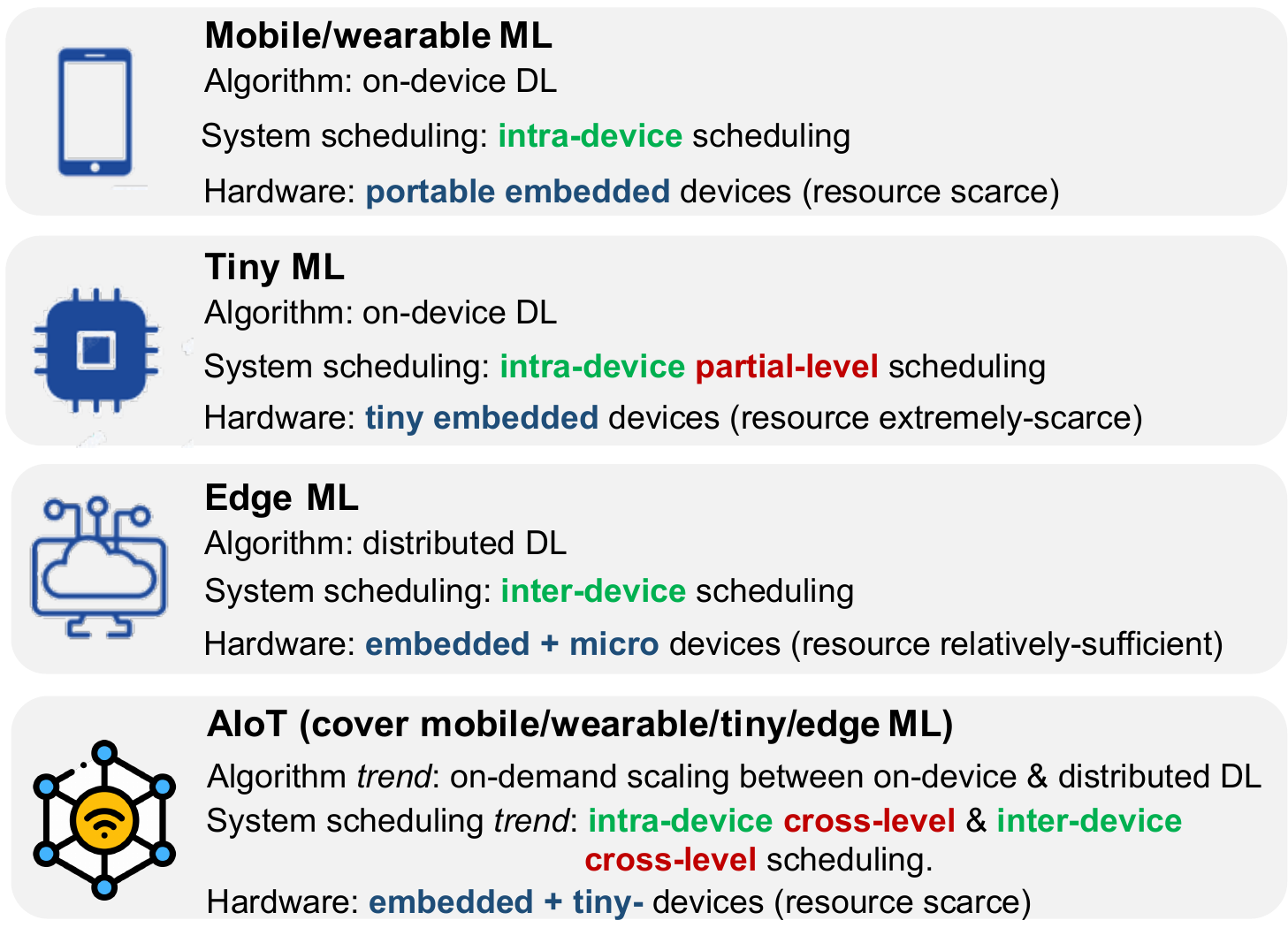}
  \caption{Difference of related paradigms.}
  \label{fig:area}
  \vspace{-3mm}
\end{figure}

\subsection{AIoT Paradigms}
\rev{
\textit{Artificial Intelligence of Things (AIoT)} refers to the integration of artificial intelligence (AI) technologies with Internet of Things (IoT) infrastructures to improve data analytics~\cite{ghosh2020aiot}.}
The primary computational task in the resource-efficient AIoT system is \textit{data analysis}.
As depicted in Figure \ref{fig:concept}, deep learning (DL) has emerged as the dominant AI methodology for learning and analyzing data since the third AI boom~\cite{aiboom}.
And there is a growing trend to incorporate DL-powered intelligence into tiny embedded AIoT devices with the advancement of embedded hardware and on-device DL technologies~\cite{saha2022machine}.
In addition, new development frameworks have been launched specifically targeting embedded devices, such as TensorFlow Lite~\cite{tensorflowlite}, Caffe2~\cite{caffe2}, and Pytorch Mobile~\cite{pytorchmobile}, in order to promote DL-based AIoT applications.
% As shown in Figure \ref{fig:concept}, DLs have become the mainstream AI method to learn and analyze data since the third AI boom, spanning deep discriminative models, deep generative models, and deep reinforcement learning models~\cite{aiboom}.
% %
% And there is a growing trend to bring DL-powered intelligence to AIoT devices with the development of embedded hardware and on-device AI technologies~\cite{saha2022machine}.
% % 
% New development frameworks targeted at embedded devices have also been launched (\eg TensorFlow Lite~\cite{tensorflowlite}, Caffe2~\cite{caffe2}, Pytorch Mobile~\cite{pytorchmobile}) to encourage DL-based AIoT applications/services. 

Several related areas utilize overlapping enabling techniques but have varying focuses, as shown in Figure~\ref{fig:concept_1} and Figure~\ref{fig:area}. 
We differentiate them below.

    \textit{1) Mobile}~\cite{nazir2019internet} and \textit{Wearable ML} ~\cite{cicceri2020deep} focus on mobile data analytic patterns and application experience (\eg real-time response) on \textit{portable embedded} mobiles and wearables. 

     \textit{2)  Tiny ML}~\cite{ray2022review} concerns machine learning aware architectures, frameworks, techniques, tools, and approaches which are capable of performing on-device analytics at \textit{tiny embedded} devices with extremely limited resources, \eg \rev{Microprogrammed control unit} (MCU).
    And TinyML project~\cite{TinyML} aims to improve the efficiency of  systems by requiring less computation, fewer engineers, and fewer data to facilitate the giant market of tiny embedded applications.
    It covers the management of data and the deployment of models.
    
    \textit{3) Edge ML} presents to keep data near where it's generated, avoiding costly and privacy-threatening data transfers~\cite{mendez2022edge}. 
    Instead of shipping data centrally to perform data analytics, edge computing analyzes data using ML at the edge while maintaining accuracy and latency.
    The edge includes embedded and micro devices with relatively sufficient resources.

As these related areas evolve, many enabling techniques and frameworks can facilitate AIoT from different aspects. 
Techniques such as DL models for mobile and edge computing~~\cite{cicceri2020deep}  and on-device/distributed DL deployment~\cite{nazir2019internet, mendez2022edge} provide various model compression and offloading algorithms for inference and training.
Frameworks such as TFlite\cite{tensorflowlite}, CMix-NN ~\cite{capotondi2020cmix}, TVM~\cite{chen2018tvm}, TensorFlow XLA~\cite{tensorflowxla}, and oneDNN~\cite{onednn} offer a range of acceleration options and memory scheduling support for DL.
The key distinction lies in the AIoT system's ability to integrate these technologies in a novel, context-aware and cross-level manner.
Specifically, we define some important concepts as follows:

\textbf{\textit{AIoT computing task and paradigm}}. 
\textit{AIoT data} mainly includes two types, \ie \textit{live sensing data} and accumulated \textit{dataset}.
And their analysis is concentrated in two stages, \ie DL \textit{inference} and \textit{training}, respectively.
\begin{itemize}
\item \textit{Near-/realtime inference of AIoT live data on resource-scare AIoT devices}. The AIoT \textit{live data} are sensed by distributed AIoT devices and should be analyzed fastly.
\item \textit{Low-cost training using AIoT dataset on resource-scare AIoT devices}. 
The AIoT \textit{dataset} are sensed and held by distributed AIoT devices and should be consumed with low data transmission and training costs, \eg memory.
\end{itemize}
Alternatively, \rrev{inference and training tasks can be performed on-device or at distributed edges within the networked system}, which enable a plethora of innovative AIoT applications that go beyond the conventional IoT paradigm in terms of accuracy, latency, bandwidth, privacy, and energy efficiency \cite{ghosh2018artificial}.

\rev{
\textbf{\textit{AIoT devices}} refer to intelligent end/edge devices equipped with advanced sensors and DL computing capabilities. 
This category includes mobiles, wearables, robots, drones, and other physical devices. 
In this context, the "end" device serves as the primary sensing source. 
Similarly to edge computing, the "edge" refers to devices located between the sensing data source and the cloud center path. 
In the realm of AIoT, there is a preference for prioritizing physically nearby edge devices to enable cost-effective near-sensing-data computing.
}

\textbf{\textit{Resource-efficient AIoT system}}.
Unlike software-hardware co-design approaches~\cite{wan2022sensor, zhou2020near}, our focus is on the system software, specifically the \textbf{\textit{algorithm-system co-design}}.
This approach allows for effective utilization of existing hardware resources in dynamic AIoT contexts.
AIoT applications like smart home and smart retail rely on a plethora of AIoT devices equipped with advanced sensors and embedded processors. 
Replacing these devices with newer hardware can often be costly and impractical. 
Therefore, adopting a strategy that optimizes the usage of available hardware resources can be an efficient and cost-effective solution in such scenarios.
We refer to the DL-based system on AIoT devices as \textbf{\textit{resource-efficient AIoT system}} in the following part for short.
The resource-efficient AIoT system is responsible for distributing and federating AIoT sensing data analytics across resource-constrained and heterogeneous end/edge devices for DL inference and training.
It is worth noting that communication networking for AIoT is outside the scope of this survey.

\subsection{Cross-level Characteristics of AIoT System}
\label{subsec:sys_layer}

\begin{figure}[t]
  \centering
  \includegraphics[width=.48\textwidth]{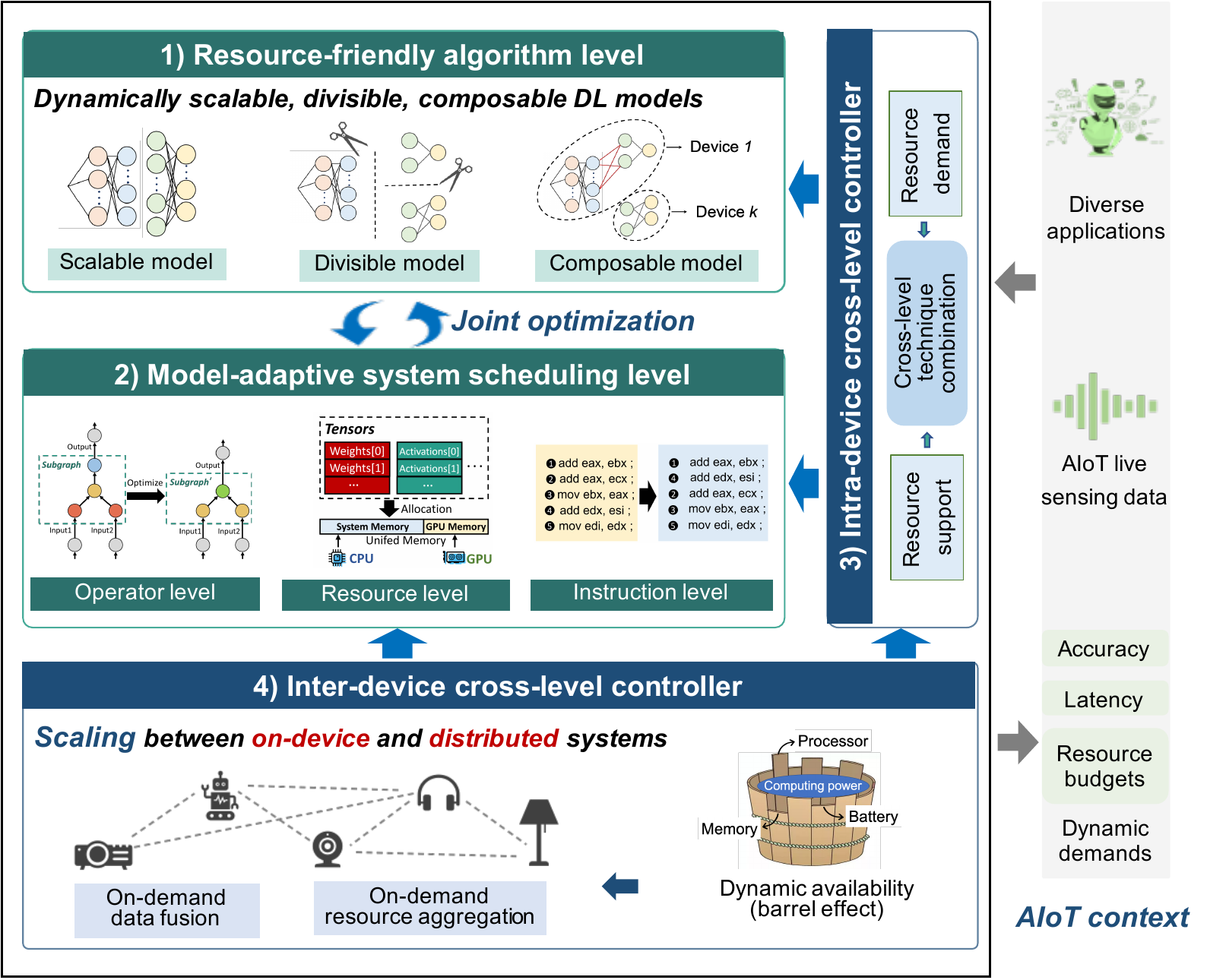}
  \caption{Resource-efficient AIoT system architecture, involving two joint-optimized levels and two context-aware controllers.}
  \label{fig:layer}
  \vspace{-3mm}
\end{figure}

We provide an overview of the remarkable cross-level characteristics and architectures of resource-efficient AIoT systems.
As illustrated in Figure \ref{fig:layer}, the resource-efficient \rev{AIoT system comprises two key levels: the \textit{resource-friendly algorithm} and \textit{model-adaptive system scheduling}. 
These levels should be \textit{co-designed} to ensure coherent resource usage.
We outline their specific features and functionality below.
}

\rev{\textit{\textbf{1) Resource-friendly algorithm level: dynamically scalable, divisible, and composable DL models}}.}
The DL models for AIoT should be \textit{dynamically scalable, divisible, and composable} in both DL inference and training tasks.
\textit{First}, the DL model (\eg structure, parameter size) should be \textit{scalable} with diverse compression degree to satisfy the platform-imposed dynamic resource constraints (\ie memory, computing, and battery) and application-specified performance demands (\ie accuracy, latency, energy cost).
This scheme will inevitably bring accuracy fluctuations if the DL models are poorly designed.
\textit{Second}, DL models' divisible and composable properties are necessary to offload computation to distributed AIoT devices.
Particularly, the divisibility of DL models depends on the internal dependency of DL layers, channels, and operators.
And the composable property of DL models is able to adjust the system to offload different DL block combinations to diverse AIoT devices for dynamic resource availability.

\rev{\textbf{\textit{2) Model-adaptive system scheduling level: maximizing runtime hardware capability}}.}
\rev{This level aims to utilize hardware resources to their fullest capacity without compromising model accuracy.}
Even for identical DL model configurations, mapping different model layers/operators onto diverse memory units in varying sequences results in different latency and resource overhead~\cite{lin2022device,rhu2016vdnn,chen2018tvm}.
For example, integrating the memory fragmentation in the tensor layout of SqueezeNet can reduce $42\%$ wasted memory fragmentation~\cite{wang2018superneurons}.
Therefore, designing appropriate strategies at computation graph, operator, and memory allocation levels to cater to the upper-level models can enhance runtime resource availability.
\rev{
Notably, this level can \textit{iteratively allow a more flexible DL model design space for the algorithm level, thereby pushing the limitations on accuracy-resource trade-offs}.
}

\rev{\textbf{\textit{3) Intra-device controller: automating the adaptive on-device cross-level optimization}}.
}
The efficacy of various optimization techniques at different levels can vary for the same DL model. 
And even within the same level of optimization techniques, there can be differences in their performance.
As a result, it is imperative to develop an extra control flow to automate the adaptive optimization of DL models and system scheduling in a cross-level manner. 
Also, adaptively adjusting the cross-level techniques based on dynamic contexts, such as input data, resource availability, and user demands, is necessary.
The controller monitors the \textit{resource availability} of the target platforms and predicts the \textit{resource requirements} of the AIoT system based on the current model configurations and scheduling strategies.
The controller automatically adjusts techniques across different levels if the resource demand exceeds the supply or does not align with the user-defined budgets.

\begin{figure}[t]
  \centering
  \includegraphics[width=.46\textwidth]{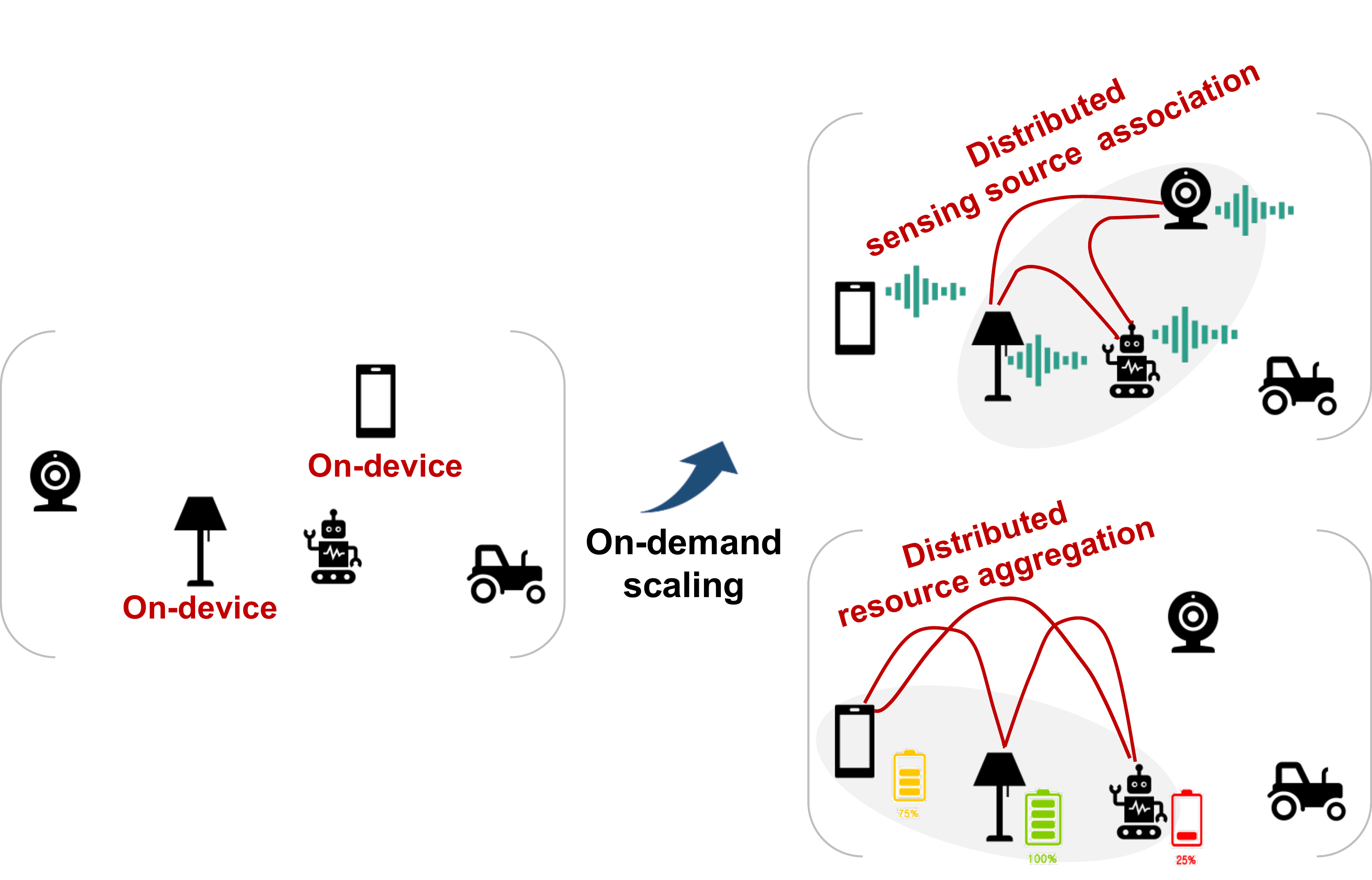}
  \caption{The inter-device controller scales the cross-level AIoT system between on-device and distributed schemes for \rev{on-demand sensing source association or resource aggregation}.}
  \label{fig:scaling}
  \vspace{-3mm}
\end{figure}

\rev{\textit{\textbf{4) Inter-device controller: automating the adaptive distributed cross-level optimization}}.
}
It scales the cross-level AIoT system from on-device to distributed schemes \rev{for achieving better performance-resource efficiency trade-off.}
As shown in Figure \ref{fig:scaling}, the distributed AIoT devices collaborate on demand for two motivations:
    
\begin{itemize}
\item \textbf{\textit{On-demand sensing source association}}. 
With the proliferation of data-rich sensors (\eg cameras, LIDAR, and hyperspectral imagers), 
\rev{different distributed sensing sources have temporal connections for specific DL training/inference tasks.}
Distributed multi-modal data benefit the environment/object recognization from different vantage points with various physical properties~\cite{li2021low, petridis2018end,tian2018audio,radu2018multimodal}.
The inter-device controller needs 
\rev{balance the necessity of sensing source association \rrev{within the networked system} to the accuracy of AIoT tasks and the overhead.}

% \TODO{here}
\item  \textbf{\textit{On-demand computing resource aggregation}}.
As mentioned above, AIoT prioritizes the on-device scheme, followed by the distributed collaboration scheme with nearby edges, to realize near-sensing-data computing for saving transmission bandwidth and protecting data privacy~\cite{ning2019deep,lin2020survey}.
Executing DL models requires large computing/memory resources that are not always available in a single AIoT device, significantly when the scale and complexity of DL models continuously increase. 
Scheduling the most suitable distributed devices within locally connected and resource-constrained edge clusters is necessary yet challenging~\cite{verbraeken2020survey,nguyen2021federated}.
    
\end{itemize}

% \begin{figure}[t]
% 	\centering 
% 	\subfloat[AIoT system for DL inference]{\label{fig:cross-device-infer}
% 		\includegraphics[width=0.45\linewidth]{cross-device-infere.pdf}}
%         \hspace{2mm}
%         \subfloat[AIoT system for DL training]{\label{fig:cross-device-train}
% 		\includegraphics[width=0.45\linewidth]{cross-device-train.pdf}}
% 	\caption{Platform scheduling for DL inference and training tasks in AIoT. Devices collaborate on-demand for two motivations, \ie distributed data fusion and resource aggregation.}
%     \label{fig:cross-device}
% \end{figure}

\begin{figure*}[t]
  \centering
  \includegraphics[width=.98\textwidth]{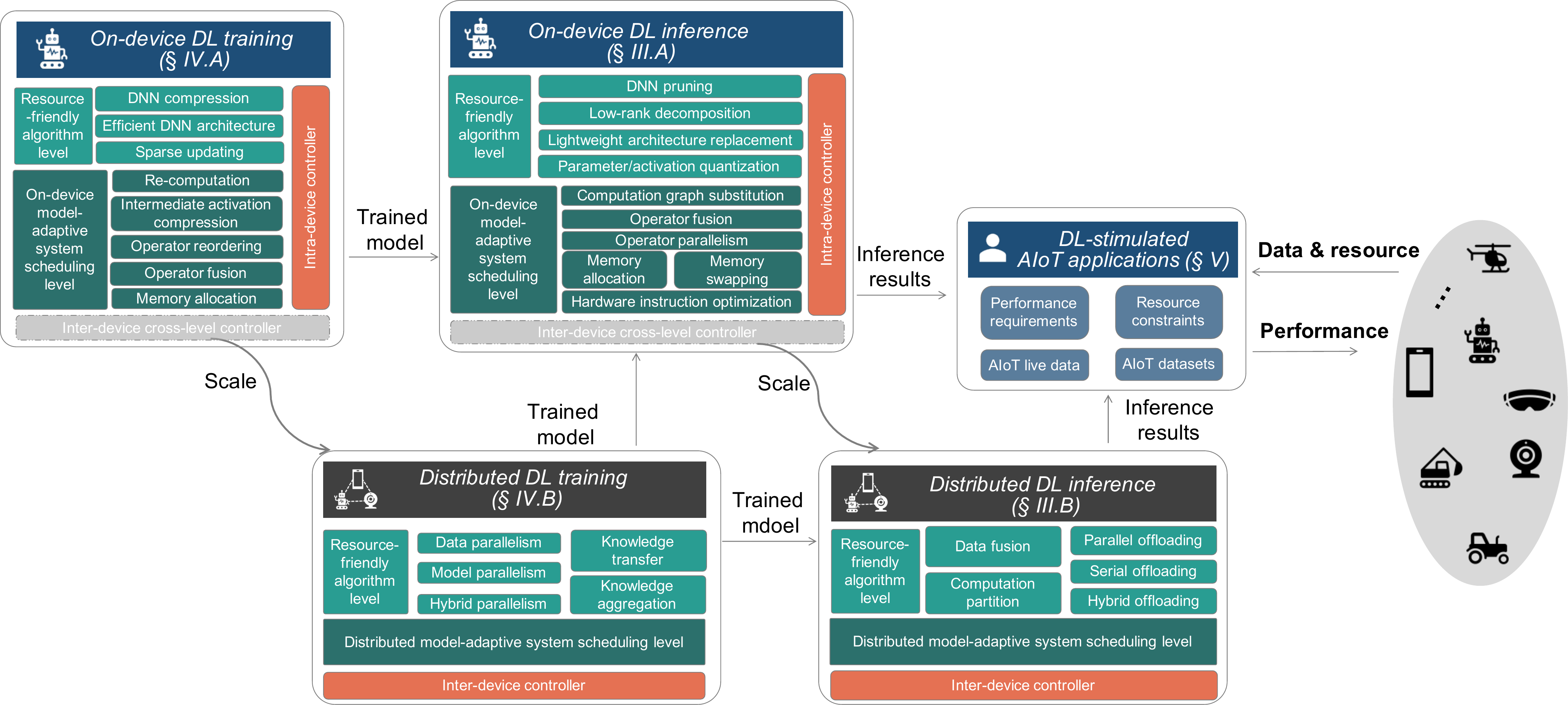} 
  \caption{\rev{Landscape of the resource-efficient AIoT system according to the proposed technique taxonomy}.}
  \label{fig:overview_4}
  \vspace{-2mm}
\end{figure*}

\subsection{Taxonomy of Enabling Techniques}
\label{sebsec:texonomy}
We identify the following essential enabling technologies for resource-efficient AIoT systems, \rev{considering the cross-level deployment issues mentioned earlier}.
Figure \ref{fig:overview_4} summarizes our taxonomy, \ie cross-level optimization for both resource-efficient \textit{DL inference and training tasks}.

\subsubsection{Cross-level Optimization for DL Inference Tasks}
There is a growing trend today to bring DL-powered intelligence into AIoT devices for various applications, \eg object recognition~\cite{tan2020equalization, zhou2020look, abu2021some, bib:xu2023edge}, semantic segmentation~\cite{yuan2020object, wang2021exploring, xie2021segformer}, object tracking~\cite{wang2019fast, zhang2022bytetrack, meinhardt2022trackformer}, natural language processing~\cite{galassi2020attention, galassi2020attention, liu2023pre}, and recommendation~\cite{wang2019sequential, naumov2019deep, wu2022graph}.

\textbf{\textit{Challenges}}.
It is non-trivial to achieve near-/real-time DL inference with limited resources in AIoT devices, which is critical for AIoT live data to satisfy the applications’ responsiveness.
First, on-device DL inference benefits user privacy and robustness.
However, mainstream DL models, such as Zero-DCE for low-light video enhancement~\cite{guo2020zero}, are still computation-intensive and fail to achieve real-time processing on the local AIoT device.
Second, distributed DL inference strives to satisfy stringent demands across multiple dimensions, \eg latency, accuracy, and transmission/resource cost.
Moreover, energy savings in DL inference is also crucial for long-term running applications since most mobile AIoT devices are  battery-powered~\cite{flinn1999energy,benini2000system}.
Also, it is desired to adapt the inference accuracy according to the resource availability and network condition for AIoT at runtime.
We detail different levels of issues in the following part.
    
\textbf{\textit{On-device DL Inference}}.  
Prior efforts explored several technologies to enable on-device DL inference, accelerate inference, and save memory occupation or energy.
\textbf{\textit{(i) resource-friendly algorithm level}} compresses DL models to reduce the resource demand without significantly compromising their accuracy.
Standalone techniques include pruning~\cite{hu2016network},\cite{luo2017thinet},\cite{he2017channel}, low-rank decomposition~\cite{kaloshin2020convolutional}, lightweight architecture replacement~\cite{lin2020mcunet, liu2021adaspring},\cite{liberis2021munas}, and parameter/activation quantization~\cite{rusci2020memory},\cite{rusci2020leveraging}.
Some research~\cite{liu2020adadeep} also automatically combines diverse compression techniques to achieve better performance-resource tradeoffs. 
Although significant progress has been made in this field, compressed models typically yield accuracy degradation.
\textbf{\textit{(ii) model-adaptive system scheduling level}} spans multiple fine-grained levels, \ie computational graph ~\cite{ding2021ios,cai2022optimus,niu2021dnnfusion}, memory scheduling~\cite{miao2021enabling,lin2020mcunet}, hardware instruction~\cite{huang1999generalized,im2001optimizing}, compiler front-end/back-end~\cite{cai2022optimus, niu2021dnnfusion,boehm2018optimizing}, and engine~\cite{lin2020mcunet}. 
For example, existing DL frameworks, \eg Tensorflow~\cite{abadi2016tensorflow}, Pytorch~\cite{paszke2017automatic}, and TVM~\cite{chen2018tvm}, provide support in computation graph and operator optimization.
IOS~\cite{ding2021ios} extensively studies the inter-operator parallelism to accelerate inference.
Miao \etal~\cite{miao2021enabling} proposed dynamically swapping data between MCU's micro-SRAM and external flash to save SRAM.
The suitable underlying scheduling can further improve resource availability than the algorithm-level compression models.
However, most existing techniques are manually designed.
\textbf{\textit{(iii) intra-device cross-level controller}} aims to adaptively select the above-mentioned cross-level optimization techniques according to the user-specified performance demands and the device-imposed resource budgets. 
For example, AdaDeep~\cite{liu2020adadeep} is an automated DNN compression framework at the algorithm level that uses deep reinforcement learning to balance performance and resources. 
MCUNet~\cite{lin2020mcunet} performs joint optimization across the algorithm and engine levels, adapting the optimization strategies according to memory constraints.
However, the combination criteria of cross-level techniques from a broader space remains a black box.
% \textbf{\textit{(iv) intra-device cross-level controller}} will scale the system from on-device inference to distributed inference once the resource limitation is not satisfied.
 \textbf{\textit{(iv) inter-device cross-level controller}} will scale up the system from the on-device to distributed scheme once the large computing/memory resources required by executing computation-intensive models locally are unavailable.
 % And optimize 

\textbf{\textit{Distributed DL Inference}}.
% Deploying different parts of DL models to multiple AIoT devices can aggregate the calculation capability of multiple devices for reducing local resource demands and improving inference efficiency.
Deploying different parts of DL models on multiple AIoT devices can harness the collective computational power of these devices, thereby decreasing local resource demands and enhancing inference efficiency.
\textbf{\textit{(i) resource-friendly algorithm level}} involves 
DL model partition~\cite{yun2022cooperative}, offloading~\cite{wu2020accuracy}, and performance  tradeoff~\cite{wu2020accuracy,he2020joint}. 
According to the operator dependency of DL models and the resource availability of edge devices, various parts of DL models can be distributed to multiple devices for either sequential or parallel processing, resulting in different performance tradeoffs.
For example, \cite{he2020joint} jointly carries out the model partition and offloading to improve accuracy, energy consumption, and latency.
\textbf{\textit{(ii) model-adaptive system scheduling level}} includes computation graph partition~\cite{mao2017modnn,mao2017mednn}, distributed operator fusion~\cite{zhao2018deepthings,stahl2021deeperthings},  separately data reuse and memory allocation.
For example, Modnn~\cite{mao2017modnn} significantly speeds up DL inference by introducing execution parallelism among multiple  devices~\cite{mao2017mednn}.
Zhang \etal~\cite{zhang2021deep} proposes jointly optimizing heterogeneous chips' computing frequency, power, and memory to achieve the optimal allocation strategy on distributed devices.
This area is relatively less explored.
\textbf{\textit{(iii) inter-device cross-level controller}} has a similar cross-level technique spectrum as the on-device inference scheme.
And the fundamental difference lies in the fact that the underlying operator and resource scheduling for distributed schemes must be \textit{optimized separately and verified globally}.
This is because different parts of the model deployed on multiple devices have separate memory pools and can not reuse data, necessitating separate optimization at diverse devices.
Meanwhile, global evaluation is required for overall performance, such as the total latency of distributed execution and transmission.
%
% x/\textbf{\textit{(iv) inter-device cross-level controller}} 
In addition, the controller selects the most suitable devices within the closely connected and resource-constrained edge.
It involves the performance profiler~\cite{zeng2020coedge}, adaptive inference serialization or parallelism~\cite{hadidi2020toward}, and autonomatic optimizer~\cite{zhang2021deepslicing}.

\subsubsection{Cross-level Optimization for DL Training Tasks}
There are many demands for DL training on resource-scarce AIoT devices.
For instance, we may need to update pre-trained DL models locally or nearby when the live sensing data drift to the original training data and an Internet connection to the cloud is unavailable.
Other requirements for DL training on AIoT devices are also widespread, such as updating models to adapt to new applications. 
These requirements can be fulfilled using techniques like transfer learning~\cite{pan2010cross}~\cite{chattopadhyay2012multisource}~\cite{duan2012domain}~\cite{oquab2014learning}, domain adaptation~\cite{wang2022continual}~\cite{shen2022connect}~\cite{gal2022stylegan}~\cite{xie2022active}, continuous learning~\cite{xu2020commerce}~\cite{irfan2021novel}~\cite{yuan2021reconfigurable}, and personalized federated learning~\cite{huang2022learn}~\cite{fang2022robust}.
Besides, DL model fine-tuning is necessary after adaptive compression, which has been demonstrated in various practical scenarios~\cite{tang2017train}~\cite{liu2017hierarchical}~\cite{tjandra2018tensor}~\cite{zhong2018practical}.

\textbf{\textit{Challenges}}.
It is intractable to realize low-cost DL training on AIoT devices.
The reasons are three-fold:
\textit{i) resource constraint}. 
The bottleneck of embedded resources in AIoT devices is the memory access bandwidth~\cite{cai2020tinytl}.
While DL training needs \textit{batched memory} chunks grouping multiple data samples for feature learning.
Besides, the computation efficiency will be low if the  memory for sufficient batch size cannot be secured.
Because computations are highly sensitive to memory access schemes, \eg Cache hit rate~\cite{liu2022distributed};
\textit{ii) irregular activation lifecycle}. 
The DL training phase involves forward propagation and backpropagation to update the model weights iteratively.
Intermediate activations pose high memory demands, produced during the forward pass and reused during the backward pass~\cite{moon2022nntrainer}.
In the case of DL inference using only forward propagation, resources occupied by activations can be directly released.
However, during DL training involving both forward and backpropagation, activations must be retained throughout the process. 
Model structures, such as control flow and branching, affect the lifecycle of activations during backpropagation, which is less regular. 
Therefore, it is difficult to determine when data will be accessed for the last time and when it is safe to release resources.
\textit{(iii) multiple iterations.} 
Unlike DL inference's "one-time" property, DL training requires optimization across multiple iterations.
Therefore, even small instabilities can be amplified over hundreds of iterations, potentially leading to the crash of DL training~\cite{lin2022device}. 
Additionally, current cloud-based DL training optimization techniques are not suitable for resource-scare AIoT devices.

\textbf{\textit{On-device DL Training}}.
Given these challenges, we summarize the enabling techniques in $\S$ \ref{subsec:train_1} that ensure sufficient resource supply and guarantee performance across different system levels.
\textbf{\textit{(i) resource-friendly algorithm level}} aims to reduce resource demands, especially memory usage, through model or training simplification.
Techniques include model quantization~\cite{deng2015reduced}, model compression~\cite{deng2015reduced, bulo2018place, jung2019restructuring}, sparse updating~\cite{liu2018dynamic,dai2020sparsetrain}, \etc
For example, TinyTL~\cite{cai2020tinytl} presents the element-wise convolution decomposition to reduce memory, not parameters, for efficient on-device DL training.
\textbf{\textit{(ii) model-adaptive system scheduling level}} mainly optimizes three objects, \ie \textit{intermediate activation}, \textit{computation graph}, and \textit{memory schedule}.
Precisely, to trim down the \textit{intermediate activation tensor} after the forward and before the backward pass, researchers present the recomputation~\cite{chen2016training,gomez2017reversible} and activation compression~\cite{jain2018gist} techniques.
They discard or compress the intermediate activation tensors to reduce the peak memory during DL training.
Optimization techniques at the \textit{computation graph} level include operator reordering~\cite{lin2021memory} and operator fusion~\cite{jia2019optimizing,zhou2020transferable}.
% \TODO{pausing..}
%
Rearranging or fusing operators in the computation graph can reduce memory usage and access delay.
As for the \textit{resource scheduling}, prior efforts mainly operate on memory, which is the lowest level optimization closest to hardware, including memory allocation~\cite{moon2022nntrainer} and memory swapping~\cite{rhu2016vdnn,chen2018modnn}.
For example, MoDNN temporarily swaps intermediate activation from graphics processing unit (GPU) memory to host memory to solve the problem of insufficient training memory and thereby realizes the optimal compromise between memory footprint and training speed~\cite{chen2018modnn}.
\textbf{\textit{(iii) intra-device cross-level controller}} jointly control multiple techniques across the above levels to achieve the best tradeoff between multiple conflicting performance goals for dynamic AIoT context. 
For example, adaptive precision training in~\cite{huang2020adaptive} dynamically allocates model precision to balance training energy cost, memory usage, and accuracy.      

\begin{figure*}[t]
  \centering
  \includegraphics[width=.95\textwidth]{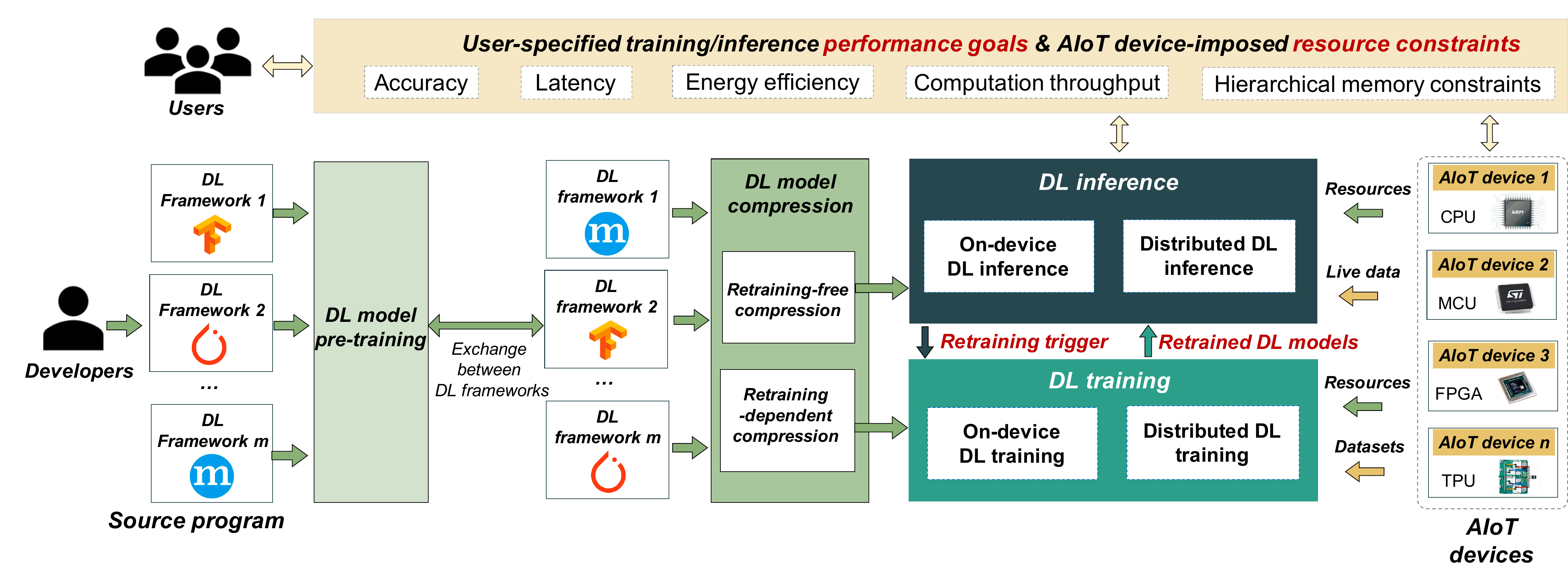}
  \caption{Workflow of the AIoT system software for resource-efficient DL training and inference tasks on AIoT devices.}
  \label{fig:workflow}
  \vspace{-2mm}
\end{figure*}
        
\textbf{\textit{Distributed DL Training}}. 
To coordinate data fusion and resource aggregation to optimize DL training efficacy and efficiency on distributed devices.
We introduce cross-level enabling techniques for distributed training in $\S$ \ref{subsec:train_m}. 
In particular,
\textbf{\textit{(i) resource-friendly algorithm level}}  
    optimizes the distributed training algorithm (\eg pipeline algorithm~\cite{huang2019gpipe} and federated learning~\cite{lim2020federated}) and changes the DL model/data partition strategy, which can significantly improve the distributed system's computing efficiency.
\textbf{\textit{(ii) model-adaptive system scheduling level}},
    like the on-device training, mainly optimize operator and memory access for training speedups and resource usage reduction, such as memory reallocation~\cite{cui2016geeps} and layer swapping~\cite{wahib2020scaling} \etc
    For example, Zico~\cite{lim2021zico} monitors the memory usage of each DL training task and reclaims the memory that is no longer needed, making it globally sharable in the  system.
\textbf{\textit{(iii) inter-device cross-level controller}} considers all cross-level factors like the on-device DL training scheme.
In addition, it considers the communication conditions, such as time-varying network throughput, to control the distributed system loop jointly.
And it schedules the cross-level techniques separately and evaluates them globally.
% \textbf{\textit{(iv) inter-device controller}} 
Specifically, it selects available devices to minimize overall training delay, ensure accuracy, and improve resource efficiency~\cite{samikwa2022ares}.
To address issues such as asynchronous processing and waiting times resulting from resource-heterogeneous AIoT devices, the inter-device controller should monitor their heterogeneity in advance and select suitable devices for collaboration.

\subsection{Workflow Overview}

Figure \ref{fig:workflow} showcases the resource-efficient AIoT system workflow and relationships of the system blocks for DL training and inference tasks.
The system takes user-provided inputs, such as DL source programs, and aims to achieve user-specified performance goals while satisfying the resource constraints imposed by the AIoT devices.
Specifically, the \textit{user/developer} specifies the \textit{DL source program} using various DL frameworks (\eg TensorFlow, Pytorch, Mindspore, \etc) and conducts model training using accumulated AIoT dataset.
After DL pre-training, the model structure files are distributed to heterogeneous AIoT devices to analyze live data and provide intelligent services based on the pre-trained model. 
To be compatible with different DL frameworks in various AIoT devices, we can convert them to a unified format and exchange them (\eg using ONNX~\cite{onnx}).
In the \textit{DL inference phase}, we can select and combine the most suitable DNN compression techniques from the \textit{model compression algorithm pools}  to reduce the model complexity and resource demands.
Most model compression techniques need to return to the training stage for several rounds of retraining to fine-tune model parameters to ensure accuracy~\cite{gudovskiy2018dnn}.
And some recent efforts also realize the retraining-free model compression techniques, \eg through ensemble training of super-nets~\cite{wu2022compiler}.

The resource-efficient \textit{DL inference} includes on-device ($\S$ \ref{subsec:inference_1}) and distributed ($\S$ \ref{subsec:inference_m}) schemes. 
Particularly, it includes the runtime compilation front-end optimization (target platform-independent), the compilation back-end, and hardware instruction optimization (target platform-dependent).
Both the compiler front-end and the back-end belong to the model-adaptive system scheduling level mentioned in $\S$ \ref{subsec:sys_layer}. 
They convert the DL models designed at the algorithm level into intermediate representations, \ie computation graph, for further optimization. 
Compiler front-end optimization focuses on platform-independent optimization, such as constant folding~\cite{tai1979constant}~\cite{glesner2004classifying} and common subexpression optimization~\cite{cocke1970global}. 
And compiler back-end optimization focuses on platform-dependent optimization, such as operator fusion~\cite{elgamal2017spoof}~\cite{jia2019optimizing}, and memory allocation~\cite{wang2022melon}~\cite{moon2022nntrainer}.

When the accuracy of the DL model drops below a certain threshold due to changes in application scenarios, data, performance requirements, \etc, the \textit{DL training} block is triggered to retrain and update the model.
Whether the DL model is re-trained locally on a single AIoT device or on multiple locally-connected yet resource-constrained edge devices depends on how well the supply of device resources (\eg memory) matches the computing requirements for training and the desired training speed.
Once the inter-device controller selects the training scheme, we can proceed to the on-device DL training optimization block (see $\S$ \ref{subsec:train_1}) or the distributed DL training block (see $\S$ \ref{subsec:train_m}).
As a separate note, in resource-efficient AIoT system, the compiler optimization is cross-device~\cite{abadi2016tensorflow}~\cite{chen2018tvm} on heterogeneous and distributed AIoT devices, \eg from GPU-based edge servers to MCUs, to support intelligent inference/training tasks. 
The inter-/intra-device controllers adaptively control the cross-level optimization strategies according to context information.

\subsection{AIoT Performance Metrics}
\label{subsec:metric}

The resource-efficient AIoT system in both DL training and inference tasks needs to optimize the user-specified performance goals (\ie accuracy, latency, energy efficiency, computation throughput) and satisfy device-imposed resource constraints (\ie hierarchical memory, processor, battery).
We listed the related metrics below:

\textit{1) Accuracy}. 
    The DL model should be accurate enough to guarantee a high-quality AIoT task. 
    The model weights at different scales are well-trained to represent the generic information of recognition objects~\cite{fenske2006top}.

\textit{2) Memory}.
    The parameters and activations of DL models should be appropriately sized to fit into the memory units of AIoT devices.
    We can directly calculate the memory needed to run a DL model using the total number of bits associated with weights and activations.
    And we should optimize how the operations access the data fetched from different memory hierarchies (\eg \rev{dynamic random access memory} (DRAM), \rev{static random access memory} (SRAM)) and how the computation is executed for latency or energy efficiency optimization~\cite{liu2021adaspring}.
        \begin{itemize}
            \item \textit{Memory budgets}. 
            The memory budget is a tough constraint; it decides whether a specific AIoT device can perform the
            DL inference or training tasks.
            % For example, 
            \item \textit{SRAM utilization}. 
            Improving SRAM utilization and reducing DRAM transmission times can improve computing efficiency~\cite{lin2001reducing}.
            \item \textit{Cache/register hit rate} measures the percentage of times that the system is able to retrieve data from the cache in the central processing unit (CPU)/GPU~\cite{raspberry} and register in MCU~\cite{mcu}, instead of accessing it from the main memory. 
            Higher cache/register hit rates mean that the system can access data more efficiently.
            We can use it to estimate the system efficiency in the presence of time-varying memory resources via run-time measurement~\cite{liu2023adaenlight}.
            % Reducing the model size to fit the on-chip memory, \eg cache in CPU/GPU~\cite{raspberry} and register in MCU~\cite{mcu}, can avoid expensive off-chip access.
        \end{itemize}
    
\textit{3) Computational cost}.
    It affects the AIoT system's latency and energy efficiency.
    We can model the computational cost of DL inference and training tasks as the total number of DL models' multiply-accumulate (MAC) operations.
    
\textit{4) Latency}.
    The complexity of the DL model for inference and training should be controllable to meet diverse user demands on latency.
    The latency of DL inference/training tasks executed in AIoT devices strongly depends on the given device's  architecture and memory hierarchy~\cite{lai2021opportunity}.
    And we can refer to the latency predictor, such as nn-Meter~\cite{zhang2021nn}, ~\cite{venieris2017latency}, for platform-aware latency estimation.

\begin{figure*}[t]
  \centering
  \includegraphics[width=.88\textwidth]{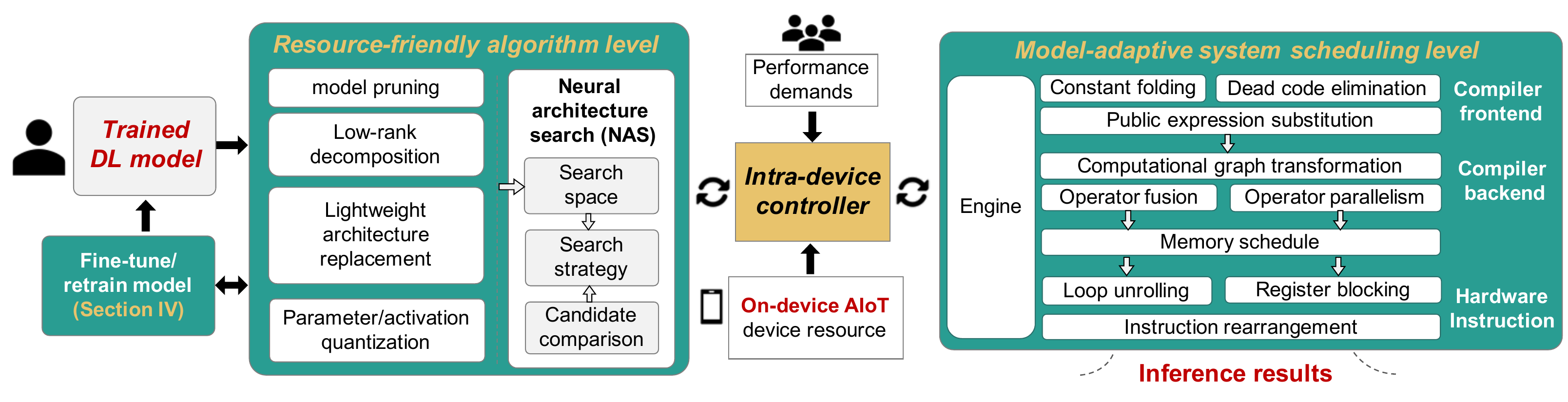}
  \caption{System loop of the algorithm, system scheduling, and intra-device controller for on-device DL inference.}
  \label{fig:inference_single}
  % \vspace{-3mm}
\end{figure*}

\textit{5) Energy efficiency}.
    It is an important metric for battery-powered AIoT devices. 
    Researchers usually formulate it offline and estimate it online since it is prohibitive to directly connect to energy monitors for measurement when the device is in service. 
    The estimation methods include two types:
    \begin{itemize}
        \item \textit{Estimation function}. 
        The energy consumption of DL inference includes computation cost and memory access cost. 
        The former can be formulated as the total energy cost of the total MACs, \ie, $E_c =\epsilon_1 C$, where $\epsilon_1$ and $C$ denote the energy cost per MAC operation and the total number of MACs, respectively.
        The latter depends on the storage scheme when executing DL models on the user-given embedded device.
        And the energy cost of fetching intermediate activations, \ie memory access, dominates in the DL inference phase.

        \item \textit{Arithmetic intensity}. The hardware-aware metric, \eg arithmetic intensity, can proxy the degree of reuse of parameters and activations and the energy cost required for processing inputs~\cite{jha2020modeling, jha2019ramifications}.
    \end{itemize}

% \TODO{add a table and cover all GPU/TPU devices} 

As a separate note, the widely used parameters number, MAC amount,
or speedup ratio is not a good approximation for hardware efficiency (\eg energy cost, latency), which heavily depends on the underlying memory movement and bandwidth bound~\cite{liu2021adaspring}.
For example, Jha \etal~\cite{jha2019ramifications} reported that although SqueezeNet \cite{iandola2016squeezenet} has $51.8\times{}$ fewer parameters than AlexNet, while it consumes $33\%$ more energy due to its larger amount of activations and data movement.
We identify that merely cutting down the parameter size may lead to an increase in activation size, which, in turn, increases the memory footprint, latency, and energy consumption~\cite{jha2020modeling}. 

The dynamic deployment context in continuously running AIoT applications often results in high levels of unpredictability and variability in terms of performance demands on the above metrics. 
This context includes factors such as agnostic AIoT sensing data, time-varying resource availability, dynamic join and exit of AIoT devices, and fluctuating inference/training request frequency triggered by real-world requirements.
Thus, the resource-efficient AIoT system should continuously evolve to balance these metrics in a context-adaptive manner~\cite{liu2021adaspring, wang2021context}.

\section{Cross-level Optimization for DL inference}
\label{sec:inference}

This section introduces existing efforts associated with the proposed challenges in cross-level AIoT systems and highlights how they have addressed some of them.

\subsection{On-device DL Inference}
\label{subsec:inference_1}

An increasing trend in the field of AIoT is integrating DL-powered intelligence into local devices, allowing for analytics to be performed where the  data is sensed. 
This approach offers advantages such as low transmission costs, network condition independence, and privacy preservation, as highlighted in~\cite{szegedy2017inception,zagoruyko2016wide}.
However, deploying DL models on resource-scarce local devices remains a challenging problem, as discussed in $\S$ \ref{sebsec:texonomy}. 
To address this issue, cross-level optimization spans multiple levels (\eg model, computation graph, memory scheduling,  hardware instruction, \etc) is essential. 
Additionally, context-aware controllers can further automate the on-device DL inference process.
Figure \ref{fig:inference_single} illustrates the on-device DL inference optimization pipeline.

% \begin{figure}[t]
% 	\centering
% 	\begin{minipage}{0.49\linewidth}
% 		\centering
% 		\includegraphics[width=0.9\linewidth]{synaptic_pruning .pdf}
% 		\caption{Synaptic pruning}
% 		\label{fig:syn_pruning}
% 	\end{minipage}
% 	%\qquad
% 	\begin{minipage}{0.49\linewidth}
% 		\centering
% 		\includegraphics[width=0.9\linewidth]{neuron_pruning.pdf}
% 		\caption{Neuron pruning}
% 		\label{fig:neu_pruning}
% 	\end{minipage}
% \end{figure}

\subsubsection{Resource-friendly algorithm level}
\label{subsub_infer_alg}
Algorithm-level optimization for on-device DL inference is extensively researched to minimize computation and memory requirements while preserving accuracy~\cite{szegedy2017inception,zagoruyko2016wide}~\cite{cheng2017survey,cheng2018recent,deng2019deep,choudhary2020comprehensive, he2018multi,gao2021pruning}.
We briefly discuss some representative ones below.
% We briefly introduce some representative compression techniques as follows:

\textbf{a. Pruning}.
It reduces the model computation cost by removing redundant parameters~\cite{srinivas2015data}, channels~\cite{he2017channel}, or connections~\cite{han2015deep}.
According to the grain size of pruning, it can be divided into synaptic pruning, neuron pruning, \etc
Synaptic pruning cuts down unimportant connections between neurons,
while neuron pruning removes neuron nodes directly.
For example, Hu \etal~\cite{hu2016network} iteratively prunes neurons and uses the average percentage of zeros to find the unimportant activation.
\cite{luo2017thinet, he2017channel} determine which channels must be pruned by minimizing the feature reconstruction error using greedy and Least Absolute Shrinkage and Selection Operator(\ie LASSO) regression optimizers, respectively.

\begin{figure}[t]
    \centering
    \includegraphics[width=0.5\textwidth,scale=1.00]{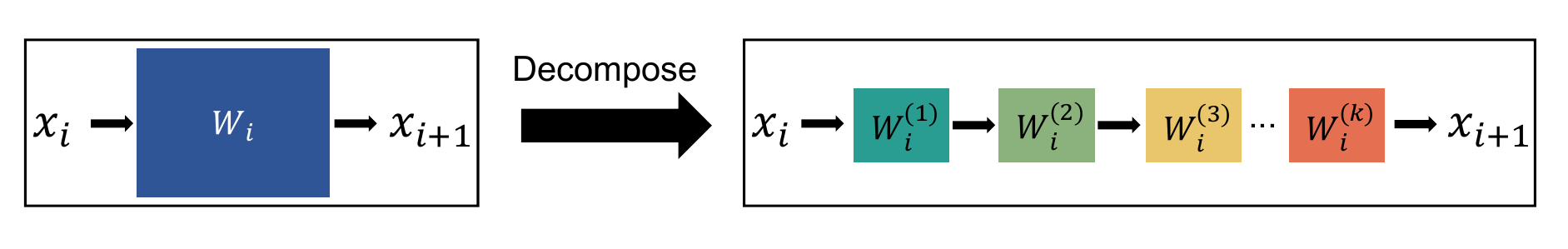}
    \caption{Illustration of low-rank decomposition technique.}
    \label{fig:low_rank}
    \vspace{-2mm}
\end{figure}

\textbf{b. Low-rank decomposition}.
Since model weight vectors are mostly distributed in low-rank subspaces, we can only use a few basis vectors to represent the convolution kernel for memory savings by combining dimensions or applying low-rank constraints~\cite{jaderberg2014speeding,tai2015convolutional}.
\rev{As shown in Figure \ref{fig:low_rank},$W_i$ is a large low-rank matrix which is decomposed into several small matrices such as $W_i^{(1)}, W_i^{(2)}, W_i^{(3)}, ..., W_i^{(k)}$.}

Pavel Kaloshin~\cite{kaloshin2020convolutional} decomposed the tensor as a sum of low-rank and sparse components, approximating the convolution weights. 
The convolutional (conv) layers take the most execution time, and fully connected (fc) layers dominate storage costs.
%
%Because of such differences, existing works always divide the acceleration of conv layers and memory reduction of fc layers into separate tasks.
%
%For example, the acceleration of conv layers uses methods such as structured pruning~\cite{wen2016learning,li2016pruning}, tensor decomposition~\cite{lebedev2014speeding,denton2014exploiting}. Methods for compressing fc layers include vector quantization~\cite{gong2014compressing}, matrix decomposition~\cite{zhang2015accelerating}, \etc Besides, 
Lin \etal~\cite{lin2018holistic} jointly accelerates conv layers and compresses fc layers by utilizing low-rank decomposition to eliminate redundancy between conv kernels and fc matrices.

\textbf{c. Lightweight architecture}.
Replacing large model blocks with lightweight architectures can adapt to resource-constrained AIoT devices.
There are generally two categories, \ie block replacement~\cite{mehta2019espnetv2} or neural architecture search (NAS)~\cite{tan2019mnasnet}. 
For example, Iandola \etal~\cite{iandola2016squeezenet} replaced conv layers by Fire block, composing of a 1$\times$1 conv layer and a conv layer with mixed 1$\times$1 and 3$\times$3 filters.
Lin \etal~\cite{lin2013network} replaced conv by a micro multi-layer perceptron embedded with multiple small kernel conv (Mlpconv). 
To realize efficient DL inference on \rev{MicroController Unit} (MCUs), with 2-3 orders of magnitude smaller memory than mobile phones, Lin \etal proposed MCUNet~\cite{lin2020mcunet}, Edgar \etal  built $\mu$NAS~\cite{liberis2021munas} to specialize model architectures for MCUs automatically.
% generating the efficient architecture by TinyNAS.
% Edgar \etal  built $\mu$NAS~\cite{liberis2021munas} to specialize model architectures for MCUs.
% with low memory usage and latency while maintaining relatively high accuracy. 

\textbf{d. Parameter/activation quantization}.
Quantization~\cite{gong2014compressing,wu2016quantized} refers to representing 32-bit floating-point model parameters with relatively low widths, including weight~\cite{courbariaux2015binaryconnect}, activation value~\cite{li2017performance}. 
The model parameters can be unified with layer-wise low-bit-widths (\eg 16-bit, 8-bit, 2-bit, 1-bit, \etc) to reduce memory usage significantly, speed up computing, and reduce power cost.
To adapt to the varying memory and computational limitations, Manuele \etal~\cite{rusci2020memory} modeled the DL inference graph by pure integer operation using mixed low-bit-width quantization.
% which is characterized by 8-bit, 4-bit, or 2-bit uniform quantization.
% a new end-to-end approach for deploying low-error model on microcontrollers.
% %
% To adapt to the memory and computational limitations of diverse devices, the inference graph is modeled by 
% %
% Given the accuracy and memory constraints, their key goal is determining the minimum bit precision for each activation and weight tensor.
%
% According to the unique memory and computational characteristics of MCUs, 
% Manuele \etal~\cite{rusci2020leveraging} proposed HAQ, an automatic hybrid quantization framework. 
Under specific RAM and FLASH memory constraints in MCUs, Manuele \etal~\cite{rusci2020leveraging} use reinforcement learning to pick the best uniform quantization bitwidths for model weights and activations.

\textbf{Discussion}. 
% Existing research strives to optimize the resource-performance trade-off of DL models by compressing the model without notably compromising its inference accuracy. 
Algorithm-level optimization techniques have shown great promise in reducing computation and memory requirements with slightly decreased accuracy.
However, the degree of accuracy decrease is bounded by the runtime resource availability. 
% Proposing new algorithms may lead to slight performance improvement upon existing techniques.
% Their suitable choice needs further exploration depending on the specific application and hardware.
While proposing new algorithms may lead to a slight improvement in performance-resource tradeoff,\eg latency, over existing techniques, their suitability to AIoT also depends on the system scheduling on the given hardware.

\begin{figure}[t]
  \centering
    \includegraphics[width=0.5\textwidth]{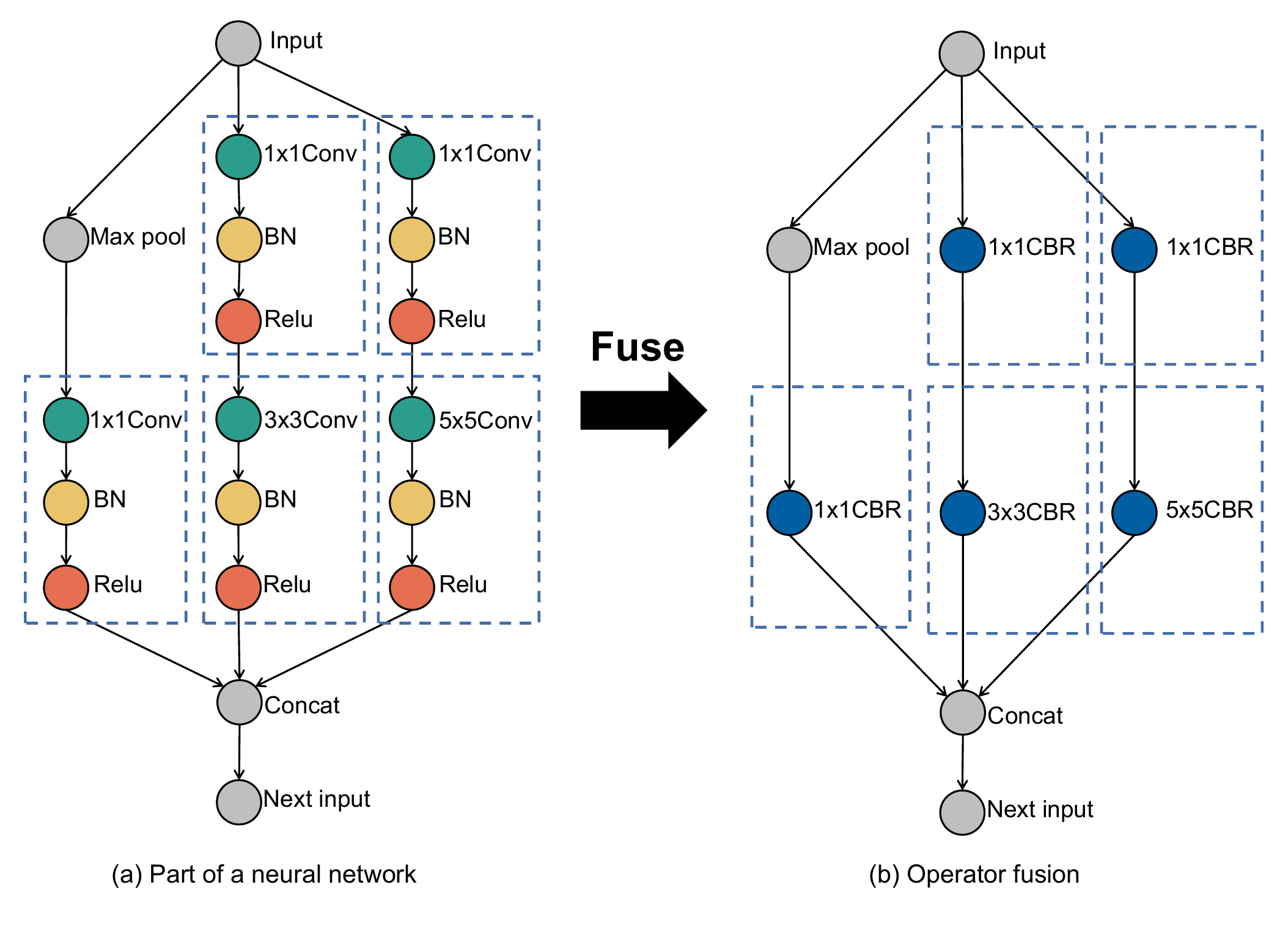}
\caption{Illustration of the operator fusion technique. (a) partial computation graph of GoogleNet, (b) longitudinally fuse \textit{convolution, batchnorm} and \textit{relu} into \textit{CBR} operator.}
\label{fig:fuse}
%\vspace{-3mm}
\end{figure}

\subsubsection{Model-adaptive system scheduling level}
\label{subsub_infer_sys}
Joint optimization of system scheduling and upper-level algorithms has the potential to overcome performance bottlenecks. 
This level aims to utilize hardware resources to their fullest capacity without compromising model accuracy.
The system scheduling for DL inference mainly includes the compiler front-end (\ie device-independent) and compiler back-end (\ie device-dependent).
Compiler front-end optimization aims to eliminate redundancy and simplify the computation of the intermediate representation (\ie computation graph) during compilation.
It is device-independent and contains constant folding~\cite{chow1984portable}, dead code elimination~\cite{chow1984portable}, \etc
And compiler back-end aims to use hardware resources to execute DL inference tasks fully.
It mainly focuses on computation graph-level optimization (\eg operator fusion~\cite{boehm2018optimizing}, computation graph substitution~\cite{jia2019optimizing,fang2020optimizing}), operator parallelism~\cite{paszke2019pytorch,ding2021ios}, memory-level optimization (\eg memory allocation~\cite{wei2019overcoming,symons2021loma}, memory swapping~\cite{miao2021enabling,ji2022task}), and hardware instruction optimization (\eg loop unrolling~\cite{murthy2010optimal}, register blocking~\cite{im2001optimizing}).
We introduce some representative enabling techniques below.

\textbf{a. Operator fusion.}
Despite the massive size of model parameters, the intermediate feature maps extracted from the input data can quickly become too large and consume significant amounts of memory.
This is especially problematic because these intermediate feature maps are often used as inputs for multiple operators in the computation graph, leading to increased memory access and processing delays.
One solution to this problem is to fuse adjacent operators in the computation graph into a new operator according to certain rules. 
Figure \ref{fig:fuse} illustrates an example in GoogleNet.
The original operators include convolution, batchnorm, relu, max pool, and concat, as shown in Figure \ref{fig:fuse}a.
Then \textit{convolution, batchnorm} and \textit{relu} operators are fused into the \textit{CBR} operator.
They have the same computation results. 
However, the number of model layers in Figure \ref{fig:fuse}b is decreased, the data channel is shortened, and the memory access frequency of the intermediate feature maps is reduced, so the inference speed is improved.
Han \etal~\cite{vanholder2016efficient} expand the CBR operator fusion to TensorRT.

Existing DL frameworks such as TensorFlow Lite~\cite{tensorflowlite}, TVM~\cite{chen2018tvm}, MNN~\cite{jiang2020mnn}, and Pytorch-Mobile~\cite{pytorchmobile} have provided \rev{application programming interfaces} (APIs) for operator fusion.
However, most of them only provide fixed operator fusion patterns, and the operator fusion types are still insufficient to support diverse DL operators and connections.
Niu \etal~\cite{niu2021dnnfusion} proposed a universal fusion framework called DNNFusion. 
It divides operators into different types according to their input and output forms, develops operator fusion plans after comparing the performance of diverse fused operators, and conducts extra optimizations such as reducing redundancy during the fusion code generation.
This approach based on general operator type greatly expands the operator fusion opportunity.
Experimental results show that the operator fusion space is expanded by 8.8$\times$, and the inference speed exceeds the advanced frameworks (\eg MNN~\cite{jiang2020mnn}, TVM~\cite{chen2018tvm}, Pytorch~\cite{paszke2019pytorch}, TensorFlow Lite~\cite{tensorflowlite}) by up to 9.3$\times$.
%
% Another problem is that the operator fusion space has cannot always guarantee optimal memory efficiency.
Another problem is that the operator fusion space and memory efficiency lack direct mapping, making it difficult to access optimal memory.
Cai \etal~\cite{cai2022optimus} proposed an operator fusion framework, Optimus, based on directed acyclic graphs.
It can be applied to several DL models running on accelerators. 
Experiments show that Optimus reduces inference memory access overhead by 17-75$\%$ and improves efficiency by 1.86-3.66$\times$.
Furthermore, automated operator fusion, instead of manually crafted fusion, can greatly enhance the performance of complicated or previously unseen operator chains.
% replacing handcraft one can significantly improve the complicated or previously unseen chain operational performance.

\begin{figure}[t]
  \centering
    \includegraphics[width=0.5\textwidth]{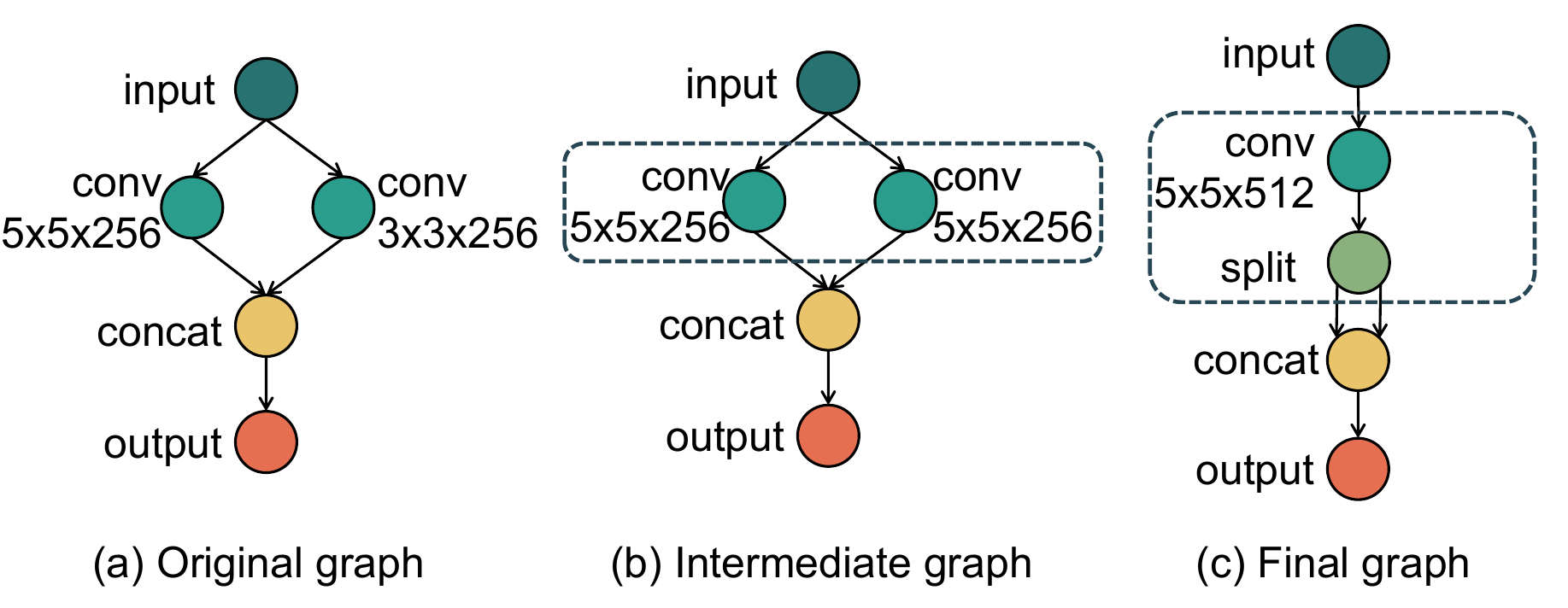}
\caption{Illustration of the computation graph substitution technique. (a) original computation graph includes \textit{convolution} operators, (b)  expanding smaller conv kernels, and (c) merging two conv operators into one.}
\label{fig:substitution}
%\vspace{-3mm}
\end{figure}

\textbf{b. Computation graph substitution.}
Computation graph substitution techniques replace the subgraph with another functionally equivalent subgraph to reduce the amount of computation and delay.
For better understanding, Figure \ref{fig:substitution} shows an example.
In the original graph (see Figure \ref{fig:substitution}a), there are two \textit{conv} operators that have 256 kernels with 3$\times$3 size and 256 kernels with 5$\times$5 size, respectively.
We can first expand all 3$\times$3 kernels to 5$\times$5 (see Figure \ref{fig:substitution}b), merge two 5$\times$256 conv operators into one 5$\times$512 conv operator, and then separate them using \textit{split} operator before executing the \textit{concat} operator (see Fig. \ref{fig:substitution}c).
Via computation graph substitution, we remove the computational-intensive conv operator. 
And the computational cost of the \textit{split} operator is almost negligible.
% compared to the removed computational-intensive conv operator. 
% So that it can greatly reduce the operator number and execution delay.

Existing DL frameworks (\eg PyTorch~\cite{paszke2019pytorch}, TensorFlow~\cite{abadi2016tensorflow}) substitute computation graphs using greedy rules or manual methods, which, however, cannot guarantee the selection or combination bring rigorous improvements.
Jia \etal~\cite{jia2019optimizing} proposed an optimizer for graph substitution.
They automatically use a cost-based search algorithm on the graph substitution space to find the optimal solution.
Jia \etal~\cite{jia2019taso} proposes an optimizer (\ie TASO) to substitute the computation graph automatically.
TASO generates several candidates for a given list of operators and picks the most suitable substitutions.
Experiments show that TASO outperforms existing DL frameworks by 2.8$\times$ and significantly reduces human labor.
Fang \etal~\cite{fang2020optimizing} formally defined the computation graph substitution problem (\ie OCGGS) and narrowed the search space to sample the best solution.
%
% However, extending with the continuously emerging new operators is prohibitive in existing DL frameworks.
% the computation graph substitution techniques in existing DL frameworks overlook many optimization opportunities. 
% For example, extending with the continuously emerging new operators is prohibitive.
%

% This work reduced computation cost, search time, and memory usage.

\begin{figure}[t]
  \centering
  \includegraphics[width=0.5\textwidth]{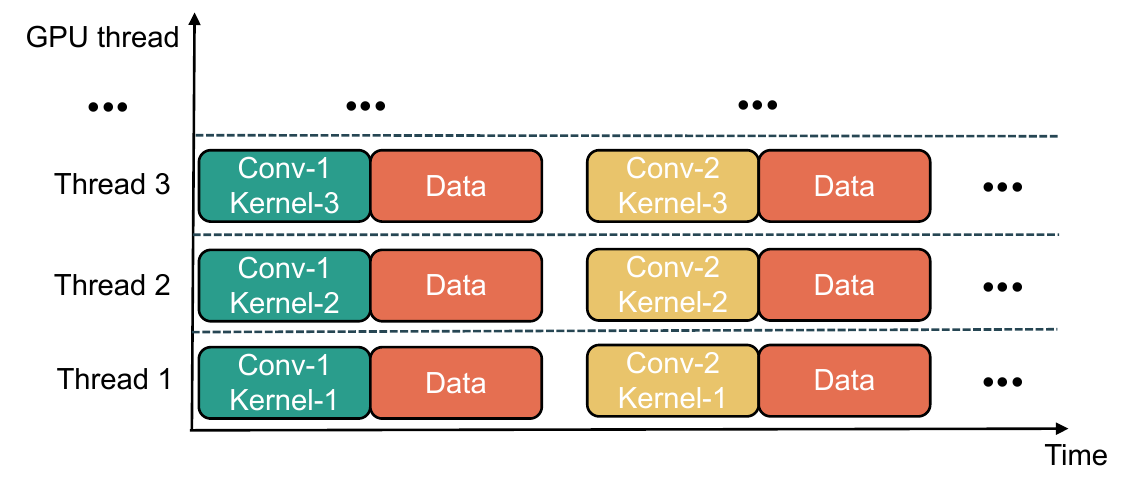}
  \caption{Illustration of 
  intra-operator parallelism technique. The conv operator kernels and data are divided into different groups and deployed on different threads to execute in parallel.}
  \label{fig:intra_para}
  % \vspace{-3mm}
\end{figure}

\begin{figure}[t]
  \centering
  \includegraphics[width=0.45\textwidth]{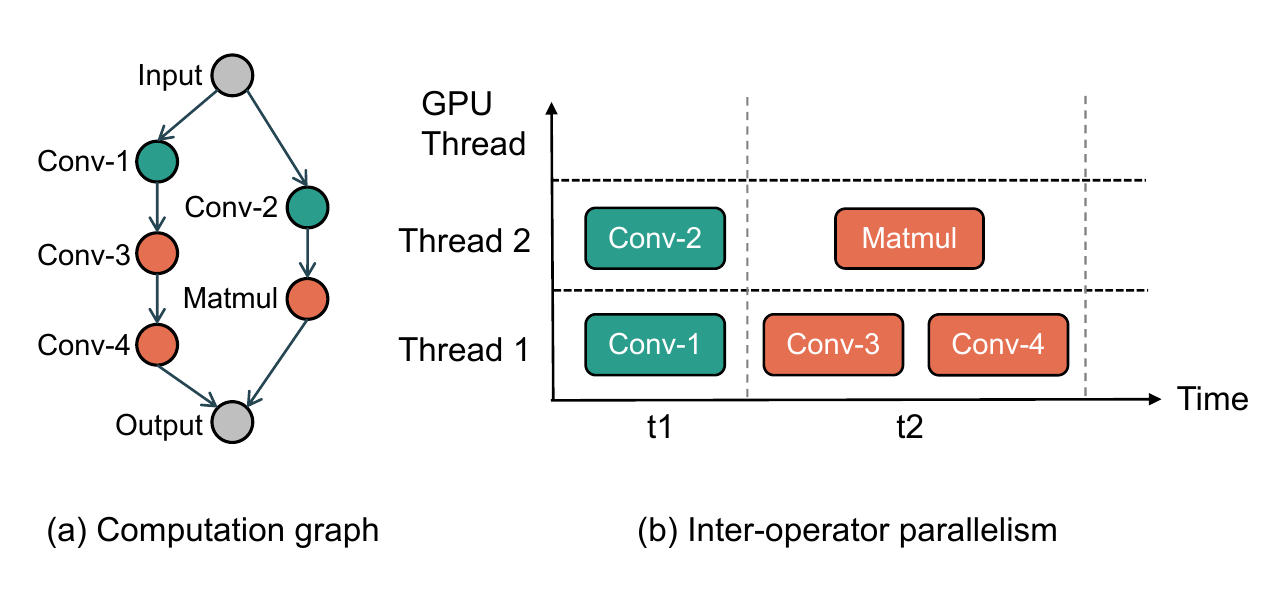}
  \caption{Illustration of 
  inter-operator parallelism technique. (a) the original computation graph, (b) two conv operators execute in parallel on different threads during the $t_1$ time period. During the $t_2$ time period, the \textit{matmul} operator executes on thread $2$, while the serial execution of two \textit{conv} operators execute on thread $1$.}
  \label{fig:inter_para}
  % \vspace{-3mm}
\end{figure}

\textbf{c. Operator parallelism.}
There are two kinds of operator parallelism techniques, \ie intra-operator and inter-operator parallelism.
Mainstream AIoT platforms always equip with multi-core CPUs and multi-core GPUs. 
Considering the increasing speed gap of diverse processors (\eg CPU and GPU), separate methods are designed for the CPU and GPU to overcome the memory access bottleneck.
For CPU devices, the cache hides delays in accessing memory to reduce the pressure on memory bandwidth.
GPUs do not use or only use relatively small caches, mainly through the parallelism of threads, to hide the memory access delay.
When some threads are stuck due to memory access, another part of the threads will continue to execute and will not let processing units idle.

\textit{Intra-operator parallelism}. 
Existing DL frameworks (\eg TensorFlow~\cite{abadi2016tensorflow}, PyTorch~\cite{paszke2019pytorch}) can support intra-operator parallelism, \ie parallelizing arithmetic operations (\eg convolution) within a single operator.
In the convolution operator, kernels that will be executed on the same thread are divided into one group, and then the corresponding data is also grouped. 
Then, different groups of conv kernels and data are deployed on diverse threads for parallel execution, as shown in Figure \ref{fig:intra_para}.
However, as high-performance hardware evolves, intra-operator parallelism is no longer efficient enough.

\textit{Inter-operator parallelism}.
Inter-operator parallelism allows multiple operators to execute on different threads in parallel, as shown in Figure \ref{fig:inter_para}.
Ding \etal~\cite{ding2021ios} proposed an inter-operator scheduler(\ie IOS). 
IOS can search highly optimized parallel schemes using a dynamic programming algorithm. 
And this approach can be generalized to existing DL frameworks. 
Experimental results show that IOS increases the inference speed by $1.1 \times \sim 1.5 \times$.

\begin{table*}[]
\caption{Summary of model-adaptive system scheduling-level enabling techniques for on-device DL inference.}
\renewcommand{\arraystretch}{1.3}
\resizebox{\linewidth}{!}{
\scriptsize
\begin{threeparttable}
\begin{tabular}{|ccc|c|c|c|c|c|c|}
\hline
\multicolumn{3}{|c|}{\textbf{Category}}                                                                                                                                                                                                                                                                                                 & \textbf{Technique highlight for improving}                                                                                                                                           & \textbf{Device}                                                 & \textbf{Year} & \textbf{Ref.}                         & \begin{tabular}[c]{@{}c@{}}\textbf{Compiler}\\ \textbf{frontend}\end{tabular} & \begin{tabular}[c]{@{}c@{}}\textbf{Compiler}\\ \textbf{backend}\end{tabular} \\ \hline
\multicolumn{1}{|c|}{\multirow{36}{*}{\begin{tabular}[c]{@{}c@{}}\textbf{Model-adaptive}\\\textbf{system}\\ \textbf{scheduling}\\ \textbf{level}\end{tabular}}} & \multicolumn{1}{c|}{\multirow{18}{*}{\begin{tabular}[c]{@{}c@{}}\textbf{Computation}\\ \textbf{graph} \textbf{level}\end{tabular}}}             & \multirow{5}{*}{\begin{tabular}[c]{@{}c@{}}\textbf{Operator}\\ \textbf{fusion}\end{tabular}}                  & \begin{tabular}[c]{@{}c@{}}Circular fusion, operator classification, reduce redundancy,\\ reduce computation latency\end{tabular}                                           & \begin{tabular}[c]{@{}c@{}}Mobile\\ phone\end{tabular} & 2021 & ~\cite{niu2021dnnfusion}     & \checkmark                                                  &                                                            \\ \cline{4-9} 
\multicolumn{1}{|c|}{}                                                                                     & \multicolumn{1}{c|}{}                                                                                               &                                                                                             & \begin{tabular}[c]{@{}c@{}}Directed acyclic graphs, accurate memory cost model,\\ reduce computation latency\end{tabular}                                                   & \begin{tabular}[c]{@{}c@{}}Cloud\\ server\end{tabular} & 2022 & ~\cite{cai2022optimus}       &                                                             & \checkmark                                                 \\ \cline{4-9} 
\multicolumn{1}{|c|}{}                                                                                     & \multicolumn{1}{c|}{}                                                                                               &                                                                                             & \begin{tabular}[c]{@{}c@{}}Candidate exploration, selection of fusion plans, \\ code generation of local and distributed operations,\\ reduce computation latency\end{tabular} & MPU                                                    & 2018 & ~\cite{boehm2018optimizing}  & \checkmark                                                  &                                                            \\ \cline{3-9} 
\multicolumn{1}{|c|}{}                                                                                     & \multicolumn{1}{c|}{}                                                                                               & \multirow{5}{*}{\begin{tabular}[c]{@{}c@{}}\textbf{Computation}\\ \textbf{graph}\\ \textbf{substitution}\end{tabular}} & \begin{tabular}[c]{@{}c@{}}Relaxed graph substitution, backtracking search algorithm, flow-\\ based graph split algorithm, reduce computation latency\end{tabular}          & \begin{tabular}[c]{@{}c@{}}Cloud\\ server\end{tabular} & 2019 & ~\cite{jia2019optimizing}    &                                                             & \checkmark                                                 \\ \cline{4-9} 
\multicolumn{1}{|c|}{}                                                                                     & \multicolumn{1}{c|}{}                                                                                               &                                                                                             & \begin{tabular}[c]{@{}c@{}}The first DL computation graph optimizer, \\ automated theorem prover, cost-based backtracking search, \\ reduce computation latency\end{tabular}   & \begin{tabular}[c]{@{}c@{}}Cloud\\ server\end{tabular} & 2019 & ~\cite{jia2019taso}          &                                                             & \checkmark                                                 \\ \cline{4-9} 
\multicolumn{1}{|c|}{}                                                                                     & \multicolumn{1}{c|}{}                                                                                               &                                                                                             & \begin{tabular}[c]{@{}c@{}}Pruning-based algorithm, sampling heuristic,\\ reduce computation latency and memory usage\end{tabular}                                          & \begin{tabular}[c]{@{}c@{}}Cloud\\ server\end{tabular} & 2020 & ~\cite{fang2020optimizing}   & \checkmark                                                  &                                                            \\ \cline{3-9} 
\multicolumn{1}{|c|}{}                                                                                     & \multicolumn{1}{c|}{}                                                                                               & \multirow{5}{*}{\begin{tabular}[c]{@{}c@{}}\textbf{Operator}\\ \textbf{parallelism}\end{tabular}}             & Intra-operator parallelism, reduce computation latency                                                                                                                      & \begin{tabular}[c]{@{}c@{}}Cloud\\ server\end{tabular} & 2016 & ~\cite{abadi2016tensorflow}  &                                                             & \checkmark                                                 \\ \cline{4-9} 
\multicolumn{1}{|c|}{}                                                                                     & \multicolumn{1}{c|}{}                                                                                               &                                                                                             & Intra-operator parallelism, reduce computation latency                                                                                                                      & \begin{tabular}[c]{@{}c@{}}Cloud\\ server\end{tabular} & 2019 & ~\cite{paszke2019pytorch}    &                                                             & \checkmark                                                 \\ \cline{4-9} 
\multicolumn{1}{|c|}{}                                                                                     & \multicolumn{1}{c|}{}                                                                                               &                                                                                             & \begin{tabular}[c]{@{}c@{}}Inter-operator parallelism, inter-operator scheduler, \\ novel dynamic programming algorithm, \\ reduce computation latency\end{tabular}             & \begin{tabular}[c]{@{}c@{}}Cloud\\ server\end{tabular} & 2021 & ~\cite{ding2021ios}          &                                                             & \checkmark                                                 \\ \cline{2-9} 
\multicolumn{1}{|c|}{}                                                                                     & \multicolumn{1}{c|}{\multirow{11}{*}{\begin{tabular}[c]{@{}c@{}}\textbf{Resource}\\ \textbf{scheduling}\\ \textbf{level}\end{tabular}}}         & \multirow{5}{*}{\begin{tabular}[c]{@{}c@{}}\textbf{Memory}\\ \textbf{allocation}\end{tabular}}                & \begin{tabular}[c]{@{}c@{}}Novel profile-guided memory optimization, heuristic algorithm,\\ reduce computation latency and memory usage\end{tabular}                        & \begin{tabular}[c]{@{}c@{}}Cloud\\ server\end{tabular} & 2018 & ~\cite{sekiyama2018profile}  &                                                             & \checkmark                                                 \\ \cline{4-9} 
\multicolumn{1}{|c|}{}                                                                                     & \multicolumn{1}{c|}{}                                                                                               &                                                                                             & \begin{tabular}[c]{@{}c@{}}Memory allocation algorithm, memory bound layers,\\ reduce computation latency and memory usage\end{tabular}                                     & FPGA                                                   & 2019 & ~\cite{wei2019overcoming}    &                                                             & \checkmark                                                 \\ \cline{4-9} 
\multicolumn{1}{|c|}{}                                                                                     & \multicolumn{1}{c|}{}                                                                                               &                                                                                             & \begin{tabular}[c]{@{}c@{}}Loop-orderbased memory allocation, fast auto-scheduling\\ methodology, reduce computation latency and memory usage\end{tabular}                  & \begin{tabular}[c]{@{}c@{}}Cloud\\ server\end{tabular} & 2021 & ~\cite{symons2021loma}       &                                                             & \checkmark                                                 \\ \cline{3-9} 
\multicolumn{1}{|c|}{}                                                                                     & \multicolumn{1}{c|}{}                                                                                               & \multirow{5}{*}{\begin{tabular}[c]{@{}c@{}}\textbf{Memory}\\ \textbf{swapping}\end{tabular}}                  & \begin{tabular}[c]{@{}c@{}}The first memory swapping on MCU, swap data block between\\ SRAM and FLASH or SD card, reduce memory usage\end{tabular}                          & MCU                                                    & 2021 & ~\cite{miao2021enabling}     &                                                             & \checkmark                                                 \\ \cline{4-9} 
\multicolumn{1}{|c|}{}                                                                                     & \multicolumn{1}{c|}{}                                                                                               &                                                                                             & \begin{tabular}[c]{@{}c@{}}Joint optimization along 3 dimensions, custom-designed genetic\\ algorithm, reduce computation latency and memory usage\end{tabular}             & \begin{tabular}[c]{@{}c@{}}Cloud\\ server\end{tabular} & 2020 & ~\cite{huang2020swapadvisor} &                                                             & \checkmark                                                 \\ \cline{4-9} 
\multicolumn{1}{|c|}{}                                                                                     & \multicolumn{1}{c|}{}                                                                                               &                                                                                             & \begin{tabular}[c]{@{}c@{}}Swap model parameters from the external storage into DRAM,\\ task bounded with subnet, reduce computation latency\end{tabular}                   & \begin{tabular}[c]{@{}c@{}}Cloud\\ server\end{tabular} & 2022 & ~\cite{ji2022task}           &                                                             & \checkmark                                                 \\ \cline{2-9} 
\multicolumn{1}{|c|}{}                                                                                     & \multicolumn{1}{c|}{\multirow{7}{*}{\begin{tabular}[c]{@{}c@{}}\textbf{Hardware}\\ \textbf{instruction}\\ \textbf{optimization}\end{tabular}}} & \multirow{3}{*}{\begin{tabular}[c]{@{}c@{}}\textbf{Loop}\\ \textbf{unrolling}\end{tabular}}                   & \begin{tabular}[c]{@{}c@{}}A generalized loop-unrolling method for any type\\ of loop construct, reduce computation latency\end{tabular}                                    & \begin{tabular}[c]{@{}c@{}}Cloud\\ server\end{tabular} & 1999 & ~\cite{huang1999generalized} &                                                             & \checkmark                                                 \\ \cline{4-9} 
\multicolumn{1}{|c|}{}                                                                                     & \multicolumn{1}{c|}{}                                                                                               &                                                                                             & \begin{tabular}[c]{@{}c@{}}A semi-automatic and compile-time approach for identifying\\ the optimal unroll factors, reduce computation latency\end{tabular}                 & \begin{tabular}[c]{@{}c@{}}Cloud\\ server\end{tabular} & 2010 & ~\cite{murthy2010optimal}    &                                                             & \checkmark                                                 \\ \cline{3-9} 
\multicolumn{1}{|c|}{}                                                                                     & \multicolumn{1}{c|}{}                                                                                               & \begin{tabular}[c]{@{}c@{}}\textbf{Register}\\ \textbf{blocking}\end{tabular}                                 & \begin{tabular}[c]{@{}c@{}}A performance model to set the appropriate register\\ block size, reduce computation latency\end{tabular}                                        & MPU                                                    & 2001 & ~\cite{im2001optimizing}     &                                                             & \checkmark                                                 \\ \cline{3-9} 
\multicolumn{1}{|c|}{}                                                                                     & \multicolumn{1}{c|}{}                                                                                               & \begin{tabular}[c]{@{}c@{}}\textbf{Instruction}\\ \textbf{reordering}\end{tabular}                            & \begin{tabular}[c]{@{}c@{}}A flexible multi-criteria instruction reordering heuristic that can\\ be adapted across architectures, reduce computation latency\end{tabular}   & \begin{tabular}[c]{@{}c@{}}Cloud\\ server\end{tabular} & 2018 & ~\cite{rawat2018associative} &                                                             & \checkmark                                                 \\ \hline
\end{tabular}
% \begin{tablenotes}
%     \item Discussion: the performance of different levels of optimization techniques on the same DL model varies. Within the same level of optimization techniques, their performance also differs. Existing DL frameworks for AIoT devices already support the techniques mentioned in this table. However, the criterion for their cross-level combination on AIoT devices needs more exploration.
% \end{tablenotes}
\end{threeparttable}
}
\label{tab:inf_sys}
\end{table*}

\begin{figure}[t]
  \centering
  \includegraphics[width=0.5\textwidth]{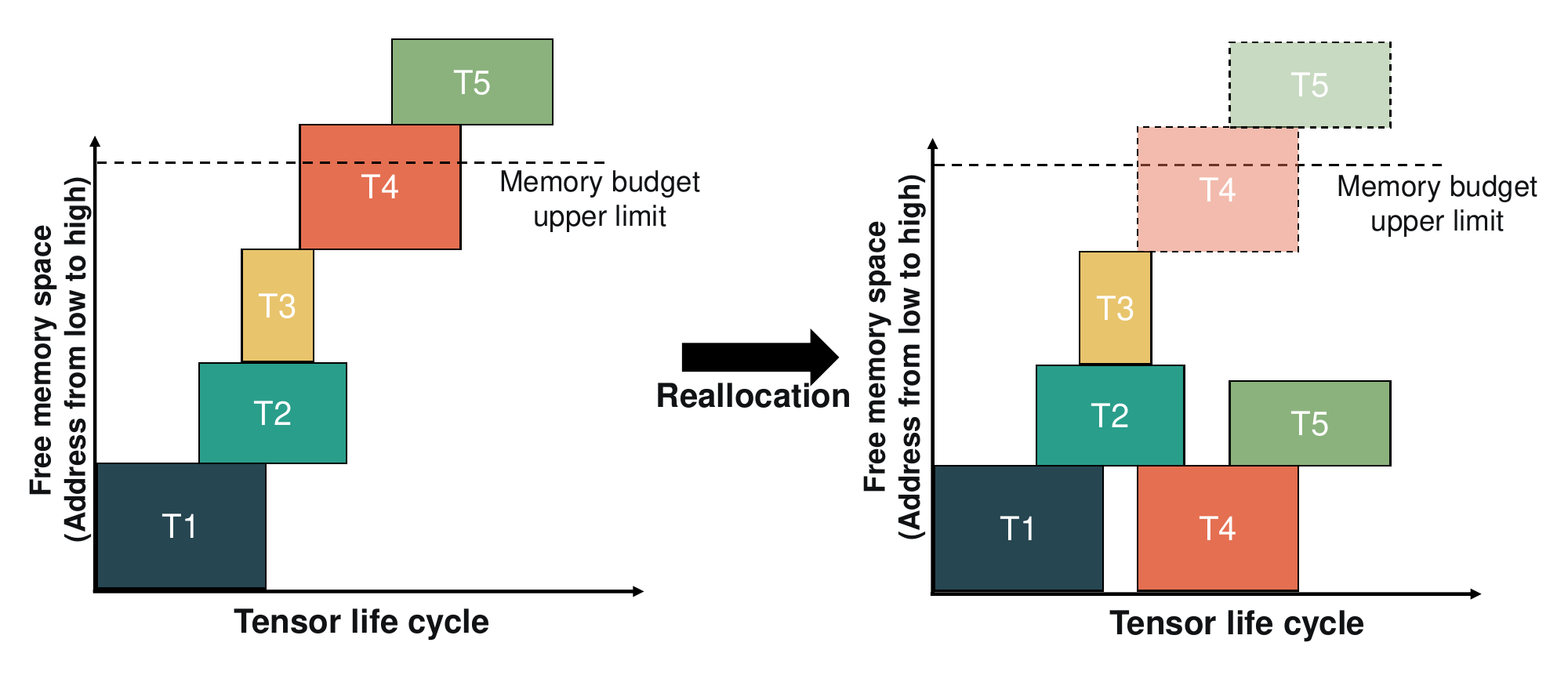}
  \caption{Illustration of memory allocation technique. After reallocating the memory position of tensors, the memory peak is reduced without conflicting access to the tensor.}
  \label{fig:train_alloc}
  % \vspace{-3mm}
\end{figure}

\textbf{d. Memory allocation}
DL model parameters and intermediate activations take up large memory resources during forward propagation as models become more complex.
The default memory allocation schemes by the operating system (\ie OS) always result in memory fragmentations, which cannot be used for other tasks. 
To solve this problem, memory reallocation is necessary.
During DL inference, input tensors, weights, and intermediate activations are all one-offs, we can release them from memory in time after use.
Also, we can allocate the same memory to different tensors that do not coincide with the usage time.
\rev{As shown in Figure~\ref{fig:train_alloc}, T1 $\sim$ T5 represent a set of tensors. 
If we allocate memory for each tensor in order, it will easily reach the memory limit, like tensors T4 and T5. 
And since the lifetime of tensor T4 is completely disjoint with T1, they can be allocated in the same memory block. 
Similarly, T5 and T2 can also share memory.}

Sekiyama \etal~\cite{sekiyama2018profile} proposed a profiler-guided memory allocation method. 
During propagation, they collected information about requested memory blocks and then allocated memory using a heuristic algorithm to reduce peak memory footprint.
By experimenting on advanced DL models, they reduced memory footprint by nearly 50$\%$ and improved inference speed by 4$\times$.
To consider the layer diversity of computation and communication, Wei \etal~\cite{wei2019overcoming} designed a layer-wise memory allocation framework on Field Programmable Gate Array(\ie field-programmable gate array(FPGA)).
They used the layer diversity and the non-overlapping lifespan information of memory buffers to schedule on-chip memory.
The hit rate on the hardware accelerator and the memory allocation policy mapping can also affect its performance.
To search for the best map fastly, many works redesigned the search space, \eg \rev{state-of-the-art}~\cite{venkatesan2019magnet, shen2017maximizing, parashar2019timeloop, stoutchinin2019optimally}.
However, \rev{the efficiency in these DL frameworks} is still slow due to exhaustive searches.
And it does not consider the user-defined or random-sampled constraints, thereby cannot guarantee the global optimum.
Also, they cannot deterministically assess the optimality because predicting the needed CPU time and peak memory in advance is prohibitive.
To address this problem, Symons \etal~\cite{symons2021loma} allocated memory  based on the loop order. 
It takes advantage of DL models' nested "FOR" loop sequence and assigns them to the most appropriate memory hierarchy.
% Thus it can be further optimized regarding latency, energy consumption, and search time.

\begin{figure}[t]
	\centering 
	\subfloat[Memory swapping process.]{\label{fig:mem_swap_a}
		\includegraphics[width=0.92\linewidth]{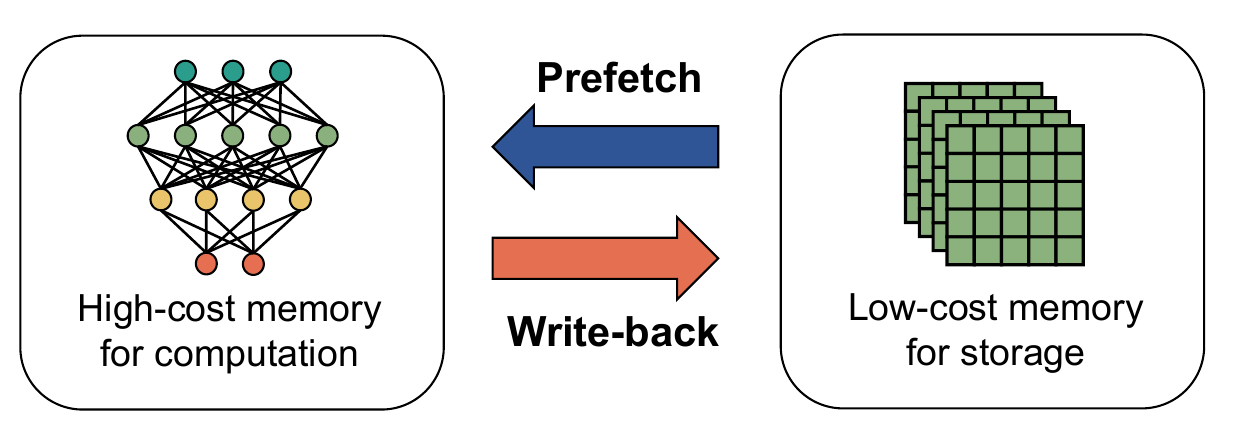}}
		\\
	\subfloat[Memory swapping delay.]{\label{fig:mem_swap_b}
		\includegraphics[width=0.92\linewidth]{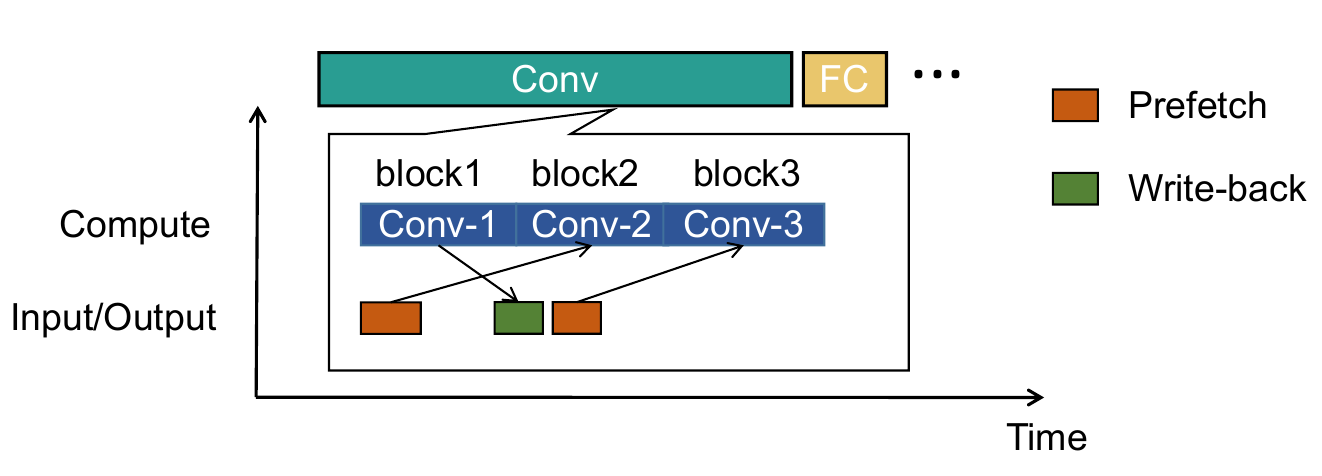}}
	\caption{Illustration of memory swapping techniques for on-device DL inference.}
	\label{fig:mem_swap}
\end{figure}

\textbf{e. Memory swapping.}
Memory swapping~\cite{huang2020swapadvisor,peng2020capuchin} refers to the exchange of tensors or data chunks between the high-cost memory and the low-cost one during computation.
For example, when the device's memory supply cannot meet the tensors' demands, we can swap out partial tensors to free up space and serve current operations.
The focus of memory swapping in the inference phase differs from that in the training phase. 
The training phase focuses more on the reuse of tensors during backpropagation. 
While the inference mainly focuses on forward propagation, \eg prefetching of tensors and the write-back after the computation is completed on memory-scarce AIoT devices, as shown in Figure \ref{fig:mem_swap_a}.
% Generally, the model is stored in high-cost memory for computation, and tensors that take up a large amount of memory can be stored in low-cost memory.
% High-cost memory has a small memory capacity and fast computation speed; low-cost memory is the opposite.
However, memory swapping brings transmission delay. 
To solve this problem, some works propose to cover the transmission delay with the computation delay so that the transmission delay does not affect the overall delay.
When the device resources are extremely limited, the model can be divided into several blocks for execution. 
For example, as shown in Figure \ref{fig:mem_swap_b}, when block 1 is executed, the parameters required by block 2 are prefetched, and the result is written back to the low-cost memory after block 1 is executed.
% \textit{block1} is prepared for the computation of \textit{block2} (\ie prefetch data).
% After the computation in \textit{block1} is completed, \textit{block2} can directly start computing. 
% Meanwhile, the result in \textit{block1} can write back to the low-cost memory.

Miao \etal~\cite{miao2021enabling} proposed a system solution for deploying DL models on MCUs.
It dynamically swapped model blocks between the SRAM and flash of MCU.
This method trades time for space, thus not affecting accuracy.
Besides, swapping tensors between GPU and CPU is also efficient because the GPU is fast with small memory. 
In contrast, the CPU has relatively large memory to store temporary tensors. 
And this method becomes more promising with the development of current GPUs.
They can realize cross-communication and computation based on generous communication bandwidth.
% At the same time, the communication bandwidth between the CPU and the GPU is also relatively enough.
Huang \etal~\cite{huang2020swapadvisor} proposed a universal swapping system called SwapAdvisor. 
It establishes search space with memory allocation and operator scheduling techniques and uses a genetic algorithm to determine exactly when and which tensors to swap before execution. 
% The system maximizes the overlap of computation and communication. 
SwapAdvisor breaks through the GPU memory limit by 12$\times$ and increases the inference speed by 4$\times$.
Ji \etal~\cite{ji2022task} proposed task-aware swapping(\ie TAS) for object detection tasks on IoT devices. 
Since the same type of task involves the same subnetwork, TAS swapped the model parameters of the corresponding subnetwork from external memory to dynamic random access memory(\ie DRAM) in time, according to different task types. 
TAS reduced the DRAM memory by 34.6$\%$ while maintaining accuracy.

\begin{figure*}[t]
  \centering
  \includegraphics[width=.8\textwidth]{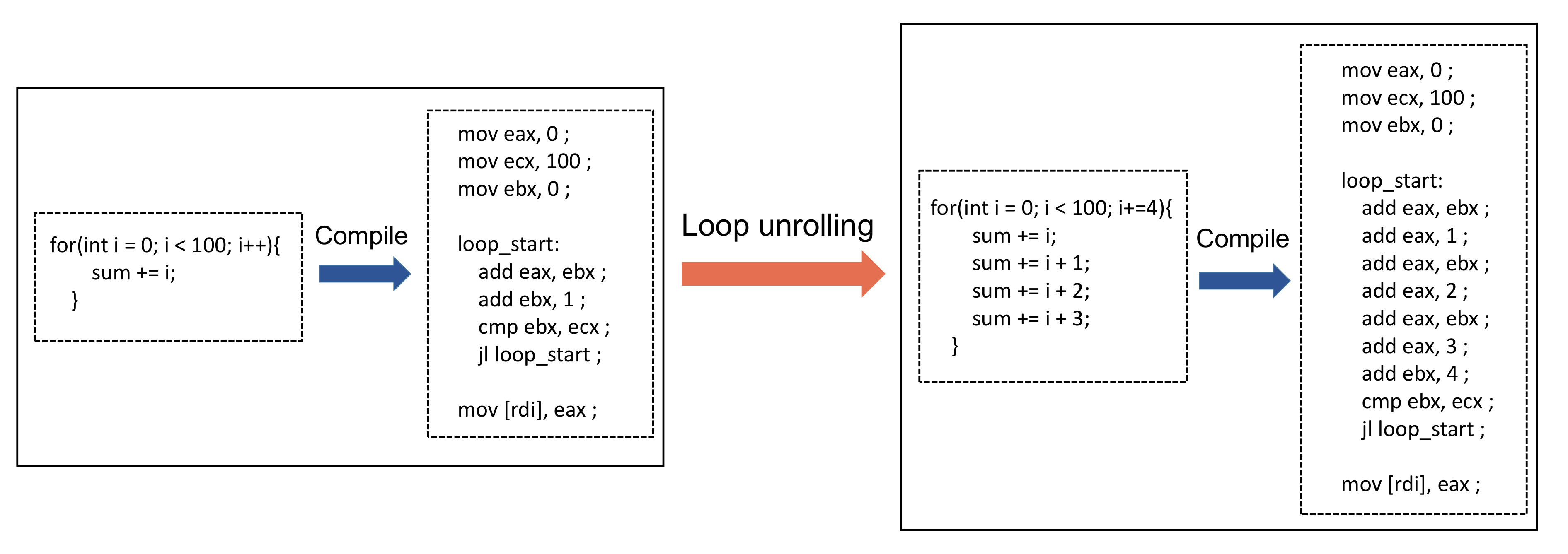}
  \caption{An example of loop unrolling technique.}
  \label{fig:loop_unroll}
  \vspace{-3mm}
\end{figure*}

\textbf{f. Hardware instruction optimization}.
DL inference optimization on AIoT devices also includes the following hardware instruction optimization techniques:

    \textit{\textbf{Loop unrolling}}. It is a widely known code conversion method to improve program execution performance.
    The unrolled loop typically executes fewer instructions than the original one.
    % Because they execute fewer branch instructions and make fewer times index variable modifications.
    % As shown in Figure \ref{fig:loop_unroll}, in the assembly code without loop unrolling on the left, the loop body is executed 100 times.
    % 4 instructions are in the loop body, and 4 instructions are out of the loop body.
    % So the total number of execution is 100$\times$4$+$4$=$404 times. 
    As shown in Figure \ref{fig:loop_unroll}, the assembly loop code on the right is unrolled, and the calculation is reduced from 404 to 229 times.
    % , the loop body is executed 25 times.
    % 9 instructions are in the loop body, and 4 instructions are out of the loop body.
    % So the total number of execution is 25$\times$9$+$4$=$229 times. 
    Thus loop unrolling can speed up calculations.
    % by increasing the step size and reducing the loop number.
    %
    %Loop unrolling can typically improve execution performance by 10$\%$-30$\%$.
    Huang \etal~\cite{huang1999generalized} proposes a generic loop unrolling approach for diverse loop structures (\eg FOR, WHILE, DO-WHILE). 
    The loop unrolling support in the GPU compiler is limited. 
    Giridhar~\cite{murthy2010optimal} developed a semi-automatic compiler to identify the optimal unrolling factor based on compile-time characteristics and the effect of loop unrolling on the program. 
    % By observing experimental results, the execution speed of the optimization method is increased by 70$\%$.

    \textit{\textbf{Register blocking}}. 
    It rationally blocks registers to reduce idleness and increase register multiplexing.
    Based on matrix-independent device features, Sparsity~\cite{im2001optimizing} proposes a performance model to set the appropriate register block size, hence optimizing the sparse matrix computation speed.

    \textit{\textbf{Instruction reordering}}. 
    It scrambles the instructions of different execution units to improve the utilization rate of the pipeline.
    Rawat \etal~\cite{rawat2018associative} proposed a flexible multi-criteria heuristic based on instruction reordering, which can be adapted across architectures.This approach alleviates register pressure while properly controlling the degree of parallelism at the instruction level.

    The compiler or CPU/GPU/TPU processor can reorder instructions to optimize the execution performance of the program.
    For example, instructions with high delay are advanced, and data dependence before and after instructions is reduced.
    From the source program to the final running instructions, there are two stages of reordering:
    
    \textit{\textbf{Compiler reordering}}.
        During compilation, without affecting the result of the program, the compiler reorders instructions based on context analysis to reduce the interaction between CPU and memory. 
        After rearranging, the CPU can read data from registers or cache rows as much as possible.
        
\begin{figure}[t]
    \centering
    \includegraphics[width=0.35\textwidth,scale=1.00]{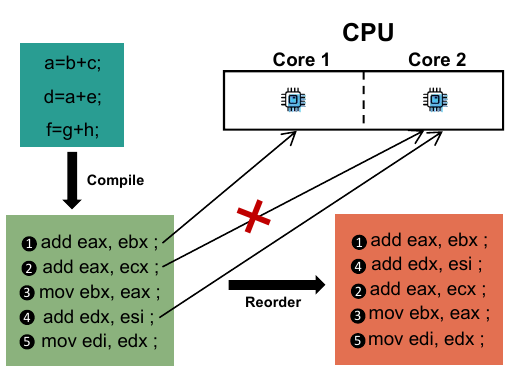}
    \caption{Illustration of parallel instruction set reordering technique.
    The digits $1 \sim 5$ represent the instruction ID, not the actual execution order.}
    \label{fig:para_reorder}
    %\vspace{-1cm}
\end{figure}

    \textit{\textbf{Processor reordering}}. 
        It includes \textit{parallel instruction set reordering} and \textit{memory system reordering}.
        First, in the \textit{parallel instruction set reordering}, modern processors use Instruction-Level Parallelism (ILP) to execute multiple instructions.
        The processor changes the order in which a statement corresponds to a device's instruction.
        As shown in Figure~\ref{fig:para_reorder}, the CPU hopes to execute the \textit{instruction 1} and \textit{instruction 2} in parallel.
        However, both them assign value to the same register \textit{eax}, so the value saved in the final register \textit{eax} is not the result of two addition computations. 
        As a result, the value saved in register \textit{ebx} is incorrect. 
        Therefore, \textit{instruction 1}, \textit{instruction 2}, and \textit{instruction 3} are executed serially. 
        At the same time, the CPU needs to select one of the subsequent instructions (\ie \textit{instruction 4}) to execute with \textit{instruction 1} in parallel.
        With the reordered instruction set, parallel execution can be more efficient.
        %%%
        Second, as for the \textit{memory system reordering}, since the processor uses cache and read/write buffers, loading and storing operations is performed out of order. 
        Therefore, memory system reordering should be well-designed.

\begin{table*}[]
\caption{Summary of related intra-device controllers for on-device DL inference.}
\renewcommand{\arraystretch}{1.3}
\resizebox{\linewidth}{!}{
\scriptsize
\begin{tabular}{|c|c|c|c|c|c|c|}
\hline
\textbf{Category}                          & \textbf{Context awareness}                                                                & \textbf{Controller}                                                                           & \textbf{Cross-level}                                                                       & \textbf{Optimizated performance}                                                                    & \textbf{Year} & \textbf{Ref.}                      \\ \hline
\multirow{6}{*}{\textbf{Controller}} & \begin{tabular}[c]{@{}c@{}}Latency,\\ memory\end{tabular}                        & DQN, DDPG                                                                            & Algorithm level                                                                   & Accuracy, energy consumption                                                               & 2020 & ~\cite{liu2020adadeep}    \\ \cline{2-7} 
                                  & Accuracy                                                                         & /                                                                                    & Algorithm level                                                                   & Energy consumption                                                                         & 2017 & ~\cite{yang2017designing} \\ \cline{2-7} 
                                  & \begin{tabular}[c]{@{}c@{}}Peak memory usage,\\ model size, latency\end{tabular} & \begin{tabular}[c]{@{}c@{}}Multiobjective constrained \\ NAS algorithm\end{tabular}  & \begin{tabular}[c]{@{}c@{}}Algorithm level\end{tabular} & \begin{tabular}[c]{@{}c@{}}Accuracy, peak memory usage,\\ model size, latency\end{tabular} & 2021 & ~\cite{liberis2021munas}  \\ \cline{2-7} 
                                  & Latency, energy, memory                                                          & \begin{tabular}[c]{@{}c@{}}Two-stage NAS for \\ tiny memory constraints\end{tabular} & \begin{tabular}[c]{@{}c@{}}Algorithm level $\&$ compiler level\end{tabular} & Latency, memory                                                                            & 2020 & ~\cite{lin2020mcunet}     \\ \hline
\end{tabular}
}
\label{tab:inf_con}
\end{table*}

% \textbf{\textit{Discussion}}. 
 
\subsubsection{Intra-device cross-level controller}
\label{subsub_infer_contr}

The controller is required upon the above optimization techniques to \rev{automatically adapt} to diverse performance requirements of AIoT applications and resource budgets of AIoT devices.
Liu \etal proposed AdaDeep~\cite{liu2020adadeep}, which leverages the deep reinforcement learning-based optimizer to automatically select the most appropriate combination of compression techniques and hyperparameters for a given DL model in a hierarchical manner.
Yang \etal~\cite{yang2017designing} proposed an energy-aware CNN pruning algorithm, which automatically guides the pruning process by using the energy estimation of DL model.
Edgar \etal~\cite{liberis2021munas} automatically design suitable DL models for MCUs, achieving high accuracy while achieving low memory usage and latency.
MCUNet~\cite{lin2020mcunet} is an example of cross-level system optimization, which can drive better optimization strategies.
They jointly optimize the DL algorithms by TinyNAS and the memory scheduling by TinyEngine to reduce memory usage.
TinyNAS automatically optimizes the search space to fit the tiny resource constraints in MCUs.
TinyEngine improves the existing inference library with code generator-based compilation methods to eliminate memory overhead. 
MCUNet has improved inference speed by 3.4$\times$ and reduced the peak memory footprint on SRAM by 4.1 $\times$.

\subsection{Distributed DL Inference}
\label{subsec:inference_m}

% \TODO{here}

\begin{figure*}[t]
  \centering
  \includegraphics[width=.85\textwidth]{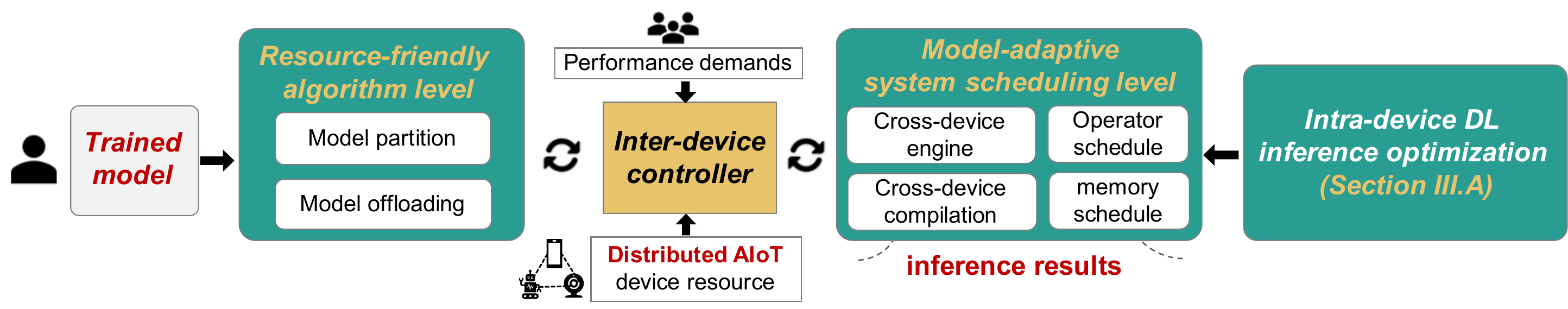}
  \caption{System loop of the algorithm, system scheduling, and inter-device controller for distributed DL inference.}
  \label{fig:inference_multiple}
  % \vspace{-3mm}
\end{figure*}

In addition to on-device optimizing to reduce the resource demands of DL models for local adaptation, distributed DL inference aims to aggregate more resources \rrev{from multiple devices within the networked AIoT systems} to improve inference efficiency. 
This is achieved by partitioning the intensive computations of DL inference tasks and offloading diver portions of them to multiple devices.
Distributed DL inference is particularly beneficial given the increasingly complex structure of modern DL models, which often require memory and computing resources far beyond the limits of a single AIoT device.
For example, MCUs typically have only 256kb of memory, while ResNet, a widely-used model, requires 7.2MB for parameter storage.

Given the challenges mentioned in ~\ref{subsec:sys_layer}, we divide distributed DL inference optimization into the resource-friendly algorithm level, model-adaptive system scheduling level, and inter-device controller, \rev{as shown in Figure~\ref{fig:inference_multiple}}.
The algorithm and system scheduling levels focus on improving the resource efficiency of given heterogeneous hardware from different aspects.
The inter-device controller automatically selects devices and cross-level techniques according to dynamic AIoT context.

\subsubsection{Resource-friendly algorithm level}
It mainly specifies the model partition and offloading schemes for satisfying distributed inference performance demands and resource budgets.
We classify them as \textit{layer-wise} and \textit{operator-wise} categories.
Specifically, the layer-wise scheme refers to partitioning and offloading the model according to model layers. 
And the operator-wise scheme goes deep inside layers, \eg optimizes the operator itself or the connection between operators. 

\begin{figure*}[t]
  \centering
    \hspace{15pt}
    \subfloat[Intermediates]{\label{fig:inter}
    \includegraphics[height=0.15\textwidth]{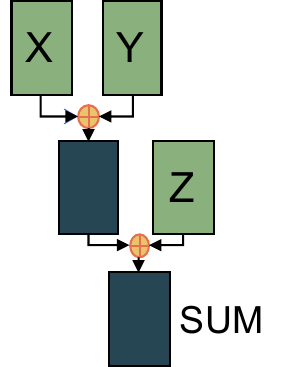}}
    \hspace{15pt}
    \subfloat[Single-pass]{\label{fig:single}
    \includegraphics[height=0.15\textwidth]{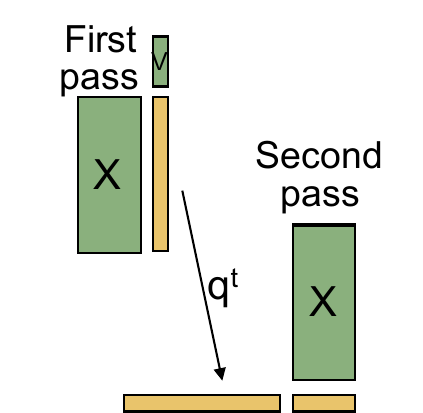}}
    \hspace{15pt}
    \subfloat[Sparsity exploitation]{\label{fig:spar}
    \includegraphics[height=0.15\textwidth]{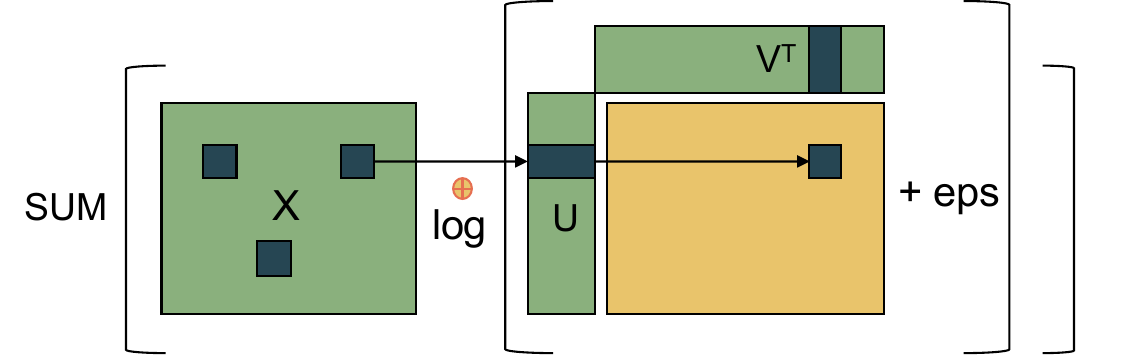}}
    %\vspace{-3mm}
\caption{Illustration of code fusion techniques. (a) Code fusion can eliminate the intermediate in the computing, \(sum(X \oplus Y \oplus Z)\); (b) code fusion can eliminate unnecessary scans of inputs, \({X^T}(Xv) \to {({(Xv)^T}X)^T}\); (c) code fusion allows sparsity exploitation across chains of operations, \(sum(X \oplus \log (U{V^T} + eps))\).
}
\label{fig:fushion}
%\vspace{-3mm}
\end{figure*}

\begin{table*}[t]
\scriptsize
\caption{Summary of cross-level optimization techniques for distributed DL inference.}
\renewcommand{\arraystretch}{1.2}
\label{Tab:co-in}
\resizebox{\linewidth}{!}{
\begin{tabular}{|c|c|c|c|c|}
\hline
\textbf{Focus level}        & \textbf{Technique highlight for improving resource efficiency} &\textbf{Offloading} & \textbf{Year} & \textbf{Ref.}               \\ \hline
\multirow{3}{*}{\textbf{DL model}}                & Multi-dimensional resource management, deep reinforcement learning              & Serial                               & 2021                           & \cite{zhang2021deep}         \\ \cline{2-5} 
                                         & Constrain Markov decision process, Lyapunov optimization                        & Serial                               & 2020                           & \cite{wu2020accuracy}        \\ \cline{2-5} 
                                         & Single batch inference, use several new model parallel methods                  & Parallel                             & 2020                           & \cite{hadidi2020toward}      \\ \hline
\multirow{6}{*}{\textbf{Operator}}                & Operator fusion, distributed code generation                                    & Hybird                               & 2018                           & \cite{boehm2018optimizing}   \\ \cline{2-5} 
                                         & Scalable convolution layer fusion, improve data reuse                           & Parallel                             & 2018                           & \cite{zhao2018deepthings}    \\ \cline{2-5} 
                                         & Partition computing layer, integer linear programming                           & Parallel                             & 2021                           & \cite{stahl2021deeperthings} \\ \cline{2-5} 
                                         & Convolution layer fusion and tiling, memory usage predictor                     & Hybird                               & 2021                           & \cite{farley2021memory}      \\ \cline{2-5} 
                                         & Partition the full connection layer, cover all layers                           & Parallel                             & 2019                           & \cite{stahl2019fully}        \\ \cline{2-5} 
                                         & Vertical partition, weight pruning technique                                    & Parallel                             & 2022                           & \cite{naveen2022memory}      \\ \hline
\multirow{4}{*}{\textbf{Computation graph}}       & Memory-constrained, joint optimization, knowledge distillation                  & Serial                               & 2022                           & \cite{yun2022cooperative}    \\ \cline{2-5} 
                                         & Delay-Sensitive, mixed integer nonlinear programming,mec server                 & Hybird                               & 2020                           & \cite{he2020joint}           \\ \cline{2-5} 
                                         & Distributed model computing system, dynamic partition                           & Parallel                             & 2020                           & \cite{zeng2020coedge}        \\ \cline{2-5} 
                                         & Reduce non-parallel data transfer time, convolution segmentation                & Hybird                               & 2017                           & \cite{mao2017modnn}          \\ \hline
\multirow{2}{*}{\textbf{Inter-device controller}} & Supports customized flexible fine-grained scheduling                            & Parallel                             & 2021                           & \cite{zhang2021deepslicing}  \\ \cline{2-5} 
                                         & Greedy two-dimensional partition, structured models are tightly deployed        & Hybird                               & 2017                           & \cite{mao2017mednn}          \\ \hline
\end{tabular}}
\label{tb:dis_in}
\end{table*}

\textit{Layer-wise scheme}.
Several studies~\cite{yun2022cooperative, zeng2019boomerang, chen2019iraf, zhang2020towards} have put forward the layer-wise distributed DL inference.
Yun \etal~\cite{yun2022cooperative} specialized the lightweight models by knowledge distillation and selected model partition points to minimize inference delay and satisfy the resource limitations of end devices (\eg Raspberry Pi 3B). 
Wu \etal~\cite{wu2020accuracy} expressed the data sampling rate and model offloading problem as a constrained Markov decision process and obtained a heuristic solution.
He~\etal~\cite{he2020joint} model the arrival process of DL inference tasks as Poisson distribution.
And they established a multi-faceted evaluation method \cite{pan2022joint,li2021throughput} to profile the inference latency, accuracy, memory usage, \etc
DeeperThings~\cite{stahl2021deeperthings} established the joint optimization of model partition and device selection as a nonlinear programming problem.
He~\etal~\cite{he2020joint} designed a CRA algorithm based on Markov approximation to search the solution quickly, \eg 350 ms for a 10-device cluster.  

\textit{Operator-wise scheme}.
Most studies focus on integrating and reallocating operators to reduce memory footprint. 
Specifically, merging more operators helps reduce the memory footprint of intermediate outputs and achieve better cross-operator sparsity development. 
For example, Boehe~\etal~\cite{boehm2018optimizing} reduced the intermediate activation of the fusion of different operators and designed a cross-operator sparsity plan to improve computational efficiency.
Stahl~\etal~\cite{stahl2019fully} conducted research on layer operator fusion and reduced the data transmission time between multiple devices by combining feature and weight division with communication-aware layer fusion methods.
Further, Farley~\etal~\cite{farley2021memory} design an independent fusion scheme for conv layer and fc layers, which reduces memory overhead through data reuse.
However, these works lead to high synchronization overhead.
To this end, Zhang~\etal~\cite{zhang2021deepslicing} presented an adaptive cooperative inference system that supports mainstream models (\eg ResNet, GoogleNet).
And it provides a set of APIs to obtain the data location for enabling fine-grained scheduling and memory reclamation.

\subsubsection{Model-adaptive system scheduling level}
This level aims to optimize the input data reuse, intermediate data movement, and memory overhead (\eg memory fragmentation and recycle) in distributed DL inference~\cite{mao2017modnn,mao2017mednn,zhao2018deepthings,stahl2021deeperthings,jeong2018ionn,eshratifar2019jointdnn}.
%\TODO{here}
It should be optimized separately at each device and verified globally cross-device.
And how DL models offload to multiple AIoT devices affects the separate system scheduling technique selection and global performance verification.
The existing DL model offloading schemes can be serial~\cite{wu2020accuracy,zhang2021deep,yun2022cooperative}, parallel~\cite{zhao2018deepthings,stahl2021deeperthings}, and hybrid~\cite{mao2017mednn,ma2018optimizing,farley2021memory}.
For example, MoDNN~\cite{mao2017modnn} partitions the conv layer and distributes them to diverse AIoT devices for parallel computing.
%
% Source nodes are the device that has input data and initiate DL inference.
It partitions the input of the conv layer according to the matrix size and balanced unloading and sends them to different devices for collaborative inference.
MeDNN~\cite{mao2017mednn} is an improvement of MoDNN, solving the problem of unbalanced model partition.
It proposes a greedy 2D model segmentation algorithm to perform static load balancing according to the computing power of each device.
%
% Also, it adopts structural pruning to reduce the model parameter size and accelerate inference. 
%
Further to Mednn, Deepthing\cite{zhao2018deepthings} proposed the improved solution.
It proposed a scalable method to merge and partition the convolution layer by dividing the input matrix into different areas and unloading it to different devices for inference.
It implemented a distributed job-stealing approach to realize dynamic workload allocation and computational efficiency balance, improving data reuse by $68\%$ and reducing latency by $\geq 1.7$ times. 
%
% Afterward, Stahl~\etal~\cite{stahl2021deeperthings} presents Deeperthing, an upgraded version of Deepthing. 
% %
% Unlike Deepthing, which focuses on optimizing the memory footprint of the convolutional layer.
%
Another study~\cite {farley2021memory} also explores the fusion and optimization of CNNs. 
% It uses a memory usage predictor and a search algorithm to provide different fusion and partition strategies for different models. 
% 
Besides, it is difficult for existing operator fusion heuristics to develop distributed operator fusion schemes for complex models, \eg DAGs.
\rev{Matthias \etal~\cite{boehm2018optimizing} proposed an optimization framework for systematic inference fusion schemes, Figure~\ref{fig:fushion} shows three main ways of code fusion.}

\begin{figure*}[t]
  \centering
  \includegraphics[width=.88\textwidth]{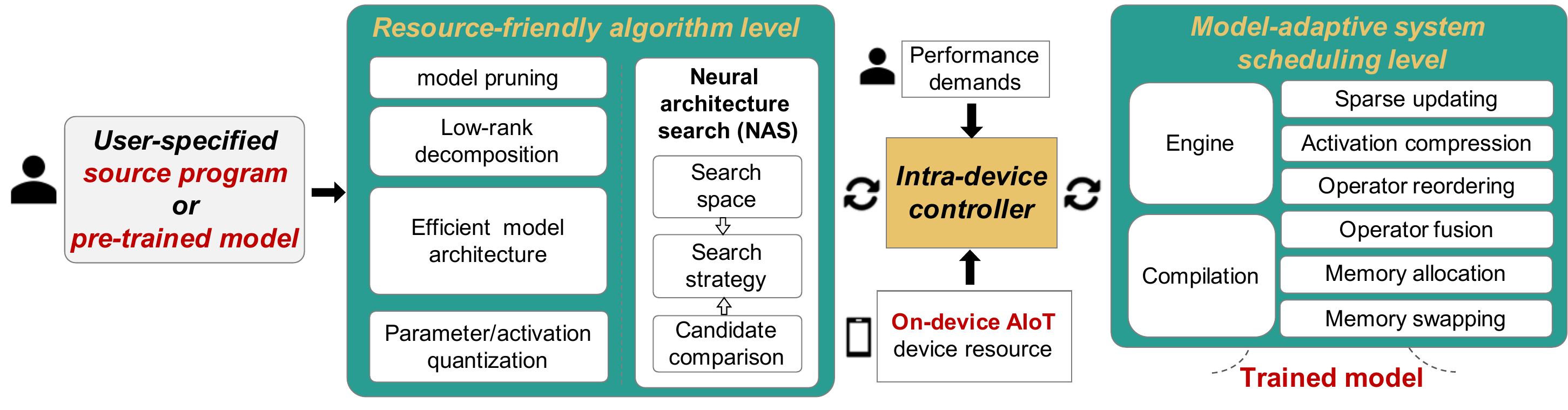}
  \caption{System loop of the algorithm, system scheduling, and intra-device controller for on-device DL training.}
  \label{fig:train_single}
  % \vspace{-3mm}
\end{figure*}

\subsubsection{Inter-device cross-level controller}
The context-aware controller can automatically adapt to varying contexts without re-designing systems, which is necessary for long-term running applications.
Adjusting the distributed inference configurations for dynamic  demands and heterogeneous devices is one of the recent research focuses~\cite{hadidi2020toward,zhang2021deepslicing,laskaridis2020spinn,jiang2020mnn}.
Haddi \etal~\cite{hadidi2020toward} integrates multiple existing parallelism inference technologies into an automated framework.
Zhang \etal~\cite{zhang2021deepslicing} proposed Deepslicing, an adaptive control system integrating multipliers and deep reinforcement learning.
It also exposes a set of APIs to users for adaptation.
However, the controller across several AIoT system levels, especially the underlying system scheduling level, is still lacking.

To be compatible with heterogeneous AIoT devices and boost the inference efficiency, prior efforts also present the inter-device schedule frameworks, such as ~\cite{zhang2020deep,zeng2020coedge,li2018edge,li2019edge}, adapting to heterogeneous AIoT platforms/clusters.
For example, Zeng \etal~\cite{zeng2020coedge} presents Coedge.
It involves a distributed DL inference framework and a fast approximate solution to determine the optimal platform scheduling strategies.
Zhang \etal~\cite{zhang2021deep} considers the embedded chips' memory limits, computing frequency, and battery and transforms the hybrid optimization problem into a Markov decision process. 
As a result, it reduces inference delay and pushes the average accuracy limit caused by pre-determined resource allocation.

\rev{
\textbf{Discussion}. 
Table \ref{tab:inf_sys} summarizes the standalone system scheduling techniques.  And Table \ref{tb:dis_in} illustrates the enabling techniques for distributed DL inference optimization across diverse levels.
The performance of different levels of optimization techniques on the same DL model varies. 
And even within the same level of optimization techniques, their performance also differs. 
Existing DL frameworks for AIoT devices already support the techniques mentioned in this table. 
However, the criterion for their cross-level combination in the context of AIoT applications need more exploration.
Moreover, combining diverse techniques across these levels with an adaptive and automatic controller is still lacking.
}

\section{Cross-level Optimization for DL training}
\label{sec:train}

DL training adopts \textit{batched memory} chunks grouping multiple data samples.
And the memory usage increases proportionately to the batch size.
Specifically, the computation efficiency will be low if not secure sufficient batch size.
Larger batch sizes can bring more accurate distribution statistics for operators like BatchNorm~\cite{ioffe2015batch} to speed up the training convergence and increase the accuracy.

% \TODO{add a comparison of principle and figure between inference and training}

\begin{figure}[t]
	\centering 
	\subfloat[DL inference]{\label{fig:train-infer_a}
		\includegraphics[width=0.95\linewidth]{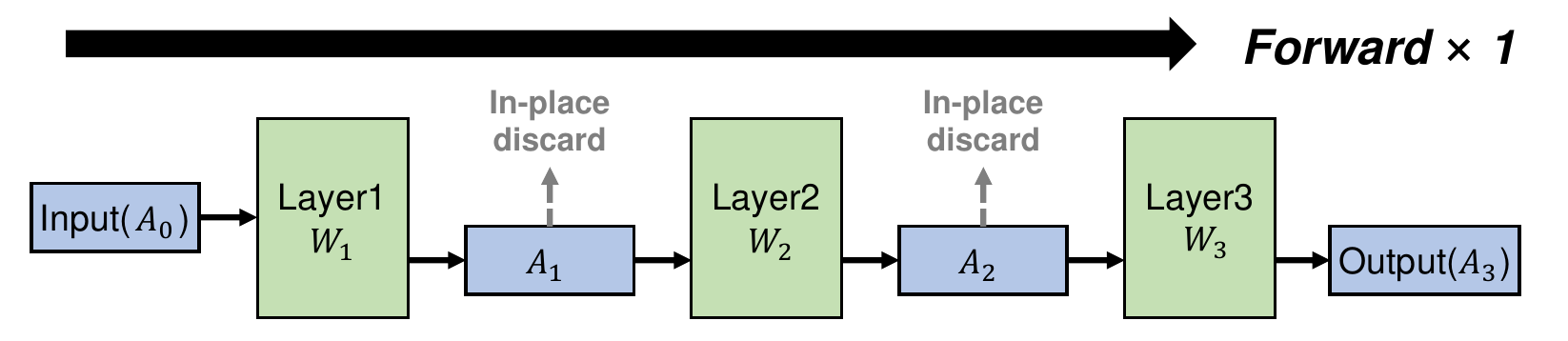}}
		\\
	\subfloat[DL training]{\label{fig::train-infer_b}
		\includegraphics[width=0.95\linewidth]{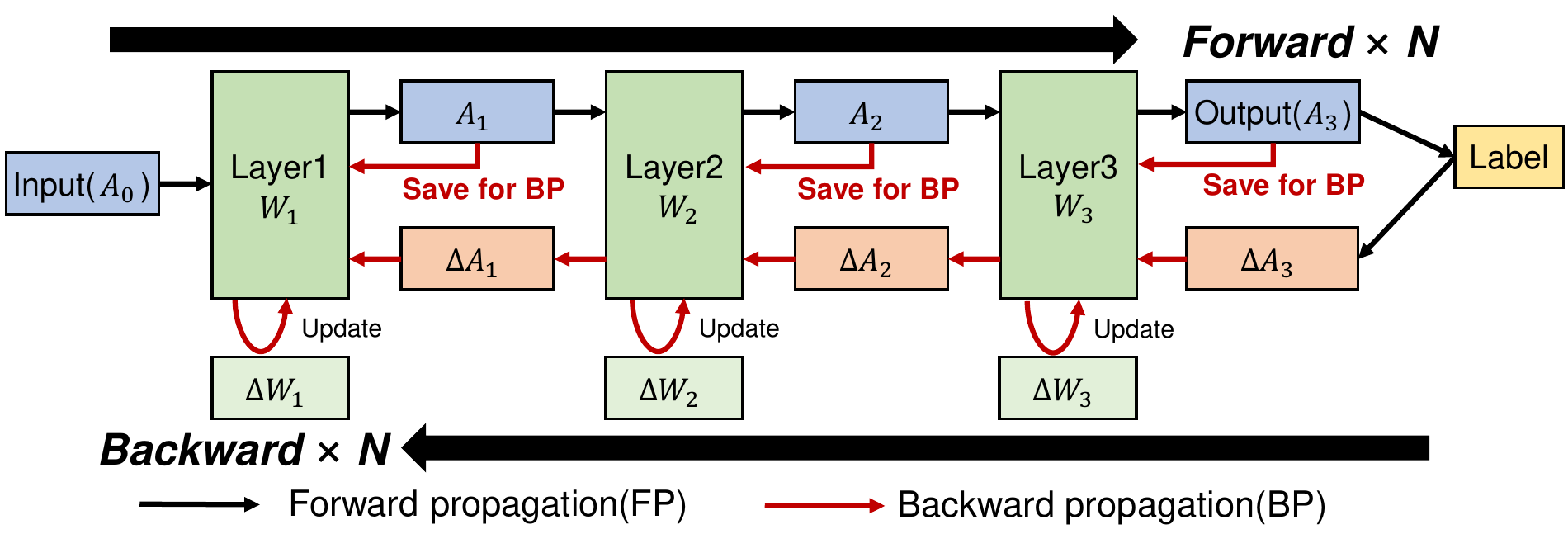}}
	\caption{Comparison illustration of DL training and inference. (a) Inference only requires once forward propagation(FP), and intermediate activation tensor $Ai$ can be discarded to reduce memory usage. (b) In training, tensor $Ai$ needs to be saved for back propagation after the forward propagation.}%, resulting in higher memory usage than that in inference. Since training is repeatedly executed, any tiny problems will be gradually amplified and even lead to a crash.}
    \label{fig:train-infer}
\end{figure}

We first \textbf{differentiate the specificities of DL training} from DL inference in terms of memory and computation demands, as shown in Figure \ref{fig:train-infer}.
\textit{(i)}
\rev{
DL training requires more computation than inference. 
% The gradients $\varDelta A$ and $\varDelta B$ of the intermediate activation A and weight W need to be calculated, respectively, which will significantly increase the delay and energy consumption.
More memory access during the training process also brings more memory access delay. 
For example, the bottleneck of GPU computing power for the GPU devices always lies in the memory access bandwidth and the upgrade of successive generations of GPU focusing on memory bandwidth proves this point~\cite{cai2020tinytl}. 
}

\textit{(ii)}
\rev{
DL training requires more memory space than inference.
According to the chain derivative rule of backpropagation, calculating the gradient and derivative of the weights in the $i^{th}$ layer requires using the derivative of the $i+1^{th}$ layer and the input of this layer. 
Therefore, the \textit{intermediate activations} $A_1$, $A_2$ produced during forward propagation need to be saved till backpropagation calculating the gradients $\varDelta A$, $\varDelta B$ and updating the weights during backpropagation. 
In contrast, $A_1$, $A_2$ in model inference process do not need to be preserved. They can be released right after they have been used, which brings a bigger memory requirement to the training process.
}

\textit{(iii)}
\rev{
DL training needs $N$ rounds of loop iteration, while inference needs only one round.
For algorithm level (\eg model parameter/activation quantization), if not carefully designed, the tiny instabilities generated will be amplified in thousands of iterations and even leads to training crash~\cite{lin2022device}.
Compared to the "one-time" property of the model inference process, DL training needs to consider optimization across \textit{multiple iterations}, which also challenges system optimization.
}

\subsection{On-device DL Training}
\label{subsec:train_1}

% \textbf{Challenge discussion}.
It is non-trivial to optimize three key performance metrics, \ie \textit{latency}, \textit{memory}, and \textit{accuracy}, simultaneously for on-device DL training.
% % 
% \textit{First}, existing studies always optimize two sides of them while sacrificing the other.
% %
% For example, reducing memory usage and maintaining accuracy may lead to higher DL training latency (\eg additional memory access and transmission latency)~\cite{rhu2016vmodel, chen2018modnn}.
% %
% On-device training takes increased time as the price to get lower memory usage and maintain the task accuracy~\cite{chen2016training}~\cite{gomez2017reversible}.
% %
% \textit{Second}, The research on the lowest-cost microcontrollers (MCUs) is still limited, although training the DL model on resource-limited devices is feasible, such as on weak edge server NVIDIA Jetson~\cite{li2019rilod}~\cite{farhadi2020enabling} or Raspberry Pi~\cite{disabato2020incremental}. 
% %
% On the one hand, popular training frameworks such as Tensorflow~\cite{abadi2016tensorflow} and Pytorch~\cite{paszke2017automatic} require huge equipment resources. 
% MCU is a bare-metal device that does not have the operating system and runtime support required by mainstream DL frameworks. 
% On the other hand, existing DL frameworks for tiny devices, like TF-Lite~\cite{tensorflowlite}, TVM~\cite{chen2018tvm}, NCNN~\cite{ncnn}, still lack support for backpropagation which is necessary for training.
%
Given challenges towards optimizing memory usage, latency, and accuracy simultaneously (see $\S$ \ref{sebsec:texonomy}), we summarize the on-device DL training optimization techniques into diverse system levels as described in $\S$ \ref{subsec:sys_layer}, \ie the resource-friendly algorithm level, model-adaptive system scheduling level, and inter-device controller.
\rev{Figure \ref{fig:train_single} shows their relationships in the system loop.}

\subsubsection{Resource-friendly algorithm level}
We first briefly introduce the algorithm-level techniques, \ie compressing DL model structure~\cite{bulo2018place}~\cite{yang2022rep}, reducing parameter/activation bit width ~\cite{deng2015reduced}\cite{zhou2016dorefa}, and sparse updating~\cite{liu2018dynamic}~\cite{dai2020sparsetrain}.

\textbf{a. Parameter/activation quantization}.
DL model quantization during training is more complicated than inference since it brings instability to  training~\cite{friesen2017deep}~\cite{zhuang2018towards}~\cite{finkelstein2019fighting}.
Deng \etal~\cite{deng2015reduced} conducted the first exploration of model quantitation in DL training.
They implemented reduced-precision memory access of parameters and saved significant memory bandwidth using an approximator.
Subsequently, Zhou \etal~\cite{zhou2016dorefa} train DL models with low-bit width weights, activations, and gradients on diverse devices, \eg CPU, GPU, application specific integrated circuit(ASIC), and FPGA, to speed up training.
Besides, Micikevicius \etal~\cite{micikevicius2017mixed} maintain a single-precision copy of weights to prevent information loss caused by quantification. 
Thus they preserve gradient values with small magnitudes and result in half-precision arithmetic.
Huang \etal~\cite{huang2020adaptive} proposed the adaptive precision training (APT) method to balance energy cost, memory usage, and accuracy in DL training.
Motivated by~\cite{gupta2015deep}, they find that starting DL training with low precision benefits energy and memory savings. 
APT dynamically allocates layer-wise bit precision, allowing model to learn faster. 
Because once the training curve reaches a plateau, increasing precision allows DL training to approach better accuracy with fewer training epochs. 
And it uses an application-specific hyperparameter to balance the abovementioned three metrics automatically.
Wang \etal~\cite{wang2022towards} find that during the backpropagation in training, mainly the activation maps' low-frequency component (LFC) is used. 
As such, they preserve the high-precision copy of LFC while compressing the high-frequency component (HFC) into a low-precision copy. 
This greatly reduces memory cost while not dramatically decreasing the precision of backpropagations.
To realize DL training quantization on MCUs lacking DL training frameworks and low-bit width APIs.
Yu \etal~\cite{yu2019tf} deploy sub-byte models on MCUs efficiently. 
They propose a training framework for low-precision models, followed by direct buffer convolution and packed sub-byte multiply-accumulation to accelerate on-device training.
Lin \etal~\cite{lin2022device} proposed quantization-aware scaling to alleviate the gradient scale mismatch issues caused by mixed bit-precision  training.
They stabilized the quantized training by calibrating the gradient scales for MCUs.
Instead of improving the quantizer functions, Lu \etal~\cite{lu2023quantization} optimize the training process from a weight-searching standpoint.

\textbf{b. Efficient model architecture}.
DL model compression is another promising method to optimize the AIoT system performance for DL training.
Bulo \etal~\cite{bulo2018place} proposed in-place activated batch normalization (BN) that eliminates intermediate results and recovers required information during the backward pass, leading to a 50$\%$ reduction in memory usage. 
It can be easily applied on existing models (\eg ResNet~\cite{he2016deep}, DenseNet~\cite{huang2017densely}) with a minor increase (about 2$\%$) in training latency.
Besides, Jung \etal~\cite{jung2019restructuring} split the BN layer into two sub-layers to restructure BN layers.
Because existing memory access optimization approaches, such as fusing convolution layers, are ineffective for accelerating BN due to their inability to optimize mini-batch-related calculations during training.
To significantly reduce main memory accesses and optimize calculations related to mini-batch, they combined the first sub-layer with the preceding conv layer and the second sub-layer with the following conv layer.
To reduce memory usage, some studies also proposed new model structures.
\eg tiny-transfer-learning (TinyTL) \cite{cai2020tinytl} added the lite residual module with low-dimensional features and group convolution to improve the model's migration ability.
And it avoids intermediate activation storage by only training bias and lite residual modules.
This approach leads to a reduction in training memory costs at near-constant precision. 
Yang \etal~\cite{yang2022rep} proposed a reprogramming network, which trains the model using the new task data to reprogram the intermediate features of a backbone model, leading to lower training memory and higher accuracy.

\textbf{c. Sparse updating}.
Like pruning in the inference phase, sparse updating does not change the model structure but selects a part of the network to update in each training epoch.
Most pruning studies focus on DL inference and are unsuitable for DL training.
Liu \etal~\cite{liu2018dynamic} utilize dynamic and sparse graphs (DSGs) in DL training, activating a few neurons during each iteration, resulting in considerable memory savings with competitive accuracy.
Dai \etal~\cite{dai2020sparsetrain} proposed SparseTrain, which employs a stochastic pruning algorithm for each layer and a sparse architecture with 1-dimensional convolution dataflow to realize implicit and artificial sparsity for training acceleration. 
Lin \etal~\cite{lin2022device} skipped the gradient calculation of insignificantly necessary layers and tensors.
And they achieved the near-optimal solution with sparse updating through an evolutionary search.
Kaplan \etal~\cite{kaplun2023subtuning} presented SubTuning, which only trains a carefully selected subset of layers depending on the fine-tuning profile while freezing the remaining. 
They proved that SubTuning fastly obtains training convergences and even outperforms full fine-tuning when training data is lacking.

\begin{table*}[]
\centering
\scriptsize
\caption{Summary of model-adaptive system scheduling techniques for on-device DL training.}
\resizebox{\textwidth}{!}{%
\renewcommand{\arraystretch}{1.1}
\setlength{\tabcolsep}{4pt}
\begin{threeparttable}
\begin{tabular}{|cccc|c|c|c|c|c|}
\hline
\multicolumn{4}{|c|}{\textbf{Category}} & \textbf{Highlight technique for improving resource efficiency} & \textbf{Year} & \textbf{Ref.} & \textbf{\begin{tabular}[c]{@{}c@{}}Compiler \\ frontend\end{tabular}} & \textbf{\begin{tabular}[c]{@{}c@{}}Compiler \\ backend\end{tabular}} \\ \hline
\multicolumn{1}{|c|}{\multirow{58}{*}{\textbf{\begin{tabular}[c]{@{}c@{}}Model-adaptive \\ system \\ scheduling level\end{tabular}}}} & \multicolumn{1}{c|}{\multirow{32}{*}{\textbf{\begin{tabular}[c]{@{}c@{}}Computation  \\  graph \\ level \end{tabular}}}} & \multicolumn{1}{c|}{\multirow{16}{*}{\textbf{\begin{tabular}[c]{@{}c@{}}Intermediate  \\  activation\end{tabular}}}} & \multirow{11}{*}{\textbf{Recomputation}} & \begin{tabular}[c]{@{}c@{}}Split model and only retain part of activations, \\ reduce memory usage with extra  computation cost\end{tabular} & 2016 & ~\cite{chen2016training} &  & \checkmark \\ \cline{5-9} 
\multicolumn{1}{|c|}{} & \multicolumn{1}{c|}{} & \multicolumn{1}{c|}{} &  & \begin{tabular}[c]{@{}c@{}}Use dynamic programming in recomputation, \\ optimize computation cost\end{tabular} & 2016 & ~\cite{gruslys2016memory} &  & \checkmark \\ \cline{5-9} 
\multicolumn{1}{|c|}{} & \multicolumn{1}{c|}{} & \multicolumn{1}{c|}{} &  & \begin{tabular}[c]{@{}c@{}}Recomputation for ResNets, reduce memory usage with \\ extra computation cost\end{tabular} & 2017 & ~\cite{gomez2017reversible} & \checkmark &  \\ \cline{5-9} 
\multicolumn{1}{|c|}{} & \multicolumn{1}{c|}{} & \multicolumn{1}{c|}{} &  & \begin{tabular}[c]{@{}c@{}}Dynamic tensor rematerialization, reduce memory \\ usage with less computation cost\end{tabular} & 2020 & ~\cite{kirisame2020dynamic} &  & \checkmark \\ \cline{5-9} 
\multicolumn{1}{|c|}{} & \multicolumn{1}{c|}{} & \multicolumn{1}{c|}{} &  & \begin{tabular}[c]{@{}c@{}}Memory-calibrated progressive recomputation on \\ mobile phone, reduce memory usage\end{tabular} & 2022 & \cite{wang2022melon} &  & \checkmark \\ \cline{5-9} 
\multicolumn{1}{|c|}{} & \multicolumn{1}{c|}{} & \multicolumn{1}{c|}{} &  & \begin{tabular}[c]{@{}c@{}}Progressive recomputation based on memory\\ worthiness of tensors, reduce memory usage\end{tabular} & 2022 & \cite{gim2022memory} &  & \checkmark \\ \cline{4-9} 
\multicolumn{1}{|c|}{} & \multicolumn{1}{c|}{} & \multicolumn{1}{c|}{} & \multirow{5}{*}{\textbf{Compression}} & \begin{tabular}[c]{@{}c@{}}Two Layer-specific encoding schemes, lossless and \\ lossy, reduce memory usage with extra latency\end{tabular} & 2018 & ~\cite{jain2018gist} & \checkmark &  \\ \cline{5-9} 
\multicolumn{1}{|c|}{} & \multicolumn{1}{c|}{} & \multicolumn{1}{c|}{} &  & \begin{tabular}[c]{@{}c@{}}Optimized JPEG method for activation compression \\ and accelerator,  reduce memory usage with extra latency\end{tabular} & 2020 & ~\cite{evans2020jpeg} &  & \checkmark \\ \cline{5-9} 
\multicolumn{1}{|c|}{} & \multicolumn{1}{c|}{} & \multicolumn{1}{c|}{} &  & \begin{tabular}[c]{@{}c@{}}A bitmap compression technique using activation \\ sparsity, reduce  memory usage with extra latency\end{tabular} & 2021 & ~\cite{hosny2021sparse} & \checkmark &  \\ \cline{3-9} 
\multicolumn{1}{|c|}{} & \multicolumn{1}{c|}{} & \multicolumn{1}{c|}{\multirow{15}{*}{\textbf{Operator}}} & \multirow{3}{*}{\textbf{Reordering}} & \begin{tabular}[c]{@{}c@{}}Discard gradients, reduce memory usage\end{tabular} & 2022 & ~\cite{lin2022device} & \checkmark &  \\ \cline{5-9} 
\multicolumn{1}{|c|}{} & \multicolumn{1}{c|}{} & \multicolumn{1}{c|}{} &  & \begin{tabular}[c]{@{}c@{}} Re-order operators to reduce the number of \\ complex operators, reduce computation cost\end{tabular} & 2022 & ~\cite{unger2022unity} & \checkmark &  \\ \cline{4-9} 
\multicolumn{1}{|c|}{} & \multicolumn{1}{c|}{} & \multicolumn{1}{c|}{} & \multirow{11}{*}{\textbf{Fusion}} & \begin{tabular}[c]{@{}c@{}}Relaxed graph substitutions by a backtracking search \\ algorithm,  reduce execution time and memory usage\end{tabular} & 2019 & ~\cite{jia2019optimizing} &  & \checkmark \\ \cline{5-9} 
\multicolumn{1}{|c|}{} & \multicolumn{1}{c|}{} & \multicolumn{1}{c|}{} &  & \begin{tabular}[c]{@{}c@{}}Transferable deep reinforcement learning for searching, \\ reduce execution time\end{tabular} & 2020 & ~\cite{zhou2020transferable} & \checkmark &  \\ \cline{5-9} 
\multicolumn{1}{|c|}{} & \multicolumn{1}{c|}{} & \multicolumn{1}{c|}{} &  & \begin{tabular}[c]{@{}c@{}}Account for memory access constraints, a unified \\ memory function, reduce execution time\end{tabular} & 2020 & ~\cite{hu2020jittor} & \checkmark &  \\ \cline{5-9} 
\multicolumn{1}{|c|}{} & \multicolumn{1}{c|}{} & \multicolumn{1}{c|}{} &  & \begin{tabular}[c]{@{}c@{}}Merge memory-intensive operators into large GPU \\ kernels, reduce execution time\end{tabular} & 2020 & ~\cite{zheng2020fusionstitching} &  & \checkmark \\ \cline{5-9} 
\multicolumn{1}{|c|}{} & \multicolumn{1}{c|}{} & \multicolumn{1}{c|}{} &  & \begin{tabular}[c]{@{}c@{}}Reduce data movement in memory for Transformer, \\ reduce execution time\end{tabular} & 2021 & ~\cite{ivanov2021data} &  & \checkmark \\ \cline{5-9} 
\multicolumn{1}{|c|}{} & \multicolumn{1}{c|}{} & \multicolumn{1}{c|}{} &  & \begin{tabular}[c]{@{}c@{}}Consider multiple optimization objectives \\ for memory-intensive computations, reduce latency\end{tabular} & 2022 & ~\cite{zheng2022astitch} &  & \checkmark \\ \cline{2-9} 
\multicolumn{1}{|c|}{} & \multicolumn{1}{c|}{\multirow{26}{*}{\textbf{\begin{tabular}[c]{@{}c@{}}Resource \\ scheduling\\ level\end{tabular}}}} & \multicolumn{2}{c|}{\multirow{5}{*}{\textbf{Memory allocation}}} & \begin{tabular}[c]{@{}c@{}}Split tensor into micro-tensor to allocate and \\ swap memory, reduce memory usage and latency\end{tabular} & 2019 & ~\cite{nie2022tsplit} &  & \checkmark \\ \cline{5-9} 
\multicolumn{1}{|c|}{} & \multicolumn{1}{c|}{} & \multicolumn{2}{c|}{} & \begin{tabular}[c]{@{}c@{}}Tensor life cycle computation and memory sharing, \\ reduce memory usage\end{tabular} & 2022 & ~\cite{moon2022nntrainer} &  & \checkmark \\ \cline{5-9} 
\multicolumn{1}{|c|}{} & \multicolumn{1}{c|}{} & \multicolumn{2}{c|}{} & \begin{tabular}[c]{@{}c@{}}Tensor-lifetime-aware memory allocation algorithm,\\  reduce memory usage, energy cost, and latency\end{tabular} & 2022 & ~\cite{wang2022melon} &  & \checkmark \\ \cline{3-9} 
\multicolumn{1}{|c|}{} & \multicolumn{1}{c|}{} & \multicolumn{2}{c|}{\multirow{21}{*}{\textbf{Memory   swapping}}} & \begin{tabular}[c]{@{}c@{}}The first work on DL memory swapping between \\ host(CPU) and   device(GPU), reduce memory usage\end{tabular} & 2016 & ~\cite{rhu2016vdnn} &  & \checkmark \\ \cline{5-9} 
\multicolumn{1}{|c|}{} & \multicolumn{1}{c|}{} & \multicolumn{2}{c|}{} & \begin{tabular}[c]{@{}c@{}}Dynamically select convolution operators and \\ memory swapping  strategy, reduce device memory usage\end{tabular} & 2018 & ~\cite{chen2018modnn} &  & \checkmark \\ \cline{5-9} 
\multicolumn{1}{|c|}{} & \multicolumn{1}{c|}{} & \multicolumn{2}{c|}{} & \begin{tabular}[c]{@{}c@{}}Combine recomputation and memory swapping, \\ reduce memory  usage\end{tabular} & 2018 & ~\cite{wang2018superneurons} &  & \checkmark \\ \cline{5-9} 
\multicolumn{1}{|c|}{} & \multicolumn{1}{c|}{} & \multicolumn{2}{c|}{} & \begin{tabular}[c]{@{}c@{}}Use categorized topological ordering to simulate \\ graph  execution, reduce memory usage\end{tabular} & 2019 & ~\cite{le2019automatic} &  & \checkmark \\ \cline{5-9} 
\multicolumn{1}{|c|}{} & \multicolumn{1}{c|}{} & \multicolumn{2}{c|}{} & \begin{tabular}[c]{@{}c@{}}Decrease amount of synchronization operations and \\ memory offload decisions, reduce memory usage\end{tabular} & 2019 & ~\cite{shriram2019dynamic} &  & \checkmark \\ \cline{5-9} 
\multicolumn{1}{|c|}{} & \multicolumn{1}{c|}{} & \multicolumn{2}{c|}{} & \begin{tabular}[c]{@{}c@{}}Joint optimization on operator scheduling, memory \\ allocation, and swap decisions, reduce memory usage\end{tabular} & 2020 & ~\cite{huang2020swapadvisor} &  & \checkmark \\ \cline{5-9} 
\multicolumn{1}{|c|}{} & \multicolumn{1}{c|}{} & \multicolumn{2}{c|}{} & \begin{tabular}[c]{@{}c@{}}Memory management based on dynamic tensor access \\ pattern, reduce device memory usage\end{tabular} & 2020 & ~\cite{peng2020capuchin} &  & \checkmark \\ \cline{5-9} 
\multicolumn{1}{|c|}{} & \multicolumn{1}{c|}{} & \multicolumn{2}{c|}{} & \begin{tabular}[c]{@{}c@{}}Optimize tensor management on heterogeneous \\ memory, reduce  memory usage\end{tabular} & 2021 & ~\cite{ren2021sentinel} &  & \checkmark \\ \cline{5-9} 
\multicolumn{1}{|c|}{} & \multicolumn{1}{c|}{} & \multicolumn{2}{c|}{} & \begin{tabular}[c]{@{}c@{}}Selectively compress tensors before memory swapping, \\ reduce memory usage\end{tabular} & 2021 & ~\cite{chen2021cswap} &  & \checkmark \\ \cline{5-9} 
\multicolumn{1}{|c|}{} & \multicolumn{1}{c|}{} & \multicolumn{2}{c|}{} & \begin{tabular}[c]{@{}c@{}}Memory swapping without  high-level information of \\ computation graphs, reduce memory usage\end{tabular} & 2022 & ~\cite{li2022application} &  & \checkmark \\ \cline{5-9} 
\multicolumn{1}{|c|}{} & \multicolumn{1}{c|}{} & \multicolumn{2}{c|}{} & \begin{tabular}[c]{@{}c@{}}Improve prefetching using correlation tables to \\ speed up memory swapping, reduce memory usage\end{tabular} & 2023 & ~\cite{deepum2023jaehoon} &  & \checkmark \\ \hline
\end{tabular}
\end{threeparttable}
}
\label{tab:single_train_system}
\end{table*}

\subsubsection{Model-adaptive system scheduling level}
It further maximizes system performance and resource efficiency beyond the optimization capabilities of the algorithm-level techniques and thus pushes the limit of performance-resource tradeoff.
As shown in Table ~\ref{tab:single_train_system}, it optimizes the intermediate activation tensors and operators in the computation graph, optimizes the runtime/compiler of the DL frameworks, and re-allocates memory/computing resources.

\begin{figure}[t]
    \centering
    \includegraphics[width=0.38\textwidth,scale=1.00]{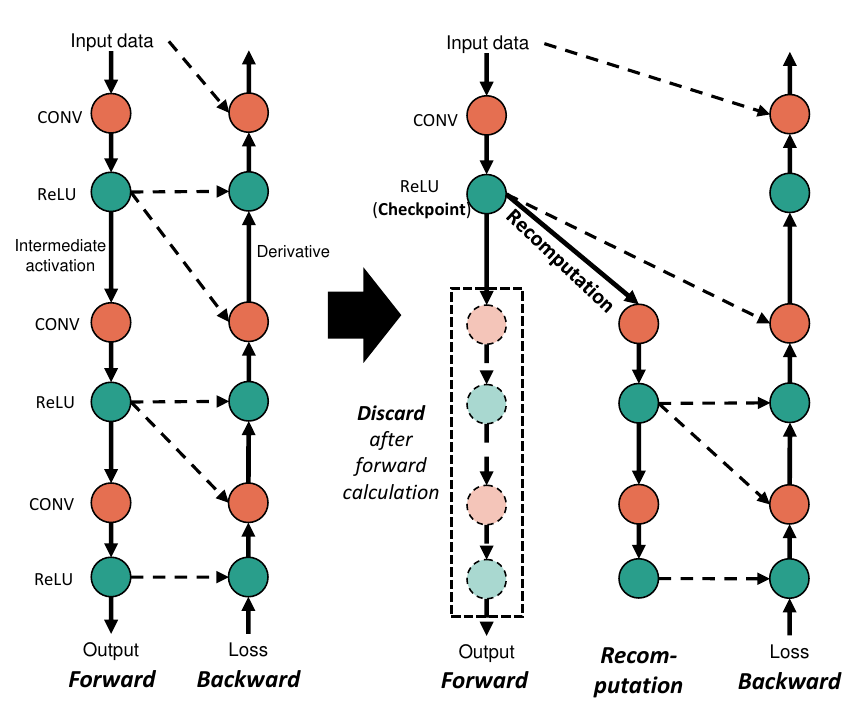}
    % \vspace{-1mm}
    \caption{Illustration of recomputation technique. The intermediate activation is discarded after forward computation and recalculated before backpropagation.}
    \label{fig:Recomputation}
%\vspace{-3mm}
\end{figure}

\textbf{a. Recomputation}.
In DL training, the intermediate activations generated during the forward propagation are typically stored until the backpropagation to calculate the gradient, which causes a large memory footprint. 
Experiments have shown that model's intermediate activations use much more memory than parameters~\cite{cai2020tinytl}. 
\rev{To this end, recomputation methods~\cite{chen2016training, gomez2017reversible, kirisame2020dynamic} discard part of intermediate activations right after forward computation to save memory usage and recalculates these activations during backpropagation for gradient calculation, as shown in Figure \ref{fig:Recomputation}. }
Chen \etal~\cite{chen2016training} is the first to present the recomputation technique, which splits the model into several parts and keeps only the first activation of each part after forward computation. 
During backpropagation, activations within each part are computed from the first activation retained.
This work realizes significant memory savings.
And the recompense of computing delay for memory space is also beneficial, especially when sufficient computing power is available. 
Following it, Gruslys \etal~\cite{gruslys2016memory} use dynamic programming to find a storage strategy that optimizes computational cost for a given memory budget.
Gomez \etal~\cite{gomez2017reversible} applied recomputation in ResNets to save memory footprint during backpropagation. 
These recomputation methods were mainly implemented on cloud servers with sufficient computing power. 
The recomputation operations may bring unacceptable additional latency overhead for devices such as Raspberry Pi and MCUs. 
\cite {wang2022melon, gim2022memory} apply recomputation strategies on mobile phones, using slight additional latency in exchange for significant memory savings.

\begin{figure}[t]
    \centering
    \includegraphics[width=0.3\textwidth,scale=0.5]{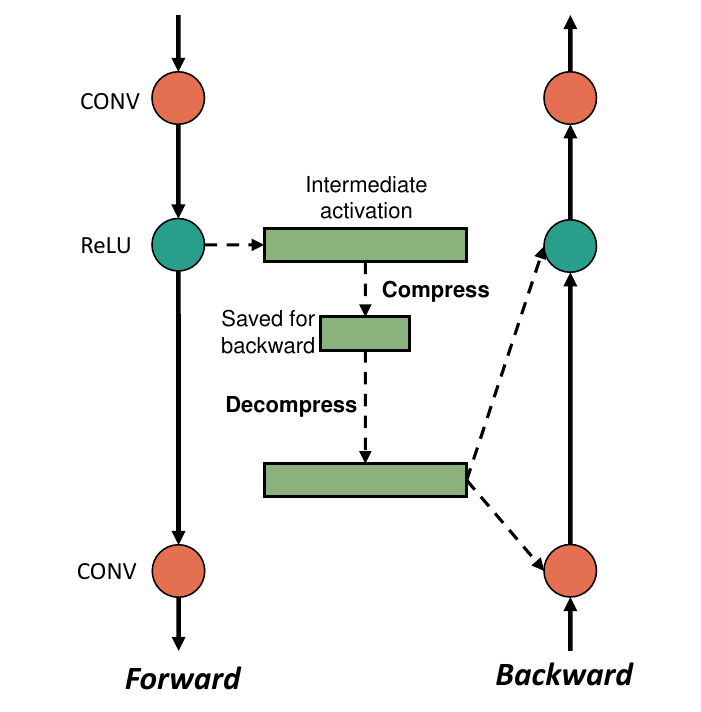}
    % \vspace{-1mm}
    \caption{Illustration of intermediate activation encoding technique. Intermediate activation is encoded after forward computation and decoded during backpropagation.}
    \label{fig:Intermediate activation compression}
%\vspace{-3mm}
\end{figure}

\textbf{b. Intermediate activation encoding}.
\rev{
Similar to but different from recomputation, intermediate activation compression does not directly discard intermediate activations but \textit{temporarily encode} activations after forward propagation and then \textit{decode} them during backpropagation to calculate gradients, as shown in Figure \ref{fig:Intermediate activation compression}.
}
%
%尽管压缩张量并不能xi解压缩一个中间激活通常比重新计算它花费更少时间，
It thereby strikes a valuable balance between computing latency and memory savings rather than simply trading latency for memory resources.
Specifically, Jain \etal~\cite{jain2018gist} proposed Gist, a system that employs layer-specific encoding schemes, lossless and lossy, to significantly reduce the memory consumption of targeted feature maps.
They store an encoded representation of feature maps and decode them for use in the backward pass; the full-fidelity feature maps are used in their forward pass and relinquished immediately.
Evans \etal~\cite{evans2020jpeg} proposed JPEG for activations (JPEGACT), a lossy activation offloading accelerator. 
They expand the well-known JPEG algorithm for 2D image encoding to compress model's activation.
And recently, Hosny \etal~\cite{hosny2021sparse} proposed BitTrain, a novel bitmap compression technique using activation sparsity to reduce the memory footprint during training.

\begin{figure}[t]
	\centering 
	\subfloat[By modifying the operator's execution order, the gradient of each layer is discarded immediately after the weight is updated.]{\label{fig:OPreorder_a}
		\includegraphics[width=0.8\linewidth]{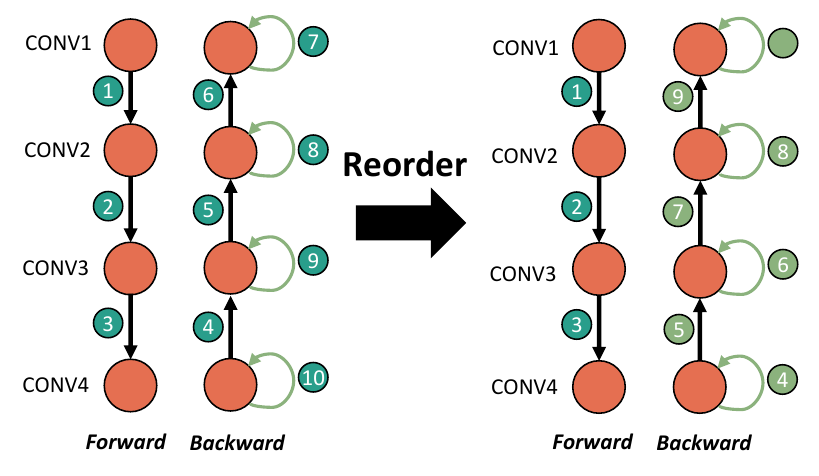}}
		\\
	\subfloat[Reording operator can also decrease the number of computation-intensive operations.]{\label{fig:OPreorder_b}
		\includegraphics[width=0.8\linewidth]{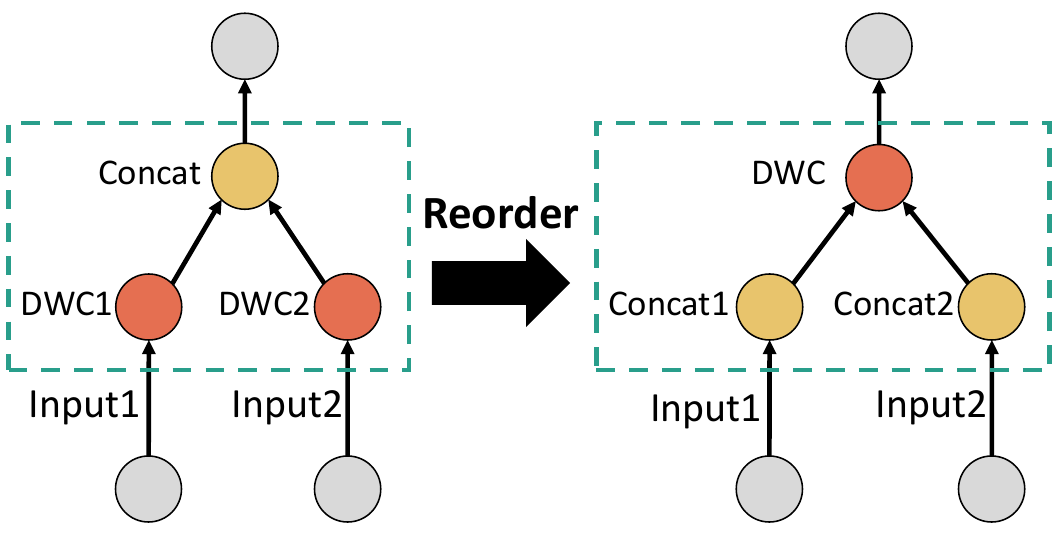}}
	\caption{Illustration of operator reordering technique. }
    \label{fig:OPreorder}
\end{figure}

\textbf{c. Operator reordering}.
The traditional DL training process retains every gradient in the backpropagation and updates weights uniformly after calculating all gradients.
Re-ordering operators within the computation graph during  backpropagation can reduce peak memory usage~\cite{lin2022device}.
Lin \etal~\cite{lin2022device} proposed the Tiny Training Engine (TTE) to reorder operators in DL training.
\rev{As shown in figure \ref{fig:OPreorder_a}, by modifying the order of operator execution(exchanging gradient computation and updating operations) in back-propagation, TTE discards the $i^{th}$ gradient immediately after updating the $i^{th}$ layer instead of keeping it throughout the whole training iteration to achieve memory saving. }
Immediately freeing up space occupied by useless tensors benefits memory savings.
Besides, Unity~\cite{unger2022unity} not only changes the operator order but also reduces the number of computation-intensive operations.
As shown in figure \ref{fig:OPreorder_b}, Unity exchanges the order of two parallel \textit{Depthwise Conv} (DWC) operators and a \textit{concat} operator to reduce the computation of the convolution process by reducing one DWC operator.

\textbf{d. Operator fusion}.
Operator fusion in DL training
focuses on  merging multiple operators into one to boost computation and memory access efficiency.
Suppose two adjacent layers ($i^{th}$ and ${i+1}^{th}$ layer) in the computational graph are merged, the intermediate activation generated by the $i^{th}$ layer can be directly applied to the ${i+1}^{th}$ layer without additional read and write transactions with the memory.
As a separate note, the \textit{computation graph optimization} that \textit{operation fusion} techniques belong to is a broad field. 
Various operator fusion techniques have been well explored in existing research~\cite{tensorflowxla,roesch2019relay} and DL frameworks, \eg TensorFlow~\cite{abadi2016tensorflow}, PyTorch\cite{paszke2019pytorch}, NCNN~\cite{ncnn}, and MindSpore~\cite{mindspore}.
Since fixed rule-based operator fusion cannot guarantee that the performance (\eg latency) is always optimal~\cite{abadi2016tensorflow, chen2018tvm, paszke2017automatic}, MetaFlow~\cite{jia2019optimizing} realized relaxed graph substitutions for operator fusion via a backtracking search algorithm to address this issue.
Zhou \etal~\cite{zhou2020transferable} proposed a transferable deep reinforcement learning-based optimizer to search for optimization policies to improve the optimization efficiency further. 
It speeds up the search drastically by making decisions based on the entire graph rather than on each node individually compared to previous techniques such as Hierarchical Device Placement(HDP)~\cite{mirhoseini2018hierarchical}, Spotlight~\cite{gao2018spotlight}, and Placeto~\cite{addanki2019placeto}.

The operator fusion for computation graph optimization can also derive novel DL \textit{compilers} or \textit{frameworks}.
For example, Zheng \etal~\cite{zheng2020fusionstitching} introduced a DL training compiler.
It merges memory-intensive operators with data dependencies and non-homogeneous parallelism into large GPU kernels, reducing global memory access and context switch overhead.
Hu \etal~\cite{hu2020jittor} proposed Jittor, a just-in-time (JIT) compiled DL framework. 
It integrated with enhanced operator fusion rules which accounts for memory access constraints as a unified function, allowing for unified management of the GPU memory.
%
% It also widens the range of operation combinations that fusion can target and reduces memory access costs. 
%
Ivanov \etal~\cite{ivanov2021data} reduced data movement for the Transformer by constructing a dataflow graph, identifying opportunities for data reuse and applying tailored fusion rules. 
Zheng \etal~\cite{zheng2022astitch} proposed AStitch, a compiler that explores the new operator-stitching optimization space for memory-intensive computations.

\textbf{e. Memory allocation}.
Memory allocation methods are typically implemented in the compiler backend, which is closer to hardware resources than the aforementioned techniques for optimizing computation graphs.
In DL training, the time between the creation and final access of a tensor is referred to as the tensor's "life cycle."
Existing DL frameworks divide a separate memory space for each tensor in the training time, much larger than most tensors' life cycles. 
Actually, it is useless to keep memory space for the tensor outside its life cycle~\cite{moon2022nntrainer}.
Recycling memory for tensors by allocating the same memory for two tensors whose life cycles do not intersect can save memory usage, as shown in Figure ~\ref{fig:train_alloc}. 
Moon \etal~\cite{moon2022nntrainer} analyzed all tensors (including input data, weights, and intermediate activations) that may appear during the training time and realized memory recycling, greatly reducing memory usage.
Wang \etal~\cite{wang2022melon} identified the memory allocation optimization problem is a 2DSP-like NP-hard problem~\cite{bortfeldt2006genetic}. 
Since tensors' life cycles can significantly impact layout effectiveness, they positioned long-lifecycle tensors beneath short-lifecycle ones to approximate the ideal solution to this problem.
Nie \etal~\cite{nie2022tsplit} presented TSPLIT, a fine-grained model memory allocation method that overcomes memory constraints while retaining DL training efficiency. 
With the tensor-splitting primitive, TSPLIT breaks the operation boundary of a tensor, allowing flexible memory allocation.

\begin{figure}[t]
	\centering 
	\subfloat[Memory swapping process.]{\label{fig:swap_a}
		\includegraphics[width=0.7\linewidth]{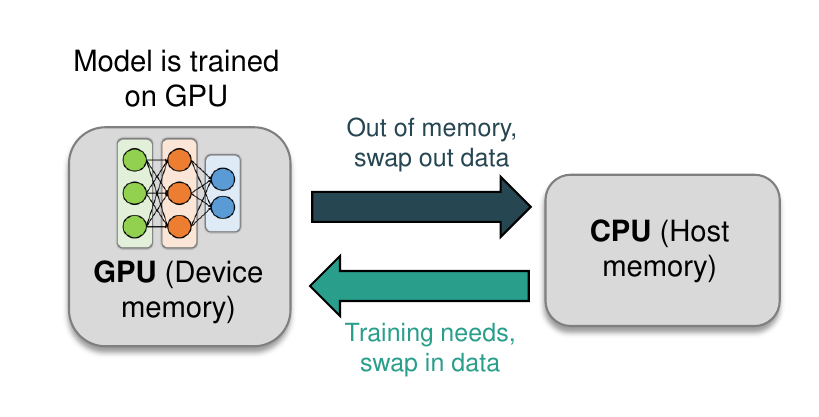}}
		\\
	\subfloat[Memory swapping delay.]{\label{fig:swap_b}
		\includegraphics[width=0.7\linewidth]{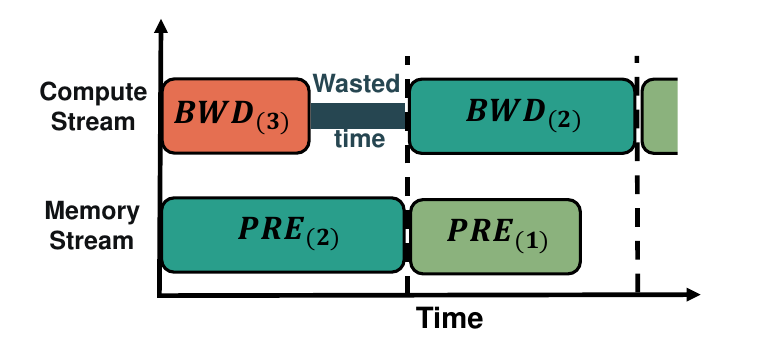}}
	\caption{Illustration of the memory swapping technique. $BWD_{\left( n \right)}$ and $PRE_{\left( n \right)}$ are the backward and prefetch computations for $layer_{\left( n \right)}$, respectively.}
    \vspace{-3mm}
	\label{fig:swap}
\end{figure}

\begin{table*}[]
\centering
\caption{Summary of related intra-device controllers for on-device DL training.}
\resizebox{\textwidth}{!}{%
\renewcommand{\arraystretch}{1.3}
\begin{tabular}{|c|c|c|c|c|c|c|}
\hline
\textbf{Category} & \textbf{Focus level} & \textbf{Context awareness} & \textbf{Controller} & \textbf{Optimized   performance} & \textbf{Year} & \textbf{Ref.} \\ \hline
\multirow{19}{*}{\textbf{Controller level}} & \begin{tabular}[c]{@{}c@{}}Resource-friendly algorithm level\\  including quantization\end{tabular} & Gradient,   energy, memory & / & Layer-wise   quantified precision & 2020 & ~\cite{huang2020adaptive} \\ \cline{2-7} 
 & \begin{tabular}[c]{@{}c@{}}Model-adaptive system \\ scheduling level   including\\  checkpoints in computation graph\end{tabular} & Memory   budget & Dynamic   programming & Total   computational cost & 2016 & ~\cite{gruslys2016memory} \\ \cline{2-7} 
 & \begin{tabular}[c]{@{}c@{}}Model-adaptive system \\ scheduling level   including\\  checkpoints in computation graph\end{tabular} & Tensor   staleness, memory budget & Greedy, heuristics & Total   computational cost & 2020 & ~\cite{kirisame2020dynamic} \\ \cline{2-7} 
 & \begin{tabular}[c]{@{}c@{}}Model-adaptive system\\  scheduling level   including\\  memory swapping\end{tabular} & Memory   budget & Genetic   algorithm & Execution   time & 2020 & ~\cite{huang2020swapadvisor} \\ \cline{2-7} 
 & \begin{tabular}[c]{@{}c@{}}Model-adaptive system\\  scheduling level   including\\  memory swapping\end{tabular} & GPU   architecture & Bayesian   optimization & (De)Compression   time & 2021 & ~\cite{chen2021cswap} \\ \cline{2-7} 
 & \begin{tabular}[c]{@{}c@{}}Co-design: resource-friendly\\  algorithm level   including \\  quantization, sparse  updating \\ $\&$ model-adaptive\\  system scheduling level\\  including   operator reordering\end{tabular} & Memory   budget & Evolutionary   search & Accuracy & 2022 & ~\cite{lin2022device} \\ \hline
\end{tabular}%
}
\label{tab:single_train_controller}
\end{table*}

\textbf{f. Memory swapping}.
In DL training, the intermediate activation of a certain model layer is \textit{temporarily used} in the layer's forward and backpropagation.
Swapping temporarily unused intermediate activations from the precious memory (\eg GPU) to the larger memory (\eg CPU) and then swapping back when gradient updating needs them can speed up computation and reduce memory usage, as shown in Figure ~\ref{fig:swap_a}.
Memory swapping is commonly supported by \textit{operating systems}, \eg virtual address spaces in Windows~\cite{MSmemoryswapping} and swaps partitions in Linux~\cite{Linuxmemoryswapping}.
When the memory capacity of a device is insufficient, data will temporarily swap out to external memory like disks. 
However, it brings extra transfer delay. 
And memory usage is difficult to predict for random programs.
Notably, this shortcoming can be well solved in DL training because the forward computation and backpropagation are in fixed orders (\ie forward computation from the first layer to the last layer and backpropagation is contrary), bringing optimization potential to memory swapping.
\rev{
For example, the data $d^{2}$ required by $layer_2$ backward propogation $BWD_{\left( 2 \right)}$ can be prefetched to GPU memory during $BWD_{\left( 2 \right)}$ computation. 
As a result, the $d^{2}$ transfer delay $PRE_{\left( 2 \right)}$ can be partially covered by $BWD_{\left( 3 \right)}$, or completely coverd like $PRE_{\left( 3 \right)}$, as shown in figure ~\ref{fig:swap_b}.
}

Rhu \etal~\cite{rhu2016vdnn} proposed vDNN, the first work on memory swapping for DL training. 
They propose a runtime memory manager that virtualizes the memory utilization of models by enabling the concurrent usage of both GPU and CPU memory during training. 
vDNN reduces the average GPU memory usage of multiple mainstream models by over 90$\%$. 
Following vDNN, Chen \etal~\cite{chen2018modnn} proposed moDNN, an intelligent solution capable of dynamically choosing convolution operators, adjusting mini-batch size, and selecting the suitable memory-swapping strategy to achieve optimal system performance.
It realized fast Fourier transforms and Winograd with improved performance and increased memory requirements.
Subsequently, Wang \etal~\cite{wang2018superneurons} combined the memory swapping and recomputation methods and realized dynamic GPU memory scheduling that enabled DL training far beyond the GPU DRAM capacity.
Le \etal~\cite{le2019automatic} formally rewritten the computation graph and inserted swap-out and swap-in operations to store intermediate results on CPU memory.
By introducing a categorized topological ordering to simulate computation graph execution, the memory consumption of a model can be profiled using operation distances in the ordering.
Shriram \etal~\cite{shriram2019dynamic} proposed vDNN++ t   o address the issues of delayed computation start, high pinned memory requirements, and GPU memory fragmentation in vDNN~\cite{rhu2016vdnn} by lowering the number of synchronization operations.
Huang \etal~\cite{huang2020swapadvisor} proposed SwapAdvisor to enhance the previously presented memory swapping methods with manual judgment.
It jointly optimizes three dimensions of given dataflow graphs, \ie operator scheduling, memory allocation, and swap decisions.
SwapAdvisor explores the wide search space through a genetic algorithm, improving the GPU's capability to accommodate large models, 
\eg WideResNet-152~\cite{zagoruyko2016wide}, NasNet-25~\cite{zoph2018learning}, and BRNN-4-8K~\cite{schuster1997bidirectional}, they need 180GB, 193GB, and 99GB memory space for training, respectively. 
Following it, Peng \etal~\cite{peng2020capuchin} conducted flexible memory management control by dynamically tracking the tensor access patterns at runtime, which decide when and how to swap memory.

Furthermore, Ren \etal~\cite{ren2021sentinel} presented Sentinel, a runtime system that automatically optimizes tensor management on heterogeneous memory.
Specifically, it enables the optimal memory \textit{co-allocation} for several tensors with similar lifetime and memory access frequencies by allocating them to the same pages to avoid unnecessary data movement.
Then Chen \etal~\cite{chen2021cswap} optimize memory swapping time by selectively compressing tensors based on their sparsity, and size.
Li \etal~\cite{li2022application} automatically trace the memory behaviors of model workloads to schedule memory swapping without perceiving high-level information of layer structures or computation graphs, which alleviate the problem of existing solutions being densely coupled with the fixed model workloads, \eg layer structures or computational graphs. 
Jaehoon \etal~\cite{deepum2023jaehoon} presented DeepUM, offering GPU memory oversubscription for models by leveraging \rev{compute unified device architecture} (CUDA) unified memory and employing CPU memory as a backing store.
They used a new correlation prefetching mechanism at UM block level to hide the fault-handling overhead. 
Thus it considerably reduces memory swapping time and increases GPU's virtual memory capacity.

\subsubsection{Intra-device cross-level controller}
The resource demands of DL training tasks vary. 
This variability, combined with the heterogeneous computing/memory resources of AIoT devices, can provide numerous optimization possibilities.
For instance, when a device has tight memory and rich computing resources, it may be necessary to frequently employ recomputation techniques, while for a device with sufficient memory, such strategies can be avoided to reduce unnecessary computation delay. 
Furthermore, different training epochs on the same device can result in diverse performance outcomes. Therefore, an adaptive controller should be employed in the resource-efficient AIoT system to adjust the technique configuration based on the availability of resources and performance demands. 
A context-aware controller can prevent memory units from being idle or overloaded, thereby enhancing efficiency.

Table ~\ref{tab:single_train_controller} summarizes existing adaptive controllers integrated with diverse levels of optimization techniques.
For example, Huang \etal~\cite{huang2020adaptive} proposed adaptive precision training in DL training quantization.
It dynamically allocated layer-wise precision and provided an application-specific hyperparameter for users to achieve the trade-off between training energy cost, memory usage, and accuracy. 
In sparse updating, Lin~\etal~\cite{lin2022device} adopted an evolutionary algorithm to search for the most important layers for updating and achieve higher training accuracy under a limited memory budget.
In recomputation, Gruslys \etal~\cite{gruslys2016memory} utilized dynamic programming to balance intermediate activations and recomputation caching. 
With a tunable memory budget, it can optimize computation costs. 
Kirisame \etal~\cite{kirisame2020dynamic} extended recomputation methods by introducing dynamic tensor rematerialization (DTR).
DTR is a greedy online algorithm that is parameterized by eviction policy.
In memory swapping, Huang \etal~\cite{huang2020swapadvisor} proposed SwapAdvisor, which uses a genetic algorithm to search for the optimal memory swapping scheme for reducing the transfer delay between CPU and GPU memory. 
Given the memory budgets, they used a dataflow engine simulator to quickly estimate the execution time of each scheme and find the best one with the lowest latency.
Chen \etal~\cite{chen2021cswap} proposed a self-tuning tensor compression framework CSWAP. 
It used bayesian optimization to search for the optimal hyperparameters for tensor compression and tackle the heterogeneity caused by different GPU device architectures and DL frameworks.

\subsection{Distributed DL Training}
\label{subsec:train_m}

\begin{figure*}[t]
  \centering
  \includegraphics[width=.85\textwidth]{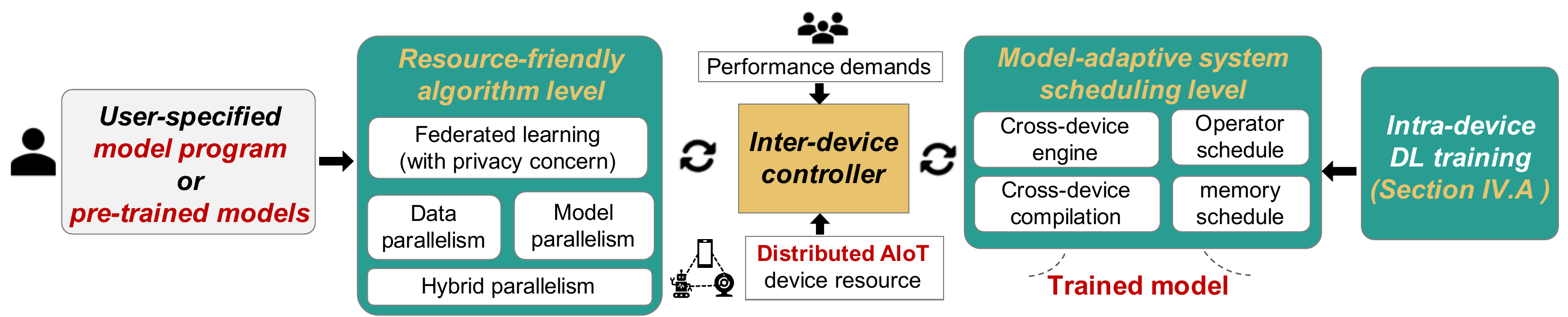}
  \caption{System loop of the algorithm, system scheduling, and inter-device controller for distributed DL training.}
  \label{fig:train_multiple}
  % \vspace{-3mm}
\end{figure*}

With the increase in DL model complexity and diversity, training models on AIoT devices at low cost is still challenging.
To realize this intractable goal, distributed DL training on multiple AIoT devices is considered as the promising way~\cite{narayanan2021memory,samikwa2022ares,liu2023map}.
This includes centralized systems such as model ensembling~\cite{mcdonald2009efficient}, decentralized systems such as tree-like topology~\cite{agarwal2014reliable} and parameter server~\cite{wei2015managed}, and fully distributed systems such as peer-to-peer~\cite{li2013parameter}.
Distributed DL training (with/without privacy concerns) partitions the data or DL models, reducing the memory load on each device compared to on-device training (as discussed in $\S$ \ref{subsec:train_1}).

\begin{figure*}[t]
  \centering
    %\hspace{15pt}
    \subfloat[Distributed DL training (without privacy concern)]{\label{fig:parallelism_mode_b}
    \includegraphics[height=0.3\textwidth]{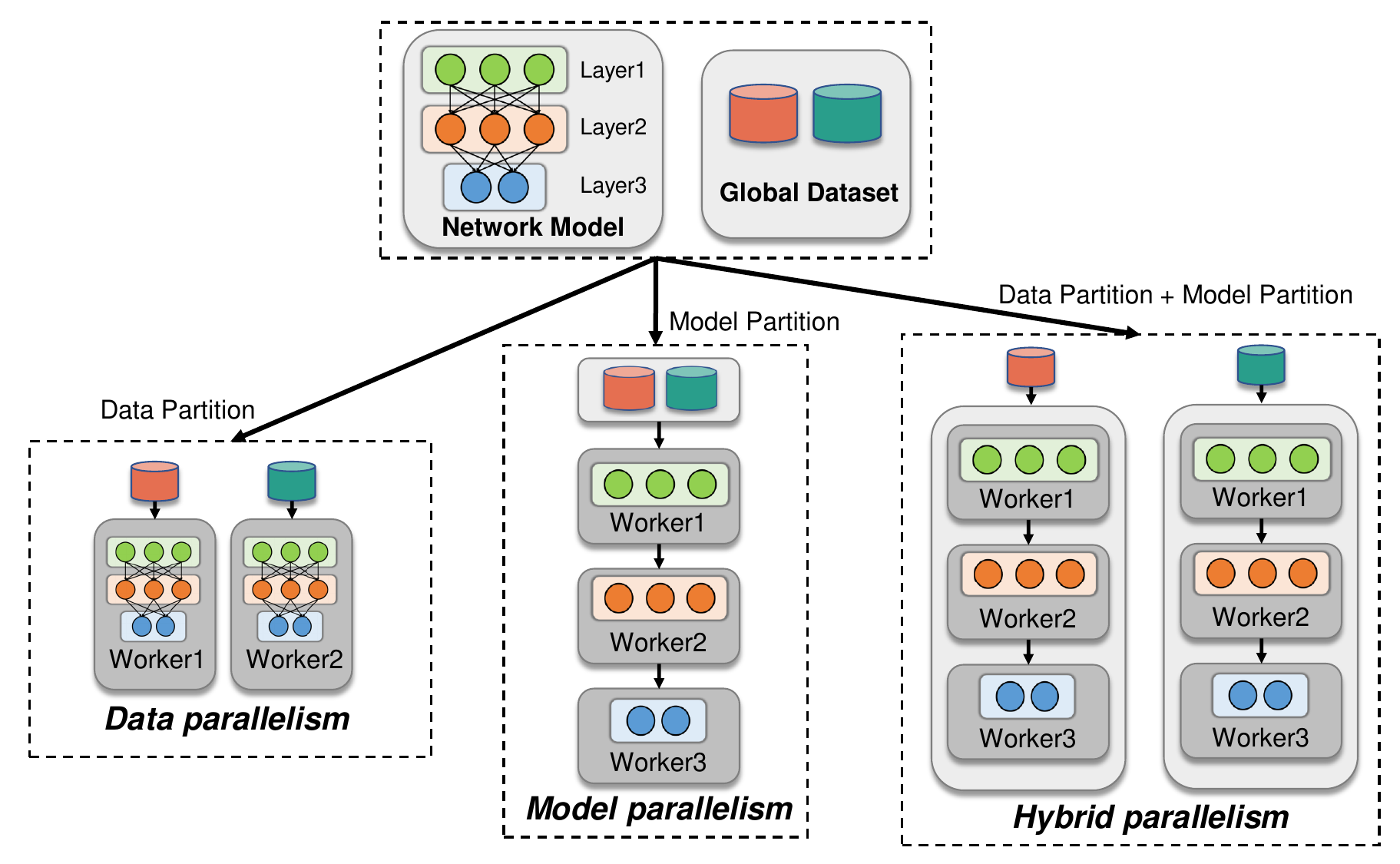}}
    \hspace{10pt}
    \subfloat[Federated learning (with privacy concern)]{\label{fig:parallelism_mode_b}
    \includegraphics[height=0.3\textwidth]{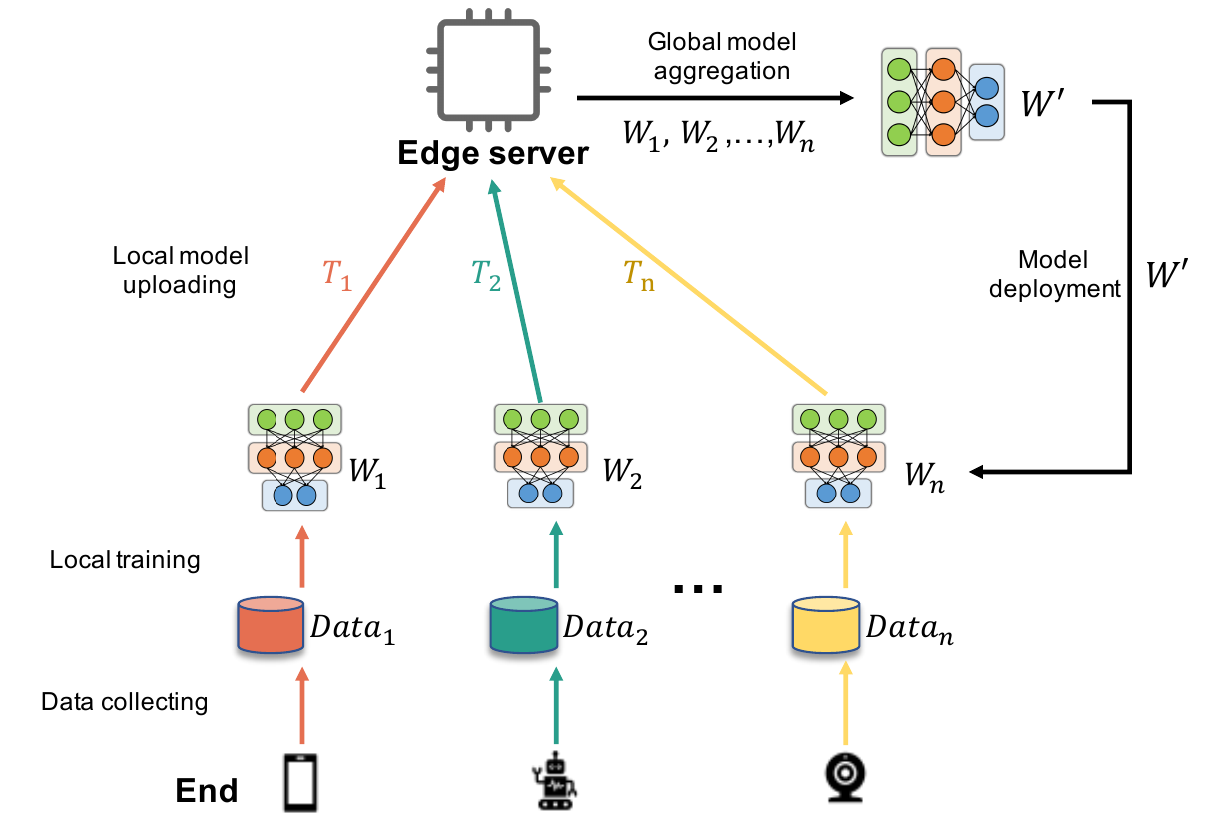}}
\caption{Illustration of different distributed DL training techniques. (a) Distributed learning (without privacy concerns) can be divided into data parallelism, model parallelism, and hybrid parallelism; (b) Considering different time points of model updating, federated learning includes asynchronous, semi-synchronous, and synchronous schemes.
}
\label{fig:parallelism_mode}
%\vspace{-3mm}
\end{figure*}

Existing effort on distributed DL training (without privacy concern) mainly includes three categories, \ie model parallelism, data parallelism, and hybrid parallelism.
Figure \ref{fig:parallelism_mode}a shows their difference.
When the DL model is too large to feed into a single device, it is necessary to adopt \textit{model parallelism} to assign different parts of the model to diverse AIoT devices for training and finally merge them into a complete model. 
\textit{Data parallelism} divides huge data into $N$ parts to deploy $N$ distributed AIoT devices without privacy concerns. 
Each AIoT device only processes $1/N$ of data and aggregates gradients to the central server for an update. 
\textit{Hybrid parallelism} combines the strengths of the above two schemes. 
Federated learning is a widely known data parallelism method with privacy concerns.
And in AIoT federated learning, the data on each AIoT device is sensed by itself and cannot be re-distributed.
It only shares the model updates among multiple devices. 
Federated learning includes asynchronous, semi-synchronous, and synchronous schemes, as shown in Figure \ref{fig:parallelism_mode}b.
In this section, from the cross-level perspective of a resource-efficient AIoT system, \rev{we introduce enabling techniques at different levels, \ie resource-friendly algorithm, model-adaptive system scheduling, and inter-device controller upon them, illustrated in Figure~\ref{fig:train_multiple}}.

\begin{table*}[]
\centering
\scriptsize
\caption{Summary of cross-level optimization techniques for distributed DL training. }
\resizebox{\textwidth}{!}{%
\renewcommand{\arraystretch}{1.3}
\begin{tabular}{|cc|c|c|c|ccc|}
\hline
\multicolumn{2}{|c|}{\multirow{2}{*}{\textbf{Category}}} & \multirow{2}{*}{\textbf{Highlight technique   for improving resource efficiency}} & \multirow{2}{*}{\textbf{Year}} & \multirow{2}{*}{\textbf{Ref.}} & \multicolumn{3}{c|}{\textbf{Parallelism   mode}} \\ \cline{6-8} 
\multicolumn{2}{|c|}{} &  &  &  & \multicolumn{1}{c|}{\textbf{Data}} & \multicolumn{1}{c|}{\textbf{Model}} & \textbf{Hybrid} \\ \hline
\multicolumn{1}{|c|}{\multirow{6}{*}{\begin{tabular}[c]{@{}c@{}}\textbf{Resource-friendly} \\ \textbf{algorithm level}\end{tabular}}} & \multirow{4}{*}{\begin{tabular}[c]{@{}c@{}}Distributed  \\  DL training  (without\\  privacy concern)\end{tabular}} & Compute   layers in pipeline to reduce latency caused by distributed training & 2019 & ~\cite{huang2019gpipe} & \multicolumn{1}{c|}{} & \multicolumn{1}{c|}{\checkmark} &  \\ \cline{3-8} 
\multicolumn{1}{|c|}{} &  & Pipelining and weight gradient coalescing to   reduce memory usage and latency & 2021 & ~\cite{narayanan2021memory} & \multicolumn{1}{c|}{} & \multicolumn{1}{c|}{} & \checkmark \\ \cline{3-8} 
\multicolumn{1}{|c|}{} &  & \begin{tabular}[c]{@{}c@{}}Dynamic programming based heuristic to estimate   memory requirements \\ more precisely, reduce latency\end{tabular} & 2022 & ~\cite{beaumont2022madpipe} & \multicolumn{1}{c|}{} & \multicolumn{1}{c|}{} & \checkmark \\ \cline{2-8} 
\multicolumn{1}{|c|}{} & \multirow{3}{*}{\begin{tabular}[c]{@{}c@{}}Federated \\ learning (with \\ privacy concern)\end{tabular}} & Asynchronous   federated learning to minimize waiting latency & 2019 & ~\cite{xie2019asynchronous} & \multicolumn{1}{c|}{\checkmark} & \multicolumn{1}{c|}{} &  \\ \cline{3-8} 
\multicolumn{1}{|c|}{} &  & Re-parameterization   to decrease the convergence time & 2020 & ~\cite{li2020federated} & \multicolumn{1}{c|}{\checkmark} & \multicolumn{1}{c|}{} &  \\ \cline{3-8} 
\multicolumn{1}{|c|}{} &  & Similarity-aware federated learning system to   reduce communication overhead & 2021 & ~\cite{ouyang2021clusterfl} & \multicolumn{1}{c|}{\checkmark} & \multicolumn{1}{c|}{} &  \\ \hline
\multicolumn{2}{|c|}{\multirow{10}{*}{\begin{tabular}[c]{@{}c@{}}\textbf{Model-adaptive}\\ \textbf{system scheduling level}\end{tabular}}} & First   GPU-specialized parameter server, reduce latency & 2016 & ~\cite{cui2016geeps} & \multicolumn{1}{c|}{\checkmark} & \multicolumn{1}{c|}{} &  \\ \cline{3-8} 
\multicolumn{2}{|c|}{} & \begin{tabular}[c]{@{}c@{}}Layer swapping and recomputing; out-of-core   training behavior analyzing,\\  reduce GPU memory usage\end{tabular} & 2020 & ~\cite{wahib2020scaling} & \multicolumn{1}{c|}{\checkmark} & \multicolumn{1}{c|}{} &  \\ \cline{3-8} 
\multicolumn{2}{|c|}{} & \begin{tabular}[c]{@{}c@{}}Eliminating memory redundancies with Zero   Redundancy Optimizer, reduce \\ memory usage\end{tabular} & 2020 & ~\cite{rajbhandari2020zero} & \multicolumn{1}{c|}{} & \multicolumn{1}{c|}{} & \checkmark \\ \cline{3-8} 
\multicolumn{2}{|c|}{} & \begin{tabular}[c]{@{}c@{}}Shared memory, minimum vertex-cut graph   partitioning algorithm, \\ delayedupdate algorithms. Reduce latency\end{tabular} & 2021 & ~\cite{md2021distgnn} & \multicolumn{1}{c|}{\checkmark} & \multicolumn{1}{c|}{} &  \\ \cline{3-8} 
\multicolumn{2}{|c|}{} & Sharing global memory among concurrent jobs,   reduce training time & 2021 & ~\cite{lim2021zico} & \multicolumn{1}{c|}{} & \multicolumn{1}{c|}{} & \checkmark \\ \cline{3-8} 
\multicolumn{2}{|c|}{} & \begin{tabular}[c]{@{}c@{}}A symbolic profiler to estimate memory and   computing statistics to improve \\ distributed training efficiency\end{tabular} & 2023 & ~\cite{liu2023map} & \multicolumn{1}{c|}{} & \multicolumn{1}{c|}{} & \checkmark \\ \hline
\end{tabular}%
}
\label{tab:multiple_train_three}
\end{table*}

\subsubsection{Resource-friendly algorithm level}

At the algorithm level, researchers explore partitioning the DL models for training parallelism or scheduling various devices in parallel to improve training efficiency.
Federated learning is one of the widely-used distributed learning schemes. 
To reduce waiting time, Xie \etal\cite{xie2019asynchronous} used asynchronous federated learning technology, adopting an asynchronous collaboration mode of upload-on-the-fly to minimize training time for each round.
Then Ouyang\etal~\cite{ouyang2021clusterfl} proposed a novel federated learning system that can identify intrinsic similarities between data points, leading to higher model accuracy and lower communication overhead. 
In distributed learning without privacy concerns, Huang \etal~\cite{huang2019gpipe} proposed GPipe, a scalable model parallelism library to boost the training efficiency of giant models, \eg \rev{generative pre-trained transformer (GPT) and bidirectional encoder representations from transformers (BERT)}. 
GPipe partitioned the model across different accelerators and split the mini-batch into smaller micro-batches for parallelism execution.
By splitting batch, GPipe provided an almost linear speed up, \ie $3.5\sim 20\times$, with no alterations to the model parameters.
PipeDream-2BW~\cite{narayanan2021memory} realized the memory-efficient DL training parallelism by utilizing a new weight gradient coalescing algorithm and weight double buffering.
It efficiently split the DL model across available hardware resources considering hardware limits such as accelerator memory capacity and interconnect topologies. 
Beaumont \etal~\cite{beaumont2022madpipe} proposed MadPipe to dramatically improve the training throughput by hybrid parallelism.
It gives a more precise estimation of the memory requirements and a dynamic programming-based heuristic algorithm, which results in efficient computation-memory allocation and schedule.

\subsubsection{Model-adaptive system scheduling level}
It is intractable to enable the scalability and adaptivity of the distributed DL training to fully use varying resources in AIoT devices (\eg drone swarms and smart camera arrays) whose resources dynamically change due to the influence of other on-device tasks.
Cui \etal~\cite{cui2016geeps} proposed GeePS, the first GPU-specialized \textit{parameter server} design for realizing the data-parallel DL training on GPUs.
Compared to previous CPU-based parameter server systems, GeePS employs GPU-friendly caching, data staging, and memory management techniques to reallocate GPU memory for intermediate layer state cache, significantly reducing training latency.
Wahib~\cite{wahib2020scaling} proposed a performance model based on the concurrent analysis of out-of-core training behavior and combined layer-wise memory swapping and redundant recomputing to decrease memory usage during training.
Regarding scalability, the proposed out-of-core data parallel method outperforms complicated hybrid model parallelism in training huge models, \eg Megatron-LM~\cite{shoeybi2019megatron} and Turning-NLG~\cite{rosset2020turing}.
Rajbhandari \etal~\cite{rajbhandari2020zero} proposed a zero redundancy optimizer (ZeRO) to optimize memory usage and improve distributed training speed while increasing the model size.
ZeRO mainly employs two techniques, \ie reducing memory state redundancy across data-parallel processes by splitting the model states rather than repeating them; and proactively allocating memory based on the lifetime of various tensors to prevent memory fragmentation.

Md \etal~\cite{md2021distgnn} proposed DistGNN, which uses a minimum vertex-cut graph partitioning algorithm to reduce the transfer time of features and gradients. 
DistGNN further accelerates the shared memory system by using cache blocking, loop reordering, and vectorization with the LIBXSMM library~\cite{heinecke2016libxsmm}, which considerably boosts distributed training throughput.
Then Lim \etal~\cite{lim2021zico} proposed Zico, the first model system which aims to reduce the memory consumption for parallelism training. 
Zico monitors the memory usage of each training task through its progress on GPU computation and reclaims the memory that is no longer needed, making it globally sharable. 
As a general approach, Zico outperforms existing GPU-sharing methods.
Recently, Liu \etal~\cite{liu2023map} presented MAP, a PyTorch-based compiler that implements memory-aware automated parallelization and provides near-real-time memory and computing statistics. 
MAP can build distributed execution plans with high training efficiency.
Because it involves a symbolic profiler to swiftly collect memory and computation overhead, a cluster detector to gather cluster hardware performance and topology, and a tense layout manager.

\begin{table*}[t]
\centering
\scriptsize
\caption{Summary of related inter-device controllers for distributed DL training.}
\resizebox{\textwidth}{!}{%
\renewcommand{\arraystretch}{1.3}
\begin{tabular}{|cc|c|c|c|c|c|c|}
\hline
\multicolumn{2}{|c|}{\textbf{Category}} & \textbf{Focus level} & \textbf{Context awareness} & \textbf{Controller} & \textbf{\begin{tabular}[c]{@{}c@{}}Optimizated \\   performance\end{tabular}} & \textbf{Year} & \textbf{Ref.} \\ \hline
\multicolumn{1}{|c|}{\multirow{13}{*}{\begin{tabular}[c]{@{}c@{}}\textbf{Inter-device} \\ \textbf{controller level}\end{tabular}}} & \multirow{3}{*}{\begin{tabular}[c]{@{}c@{}}Federated learning (with\\  privacy concern)\end{tabular}} & \begin{tabular}[c]{@{}c@{}}Resource-friendly algorithm level including \\ data distribution and device training latency\end{tabular} & \begin{tabular}[c]{@{}c@{}}Heterogeneous   device \\ resources, sample \\ importance\end{tabular} & Greedy   heuristic & Training   time & 2021 & ~\cite{lai2021oort} \\ \cline{3-8} 
\multicolumn{1}{|c|}{} &  & \begin{tabular}[c]{@{}c@{}}Resource-friendly algorithm \\ level including device conditions\end{tabular} & \begin{tabular}[c]{@{}c@{}}Network   bandwidth,\\  heterogeneous devices\end{tabular} & Synchronization   scheduler & \begin{tabular}[c]{@{}c@{}}Energy   consumption, \\ training time, accuracy\end{tabular} & 2022 & ~\cite{sun2022fedsea} \\ \cline{2-8} 
\multicolumn{1}{|c|}{} & \multirow{10}{*}{\begin{tabular}[c]{@{}c@{}}Distributed learning \\ (without privacy concern)\end{tabular}} & \begin{tabular}[c]{@{}c@{}}Resource-friendly algorithm \\ level including device conditions\end{tabular} & \begin{tabular}[c]{@{}c@{}}Network   throughput,\\  computing resources\end{tabular} & / & \begin{tabular}[c]{@{}c@{}}Energy   consumption, \\ training time\end{tabular} & 2022 & ~\cite{samikwa2022ares} \\ \cline{3-8} 
\multicolumn{1}{|c|}{} &  & \begin{tabular}[c]{@{}c@{}}Resource-friendly algorithm level including memory\\  estimation and   non-contiguous allocations of DL layers\end{tabular} & \begin{tabular}[c]{@{}c@{}}Memory   budget, \\ network bandwidth\end{tabular} & Dynamic   programming & Execution   time & 2022 & ~\cite{beaumont2022madpipe} \\ \cline{3-8} 
\multicolumn{1}{|c|}{} &  & \begin{tabular}[c]{@{}c@{}}Model-adaptive system scheduling including\\ checkpoints in computation graph\end{tabular} & Memory   budget & \begin{tabular}[c]{@{}c@{}}Markov  Chain Monte\\  Carlo search algorithm\end{tabular} & Execution   time & 2020 & ~\cite{band2020memflow} \\ \cline{3-8} 
\multicolumn{1}{|c|}{} &  & \begin{tabular}[c]{@{}c@{}}Model-adaptive system scheduling including \\ checkpoints in computation graph\end{tabular} & Memory   budget & Dynamic   programming & \begin{tabular}[c]{@{}c@{}}The   number of\\  recomputation\end{tabular} & 2020 & ~\cite{beaumont2020optimal} \\ \cline{3-8} 
\multicolumn{1}{|c|}{} &  & \begin{tabular}[c]{@{}c@{}}Model-adaptive system scheduling including \\ model   statistics collection and \\ computation graph static planning\end{tabular} & Memory   budget & / & Execution   time & 2023 & ~\cite{liu2023map} \\ \hline
\end{tabular}
}
\label{tab:multiple_train_controller}
\end{table*}

\subsubsection{Inter-device cross-level controller}
\label{subsec:train_platform}
Like on-device DL training, distributed DL training also benefits from the context-aware controller.
On the one hand, imbalances in completion time among different devices result in significant synchronization overheads.
On the other hand, DL training parallelism depends on the degree of the decoupled model layer, channel, or operator.
Table ~\ref{tab:multiple_train_controller} summarizes some prior explorations.
In federated learning, the disparate computational capabilities of the devices involved can result in differences in training time, which can impact the overall system efficiency. 
Lai \etal\cite{lai2021oort} introduced Oort, a method that enhances the performance of federated training and testing through guided participant selection.
Oort employs a synchronous federated learning scheme, selecting participants with similar device performance for training in each round to mitigate the impact of hardware differences among devices.
To further reduce the impact of heterogeneous devices, Sun \etal\cite{sun2022fedsea} proposed FedSEA, a semi-asynchronous FL framework for extremely heterogeneous devices. FedSEA adjusts training parameters for device heterogeneity and balances the trade-off between waiting costs and data benefits.
MemFlow~\cite{band2020memflow} jointly optimizes memory usage and computation time when searching for an optimal parallelism training strategy.
In detail, it adopts a Markov chain monte Carlo search algorithm to select the most suitable degrees of recomputation.
Olivier \etal ~\cite{beaumont2020optimal} employed memory-aware scheduling and automatic differentiation to execute a back-propagation graph within the bounded memory requirement at the cost of computation.

Recently, Liu \etal~\cite{liu2023map} proposed MAP to search for the optimal distributed training strategy in two ways, \ie intra-operator parallelism and activation checkpoint.
Also, to automate intra-operator parallel training for large models, MAP adopted an efficient strategy using a two-stage solver and recompiled it into a module instance.
However, the cross-level adaptive controller in a broader space for distributed DL training is still underexplored.
According to the dynamically available resources of different AIoT devices in the distributed system without privacy concerns, the split data and model sizes can be dynamically adjusted for diverse devices.
To better control different devices' training cycles and synchronization strategies, Samikwa \etal~\cite{samikwa2022ares} introduced resource-aware split-learning (ARES) for adaptive DL distributed training.
ARES presented the device-oriented model partition method, which adaptively deploys tasks for heterogeneous devices to reduce the impact of stragglers.
To address insufficient memory issues on individual end devices, Beaumont \etal~\cite{beaumont2022madpipe} uses a dynamic programming-based heuristic to find the best strategy for distributed memory allocation.

\rev{
\textbf{Discussion}.
The cross-layer optimization of DL training is essential for achieving efficient and effective AIoT deployments.
While both DL training and DL inference ($\S$ III) can benefit from similar optimization approaches and automated controllers, training poses additional constraints and challenges due to higher computational and storage requirements and increased sensitivity to errors. 
On one hand, both DL training and DL inference share the same cross-layer optimization space and require automated controllers, whether implemented on-device or in a distributed manner. 
Table \ref{tab:single_train_system} presents a range of individual optimization techniques that can be employed for DL training task optimization in AIoT systems. 
Similarly, the performance of optimization techniques can vary significantly, necessitating careful selection and exploration of their combined usage on AIoT devices.
On the other hand, DL training has higher computational and storage requirements compared to DL inference, along with increased sensitivity to errors due to multiple iterations. 
This implies that DL training imposes more constraints and challenges compared to DL inference.
}

\section{Resource-efficient AIoT Applications}
\label{sec:system}

\begin{table*}[]
\centering
\caption{Comparison of two types of resource-efficient DL engines for AIoT.}
\resizebox{\textwidth}{!}{
\renewcommand{\arraystretch}{1.1}
\scriptsize
\begin{tabular}{|c|c|c|c|c|c|}
\hline
\textbf{Type}               & \textbf{Engine}            & \textbf{DL model parser} & \textbf{DL model interpreter} & \textbf{DL model optimizer} & \textbf{DL model compiler} \\ \hline
\textbf{Compiled engine}    & TVM~\cite{chen2018tvm}               & \checkmark            & \checkmark                 & \checkmark               & \checkmark              \\ \cline{2-6} 
                   & OneDNN~\cite{onednn}            &             & \checkmark                 & \checkmark               & \checkmark              \\ \cline{2-6} 
                   & TensorflowXLA~\cite{tensorflowxla}     & \checkmark            &                  & \checkmark               & \checkmark              \\ \cline{2-6} 
                   & TensorflowRuntime~\cite{tensorflowruntime} & \checkmark            &                  & \checkmark               & \checkmark              \\ \hline
\textbf{Interpreted engine} & TensorflowLite~\cite{tensorflowlite}    & \checkmark            & \checkmark                 & \checkmark               &               \\ \cline{2-6} 
                   & CMix-NN~\cite{capotondi2020cmix}           &             & \checkmark                 & \checkmark               &               \\ \cline{2-6} 
                   & CMSIS-NN~\cite{lai2018cmsis}          & \checkmark            & \checkmark                 & \checkmark               &               \\ \hline
\end{tabular}}
\label{Tab:engine}
\end{table*}

This section briefly introduces some enabling systems and the potential application scenarios in AIoT.

\subsection{DL Engines for AIoT}
\begin{figure}
    \centering
    \includegraphics[width=0.85\linewidth]{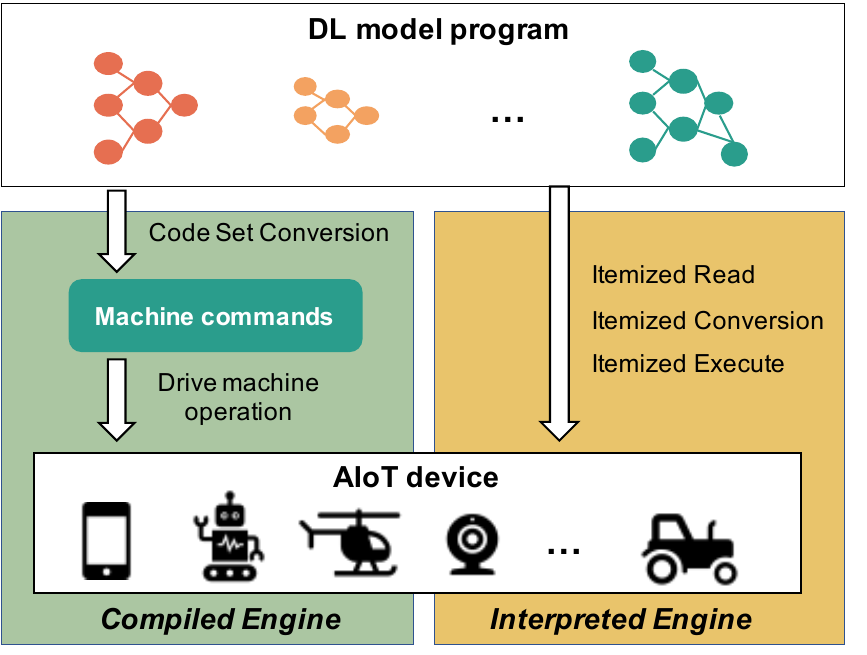}
    \caption{Difference of interpreted and compiled engines.}
    \label{fig:engines}
    \vspace{-3mm}
\end{figure}

The engine for resource-efficient DL deployment at AIoT devices mainly aims at improving the flexibility of memory scheduling, optimizing the computational efficiency of operators, and enhancing the underlying adaptability  to heterogeneous devices.
It creates a paradigm shift in how operators are carried out.
\rev{As shown in Figure \ref{fig:engines}, they include \textit{interpreted} and \textit{compiled engines},We demonstrated how these two different deep learning engines play a role in deploying models to devices.}

\textbf{\textit{Interpreted engines}} generally include the DL model parser, interpreter, and optimizer.
The DL model parser is responsible for reading and parsing the model file and converting it into a format suitable for processing by the interpreter.
The DL model optimizer is responsible for transforming the original DL model into an equivalent model with more efficient inference/training speed;
The DL model interpreter accepts the input data from the application scenarios and executes the corresponding internal operators of the DNN in sequence according to the model architecture and target device's hardware schedule configurations to finally output results.

\textbf{\textit{Compiled engines}} mainly involve the DL model parser and compiler.
The role of the model parser is the same as that of interpreted engines.
The compiler transforms models into machine code that can be directly processed by the target deployment platform (such as CPU, GPU, \etc
Also, it can apply various optimization methods during the compilation process to improve the operating efficiency of machine codes, such as automatic computing graph scheduling~\cite{chen2018tvm}.

Table \ref{Tab:engine} summarizes the state-of-the-art DL deployment engines on AIoT devices.
Recent representative compiled inference engines mainly include TVM~\cite{chen2018tvm}, oneDNN~\cite{onednn}, TensorFlow XLA~\cite{tensorflowxla}, TensorFlow Runtime~\cite{tensorflowruntime}, \etal.
Different engines focus on different aspects.
For example, TVM~\cite{chen2018tvm} is a cross-platform DL development framework. 
Compared with commonly used frameworks such as Tensorflow and Pytorch, TVM optimizes DNN operators at the computation graph level by converting operators into descriptions of tensor changes, generating code, and transmitting it to the CUDA compiler.
TVM does not rely on a specific framework's compute library and can be deployed on diverse hardware platforms.
Meanwhile, TVM supports integrating new operators.
Unlike the cross-platform characteristics of TVM, OneDNN~\cite{onednn} emphasizes on DL performance optimization with Intel-architecture processors, Intel-architecture graphics cards, and Xe-architecture graphics cards.
The oneDNN engine executes DNN primitives to process data in several memory objects, reducing memory usage of DL inference. 
The primitives are objects encapsulating specific computations, \eg forward convolution, data transformation, \etc 
Compared to purely functional operations (\eg conv), primitives can be specialized for a subset of input parameters. 
Both TVM and OneDNN are compatible with various DL frameworks.

Some other engines are deeply optimized for one framework, such as Tensorflow XLA~\cite{tensorflowxla} and Tensorflow Runtime~\cite{tensorflowruntime}.
TensorflowXLA~\cite{tensorflowxla} is a linear algebra compiler engine for the TensorFlow framework.
In the regular TensorFlow framework, when a program runs, it invokes a pre-compiled GPU kernel for each operation, causing the computing kernel redundancy problem.
TensorflowXLA solve this problem by compiling the computation graph of each DNN into a series of specially generated computation kernels to speed up DL inference and improve memory usage with model-specific information.
%%%%
Tensorflow Runtime (TFRT)~\cite{tensorflowruntime} mainly optimizes the DL inference for specific hardware in various fields to enable scalability. 
It implements efficient execution of the computing kernel through specific primitives on the underlying devices. 
Meanwhile, it optimizes the parallel operation of the existing graph and reduces the synchronization overhead. 
Also, TFRT provides a lightweight just-in-time operator distribution stack for asynchronous API calls to improve computing efficiency.

TFlite\cite{tensorflowlite} is a lightweight engine that is mainly applied to DL inference on mobile devices.
TFlite not only provides a series of core operators according to the requirements of mobile platforms but also supports custom operators. 
And TFlite defines a new DNN file format, removing the parsing step before revisiting  data, and greatly reduces the memory footprint of the code.
In addition, TFlite designed an optimized interpreter that uses static graphic sorting and a custom (less dynamic) memory allocator to ensure minimal load, and execution latency.
CMix-NN \cite{capotondi2020cmix} is a framework specifically supporting DNN inference on MCU.
Compared with other frameworks, CMix-NN focuses on effectively compressing DNNs.
Specifically, CMix-NN supports DNN quantization strategies for diverse DNN layer,  filter channel, and activation.
% It designs a mixed low-precision compression algorithm.
%
% Through the fine-grained DNN compression techniques, DNN model can maintain accuracy while reduce memory consumption so it can be better deployed on MCU.
%
CMSIS-NN\cite{lai2018cmsis} is an edge DNN inference framework for the internet of things (IOT) scenarios.
% , which is an optimization scheme for the ARM architecture. 
% \TODO{frame or engine? please be consistent}
% The engine is mainly composed of two parts, \ie NNFunctions and NNSupport functions. 
% NNFunctions include the operators required to implement DNN, including convolution, deep separable convolution, full join, pooling, activation, \etal. 
% NNSupportFuncions contains different utility functions, including data conversion, \etal.
It can directly interact with the underlying hardware, improving computing efficiency by up to $5\times{}$  in the MCU tests.

\rev{
\textbf{Discussion}.
Building upon the aforementioned techniques, engine systems can assist the algorithm layer in achieving cross-layer optimization of AIoT systems.
Compiled engines offer advantages such as reduced memory consumption and simplified deployment, while interpreted engines excel in adapting to various hardware architectures at runtime.
The choice between them depends on the specific requirements and constraints of the AIoT system.
Compiled engines, with their lower memory footprint and absence of additional graph interpretation, offer improved support for heterogeneous AIoT devices. 
However, their runtime and dynamic adaptation capabilities remain limited.
On the other hand, interpreted engines excel in executing DL inference and training tasks across diverse AIoT devices. 
They possess the ability to interpret the computation graph based on specific hardware instructions during runtime. 
This flexibility allows for efficient utilization of the underlying hardware resources, enabling optimal performance across different AIoT device configurations.
}

\begin{table*}[]
\caption{Summary of resource-efficient AIoT system optimization for enabling diverse resource-aware applications. }
\label{Tab:multi-task}
\resizebox{\textwidth}{!}{
\renewcommand{\arraystretch}{0.75}
\scriptsize
\begin{tabular}{|c|c|c|c|c|c|}
\hline
\multirow{2}{*}{\textbf{Applications}}     & \multirow{2}{*}{\textbf{Technique highlight for improving resource efficiency}} & \multirow{2}{*}{\textbf{Focus level}}      & \multirow{2}{*}{\textbf{Optimization}}            & \multirow{2}{*}{\textbf{Year}} & \multirow{2}{*}{\textbf{Ref.}}                    \\
                                       &                                                                        &                             &                                          &                       &                                          \\ \hline
\multirow{8}{*}{\textbf{Image classification}}  & \multirow{2}{*}{Hyperspectral analysis, less parameters}                    & \multirow{2}{*}{Computation graph}     & \multirow{2}{*}{Improve accuracy}        & \multirow{2}{*}{2020} & \multirow{2}{*}{\cite{paoletti2020flop}}        \\
                                       &                                                                        &                             &                                          &                       &                                          \\ \cline{2-6} 
                                       & \multirow{2}{*}{Quickly reduce the size of the image}                  & \multirow{2}{*}{DL model}  & \multirow{2}{*}{Reduce parameters}       & \multirow{2}{*}{2020} & \multirow{2}{*}{\cite{saha2020rnnpool}}         \\
                                       &                                                                        &                             &                                          &                       &                                          \\ \cline{2-6} 
                                       & \multirow{2}{*}{Reduce early activation}                               & \multirow{2}{*}{Operetor}   & \multirow{2}{*}{Increase inputs}         & \multirow{2}{*}{2021} & \multirow{2}{*}{\cite{oh2021quantum}}           \\
                                       &                                                                        &                             &                                          &                       &                                          \\ \cline{2-6} 
                                       & \multirow{2}{*}{Quantum computing,few memory}                          & \multirow{2}{*}{Inter-device controller} & \multirow{2}{*}{Reduce data storage}     & \multirow{2}{*}{2022} & \multirow{2}{*}{\cite{liang2022distrihd}}       \\
                                       &                                                                        &                             &                                          &                       &                                          \\ \hline
\multirow{8}{*}{\textbf{Semantic segmentation}} & \multirow{2}{*}{Space pyramid}                                         & \multirow{2}{*}{DL model}  & \multirow{2}{*}{Improve computing speed} & \multirow{2}{*}{2019} & \multirow{2}{*}{\cite{emara2019liteseg}}        \\
                                       &                                                                        &                             &                                          &                       &                                          \\ \cline{2-6} 
                                       & \multirow{2}{*}{Asymmetric codec structure}                            & \multirow{2}{*}{Computation graph}     & \multirow{2}{*}{Reduce memory}           & \multirow{2}{*}{2019} & \multirow{2}{*}{\cite{wang2019lednet}}          \\
                                       &                                                                        &                             &                                          &                       &                                          \\ \cline{2-6} 
                                       & \multirow{2}{*}{Trapezoidal up-sampling}                               & \multirow{2}{*}{Operator}  & \multirow{2}{*}{Improve computing speed} & \multirow{2}{*}{2020} & \multirow{2}{*}{\cite{krevso2020efficient}}     \\
                                       &                                                                        &                             &                                          &                       &                                          \\ \cline{2-6} 
                                       & \multirow{2}{*}{Built-in memory module}                         & \multirow{2}{*}{Memory scheduling}  & \multirow{2}{*}{Improve accuracy}        & \multirow{2}{*}{2021} & \multirow{2}{*}{\cite{jin2021memory}}     \\
                                       &                                                                        &                             &                                          &                       &                                          \\ \hline
\multirow{8}{*}{\textbf{Speech recognition}}    & \multirow{2}{*}{End-to-end neural network architecture}                & \multirow{2}{*}{Computation graph}     & \multirow{2}{*}{Reduce parameters}       & \multirow{2}{*}{2019} & \multirow{2}{*}{\cite{shangguan2019optimizing}} \\
                                       &                                                                        &                             &                                          &                       &                                          \\ \cline{2-6} 
                                       & \multirow{2}{*}{Finite-size beam search decoding}                      & \multirow{2}{*}{Computation graph}   & \multirow{2}{*}{Improve accuracy}        & \multirow{2}{*}{2020} & \multirow{2}{*}{\cite{guo2020efficient}}        \\
                                       &                                                                        &                             &                                          &                       &                                          \\ \cline{2-6} 
                                       & \multirow{2}{*}{Low rank matrix substitution}                          & \multirow{2}{*}{DL model}  & \multirow{2}{*}{Reduce parameters}       & \multirow{2}{*}{2020} & \multirow{2}{*}{\cite{winata2020lightweight}}   \\
                                       &                                                                        &                             &                                          &                       &                                          \\ \cline{2-6} 
                                       & \multirow{2}{*}{Streaming oriented speech separation technology}       & \multirow{2}{*}{Inter-device controller} & \multirow{2}{*}{Improve computing speed} & \multirow{2}{*}{2020} & \multirow{2}{*}{\cite{wang2020voicefilter}}     \\
                                       &                                                                        &                             &                                          &                       &                                          \\ \hline
\end{tabular}}
\end{table*}

\subsection{Resource-efficient AIoT system for Diverse Applications}
\label{subsec:sys_application}
The remarkable success of DL has fostered a growing number of intelligent applications/services on AIoT devices~\cite{dosovitskiy2020image,li2019deep,zoph2018learning, weng2021semantic,das2021fundamentals}.
Table \ref{Tab:multi-task} summarizes three typical AIoT applications, \eg image classification, semantic segmentation, and speech recognition.
And we note that the algorithm-system co-design that jointly optimizes the resource-friendly DL models
and model-adaptive resource scheduling can improve the runtime resource availability and thus pushes the limit of performance-resource tradeoff set by standalone levels.
% \TODO{add ref}
%

\subsubsection{Cross-level optimization for image classification}
Image classification has a wide range of applications, including object classification~\cite{zhao2017survey}, human face recognition~\cite{gupta2018deep}, remote-sensing image recognition~\cite{you2019pixel}, image spectrum analysis~\cite{protasov2016fracture} \etc 
%
% In the hyperspectral image analysis field, CNN plays a critical role, whose conv kernels can naturally contain spatial information to smooth the spectral variability and noise existing in the digital image information. 
% However, large CNN models involving huge parameter sizes limit computing efficiency. 
There are many attempts at the algorithm and system scheduling levels.
Qian~\etal~\cite{qian2022rex} proposed the recurrent aggregation operator (ReX) to extract informative information and avoid memory-intensive large-scale early activations.
ReX integrates important features of intermediate activations by using two RNNs and compresses them into a low-dimensional vector, which greatly reduces the memory footprint in the early exit module.

\begin{figure*}
    \centering
    \includegraphics[width=\linewidth]{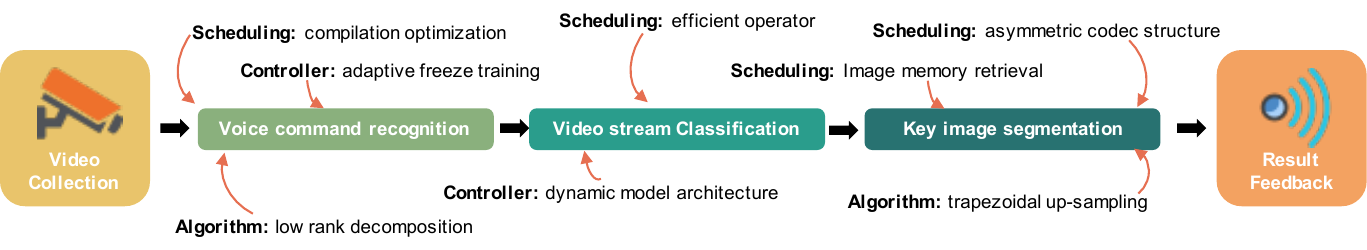}
    \caption{An example of cross-level optimization for resource-efficient video analysis in smart city scenarios.}
    \label{fig:smart-system}
\end{figure*}

\subsubsection{Cross-level optimization for semantic segmentation}
In contrast to image classification, which only requires a single label for the entire image, semantic segmentation requires labels for each pixel.
It is the pixel-level segmentation of different objects.
% in the image through pixel-level classification.
% It requires labels for each pixel
%
% Thus, 
% the bottleneck of semantic segmentation to reduce memory is the amount of computation because each pixel in the image needs to be classified individually, resulting in a large amount of computation. 
Therefore, semantic segmentation is more memory exhaustive than image classification, especially on large images~\cite{sun2018fully}.
To realize on-device semantic segmentation on the AIoT device, 
% Emara ~\etal \cite{emara2019liteseg} proposed a deeper Atrous spatial pyramid pooling model at the algorithm level.
Jin \etal~\cite{jin2021memory} designed an adaptive built-in memory module to decrease memory usage.
We note that cross-level algorithm and system scheduling co-design will be beneficial.
Wang \etal~\cite{wang2019lednet} presents Lednet based on an asymmetric encoding and decoding structure.
And they designed two new operators, \ie channel splitting and shuffling, which reduced the computing cost of the entire network and improved the computing speed.

\subsubsection{Cross-level optimization for speech recognition}
Speech recognition has a broad application area, such as smart home~\cite{stojkoska2017review,alaa2017review} and intelligent driving~\cite{feng2021intelligent,khan2021modelling}.  
% Speech recognition provides basic support for the user's command recognition and subsequent operation control.
%
% In speech recognition, it is necessary to sink computing to AIoT edge/end devices, which can protect users' speech privacy and guarantee application quality under the local area network (LAN).
%
Diverse studies have investigated compressing memory for various aspects of speech recognition applications, including data separation~\cite{yoshioka2018multi}, lightweight models~\cite{wang2020voicefilter,winata2020lightweight}, and system deployment~\cite{pratap2020scaling}. 
For example, Shangguan \etal~\cite{shangguan2019optimizing} 
% proposed an edge speech recognition framework and constructed the end-to-end system, parallel network topology.It 
supported the lightweight conversion of various speech recognition models, such as RNN and LSTM, by integrating different computational graph optimization techniques.
Han \etal \cite{han2017ese} proposed load-balance-aware pruning to ensure high hardware utilization. 
And they design a system scheduler that encodes the compressed model to multiple PEs for parallelism and schedules the complicated LSTM data flow. 

% \eg pruning, and parameter quantization.
% Guo \etal~\cite{guo2020efficient} presented a new algorithm for training the RNN-Transducer model that aims to minimize the system's word error rate (WER). 
% They adopted a variant of the forward-backward algorithm to compute the gradient with respect to the model parameters efficiently.
% And they employ the parallel decoding and training process to reduce memory usage. 
%
% Winata \etal~\cite{winata2020lightweight} proposed LRT for joint optimization of DL inference and training on memory-constrained mobile devices.
% %
% Replacing the high-rank matrix with a low-rank matrix reduces the number of parameters and eliminates the memory bottleneck.
% Also, they adopted the automatic encoder in LRT to compress high-dimensional input and used low-dimensional vector representation.
% 
% VocieFilter-Lite~\cite{wang2020voicefilter} designed an asymmetric loss function for speech separation model training and an adaptive scheduling method for energy and latency reduction.

\subsubsection{AIoT-powered application scenarios}
Above advances in applications have driven increasing solutions in various AIoT scenarios, including smart homes~\cite{liciotti2020sequential,mekruksavanich2021lstm,schneider2019wav2vec,radford2022robust}, smart factories~\cite{liu2020privacy,hoffmann2019model,yi2018deep}, and smart cities~\cite{manogaran2021blockchain,shahzad2020internet,andronie2021artificial}.

\textbf{Smart home}.
Current smart home scenarios involve video/voice recognition and environmental awareness to realize automated appliance operations, \eg curtain controlling and TV turning on/off.
Deeperthings~\cite{stahl2021deeperthings} proposed a collaborative optimization framework at three levels: communication, computing, and memory usage. 
By integrating communication and perception layers to achieve cross-layer overall optimization and balancing memory usage between different devices, the efficient inference is ultimately achieved in resource-constrained situations.
Given the diversity and complexity of scenarios, the resource-efficient AIoT system deployed in open environments should dynamically adjust itself.
Also, cross-level optimization must be jointly explored.

\textbf{Smart city}.
In urban construction and city life, resource-efficient AIoT systems have stimulated plenty of applications, \eg traffic flow prediction~\cite{liu2020privacy}, street map estimation~\cite{hoffmann2019model}, and air quality prediction~\cite{yi2018deep}.
%%%
Specifically, accurate traffic flow prediction serves lane planning~\cite{liebig2017dynamic} and indicator time regulation~\cite{polson2017deep}. 
\rev{As shown in Figure \ref{fig:smart-system}, We used monitoring as an example to analyze how optimization technologies at different levels are applied to the systems of smart cities}
%
% Liu \etal~\cite{liu2020privacy} realized the traffic flow prediction based on federated learning.
% ensuring prediction accuracy and preserving privacy. 
%
% Street view mapping fuses images collected by diverse devices for remote sensing and aerial photography~\cite{hoffmann2019model}. 
% Hoffmann \etal~\cite{hoffmann2019model} solved the influence of spatial dislocation of features through a decision-level fusion. 
%
% Yi \etal \cite{yi2018deep} combined a spatial transformation component for processing sparse "air quality" data and a distributed fusion network for integrating city-level data.
% \rev{To sum up, the development of smart cities is mainly used to promote better resource allocation and urban construction by the government, which can provide a decision-making basis for government work from a macro level}
Summarily, resource-efficient AIoT systems in smart cities usually comprise heterogeneous devices with a large physical span. 
To ensure efficient collaboration, it is necessary to provide unified management of cross-device resources at the algorithm and underlying system levels.

\textbf{Smart industry}.
Resource-efficient AIoT systems are gradually integrated into engineering management and process optimization to in the industrial field.
The specific functionality includes product defect detection, manufacturing process optimization, predictive operation and maintenance, equipment failure warning, production process planning, \etc 
To avoid the fragmented ecosystem of DL models in AIoT industry, Ren \etal ~\cite{ren2022manage} proposed a framework using Semantic Web technologies to enable the joint management of TinyML models and IoT devices at scale, from modeling information to discovering possible combinations and benchmarking, and eventually facilitate TinyML component exchange and reuse.
%
% Manogaran \etal~\cite{manogaran2021blockchain} designed a security data sharing model based on blockchain.
% It is responsible for managing security issues of data acquisition and transmission in data classification and integrity verification.
%
% Shahzad \etal~\cite{shahzad2020internet} implemented decisions and monitoring in automated processes.
%
% Good process planning can accelerate production speed and reduce production costs. 
% Andronie \etal ~\cite{andronie2021artificial} summarized the complexity and flexibility of intelligent process planning in production lines. 

\subsection{Resource-efficient AIoT System on Heterogeneous Devices}
Diverse AIoT devices are always heterogeneous regarding memory, computing, and battery resources.
And it is non-trivial to deploy resource-efficient AIoT system on heterogeneous devices.
Specifically, we select three types of representative AIoT devices, \ie the cheapest but with extremely constrained MCU devices, the most ubiquitous ARM-based mobile devices (\eg smartphones, wearables, robots), and typical GPU-based edge servers with weak resources (\eg NVIDIA DGX, HPE ProLiant DL380 Gen10).
As an indispensable interface, smartphones often serve as the control center of AIoT clusters. 
The MCU is very cheap but has extremely limited computing and memory resources. 
And the FPGA chip stands out with the advantages of high flexibility, low power consumption, and strong expansibility.
Table~\ref{tb:device3} summarizes how to optimize resource-efficient AIoT systems in MCU, ARM, and FPGA devices.

\subsubsection{Resource-efficient AIoT system on MCUs}
MCU is widely used for simple tasks in autopilot, medical care, office equipment, \etc 
And its small size and low cost make it suitable for large-scale deployment, \eg the daily inspection of the lake surface~\cite{yang2020analysis}.
However, the MCU's tightly-limited memory resources constrain the efficient execution of DL inference/training tasks. 
Many studies~\cite{lin2020mcunet, hosny2021sparse, banbury2021micronets, liberis2021munas, lin2021memory, sudharsan2020rce, maskey2020cubesatnet, hung2021end, wu2020enabling} have been conducted for DL deployment on MCUs.
%
% Existing research is mainly devoted to improving NAS for MCUs in two aspects, \ie compressing the search space~\cite{lin2020mcunet}, and reducing the search cost~\cite{banbury2021micronets,liberis2021munas}.
% \rev{and reduce their memory usage}. 
SpArSe~\cite{fedorov2019sparse} is a network architecture search (NAS) framework designed for DL inference on MCUs. 
% Sparse NAS combined with DNN structure trimming greatly reduced the memory requirement.
%
% Therefore, 
%
Micronets~\cite{banbury2021micronets} also leverage NAS for specifying DNN architectures.
%
% It imposes strict limits and constraints on MCU memory, latency, and energy. 
To map the MCU-imposed memory, latency, and energy constraints to the NAS search framework, they found an important basis, \ie model latency varies linearly with the model operation (op) counts before searching models in space. 
And they used these discoveries in NAS to reduce memory utilization and operands. 
$\mu$NAS \cite{liberis2021munas} is another NAS system for MCNS, emphasizing RAM size, memory size, and processor speed. 
% $\mu$NAS is optimized for speeding up matrix multiplication, reducing memory usage, and achieving a balance between memory usage and inference delay.
%
MCUNets~\cite{lin2020mcunet} integrates Tiny NAS and Tiny Engine to optimize the resource-efficient AIoT systems on MCUs. 
They find that for a DNN with the same size, the larger the amount of computation, the higher the accuracy.
With this insight, Tiny NAS greatly reduces the search space and improves search efficiency.
TinyEngine is a compilation engine that reduces the memory occupation in DNN operation by optimizing the loop operation.
% , and provides Tiny NAS with a larger search space to search the depth model with better effect.
%
% MCUNetsV2~\cite{lin2021memory} solves the problem of peak memory of the CNN models and greatly improves the accuracy with only a small increase in time cost.
%
Facing the huge memory cost of the first few CNN layers, MCUNetsV2~\cite{lin2021memory} makes longitudinal cutting to execute a small part of models and finally integrate the results of all cut parts.
% Then the calculation continues to achieve the purpose of making the large model run on MCU. 
% To face more complex calculation tasks and higher accuracy requirements.

\begin{figure}[t]
  \centering
    \subfloat[DLAU]{\label{fig:DLAU}
    \includegraphics[height=0.245\textwidth]{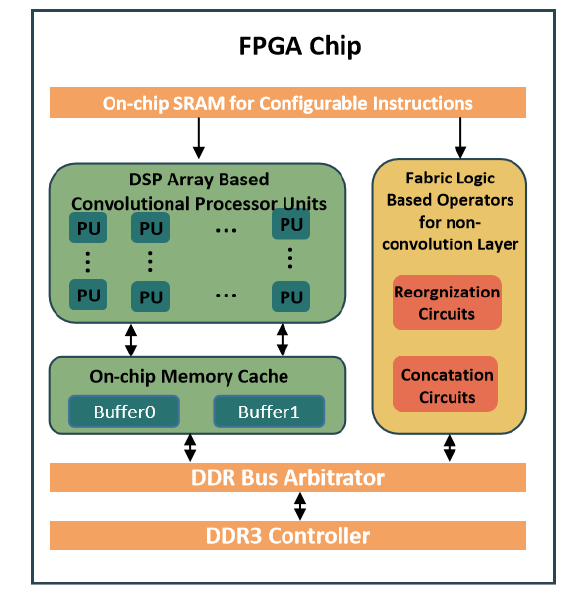}}
    \hspace{2pt}
    \subfloat[WPU]{\label{fig:WPU}
    \includegraphics[height=0.245\textwidth]{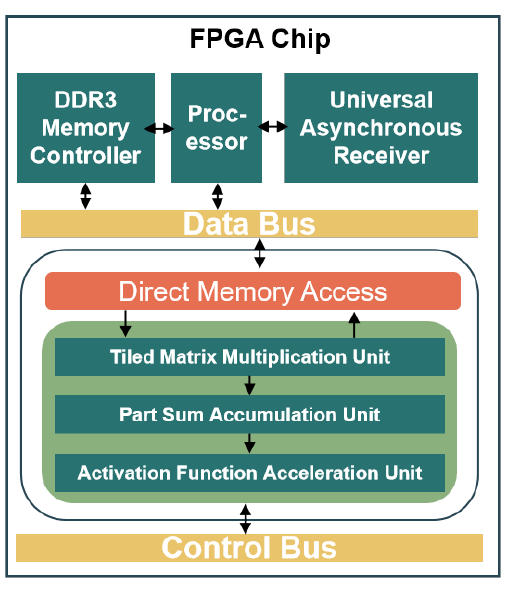}}
    %\vspace{-3mm}
\caption{Two different FPGA architectures for convolution acceleration, in which DLAU mainly adopts a three-layer pipeline, and WPU uses a sparse network design.}
\label{fig:pruning}
%\vspace{-3mm}
\end{figure}

% \usepackage{multirow}
\begin{table*}[]
\scriptsize
\renewcommand{\arraystretch}{0.8}
\caption{Summary of resource-efficient system software over heterogeneous AIoT devices. }
\resizebox{\linewidth}{!}{
\begin{tabular}{|c|c|c|c|c|c|}
\hline
\multirow{2}{*}{\textbf{Device type}} & \multirow{2}{*}{\textbf{Technique highlight}}                                             & \multirow{2}{*}{\textbf{Focus level}}      & \multirow{2}{*}{\textbf{Resource efficiency improvements}}             & \multirow{2}{*}{\textbf{Year}} & \multirow{2}{*}{\textbf{Ref.}}                 \\
                             &                                                                                  &                             &                                                               &                       &                                       \\ \hline
\multirow{10}{*}{\textbf{MCU}}        & \multirow{2}{*}{Sparse architecture search, NN-structure pruning}                & \multirow{2}{*}{Computation graph}   & \multirow{2}{*}{Reduce memory}                                & \multirow{2}{*}{2019} & \multirow{2}{*}{\cite{hosny2021sparse}}      \\
                             &                                                                                  &                             &                                                               &                       &                                       \\ \cline{2-6} 
                             & \multirow{2}{*}{Constraint mapping, NAS}                                         & \multirow{2}{*}{DL model}  & \multirow{2}{*}{Reduce memory, reduce latency, reduce energy} & \multirow{2}{*}{2021} & \multirow{2}{*}{\cite{banbury2021micronets}} \\
                             &                                                                                  &                             &                                                               &                       &                                       \\ \cline{2-6} 
                             & \multirow{2}{*}{Matrix multiplication optimization}                              & \multirow{2}{*}{DL model}   & \multirow{2}{*}{Reduce memory,reduce latency}                 & \multirow{2}{*}{2021} & \multirow{2}{*}{\cite{liberis2021munas}}     \\
                             &                                                                                  &                             &                                                               &                       &                                       \\ \cline{2-6} 
                             & \multirow{2}{*}{Cooperation of NAS and Engine, calculation library optimization} & \multirow{2}{*}{Memory scheduling}     & \multirow{2}{*}{Reduce Memory}                                & \multirow{2}{*}{2020} & \multirow{2}{*}{\cite{lin2020mcunet}}        \\
                             &                                                                                  &                             &                                                               &                       &                                       \\ \cline{2-6} 
                             & \multirow{2}{*}{Patch-to-patch inference, optimization of receptive field}       & \multirow{2}{*}{Operator}  & \multirow{2}{*}{Reduce memory}                                & \multirow{2}{*}{2021} & \multirow{2}{*}{\cite{lin2021memory}}        \\
                             &                                                                                  &                             &                                                               &                       &                                       \\ \hline
\multirow{12}{*}{\textbf{ARM}}        & \multirow{2}{*}{Parallel convolution algorithm based on Fast Fourier transform}  & \multirow{2}{*}{Compiler}   & \multirow{2}{*}{Reduce memory, reduce latency}                & \multirow{2}{*}{2017} & \multirow{2}{*}{\cite{ncnn}}                 \\
                             &                                                                                  &                             &                                                               &                       &                                       \\ \cline{2-6} 
                             & \multirow{2}{*}{Data rearrangement, complex matrix multiplication}               & \multirow{2}{*}{DL model}  & \multirow{2}{*}{Reduce memory, reduce latency}                & \multirow{2}{*}{2020} & \multirow{2}{*}{\cite{wang2020optimizing}}   \\
                             &                                                                                  &                             &                                                               &                       &                                       \\ \cline{2-6} 
                             & \multirow{2}{*}{Code automatically generates open source library}                & \multirow{2}{*}{Compiler}     & \multirow{2}{*}{Reduce memory, improve accuracy}              & \multirow{2}{*}{2022} & \multirow{2}{*}{\cite{meng2022automatic}}    \\
                             &                                                                                  &                             &                                                               &                       &                                       \\ \cline{2-6} 
                             & \multirow{2}{*}{Adjust data layout, convolution based on Winograd}               & \multirow{2}{*}{Intra-device controller} & \multirow{2}{*}{Reduce memory, reduce latency}                & \multirow{2}{*}{2021} & \multirow{2}{*}{\cite{li2021optimizing}}     \\
                             &                                                                                  &                             &                                                               &                       &                                       \\ \cline{2-6} 
                             & \multirow{2}{*}{Pipeline strategy, improve data reuse}                           & \multirow{2}{*}{Computation graph}  & \multirow{2}{*}{Reduce memory, reduce latency}                & \multirow{2}{*}{2021} & \multirow{2}{*}{\cite{huang2021numa}}        \\
                             &                                                                                  &                             &                                                               &                       &                                       \\ \cline{2-6} 
                             & \multirow{2}{*}{Fine memory management and data structure design}                & \multirow{2}{*}{Memory scheduling}   & \multirow{2}{*}{Reduce memory, improve accuracy}              & \multirow{2}{*}{2022} & \multirow{2}{*}{\cite{zhou2022pipelining}}   \\
                             &                                                                                  &                             &                                                               &                       &                                       \\ \hline
\multirow{8}{*}{\textbf{FPGA}}        & \multirow{2}{*}{Pipeline strategy, scalable accelerator architecture}            & \multirow{2}{*}{Computation graph}     & \multirow{2}{*}{Reduce latency; reduce energy}                & \multirow{2}{*}{2016} & \multirow{2}{*}{\cite{wang2016dlau}}         \\
                             &                                                                                  &                             &                                                               &                       &                                       \\ \cline{2-6} 
                             & \multirow{2}{*}{Simpler sparse data selection logic}                             & \multirow{2}{*}{Memory scheduling}     & \multirow{2}{*}{Improve resource utilization}                 & \multirow{2}{*}{2018} & \multirow{2}{*}{\cite{ma2018optimizing}}     \\
                             &                                                                                  &                             &                                                               &                       &                                       \\ \cline{2-6} 
                             & \multirow{2}{*}{Support high sparsity, low precision calculation}                & \multirow{2}{*}{Computation graph}  & \multirow{2}{*}{Reduce latency}                               & \multirow{2}{*}{2021} & \multirow{2}{*}{\cite{xie2021wpu}}           \\
                             &                                                                                  &                             &                                                               &                       &                                       \\ \cline{2-6} 
                             & \multirow{2}{*}{Two-bit convolution accelerator}                                 & \multirow{2}{*}{DL model}  & \multirow{2}{*}{Reduce latency}                               & \multirow{2}{*}{2021} & \multirow{2}{*}{\cite{meng2021fixyfpga}}     \\
                             &                                                                                  &                             &                                                               &                       &                                       \\ \hline
\end{tabular}}
\label{tb:device3}
\end{table*}

\subsubsection{Resource-efficient AIoT system on ARMs}
Compared to MCUs, ARM-based devices have more powerful computing power and are widely used in mobile devices such as robots and drones.
% DNNs always have large-scale matrix addition and multiplication, causing resource overhead.
%
To accelerate DNN inference/training on ARM devices, many researches~\cite{meng2022automatic,li2021optimizing,zhou2022pipelining,ncnn,chang2019memory,rios2018performance,fu2016visual,wang2019high} have been proposed.
It is reported that the conv layers in DNNs consume most of the computation due to the high computational amount~\cite{2014Going}.
Huang~\etal \cite{wang2020optimizing} realized the parallel conv computing based on the fast Fourier transform, which optimized the memory occupation and reduced the latency of multi-core parallel.
To further improve non-uniform memory access in many-core CPUs, Huang \etal~\cite{huang2021numa} proposed a fast Fourier convolution method based on NUMA awareness, conducting data rearrangement and parallelizing complex matrix multiplications to reduce remote memory accesses and thus improves the calculation of CNNs. 
Meng \etal~\cite{meng2022automatic} proposed FastConv, a template-based open-source library for automatic code generation, which can automatically generate high-performance CNN kernel and improve the performance of the conv layers on ARM devices. 
% Fastconv's convolution calculation method is mainly based on the Winograd algorithm. 
% The Winograd algorithm is a fast mathematical technique for performing matrix multiplication, commonly used in deep learning and other areas of scientific computing. This algorithm is based on a set of precomputed matrices called Winograd transformation matrices, which are designed to reduce the number of multiplication operations required to compute the product of two matrices.
% Fastcon uses this matrix multiplication to speed up the convolution operation.
% FastConv solves the adaptation of various convolution layers to various heterogeneous devices by automatically generating tuned kernel variants of various shapes. 
% And it automatically adjusts the execution parameters of the CNNs by searching for the best combination of kernel shape, cache size, loop scheduling, packaging strategy, access model, and online/offline combination. 
Meanwhile, Li \etal~\cite{li2021optimizing} focued on maximizing the parallelism of the algorithm to fully utilize the multiple processing cores available on modern ARM CPUs. 
They propose several optimizations, including a parallel tile processing scheme, a memory layout optimization, and an efficient data rearrangement technique.
%
% The optimized implementation is evaluated on a variety of ARM processor architectures, and the results demonstrate significant performance gains over the original Winograd algorithm and other state-of-the-art convolution methods.
%
Zhou \etal~\cite{zhou2022pipelining} optimized convolution layers through pipeline strategy. 
It computed the $3\times{3}$ convolution on the ARM CPU by means of a single instruction and multiple data. 
And it improved the computational efficiency by increasing the data reuse rate.
Tencent \etal~\cite{ncnn} presented NCNN framework, an ARM-based DNN optimization framework that integrates various memory management techniques.
It utilizes multi-core parallel to accelerate the DNN calculation. 
The cross-platform characteristics of NCNN greatly benefit users in transplanting the DL models to the AIoT terminals and reducing the latency and memory occupation of DNNs.

\subsubsection{Resource-efficient AIoT system on FGPAs}

FGPA is developed based on programmable array logic and universal array logic.
The main feature of FGPAs is that they are customizable and can independently change the circuit structure and expand the chips according to the amount and way of calculation~\cite{wang2016dlau,ma2018optimizing}. 
When deploying DNNs on FPGAs, such special features can maximize computational efficiency and reduce energy costs. 
Therefore, in AIoT application scenarios, FPGA is becoming a promising intelligent perception, computing, and control platform~\cite{liu2021enterprise}.
Existing efforts ~\cite{xie2021wpu,meng2021fixyfpga,peng2021binary,korol2019fpga,zainab2019fpga,petrica2020memory} have begun to study resource-efficient DNN deployments with FGPAs. 
Wang \etal~\cite{wang2016dlau} proposed DLAU, a scalable accelerator architecture for the large-scale deployment of DNNs on FGPA.  
In previous studies, the DL acceleration mainly includes loop unrolling~\cite{yang2020interstellar}, tiling~\cite{parashar2019timeloop}, switching~\cite{udagawa2022human}, \etc 
These methods, however, without hardware redesigning, can hardly give full play to the programmable characteristics of FGPA. 
To this end, \rev{Ma \etal~\cite{ma2018optimizing} designed the specific data stream to accelerate DL execution, as shown in Figure \ref{fig:pruning}(a).} 
To optimize memory accesses, they minimize data traffic and improve computation performance through quantitative analysis of multiple design variables, \eg energy, and latency. 
Xie \etal~\cite{xie2021wpu} proposed \rev{the DL model construction algorithm and the accelerator for realizing sparse data selection logic on FGPAs, as shown in Figure \ref{fig:pruning}(b).}
Most of the existing work requires the off-chip DDR memory to store parameters and the expensive DSP module for DL computation on FGPAs.
To overcome this issue, Meng \etal ~\cite{meng2021fixyfpga} proposed FixyFGPA, a DL inference accelerator, which supports high sparsity and low precision computation by integrating dense and sparse computation units in FPGAs. 
Still, it performs poorly in complex computation. 
Peng \etal ~\cite{peng2021binary} presents a novel two-position accelerator by combining the deep complex network with the binary neural network to speed up inference on FGPAs.
\section{Open Issues and Future Directions}
\label{sec:issue}
This section discusses existing challenges and potential misleading directions related to resource-efficient AIoT systems and enabling technologies.

\subsection{Cross-level AIoT System}
Despite various optimization techniques at a single level, \eg algorithm, computation graph, compiler, operator, hardware instruction, without considering their cooperation,
we discuss the open issues as below.

\textit{(i) Cross-level co-design.}
Prior efforts in cross-level co-design mainly include algorithm-hardware~\cite{hao2019fpga, li2020edd, jiang2019accuracy}, compiler-hardware~\cite{wang2020fann, garofalo2020pulp}, and algorithm-system~\cite{niu2020patdnn, liu2020cocopie, lin2020mcunet} co-design. 
The algorithm-system co-design is a promising way to address the increasing complexity of DL models for complicated application problems and optimize the runtime execution of those models on AIoT hardware.
While the hardware re-design in the first two types is costly for existing AIoT applications and suitable for brand-new construction.
Some of the key open issues in cross-level algorithm-system co-design include:
First, \textit{Runtime resource-efficient algorithm}. The challenge is how to propagate the feedback of the runtime system execution to the algorithm design.
State-of-the-art on algorithm-system co-design like PCONV~\cite{ma2020pconv}, PatDNN~\cite{niu2020patdnn}, and CoCoPIE~\cite{liu2020cocopie} simultaneously optimize the model compression algorithms and runtime operator/memory scheduling mechanisms.
They leverage some fixed resource-friendly patterns to guide model design.
Specifically, DL models are specified with a specific shape to maximize and sustain instruction-level and thread-level parallelism.
Second, \textit{Model-adaptive runtime compiler/engine}.
Most of the existing works at the compiler and engine level for operator/memory scheduling only manually optimize partial factors.
For example, the co-design MCUNet is composed of TinyNAS and TinyEngine. 
TinyNAS searches for the most efficient model architecture running on TinyEngine. 
At the same time, the TinyEngine library generates codes for the network search space of TinyNAS to eliminate instruction and memory redundancy. 
Moreover, 
it is non-trivial to make the compiler/engine to be compatible with various hardware backends of AIoT devices.
They consist of multiple processing units with different architectures and capabilities, \eg Jetsons~\cite{jetson}, Raspberry Pis~\cite{raspberry}, FPGAs~\cite{xilinx}.

\textit{(ii) Cross-level adaptive controller.}
Given heterogeneous AIoT devices and diverse performance demands, it is necessary to automate the adaptive optimization control across multiple levels, \eg DL model, operator, memory allocation, compiler, engine, and hardware instructions.
The adaptive controller at the algorithm level with automatic search, \eg neural architecture search (NAS), is widely explored~\cite{baymurzina2022review,kim2022neural,liu2022federated}.
While the adaptive controller across algorithm and system levels is less explored.
The challenges are from three aspects: 
First, the \textit{combined search space} should consider the complex cross-level dependency and collaborations in different techniques, as mentioned in ~\ref{sec:inference} and \ref{sec:train}.
Second, the \textit{performance validation} of candidates is non-trivial because the cross-level algorithm and system co-design cover multiple phases, \ie offline model design and parameter training, runtime compiler, and accuracy testing.
Integrating their validation and verification together is desired yet challenging, especially as the design becomes more complex.
Third, the \textit{runtime search algorithm}, \ie boosting the efficiency and efficacy of solving constrained multi-objective optimization problems, is a long-term open problem~\cite{yu2019evaluating}~\cite{li2020random}.

\subsection{Context-aware AIoT System Evolution}
As discussed in $\S$ \ref{subsec:metric}, 
to ensure that the deployed AIoT system
% , such as mobile VR/AR~\cite{mangiante2017vr,hou2017wireless,schmoll2018demonstration}, autonomous human-following drones~\cite{le2019human,yamashita2021autonomous,piquero2021novel}, vision-based robot navigation~\cite{zhou2019vision,khairudin2020vision,zielinski20213d}, and autonomous driving cars~\cite{yaqoob2019autonomous,zhou2019hidden,fujiyoshi2019deep}, 
can maintain continuous and stable high-quality services in the long-term life cycle, the resource demands of the AIoT system should be evolvable.
That is, the AIoT system can be compatible with the  \textit{dynamic nature of the deployment context} (\eg the dynamic resource availability, diverse performance demands) and integrates the adaptation loop to adjust all the system blocks.
The AIoT system evolution requirements include \textit{DL model structure scaling} and \textit{weight retraining}. 
The former is always caused by mismatching between model resource demand and device resource supply, and the latter is usually required by model accuracy degradation in live sensing data.
In particular, the evolvable AIoT system may need to deal with the following problems.

\textit{(i) Resource availability monitoring.}
The key resources affecting DL model deployment (\eg memory budget) and performance outcome (\eg latency, energy efficiency) in AIoT devices include memory, computing, and battery.
All of them exhibit high dynamics over time~\cite{liu2021adaspring}.
For example, the battery-powered device will gradually consume energy as the device runs, and the available memory or computing resources will be encroached upon by other applications/tasks running on a device.
The dynamic nature of available resources in AIoT devices is also related to the operating system (OS) resource scheduling.
A fast and accurate resource monitor that can interact with heterogeneous and cross-platform OS is needed.

\textit{(ii) Resource demand prediction and performance profiling of DL inference/training tasks}.
The resource demand of deployed AIoT systems contains energy cost, computation, and memory usage. 
    And systematic formulation of these resource demands can predict the most suitable DL inference/training configurations in advance.
    The AIoT system performance involves DL inference accuracy, inference latency, and training convergence latency.
    The inference accuracy is coupled with the DNN parameter weights, and dramatic accuracy drop trigger DNN retraining for weight evolving.
    As mentioned in $\S$ \ref{subsec:metric}, the total computation amount and memory usage for DL inference and training tasks can be directly calculated by the DNN structure and training configurations (\eg iteration number, batch size). 
    While the prediction of energy demand and latency is non-trivial since its measurement is not straightforward. 
    They heavily depend on the underlying operator scheduling, data flow, and memory bandwidth bound~\cite{liu2023adaenlight}.
    % Accurate and timely energy prediction is still intractable.
    % The DL inference/training latency depends on the deployment devices' hardware configurations.
    Measuring energy cost and latency in real-world devices is infeasible or too costly~\cite{zhang2021nn}. 
    However, it is highly desirable for many tasks, \eg searching for the most suitable DNN structure with latency demands from a vast space.

\textit{(iii) Dynamic context awareness and system evolution trigger}.
    The dynamic context awareness block detects the mismatch between resource supply and demand or the dramatic accuracy drop to trigger the adaptation block.
    The triggering station for resource supply-demand mismatch can be further modeled as the noticeable context changes.
    We note that the onset of the accuracy drop is not always the optimal trigger time point for DNN retraining, which may increase unnecessary retraining workload and even lag some necessary evolution tasks.
    A suitable trigger is also desired for evolving efficiency and efficacy.

\textit{(iv) Automated system loop}.
    The self-evolutionary AIoT system needs an automated loop consisting of the \textit{resource monitor}, the \textit{runtime resource demand and performance profiler}, the \textit{evolution trigger}, and the \textit{optimizer}.
    The resource monitor tracks the memory/computing resource supply of the available AIoT devices. 
    The resource demand profiler predicts the DL inference tasks' memory, computing, and energy resource demands with the current configurations. 
    If the resource demand exceeds the supply or the accuracy drop exceeds a pre-defined threshold, the evolution trigger notifies the optimizer, adjusting the DL inference/training configurations.
    The automated control loop routinely checks for system changes and performs on-demand evolution.

\subsection{Distributed AIoT Resource Aggregation in DL inference/training Tasks}
\rev{Efficiently aggregating memory and computing resources \rrev{within the networked AIoT system} for seamless communication and the provision of complex services is an active and research-intensive domain, which has several open issues.}

\begin{itemize}
\item How to uniformly manage heterogeneous resources (\eg SRAM, DRAM, CPU/GPU, battery) \rrev{within the networked AIoT system}? 
They should be organized and managed in a unified manner to enable efficient DNN training or inference tasks.
\item How to combine operator scheduling, memory allocation, and hardware instruction mechanisms \rrev{within the networked AIoT system}.
Specifically, the computing power of AIoT devices is subject to the \textit{barrel effect} of memory, computing, and battery resources. 
The shortest board of the barrel restricts the overall computing capability.
Thus, monitoring the AIoT device resources with \textit{normalized assessment necessary} is necessary.
\item How to balance multiple performance goals in the resource aggregation process (\eg energy efficiency and latency)? 
We need to establish the hierarchical dependence between several internal factors (\eg DL model, operator, memory, instructions) of the AIoT system and the external environment (\eg input).
\item How to develop hybrid distributed resource aggregation methods that \rev{combine both synchronous and asynchronous communication schemes to balance the trade-off between convergence speed and system stability.} 
\item How to deal with the opportunistic connection problem of AIoT devices for guaranteeing resource availability?
\end{itemize}

Advanced research in areas such as edge intelligence~\cite{bib:hua2023edge} and federated learning~\cite{huang2022learn} intersects with some of the issues mentioned above, but it cannot comprehensively address them. To tackle these challenges, it is important to decouple resource aggregation methods from the DL inference/training tasks and develop them as a general-purpose middleware function that can actively perceive, analyze, and select AIoT resources to meet diverse demands.

\subsection{Intelligence Enhancement in Distributed DL Training}
\rev{
AIoT devices have been extensively deployed in numerous domains. 
Their wide array of sensors allows for the collection of massive amounts of data, facilitating the execution of DL tasks and enabling different intelligent applications.
In practical AIoT systems, a large number of distributed devices continuously sense the environment and hold accumulated datasets.
\rrev{However, real-world scenarios present several challenges that must be addressed to ensure reliable and efficient data transmission, synchronization, and coordination among these devices. These challenges include:
}
}
    
\textit{(i) Intrinsic linkages of distributed holding datasets.}.
To fully leverage the distributed devices' computing power and accumulated datasets, studying the intrinsic linkages between distributed datasets in different AIoT sub-clusters and designing corresponding distributed DL training systems is essential.
In particular, the inherent correlation, redundancy, and hysteresis of distributed datasets can drive the design of distributed training systems, leading to high-quality, responsive, and low-cost distributed DL training.

\textit{(ii) Temporal collaboration of distributed devices}.
    In real-world AIoT systems, each device may have different training power and speed.
    \rev{Asynchronous communication methods have been proposed to mitigate this problem to some extent~\cite{chen2019communication}, but extreme temporal differences still pose a significant challenge to participant collaboration.} 
    Specifically, the challenge is effectively promoting collaboration among participant devices with different temporal patterns.
    For example, existing distributed DL training systems often partition the dataset by data category to highlight the Non-IID nature of the data~\cite{huang2021personalized}. 
    However, the data distribution collected by AIoT devices is clearly more diverse and challenging.
    Possible directions for distributed DL training can consider: 
    \begin{itemize}
        \item The temporal correlation of the distributed data at diverse sub-clusters of AIoT devices varies. 
        \rev{Distributed DL model/data aggregation and communication mechanisms in AIoT system's distributed learning for syn-/asyn-chronic datasets/models should be different.}
        \item The spatial correlation of the distributed data in distributed DNN training, \eg physically nearby devices have the potential for better collaboration gains.
        \item We can divide the participating AIoT devices into diverse roles in distributed DL training, \eg collaborators, competitors, and supervisors.
    \end{itemize}

\subsection{Inference and Training Task Balance in Resource-constrained AIoT Devices}
Resouce-constrained AIoT devices always load lightweight Dl models, such as compressed DL models.
However, DL models with shallow or sparse structures are highly susceptible to data drift, which occurs when the live data stream captured by the devices diverges from the data used for training, leading to a drop in accuracy in real-world applications.
Some efforts~\cite{bhardwaj2022ekya,mullapudi2019online,khani2021real} have proven that DL models deployed in AIoT devices should be continuously retrained using newly captured live data to maintain accuracy.

However, AIoT devices provide limited resources to execute computation. 
Introducing training tasks into AIoT devices will likely \textit{deprive resources of the inference tasks for training}, leading to decreased inference performance.
The more resources we allocate to the task of DL retraining, the faster the training process will be, and we can obtain a high-accuracy inference model earlier. 
However, the limited resources allocated to the inference task during the training process may decrease the overall performance of DL inference. 
On the other hand, allocating too few resources to the retraining task may slow down the retraining speed and delay the accuracy improvement of the updated model, which could impact the timely inference accuracy gain.
Therefore, it is crucial to trade off the accuracy improvement from training with real-time inference performance. 
We discuss open issues in more detail:
% In this case, we can further optimize the AIoT system to consider the following challenging problems:
\begin{itemize}
    % \item How to evaluate AIoT system accuracy for real-world deployment using unlabeled live data streams to trigger the training of DNN in the inference window?
    \item How to allocate resources for maximizing the inference accuracy when the training task deprives a certain amount of memory, regarding the tunable data flow in memory units.
    For example, we can share memory resources between the inference and training tasks via memory reallocation, recomputation, and swapping techniques. 
    % \item How to allocate memory resources among diverse DL training and inference tasks ?
    \item How to select promising DL training tasks and allocate computing resources among them in AIoT devices.
    \item How to select DL training configurations (\eg epoch) and inference configurations (\eg sampling rate) to maximize inference accuracy with minimum resource usage.
\end{itemize}

% Like hyperparametric search, these problems are complex, and the decision spaces between memory/computing resources and cross-level system configurations are ample.
%
% It is hard to accurately assess each resource and performance to assist decision-making without performing DNN training. 
% % 
% Also, the optimal resources and configurations in dynamically changing AIoT scenes are not set in stone, \ie the best decision at the current training may be only sub-optimal at the next training. 
% Adaptive adjustment of the decision scheme according to the different scenes is worth further research and exploration.

Making accurate decisions on resource allocation and performance assessment is challenging in advance DL training. 
Furthermore, due to the dynamic nature of AIoT context, the optimal resources and configurations may change over time. 
The best decision at the current training may become sub-optimal for future training phases.

\subsection{Memory-computation Feedback and Joint-optimization}
Memory and computation cost of DL models are two crucial metrics tunable in AIoT systems to affect overall performance, \eg delay, and energy cost. 
However, previous studies have shown that memory and computation optimization are often conflicting goals. 
For instance, recomputation techniques can save memory space by discarding and recalculating intermediate activations, but they can also introduce extra computing delay~\cite{chen2016training, gomez2017reversible, kirisame2020dynamic}. 
Further research is needed to balance these two metrics best and enable memory-computation joint optimization in AIoT systems.

\textit{(i) Near- and in-memory computing}.
The underlying memory schedule techniques have not kept up with computation optimization advances in latency and energy reduction over the years, referred to as the memory wall~\cite{wulf1995hitting}. 
The development of DL has exacerbated this problem as the frequent movement of a large amount of data, including input data and intermediate activations between memory and compute units. They seriously impacted the latency and energy cost of DL training and inference tasks.
To address this issue, some studies~\cite{wordeman20123d, zhu20133d, zhu2013accelerating, seshadri2015fast, seshadri2013rowclone} physically relocate compute units (multi-core, GPU, FPGA) closer to memory to reduce data transport costs. 
Near-memory and in-memory computing embed computation in the memory array. 
As compute units become more intimately connected with memory, finer-grained parallelism can be proposed to improve energy efficiency and latency~\cite{jerger2008virtual, kang2012flexram, wang2015energy, liu2015spiking}.
However, these compute units placed next to memory have problems such as less supporting computing types and weak computing power. 
Thus, designing at the system level, rather than the hardware level, to deploy data-intensive operators (such as ReLU) in near-memory compute units is promising, especially for ubiquitous AIoT devices without hardware replacement. 
DL operators need to access a large amount of data for computing, which causes considerable data transfer delay and lower computing demands. 
Also, we can carefully allocate DL operators to the near- and in-memory units to reduce memory access costs and improve computation efficiency at the operator and instructor levels.

\textit{(ii) Memory-aware computing in runtime/compiler optimization}.
Memory-computation joint optimization for AIoT can be achieved by co-designing the DL algorithm and engine.
At the algorithm level, we can interleave computation-intensive and data-intensive operators to balance the workload of compute units near memory. 
At the engine level, we can optimize the computation graph and generate execution code to reduce access delay for data-intensive operators. 
These can be achieved by optimizing the inner layers of the loop.

\section{Conclusion}
\label{sec:conclu}

Artificial Intelligence of Things (AIoT) combines AI technologies with IoT infrastructures, enhancing the efficiency and efficacy of data analysis. 
However, due to the heterogeneous and dynamic nature of AIoT hardware, co-designed cross-level AIoT system and adaptive controllers are needed to expand the boundaries of system performance beyond what can be achieved by algorithm-level techniques alone. 
These co-designed systems can push the boundaries of resource-performance tradeoffs for AIoT.
Specifically, the cross-level AIoT system spans on-device and distributed DL training/inference algorithms, computation graphs, operators, memory schedules, hardware instructions, \etc
With the continuous development of DL technologies and AIoT devices, the AIoT system expands the cyber-physical space to the human space, providing low-cost, high-quality, and inclusive ubiquitous intelligence for a wide range of AIoT application domains.
Given the heterogeneous and dynamic nature of AIoT hardware, co-designed cross-level AIoT systems, and adaptive controllers can further expand the boundaries of system performance, including accuracy and resource consumption, beyond what can be achieved with algorithm-level or system-level techniques alone.
We hope this survey will raise awareness and stimulate discussion among researchers and developers on AIoT system.
\rrev{
This paper elucidates many aspects of distributed device collaboration (\ie network topology establishment), and data exchange in real-world inference and training tasks. 
These insights are invaluable for communication researchers, as they provide a deeper understanding of the intricacies involved. 
More heuristics and insights are needed to ensure resource-efficient AIoT systems.
For example, reliable and real-time communication can further enhance network efficiency in AIoT deployments.}

\section*{Acknowledgments}
This work was partially supported by the National Science Fund for Distinguished Young Scholars (62025205) and the National Natural Science Foundation of China (No. 62032020, 62102317).

\bibliography{IEEEart}

\begin{thebibliography}{100}
\providecommand{\url}[1]{#1}
\csname url@samestyle\endcsname
\providecommand{\newblock}{\relax}
\providecommand{\bibinfo}[2]{#2}
\providecommand{\BIBentrySTDinterwordspacing}{\spaceskip=0pt\relax}
\providecommand{\BIBentryALTinterwordstretchfactor}{4}
\providecommand{\BIBentryALTinterwordspacing}{\spaceskip=\fontdimen2\font plus
\BIBentryALTinterwordstretchfactor\fontdimen3\font minus \fontdimen4\font\relax}
\providecommand{\BIBforeignlanguage}[2]{{%
\expandafter\ifx\csname l@#1\endcsname\relax
\typeout{** WARNING: IEEEtran.bst: No hyphenation pattern has been}%
\typeout{** loaded for the language `#1'. Using the pattern for}%
\typeout{** the default language instead.}%
\else
\language=\csname l@#1\endcsname
\fi
#2}}
\providecommand{\BIBdecl}{\relax}
\BIBdecl

\bibitem{ghosh2018artificial}
A.~Ghosh, D.~Chakraborty, and A.~Law, ``Artificial intelligence in internet of things,'' \emph{CAAI Transactions on Intelligence Technology}, vol.~3, no.~4, pp. 208--218, 2018.

\bibitem{cisco}
``Cisco annual internet report (2018–2023) white paper,'' \url{https://www.cisco.com/c/en/us/solutions/collateral/executive-perspectives/annual-internet-report/white-paper-c11-741490.html}.

\bibitem{hassan2019privacy}
M.~U. Hassan, M.~H. Rehmani, and J.~Chen, ``Privacy preservation in blockchain based iot systems: Integration issues, prospects, challenges, and future research directions,'' \emph{Future Generation Computer Systems}, vol.~97, pp. 512--529, 2019.

\bibitem{howard2017mobilenets}
A.~G. Howard, M.~Zhu, B.~Chen, D.~Kalenichenko, W.~Wang, T.~Weyand, M.~Andreetto, and H.~Adam, ``Mobilenets: Efficient convolutional neural networks for mobile vision applications,'' \emph{arXiv preprint arXiv:1704.04861}, 2017.

\bibitem{ray2021review}
P.~P. Ray, ``A review on tinyml: State-of-the-art and prospects,'' \emph{Journal of King Saud University-Computer and Information Sciences}, 2021.

\bibitem{guo2021mistify}
P.~Guo, B.~Hu, and W.~Hu, ``Mistify: Automating dnn model porting for on-device inference at the edge,'' in \emph{NSDI}, 2021.

\bibitem{liu2019edge}
L.~Liu, H.~Li, and M.~Gruteser, ``Edge assisted real-time object detection for mobile augmented reality,'' in \emph{The 25th annual international conference on mobile computing and networking}, 2019, pp. 1--16.

\bibitem{liu2021adaspring}
S.~Liu, B.~Guo, K.~Ma, Z.~Yu, and J.~Du, ``Adaspring: Context-adaptive and runtime-evolutionary deep model compression for mobile applications,'' \emph{Proceedings of the ACM on Interactive, Mobile, Wearable and Ubiquitous Technologies}, vol.~5, no.~1, pp. 1--22, 2021.

\bibitem{chen2018tvm}
T.~Chen, T.~Moreau, Z.~Jiang, L.~Zheng, E.~Yan, H.~Shen, M.~Cowan, L.~Wang, Y.~Hu, L.~Ceze \emph{et~al.}, ``$\{$TVM$\}$: An automated $\{$End-to-End$\}$ optimizing compiler for deep learning,'' in \emph{13th USENIX Symposium on Operating Systems Design and Implementation (OSDI 18)}, 2018, pp. 578--594.

\bibitem{tensorflowruntime}
tensorflow, ``Tensorflowruntime,'' \url{https://blog.tensorflow.org/2020/04/tfrt-new-tensorflow-runtime.html}.

\bibitem{tensorflowxla}
------, ``tensorflowxla,'' \url{https://www.tensorflow.org/xla?hl=zh-cn}.

\bibitem{lin2020mcunet}
J.~Lin, W.-M. Chen, Y.~Lin, C.~Gan, S.~Han \emph{et~al.}, ``Mcunet: Tiny deep learning on iot devices,'' \emph{Advances in Neural Information Processing Systems}, vol.~33, pp. 11\,711--11\,722, 2020.

\bibitem{benmeziane2021hardware}
H.~Benmeziane, K.~El~Maghraoui, H.~Ouarnoughi, S.~Niar, M.~Wistuba, and N.~Wang, ``Hardware-aware neural architecture search: Survey and taxonomy.'' in \emph{IJCAI}, 2021, pp. 4322--4329.

\bibitem{joshi2022enabling}
P.~Joshi, H.~Afli, M.~Hasanuzzaman, C.~Thapa, and T.~Scully, ``Enabling deep learning for all-in edge paradigm,'' \emph{arXiv preprint arXiv:2204.03326}, 2022.

\bibitem{wang2020convergence}
X.~Wang, Y.~Han, V.~C. Leung, D.~Niyato, X.~Yan, and X.~Chen, ``Convergence of edge computing and deep learning: A comprehensive survey,'' \emph{IEEE Communications Surveys \& Tutorials}, vol.~22, no.~2, pp. 869--904, 2020.

\bibitem{chen2019deep}
J.~Chen and X.~Ran, ``Deep learning with edge computing: A review,'' \emph{Proceedings of the IEEE}, vol. 107, no.~8, pp. 1655--1674, 2019.

\bibitem{li2020deep}
M.~Li, Y.~Liu, X.~Liu, Q.~Sun, X.~You, H.~Yang, Z.~Luan, L.~Gan, G.~Yang, and D.~Qian, ``The deep learning compiler: A comprehensive survey,'' \emph{IEEE Transactions on Parallel and Distributed Systems}, vol.~32, no.~3, pp. 708--727, 2020.

\bibitem{zhang2021compacting}
K.~Zhang, H.~Ying, H.-N. Dai, L.~Li, Y.~Peng, K.~Guo, and H.~Yu, ``Compacting deep neural networks for internet of things: Methods and applications,'' \emph{IEEE Internet of Things Journal}, vol.~8, no.~15, pp. 11\,935--11\,959, 2021.

\bibitem{lecun2015deep}
Y.~LeCun, Y.~Bengio, and G.~Hinton, ``Deep learning,'' \emph{nature}, vol. 521, no. 7553, pp. 436--444, 2015.

\bibitem{tang2022computational}
S.~Tang, L.~Chen, K.~He, J.~Xia, L.~Fan, and A.~Nallanathan, ``Computational intelligence and deep learning for next-generation edge-enabled industrial iot,'' \emph{IEEE Transactions on Network Science and Engineering}, 2022.

\bibitem{zhou2019edge}
Z.~Zhou, X.~Chen, E.~Li, L.~Zeng, K.~Luo, and J.~Zhang, ``Edge intelligence: Paving the last mile of artificial intelligence with edge computing,'' \emph{Proceedings of the IEEE}, vol. 107, no.~8, pp. 1738--1762, 2019.

\bibitem{nauth2009embedded}
P.~Nauth, \emph{Embedded intelligent systems}.\hskip 1em plus 0.5em minus 0.4em\relax Walter de Gruyter, 2009.

\bibitem{ghosh2020aiot}
I.~Ghosh, ``Aiot: when artificial intelligence meets the internet of things,'' \emph{Visual Capitalist}, vol.~12, 2020.

\bibitem{aiboom}
``The third wave of ai,'' \url{https://elleknowsmachines.com/third-wave-of-ai/}.

\bibitem{saha2022machine}
S.~S. Saha, S.~S. Sandha, and M.~Srivastava, ``Machine learning for microcontroller-class hardware-a review,'' \emph{IEEE Sensors Journal}, 2022.

\bibitem{tensorflowlite}
``Tensorflow lite,'' \url{https://www.tensorflow.org/lite.}

\bibitem{caffe2}
``Caffe2,'' \url{https://caffe2.ai/}.

\bibitem{pytorchmobile}
``Pytorch mobile,'' \url{https://pytorch.org/mobile/home/}.

\bibitem{nazir2019internet}
S.~Nazir, Y.~Ali, N.~Ullah, and I.~Garc{\'\i}a-Magari{\~n}o, ``Internet of things for healthcare using effects of mobile computing: a systematic literature review,'' \emph{Wireless Communications and Mobile Computing}, vol. 2019, pp. 1--20, 2019.

\bibitem{cicceri2020deep}
G.~Cicceri, F.~De~Vita, D.~Bruneo, G.~Merlino, and A.~Puliafito, ``A deep learning approach for pressure ulcer prevention using wearable computing,'' \emph{Human-centric Computing and Information Sciences}, vol.~10, no.~1, pp. 1--21, 2020.

\bibitem{ray2022review}
P.~P. Ray, ``A review on tinyml: State-of-the-art and prospects,'' \emph{Journal of King Saud University-Computer and Information Sciences}, vol.~34, no.~4, pp. 1595--1623, 2022.

\bibitem{TinyML}
\BIBentryALTinterwordspacing
S.~Han. Tinyml project. [Online]. Available: \url{https://tinyml.mit.edu/}
\BIBentrySTDinterwordspacing

\bibitem{mendez2022edge}
J.~Mendez, K.~Bierzynski, M.~Cu{\'e}llar, and D.~P. Morales, ``Edge intelligence: Concepts, architectures, applications and future directions,'' \emph{ACM Transactions on Embedded Computing Systems}, 2022.

\bibitem{capotondi2020cmix}
A.~Capotondi, M.~Rusci, M.~Fariselli, and L.~Benini, ``Cmix-nn: Mixed low-precision cnn library for memory-constrained edge devices,'' \emph{IEEE Transactions on Circuits and Systems II: Express Briefs}, vol.~67, no.~5, pp. 871--875, 2020.

\bibitem{onednn}
Intel, ``onednn,'' \url{https://github.com/oneapi-src/oneDNN}.

\bibitem{wan2022sensor}
T.~Wan, B.~Shao, S.~Ma, Y.~Zhou, Q.~Li, and Y.~Chai, ``In-sensor computing: Materials, devices, and integration technologies,'' \emph{Advanced Materials}, p. 2203830, 2022.

\bibitem{zhou2020near}
F.~Zhou and Y.~Chai, ``Near-sensor and in-sensor computing,'' \emph{Nature Electronics}, vol.~3, no.~11, pp. 664--671, 2020.

\bibitem{lin2022device}
J.~Lin, L.~Zhu, W.-M. Chen, W.-C. Wang, C.~Gan, and S.~Han, ``On-device training under 256kb memory,'' \emph{arXiv preprint arXiv:2206.15472}, 2022.

\bibitem{rhu2016vdnn}
M.~Rhu, N.~Gimelshein, J.~Clemons, A.~Zulfiqar, and S.~W. Keckler, ``vdnn: Virtualized deep neural networks for scalable, memory-efficient neural network design,'' in \emph{2016 49th Annual IEEE/ACM International Symposium on Microarchitecture (MICRO)}.\hskip 1em plus 0.5em minus 0.4em\relax IEEE, 2016, pp. 1--13.

\bibitem{wang2018superneurons}
L.~Wang, J.~Ye, Y.~Zhao, W.~Wu, A.~Li, S.~L. Song, Z.~Xu, and T.~Kraska, ``Superneurons: Dynamic gpu memory management for training deep neural networks,'' in \emph{Proceedings of the 23rd ACM SIGPLAN symposium on principles and practice of parallel programming}, 2018, pp. 41--53.

\bibitem{li2021low}
T.~Li, J.~Huang, E.~Risinger, and D.~Ganesan, ``Low-latency speculative inference on distributed multi-modal data streams,'' in \emph{Proceedings of the 19th Annual International Conference on Mobile Systems, Applications, and Services}, 2021, pp. 67--80.

\bibitem{petridis2018end}
S.~Petridis, T.~Stafylakis, P.~Ma, F.~Cai, G.~Tzimiropoulos, and M.~Pantic, ``End-to-end audiovisual speech recognition,'' in \emph{2018 IEEE international conference on acoustics, speech and signal processing (ICASSP)}.\hskip 1em plus 0.5em minus 0.4em\relax IEEE, 2018, pp. 6548--6552.

\bibitem{tian2018audio}
Y.~Tian, J.~Shi, B.~Li, Z.~Duan, and C.~Xu, ``Audio-visual event localization in unconstrained videos,'' in \emph{Proceedings of the European Conference on Computer Vision (ECCV)}, 2018, pp. 247--263.

\bibitem{radu2018multimodal}
V.~Radu, C.~Tong, S.~Bhattacharya, N.~D. Lane, C.~Mascolo, M.~K. Marina, and F.~Kawsar, ``Multimodal deep learning for activity and context recognition,'' \emph{Proceedings of the ACM on Interactive, Mobile, Wearable and Ubiquitous Technologies}, vol.~1, no.~4, pp. 1--27, 2018.

\bibitem{ning2019deep}
Z.~Ning, P.~Dong, X.~Wang, M.~S. Obaidat, X.~Hu, L.~Guo, Y.~Guo, J.~Huang, B.~Hu, and Y.~Li, ``When deep reinforcement learning meets 5g-enabled vehicular networks: A distributed offloading framework for traffic big data,'' \emph{IEEE Transactions on Industrial Informatics}, vol.~16, no.~2, pp. 1352--1361, 2019.

\bibitem{lin2020survey}
H.~Lin, S.~Zeadally, Z.~Chen, H.~Labiod, and L.~Wang, ``A survey on computation offloading modeling for edge computing,'' \emph{Journal of Network and Computer Applications}, vol. 169, p. 102781, 2020.

\bibitem{verbraeken2020survey}
J.~Verbraeken, M.~Wolting, J.~Katzy, J.~Kloppenburg, T.~Verbelen, and J.~S. Rellermeyer, ``A survey on distributed machine learning,'' \emph{Acm computing surveys (csur)}, vol.~53, no.~2, pp. 1--33, 2020.

\bibitem{nguyen2021federated}
D.~C. Nguyen, M.~Ding, P.~N. Pathirana, A.~Seneviratne, J.~Li, and H.~V. Poor, ``Federated learning for internet of things: A comprehensive survey,'' \emph{IEEE Communications Surveys \& Tutorials}, vol.~23, no.~3, pp. 1622--1658, 2021.

\bibitem{tan2020equalization}
J.~Tan, C.~Wang, B.~Li, Q.~Li, W.~Ouyang, C.~Yin, and J.~Yan, ``Equalization loss for long-tailed object recognition,'' in \emph{Proceedings of the IEEE/CVF conference on computer vision and pattern recognition}, 2020, pp. 11\,662--11\,671.

\bibitem{zhou2020look}
M.~Zhou, Y.~Bai, W.~Zhang, T.~Zhao, and T.~Mei, ``Look-into-object: Self-supervised structure modeling for object recognition,'' in \emph{Proceedings of the IEEE/CVF conference on computer vision and pattern recognition}, 2020, pp. 11\,774--11\,783.

\bibitem{abu2021some}
A.~T. Abu-Jassar, Y.~M. Al-Sharo, V.~Lyashenko, and S.~Sotnik, ``Some features of classifiers implementation for object recognition in specialized computer systems,'' \emph{TEM Journal}, vol.~10, no.~4, p. 1645, 2021.

\bibitem{bib:xu2023edge}
W.~Xu, Z.~Yang, D.~W.~K. Ng, M.~Levorato, Y.~C. Eldar, and M.~Debbah, ``Edge learning for b5g networks with distributed signal processing: Semantic communication, edge computing, and wireless sensing,'' \emph{IEEE Journal of Selected Topics in Signal Processing}, 2023.

\bibitem{yuan2020object}
Y.~Yuan, X.~Chen, and J.~Wang, ``Object-contextual representations for semantic segmentation,'' in \emph{Computer Vision--ECCV 2020: 16th European Conference, Glasgow, UK, August 23--28, 2020, Proceedings, Part VI 16}.\hskip 1em plus 0.5em minus 0.4em\relax Springer, 2020, pp. 173--190.

\bibitem{wang2021exploring}
W.~Wang, T.~Zhou, F.~Yu, J.~Dai, E.~Konukoglu, and L.~Van~Gool, ``Exploring cross-image pixel contrast for semantic segmentation,'' in \emph{Proceedings of the IEEE/CVF International Conference on Computer Vision}, 2021, pp. 7303--7313.

\bibitem{xie2021segformer}
E.~Xie, W.~Wang, Z.~Yu, A.~Anandkumar, J.~M. Alvarez, and P.~Luo, ``Segformer: Simple and efficient design for semantic segmentation with transformers,'' \emph{Advances in Neural Information Processing Systems}, vol.~34, pp. 12\,077--12\,090, 2021.

\bibitem{wang2019fast}
Q.~Wang, L.~Zhang, L.~Bertinetto, W.~Hu, and P.~H. Torr, ``Fast online object tracking and segmentation: A unifying approach,'' in \emph{Proceedings of the IEEE/CVF conference on Computer Vision and Pattern Recognition}, 2019, pp. 1328--1338.

\bibitem{zhang2022bytetrack}
Y.~Zhang, P.~Sun, Y.~Jiang, D.~Yu, F.~Weng, Z.~Yuan, P.~Luo, W.~Liu, and X.~Wang, ``Bytetrack: Multi-object tracking by associating every detection box,'' in \emph{Computer Vision--ECCV 2022: 17th European Conference, Tel Aviv, Israel, October 23--27, 2022, Proceedings, Part XXII}.\hskip 1em plus 0.5em minus 0.4em\relax Springer, 2022, pp. 1--21.

\bibitem{meinhardt2022trackformer}
T.~Meinhardt, A.~Kirillov, L.~Leal-Taixe, and C.~Feichtenhofer, ``Trackformer: Multi-object tracking with transformers,'' in \emph{Proceedings of the IEEE/CVF conference on computer vision and pattern recognition}, 2022, pp. 8844--8854.

\bibitem{galassi2020attention}
A.~Galassi, M.~Lippi, and P.~Torroni, ``Attention in natural language processing,'' \emph{IEEE transactions on neural networks and learning systems}, vol.~32, no.~10, pp. 4291--4308, 2020.

\bibitem{liu2023pre}
P.~Liu, W.~Yuan, J.~Fu, Z.~Jiang, H.~Hayashi, and G.~Neubig, ``Pre-train, prompt, and predict: A systematic survey of prompting methods in natural language processing,'' \emph{ACM Computing Surveys}, vol.~55, no.~9, pp. 1--35, 2023.

\bibitem{wang2019sequential}
S.~Wang, L.~Hu, Y.~Wang, L.~Cao, Q.~Z. Sheng, and M.~Orgun, ``Sequential recommender systems: challenges, progress and prospects,'' \emph{arXiv preprint arXiv:2001.04830}, 2019.

\bibitem{naumov2019deep}
M.~Naumov, D.~Mudigere, H.-J.~M. Shi, J.~Huang, N.~Sundaraman, J.~Park, X.~Wang, U.~Gupta, C.-J. Wu, A.~G. Azzolini \emph{et~al.}, ``Deep learning recommendation model for personalization and recommendation systems,'' \emph{arXiv preprint arXiv:1906.00091}, 2019.

\bibitem{wu2022graph}
S.~Wu, F.~Sun, W.~Zhang, X.~Xie, and B.~Cui, ``Graph neural networks in recommender systems: a survey,'' \emph{ACM Computing Surveys}, vol.~55, no.~5, pp. 1--37, 2022.

\bibitem{guo2020zero}
C.~Guo, C.~Li, J.~Guo, C.~C. Loy, J.~Hou, S.~Kwong, and R.~Cong, ``Zero-reference deep curve estimation for low-light image enhancement,'' in \emph{Proceedings of the IEEE/CVF conference on computer vision and pattern recognition}, 2020, pp. 1780--1789.

\bibitem{flinn1999energy}
J.~Flinn and M.~Satyanarayanan, ``Energy-aware adaptation for mobile applications,'' \emph{ACM SIGOPS Operating Systems Review}, vol.~33, no.~5, pp. 48--63, 1999.

\bibitem{benini2000system}
L.~Benini and G.~d. Micheli, ``System-level power optimization: techniques and tools,'' \emph{ACM Transactions on Design Automation of Electronic Systems (TODAES)}, vol.~5, no.~2, pp. 115--192, 2000.

\bibitem{hu2016network}
H.~Hu, R.~Peng, Y.-W. Tai, and C.-K. Tang, ``Network trimming: A data-driven neuron pruning approach towards efficient deep architectures,'' \emph{arXiv preprint arXiv:1607.03250}, 2016.

\bibitem{luo2017thinet}
J.-H. Luo, J.~Wu, and W.~Lin, ``Thinet: A filter level pruning method for deep neural network compression,'' in \emph{Proceedings of the IEEE international conference on computer vision}, 2017, pp. 5058--5066.

\bibitem{he2017channel}
Y.~He, X.~Zhang, and J.~Sun, ``Channel pruning for accelerating very deep neural networks,'' in \emph{Proceedings of the IEEE international conference on computer vision}, 2017, pp. 1389--1397.

\bibitem{kaloshin2020convolutional}
P.~Kaloshin, ``Convolutional neural networks compression with low rank and sparse tensor decompositions,'' \emph{arXiv preprint arXiv:2006.06443}, 2020.

\bibitem{liberis2021munas}
E.~Liberis, {\L}.~Dudziak, and N.~D. Lane, ``$\mu$nas: Constrained neural architecture search for microcontrollers,'' in \emph{Proceedings of the 1st Workshop on Machine Learning and Systems}, 2021, pp. 70--79.

\bibitem{rusci2020memory}
M.~Rusci, A.~Capotondi, and L.~Benini, ``Memory-driven mixed low precision quantization for enabling deep network inference on microcontrollers,'' \emph{Proceedings of Machine Learning and Systems}, vol.~2, pp. 326--335, 2020.

\bibitem{rusci2020leveraging}
M.~Rusci, M.~Fariselli, A.~Capotondi, and L.~Benini, ``Leveraging automated mixed-low-precision quantization for tiny edge microcontrollers,'' in \emph{IoT Streams for Data-Driven Predictive Maintenance and IoT, Edge, and Mobile for Embedded Machine Learning}.\hskip 1em plus 0.5em minus 0.4em\relax Springer, 2020, pp. 296--308.

\bibitem{liu2020adadeep}
S.~Liu, J.~Du, K.~Nan, Z.~Zhou, H.~Liu, Z.~Wang, and Y.~Lin, ``Adadeep: a usage-driven, automated deep model compression framework for enabling ubiquitous intelligent mobiles,'' \emph{IEEE Transactions on Mobile Computing}, vol.~20, no.~12, pp. 3282--3297, 2020.

\bibitem{ding2021ios}
Y.~Ding, L.~Zhu, Z.~Jia, G.~Pekhimenko, and S.~Han, ``Ios: Inter-operator scheduler for cnn acceleration,'' \emph{Proceedings of Machine Learning and Systems}, vol.~3, pp. 167--180, 2021.

\bibitem{cai2022optimus}
X.~Cai, Y.~Wang, and L.~Zhang, ``Optimus: An operator fusion framework for deep neural networks,'' \emph{ACM Transactions on Embedded Computing Systems (TECS)}, 2022.

\bibitem{niu2021dnnfusion}
W.~Niu, J.~Guan, Y.~Wang, G.~Agrawal, and B.~Ren, ``Dnnfusion: accelerating deep neural networks execution with advanced operator fusion,'' in \emph{Proceedings of the 42nd ACM SIGPLAN International Conference on Programming Language Design and Implementation}, 2021, pp. 883--898.

\bibitem{miao2021enabling}
H.~Miao and F.~X. Lin, ``Enabling large neural networks on tiny microcontrollers with swapping,'' \emph{arXiv preprint arXiv:2101.08744}, 2021.

\bibitem{huang1999generalized}
J.-C. Huang and T.~Leng, ``Generalized loop-unrolling: a method for program speedup,'' in \emph{Proceedings 1999 IEEE Symposium on Application-Specific Systems and Software Engineering and Technology. ASSET'99 (Cat. No. PR00122)}.\hskip 1em plus 0.5em minus 0.4em\relax IEEE, 1999, pp. 244--248.

\bibitem{im2001optimizing}
E.-J. Im and K.~Yelick, ``Optimizing sparse matrix computations for register reuse in sparsity,'' in \emph{Computational Science—ICCS 2001: International Conference San Francisco, CA, USA, May 28--30, 2001 Proceedings, Part I 1}.\hskip 1em plus 0.5em minus 0.4em\relax Springer, 2001, pp. 127--136.

\bibitem{boehm2018optimizing}
M.~Boehm, B.~Reinwald, D.~Hutchison, A.~V. Evfimievski, and P.~Sen, ``On optimizing operator fusion plans for large-scale machine learning in systemml,'' \emph{arXiv preprint arXiv:1801.00829}, 2018.

\bibitem{abadi2016tensorflow}
M.~Abadi, P.~Barham, J.~Chen, Z.~Chen, A.~Davis, J.~Dean, M.~Devin, S.~Ghemawat, G.~Irving, M.~Isard \emph{et~al.}, ``$\{$TensorFlow$\}$: a system for $\{$Large-Scale$\}$ machine learning,'' in \emph{12th USENIX symposium on operating systems design and implementation (OSDI 16)}, 2016, pp. 265--283.

\bibitem{paszke2017automatic}
A.~Paszke, S.~Gross, S.~Chintala, G.~Chanan, E.~Yang, Z.~DeVito, Z.~Lin, A.~Desmaison, L.~Antiga, and A.~Lerer, ``Automatic differentiation in pytorch,'' 2017.

\bibitem{yun2022cooperative}
S.~Yun, W.~Choi, and I.-M. Kim, ``Cooperative inference of dnns for delay-and memory-constrained wireless iot systems,'' \emph{IEEE Internet of Things Journal}, 2022.

\bibitem{wu2020accuracy}
W.~Wu, P.~Yang, W.~Zhang, C.~Zhou, and X.~Shen, ``Accuracy-guaranteed collaborative dnn inference in industrial iot via deep reinforcement learning,'' \emph{IEEE Transactions on Industrial Informatics}, vol.~17, no.~7, pp. 4988--4998, 2020.

\bibitem{he2020joint}
W.~He, S.~Guo, S.~Guo, X.~Qiu, and F.~Qi, ``Joint dnn partition deployment and resource allocation for delay-sensitive deep learning inference in iot,'' \emph{IEEE Internet of Things Journal}, vol.~7, no.~10, pp. 9241--9254, 2020.

\bibitem{mao2017modnn}
J.~Mao, X.~Chen, K.~W. Nixon, C.~Krieger, and Y.~Chen, ``Modnn: Local distributed mobile computing system for deep neural network,'' in \emph{Design, Automation \& Test in Europe Conference \& Exhibition (DATE), 2017}.\hskip 1em plus 0.5em minus 0.4em\relax IEEE, 2017, pp. 1396--1401.

\bibitem{mao2017mednn}
J.~Mao, Z.~Yang, W.~Wen, C.~Wu, L.~Song, K.~W. Nixon, X.~Chen, H.~Li, and Y.~Chen, ``Mednn: A distributed mobile system with enhanced partition and deployment for large-scale dnns,'' in \emph{2017 IEEE/ACM International Conference on Computer-Aided Design (ICCAD)}.\hskip 1em plus 0.5em minus 0.4em\relax IEEE, 2017, pp. 751--756.

\bibitem{zhao2018deepthings}
Z.~Zhao, K.~M. Barijough, and A.~Gerstlauer, ``Deepthings: Distributed adaptive deep learning inference on resource-constrained iot edge clusters,'' \emph{IEEE Transactions on Computer-Aided Design of Integrated Circuits and Systems}, vol.~37, no.~11, pp. 2348--2359, 2018.

\bibitem{stahl2021deeperthings}
R.~Stahl, A.~Hoffman, D.~Mueller-Gritschneder, A.~Gerstlauer, and U.~Schlichtmann, ``Deeperthings: Fully distributed cnn inference on resource-constrained edge devices,'' \emph{International Journal of Parallel Programming}, vol.~49, no.~4, pp. 600--624, 2021.

\bibitem{zhang2021deep}
W.~Zhang, D.~Yang, H.~Peng, W.~Wu, W.~Quan, H.~Zhang, and X.~Shen, ``Deep reinforcement learning based resource management for dnn inference in industrial iot,'' \emph{IEEE Transactions on Vehicular Technology}, vol.~70, no.~8, pp. 7605--7618, 2021.

\bibitem{zeng2020coedge}
L.~Zeng, X.~Chen, Z.~Zhou, L.~Yang, and J.~Zhang, ``Coedge: Cooperative dnn inference with adaptive workload partitioning over heterogeneous edge devices,'' \emph{IEEE/ACM Transactions on Networking}, vol.~29, no.~2, pp. 595--608, 2020.

\bibitem{hadidi2020toward}
R.~Hadidi, J.~Cao, M.~S. Ryoo, and H.~Kim, ``Toward collaborative inferencing of deep neural networks on internet-of-things devices,'' \emph{IEEE Internet of Things Journal}, vol.~7, no.~6, pp. 4950--4960, 2020.

\bibitem{zhang2021deepslicing}
S.~Zhang, S.~Zhang, Z.~Qian, J.~Wu, Y.~Jin, and S.~Lu, ``Deepslicing: collaborative and adaptive cnn inference with low latency,'' \emph{IEEE Transactions on Parallel and Distributed Systems}, vol.~32, no.~9, pp. 2175--2187, 2021.

\bibitem{pan2010cross}
S.~J. Pan, X.~Ni, J.-T. Sun, Q.~Yang, and Z.~Chen, ``Cross-domain sentiment classification via spectral feature alignment,'' in \emph{Proceedings of the 19th international conference on World wide web}, 2010, pp. 751--760.

\bibitem{chattopadhyay2012multisource}
R.~Chattopadhyay, Q.~Sun, W.~Fan, I.~Davidson, S.~Panchanathan, and J.~Ye, ``Multisource domain adaptation and its application to early detection of fatigue,'' \emph{ACM Transactions on Knowledge Discovery from Data (TKDD)}, vol.~6, no.~4, pp. 1--26, 2012.

\bibitem{duan2012domain}
L.~Duan, I.~W. Tsang, and D.~Xu, ``Domain transfer multiple kernel learning,'' \emph{IEEE Transactions on Pattern Analysis and Machine Intelligence}, vol.~34, no.~3, pp. 465--479, 2012.

\bibitem{oquab2014learning}
M.~Oquab, L.~Bottou, I.~Laptev, and J.~Sivic, ``Learning and transferring mid-level image representations using convolutional neural networks,'' in \emph{Proceedings of the IEEE conference on computer vision and pattern recognition}, 2014, pp. 1717--1724.

\bibitem{wang2022continual}
Q.~Wang, O.~Fink, L.~Van~Gool, and D.~Dai, ``Continual test-time domain adaptation,'' in \emph{Proceedings of the IEEE/CVF Conference on Computer Vision and Pattern Recognition}, 2022, pp. 7201--7211.

\bibitem{shen2022connect}
K.~Shen, R.~M. Jones, A.~Kumar, S.~M. Xie, J.~Z. HaoChen, T.~Ma, and P.~Liang, ``Connect, not collapse: Explaining contrastive learning for unsupervised domain adaptation,'' in \emph{International Conference on Machine Learning}.\hskip 1em plus 0.5em minus 0.4em\relax PMLR, 2022, pp. 19\,847--19\,878.

\bibitem{gal2022stylegan}
R.~Gal, O.~Patashnik, H.~Maron, A.~H. Bermano, G.~Chechik, and D.~Cohen-Or, ``Stylegan-nada: Clip-guided domain adaptation of image generators,'' \emph{ACM Transactions on Graphics (TOG)}, vol.~41, no.~4, pp. 1--13, 2022.

\bibitem{xie2022active}
B.~Xie, L.~Yuan, S.~Li, C.~H. Liu, X.~Cheng, and G.~Wang, ``Active learning for domain adaptation: An energy-based approach,'' in \emph{Proceedings of the AAAI Conference on Artificial Intelligence}, vol.~36, no.~8, 2022, pp. 8708--8716.

\bibitem{xu2020commerce}
F.~Xu, Z.~Pan, and R.~Xia, ``E-commerce product review sentiment classification based on a na{\"\i}ve bayes continuous learning framework,'' \emph{Information Processing \& Management}, vol.~57, no.~5, p. 102221, 2020.

\bibitem{irfan2021novel}
M.~Irfan, Z.~Jiangbin, M.~Iqbal, and M.~H. Arif, ``A novel lifelong learning model based on cross domain knowledge extraction and transfer to classify underwater images,'' \emph{Information Sciences}, vol. 552, pp. 80--101, 2021.

\bibitem{yuan2021reconfigurable}
J.~Yuan, S.~E. Liu, A.~Shylendra, W.~A. Gaviria~Rojas, S.~Guo, H.~Bergeron, S.~Li, H.-S. Lee, S.~Nasrin, V.~K. Sangwan \emph{et~al.}, ``Reconfigurable mos2 memtransistors for continuous learning in spiking neural networks,'' \emph{Nano letters}, vol.~21, no.~15, pp. 6432--6440, 2021.

\bibitem{huang2022learn}
W.~Huang, M.~Ye, and B.~Du, ``Learn from others and be yourself in heterogeneous federated learning,'' in \emph{Proceedings of the IEEE/CVF Conference on Computer Vision and Pattern Recognition}, 2022, pp. 10\,143--10\,153.

\bibitem{fang2022robust}
X.~Fang and M.~Ye, ``Robust federated learning with noisy and heterogeneous clients,'' in \emph{Proceedings of the IEEE/CVF Conference on Computer Vision and Pattern Recognition}, 2022, pp. 10\,072--10\,081.

\bibitem{tang2017train}
W.~Tang, G.~Hua, and L.~Wang, ``How to train a compact binary neural network with high accuracy?'' in \emph{Proceedings of the AAAI conference on artificial intelligence}, vol.~31, no.~1, 2017.

\bibitem{liu2017hierarchical}
H.~Liu, K.~Simonyan, O.~Vinyals, C.~Fernando, and K.~Kavukcuoglu, ``Hierarchical representations for efficient architecture search,'' \emph{arXiv preprint arXiv:1711.00436}, 2017.

\bibitem{tjandra2018tensor}
A.~Tjandra, S.~Sakti, and S.~Nakamura, ``Tensor decomposition for compressing recurrent neural network,'' in \emph{2018 International Joint Conference on Neural Networks (IJCNN)}.\hskip 1em plus 0.5em minus 0.4em\relax IEEE, 2018, pp. 1--8.

\bibitem{zhong2018practical}
Z.~Zhong, J.~Yan, W.~Wu, J.~Shao, and C.-L. Liu, ``Practical block-wise neural network architecture generation,'' in \emph{Proceedings of the IEEE conference on computer vision and pattern recognition}, 2018, pp. 2423--2432.

\bibitem{cai2020tinytl}
H.~Cai, C.~Gan, L.~Zhu, and S.~Han, ``Tinytl: Reduce memory, not parameters for efficient on-device learning,'' \emph{Advances in Neural Information Processing Systems}, vol.~33, pp. 11\,285--11\,297, 2020.

\bibitem{liu2022distributed}
S.~Liu, C.~Zheng, Y.~Huang, and T.~Q. Quek, ``Distributed reinforcement learning for privacy-preserving dynamic edge caching,'' \emph{IEEE Journal on Selected Areas in Communications}, vol.~40, no.~3, pp. 749--760, 2022.

\bibitem{moon2022nntrainer}
J.~J. Moon, P.~Kapoor, J.~H. Lee, M.~J. Ham, and H.~S. Lee, ``Nntrainer: Light-weight on-device training framework,'' \emph{arXiv preprint arXiv:2206.04688}, 2022.

\bibitem{deng2015reduced}
Z.~Deng, C.~Xu, Q.~Cai, P.~Faraboschi, and H.~Packard, ``Reduced-precision memory value approximation for deep learning,'' \emph{Hewlett Packard Labs, HPL-2015-100}, 2015.

\bibitem{bulo2018place}
S.~R. Bulo, L.~Porzi, and P.~Kontschieder, ``In-place activated batchnorm for memory-optimized training of dnns,'' in \emph{Proceedings of the IEEE Conference on Computer Vision and Pattern Recognition}, 2018, pp. 5639--5647.

\bibitem{jung2019restructuring}
W.~Jung, D.~Jung, B.~Kim, S.~Lee, W.~Rhee, and J.~H. Ahn, ``Restructuring batch normalization to accelerate cnn training,'' \emph{Proceedings of Machine Learning and Systems}, vol.~1, pp. 14--26, 2019.

\bibitem{liu2018dynamic}
L.~Liu, L.~Deng, X.~Hu, M.~Zhu, G.~Li, Y.~Ding, and Y.~Xie, ``Dynamic sparse graph for efficient deep learning,'' \emph{arXiv preprint arXiv:1810.00859}, 2018.

\bibitem{dai2020sparsetrain}
P.~Dai, J.~Yang, X.~Ye, X.~Cheng, J.~Luo, L.~Song, Y.~Chen, and W.~Zhao, ``Sparsetrain: Exploiting dataflow sparsity for efficient convolutional neural networks training,'' in \emph{2020 57th ACM/IEEE Design Automation Conference (DAC)}.\hskip 1em plus 0.5em minus 0.4em\relax IEEE, 2020, pp. 1--6.

\bibitem{chen2016training}
T.~Chen, B.~Xu, C.~Zhang, and C.~Guestrin, ``Training deep nets with sublinear memory cost,'' \emph{arXiv preprint arXiv:1604.06174}, 2016.

\bibitem{gomez2017reversible}
A.~N. Gomez, M.~Ren, R.~Urtasun, and R.~B. Grosse, ``The reversible residual network: Backpropagation without storing activations,'' \emph{Advances in neural information processing systems}, vol.~30, 2017.

\bibitem{jain2018gist}
A.~Jain, A.~Phanishayee, J.~Mars, L.~Tang, and G.~Pekhimenko, ``Gist: Efficient data encoding for deep neural network training,'' in \emph{2018 ACM/IEEE 45th Annual International Symposium on Computer Architecture (ISCA)}.\hskip 1em plus 0.5em minus 0.4em\relax IEEE, 2018, pp. 776--789.

\bibitem{lin2021memory}
J.~Lin, W.-M. Chen, H.~Cai, C.~Gan, and S.~Han, ``Memory-efficient patch-based inference for tiny deep learning,'' \emph{Advances in Neural Information Processing Systems}, vol.~34, pp. 2346--2358, 2021.

\bibitem{jia2019optimizing}
Z.~Jia, J.~Thomas, T.~Warszawski, M.~Gao, M.~Zaharia, and A.~Aiken, ``Optimizing dnn computation with relaxed graph substitutions,'' \emph{Proceedings of Machine Learning and Systems}, vol.~1, pp. 27--39, 2019.

\bibitem{zhou2020transferable}
Y.~Zhou, S.~Roy, A.~Abdolrashidi, D.~Wong, P.~Ma, Q.~Xu, H.~Liu, P.~Phothilimtha, S.~Wang, A.~Goldie \emph{et~al.}, ``Transferable graph optimizers for ml compilers,'' \emph{Advances in Neural Information Processing Systems}, vol.~33, pp. 13\,844--13\,855, 2020.

\bibitem{chen2018modnn}
X.~Chen, D.~Z. Chen, and X.~S. Hu, ``modnn: Memory optimal dnn training on gpus,'' in \emph{2018 Design, Automation \& Test in Europe Conference \& Exhibition (DATE)}.\hskip 1em plus 0.5em minus 0.4em\relax IEEE, 2018, pp. 13--18.

\bibitem{huang2020adaptive}
T.~Huang, L.~Tao, and J.~T. Zhou, ``Adaptive precision training for resource constrained devices,'' in \emph{2020 IEEE 40th International Conference on Distributed Computing Systems (ICDCS)}.\hskip 1em plus 0.5em minus 0.4em\relax IEEE, 2020, pp. 1403--1408.

\bibitem{huang2019gpipe}
Y.~Huang, Y.~Cheng, A.~Bapna, O.~Firat, D.~Chen, M.~Chen, H.~Lee, J.~Ngiam, Q.~V. Le, Y.~Wu \emph{et~al.}, ``Gpipe: Efficient training of giant neural networks using pipeline parallelism,'' \emph{Advances in neural information processing systems}, vol.~32, 2019.

\bibitem{lim2020federated}
W.~Y.~B. Lim, N.~C. Luong, D.~T. Hoang, Y.~Jiao, Y.-C. Liang, Q.~Yang, D.~Niyato, and C.~Miao, ``Federated learning in mobile edge networks: A comprehensive survey,'' \emph{IEEE Communications Surveys \& Tutorials}, vol.~22, no.~3, pp. 2031--2063, 2020.

\bibitem{cui2016geeps}
H.~Cui, H.~Zhang, G.~R. Ganger, P.~B. Gibbons, and E.~P. Xing, ``Geeps: Scalable deep learning on distributed gpus with a gpu-specialized parameter server,'' in \emph{Proceedings of the eleventh european conference on computer systems}, 2016, pp. 1--16.

\bibitem{wahib2020scaling}
M.~Wahib, H.~Zhang, T.~T. Nguyen, A.~Drozd, J.~Domke, L.~Zhang, R.~Takano, and S.~Matsuoka, ``Scaling distributed deep learning workloads beyond the memory capacity with karma,'' in \emph{SC20: International Conference for High Performance Computing, Networking, Storage and Analysis}.\hskip 1em plus 0.5em minus 0.4em\relax IEEE, 2020, pp. 1--15.

\bibitem{lim2021zico}
G.~Lim, J.~Ahn, W.~Xiao, Y.~Kwon, and M.~Jeon, ``Zico: Efficient gpu memory sharing for concurrent dnn training.'' in \emph{USENIX Annual Technical Conference}, 2021, pp. 161--175.

\bibitem{samikwa2022ares}
E.~Samikwa, A.~Di~Maio, and T.~Braun, ``Ares: Adaptive resource-aware split learning for internet of things,'' \emph{Computer Networks}, vol. 218, p. 109380, 2022.

\bibitem{onnx}
``Open neural network exchange,'' \url{https://onnx.ai/}.

\bibitem{gudovskiy2018dnn}
D.~Gudovskiy, A.~Hodgkinson, and L.~Rigazio, ``Dnn feature map compression using learned representation over gf (2),'' in \emph{Proceedings of the European Conference on Computer Vision (ECCV) Workshops}, 2018, pp. 0--0.

\bibitem{wu2022compiler}
Y.~Wu, Y.~Gong, P.~Zhao, Y.~Li, Z.~Zhan, W.~Niu, H.~Tang, M.~Qin, B.~Ren, and Y.~Wang, ``Compiler-aware neural architecture search for on-mobile real-time super-resolution,'' in \emph{Computer Vision--ECCV 2022: 17th European Conference, Tel Aviv, Israel, October 23--27, 2022, Proceedings, Part XIX}.\hskip 1em plus 0.5em minus 0.4em\relax Springer, 2022, pp. 92--111.

\bibitem{tai1979constant}
K.-C. Tai, ``Constant folding within an expression by semantic attributes,'' \emph{Computer Languages}, vol.~4, no. 3-4, pp. 131--137, 1979.

\bibitem{glesner2004classifying}
S.~Glesner and J.~O. Blech, ``Classifying and formally verifying integer constant folding,'' \emph{Electronic Notes in Theoretical Computer Science}, vol.~82, no.~2, pp. 410--425, 2004.

\bibitem{cocke1970global}
J.~Cocke, ``Global common subexpression elimination,'' in \emph{Proceedings of a symposium on Compiler optimization}, 1970, pp. 20--24.

\bibitem{elgamal2017spoof}
T.~Elgamal, S.~Luo, M.~Boehm, A.~V. Evfimievski, S.~Tatikonda, B.~Reinwald, and P.~Sen, ``Spoof: Sum-product optimization and operator fusion for large-scale machine learning.'' in \emph{CIDR}, 2017.

\bibitem{wang2022melon}
Q.~Wang, M.~Xu, C.~Jin, X.~Dong, J.~Yuan, X.~Jin, G.~Huang, Y.~Liu, and X.~Liu, ``Melon: Breaking the memory wall for resource-efficient on-device machine learning,'' in \emph{Proceedings of the 20th Annual International Conference on Mobile Systems, Applications and Services}, 2022, pp. 450--463.

\bibitem{fenske2006top}
M.~J. Fenske, E.~Aminoff, N.~Gronau, and M.~Bar, ``Top-down facilitation of visual object recognition: object-based and context-based contributions,'' \emph{Progress in brain research}, vol. 155, pp. 3--21, 2006.

\bibitem{lin2001reducing}
W.-F. Lin, S.~K. Reinhardt, and D.~Burger, ``Reducing dram latencies with an integrated memory hierarchy design,'' in \emph{Proceedings HPCA Seventh International Symposium on High-Performance Computer Architecture}.\hskip 1em plus 0.5em minus 0.4em\relax IEEE, 2001, pp. 301--312.

\bibitem{raspberry}
``Raspberry pi series processors,'' \url{https://www.raspberrypi.com/products/}.

\bibitem{mcu}
``Microcontrollers and microprocessors,'' \url{https://www.microchip.com/en-us/products/microcontrollers-and-microprocessors}.

\bibitem{liu2023adaenlight}
S.~Liu, X.~Li, Z.~Zhou, B.~Guo, M.~Zhang, H.~Shen, and Z.~Yu, ``Adaenlight: Energy-aware low-light video stream enhancement on mobile devices,'' \emph{Proceedings of the ACM on Interactive, Mobile, Wearable and Ubiquitous Technologies}, vol.~6, no.~4, pp. 1--26, 2023.

\bibitem{lai2021opportunity}
J.-W. Lai, ``Opportunity and challenge of chiplet-based hpc and aiot,'' in \emph{2021 International Symposium on VLSI Design, Automation and Test (VLSI-DAT)}.\hskip 1em plus 0.5em minus 0.4em\relax IEEE, 2021, pp. 1--2.

\bibitem{zhang2021nn}
L.~L. Zhang, S.~Han, J.~Wei, N.~Zheng, T.~Cao, Y.~Yang, and Y.~Liu, ``Nn-meter: Towards accurate latency prediction of deep-learning model inference on diverse edge devices,'' in \emph{Proceedings of the 19th Annual International Conference on Mobile Systems, Applications, and Services}, 2021, pp. 81--93.

\bibitem{venieris2017latency}
S.~I. Venieris and C.-S. Bouganis, ``Latency-driven design for fpga-based convolutional neural networks,'' in \emph{2017 27th International Conference on Field Programmable Logic and Applications (FPL)}.\hskip 1em plus 0.5em minus 0.4em\relax IEEE, 2017, pp. 1--8.

\bibitem{jha2020modeling}
N.~K. Jha and S.~Mittal, ``Modeling data reuse in deep neural networks by taking data-types into cognizance,'' \emph{IEEE Transactions on Computers}, vol.~70, no.~9, pp. 1526--1538, 2020.

\bibitem{jha2019ramifications}
N.~K. Jha, S.~Mittal, and G.~Mattela, ``The ramifications of making deep neural networks compact,'' in \emph{2019 32nd International Conference on VLSI Design and 2019 18th International Conference on Embedded Systems (VLSID)}.\hskip 1em plus 0.5em minus 0.4em\relax IEEE, 2019, pp. 215--220.

\bibitem{iandola2016squeezenet}
F.~N. Iandola, S.~Han, M.~W. Moskewicz, K.~Ashraf, W.~J. Dally, and K.~Keutzer, ``Squeezenet: Alexnet-level accuracy with 50x fewer parameters and< 0.5 mb model size,'' \emph{arXiv preprint arXiv:1602.07360}, 2016.

\bibitem{wang2021context}
H.~Wang, B.~Guo, J.~Liu, S.~Liu, Y.~Wu, and Z.~Yu, ``Context-aware adaptive surgery: A fast and effective framework for adaptative model partition,'' \emph{Proceedings of the ACM on Interactive, Mobile, Wearable and Ubiquitous Technologies}, vol.~5, no.~3, pp. 1--22, 2021.

\bibitem{szegedy2017inception}
C.~Szegedy, S.~Ioffe, V.~Vanhoucke, and A.~Alemi, ``Inception-v4, inception-resnet and the impact of residual connections on learning,'' in \emph{Proceedings of the AAAI conference on artificial intelligence}, vol.~31, no.~1, 2017.

\bibitem{zagoruyko2016wide}
S.~Zagoruyko and N.~Komodakis, ``Wide residual networks,'' \emph{arXiv preprint arXiv:1605.07146}, 2016.

\bibitem{cheng2017survey}
Y.~Cheng, D.~Wang, P.~Zhou, and T.~Zhang, ``A survey of model compression and acceleration for deep neural networks,'' \emph{arXiv preprint arXiv:1710.09282}, 2017.

\bibitem{cheng2018recent}
J.~Cheng, P.-s. Wang, G.~Li, Q.-h. Hu, and H.-q. Lu, ``Recent advances in efficient computation of deep convolutional neural networks,'' \emph{Frontiers of Information Technology \& Electronic Engineering}, vol.~19, pp. 64--77, 2018.

\bibitem{deng2019deep}
Y.~Deng, ``Deep learning on mobile devices: a review,'' in \emph{Mobile Multimedia/Image Processing, Security, and Applications 2019}, vol. 10993.\hskip 1em plus 0.5em minus 0.4em\relax SPIE, 2019, pp. 52--66.

\bibitem{choudhary2020comprehensive}
T.~Choudhary, V.~Mishra, A.~Goswami, and J.~Sarangapani, ``A comprehensive survey on model compression and acceleration,'' \emph{Artificial Intelligence Review}, vol.~53, pp. 5113--5155, 2020.

\bibitem{he2018multi}
X.~He, Z.~Zhou, and L.~Thiele, ``Multi-task zipping via layer-wise neuron sharing,'' in \emph{Advances in Neural Information Processing Systems}, 2018, pp. 6019--6029.

\bibitem{gao2021pruning}
D.~Gao, X.~He, Z.~Zhou, Y.~Tong, and L.~Thiele, ``Pruning meta-trained networks for on-device adaptation,'' in \emph{Proceedings of the ACM International Conference on Information \& Knowledge Management}, 2021, pp. 514--523.

\bibitem{srinivas2015data}
S.~Srinivas and R.~V. Babu, ``Data-free parameter pruning for deep neural networks,'' \emph{arXiv preprint arXiv:1507.06149}, 2015.

\bibitem{han2015deep}
S.~Han, H.~Mao, and W.~J. Dally, ``Deep compression: Compressing deep neural networks with pruning, trained quantization and huffman coding,'' \emph{arXiv preprint arXiv:1510.00149}, 2015.

\bibitem{jaderberg2014speeding}
M.~Jaderberg, A.~Vedaldi, and A.~Zisserman, ``Speeding up convolutional neural networks with low rank expansions,'' \emph{arXiv preprint arXiv:1405.3866}, 2014.

\bibitem{tai2015convolutional}
C.~Tai, T.~Xiao, Y.~Zhang, X.~Wang \emph{et~al.}, ``Convolutional neural networks with low-rank regularization,'' \emph{arXiv preprint arXiv:1511.06067}, 2015.

\bibitem{lin2018holistic}
S.~Lin, R.~Ji, C.~Chen, D.~Tao, and J.~Luo, ``Holistic cnn compression via low-rank decomposition with knowledge transfer,'' \emph{IEEE transactions on pattern analysis and machine intelligence}, vol.~41, no.~12, pp. 2889--2905, 2018.

\bibitem{mehta2019espnetv2}
S.~Mehta, M.~Rastegari, L.~Shapiro, and H.~Hajishirzi, ``Espnetv2: A light-weight, power efficient, and general purpose convolutional neural network,'' in \emph{Proceedings of the IEEE/CVF conference on computer vision and pattern recognition}, 2019, pp. 9190--9200.

\bibitem{tan2019mnasnet}
M.~Tan, B.~Chen, R.~Pang, V.~Vasudevan, M.~Sandler, A.~Howard, and Q.~V. Le, ``Mnasnet: Platform-aware neural architecture search for mobile,'' in \emph{Proceedings of the IEEE/CVF conference on computer vision and pattern recognition}, 2019, pp. 2820--2828.

\bibitem{lin2013network}
M.~Lin, Q.~Chen, and S.~Yan, ``Network in network,'' \emph{arXiv preprint arXiv:1312.4400}, 2013.

\bibitem{gong2014compressing}
Y.~Gong, L.~Liu, M.~Yang, and L.~Bourdev, ``Compressing deep convolutional networks using vector quantization,'' \emph{arXiv preprint arXiv:1412.6115}, 2014.

\bibitem{wu2016quantized}
J.~Wu, C.~Leng, Y.~Wang, Q.~Hu, and J.~Cheng, ``Quantized convolutional neural networks for mobile devices,'' in \emph{Proceedings of the IEEE conference on computer vision and pattern recognition}, 2016, pp. 4820--4828.

\bibitem{courbariaux2015binaryconnect}
M.~Courbariaux, Y.~Bengio, and J.-P. David, ``Binaryconnect: Training deep neural networks with binary weights during propagations,'' \emph{Advances in neural information processing systems}, vol.~28, 2015.

\bibitem{li2017performance}
Z.~Li, B.~Ni, W.~Zhang, X.~Yang, and W.~Gao, ``Performance guaranteed network acceleration via high-order residual quantization,'' in \emph{Proceedings of the IEEE international conference on computer vision}, 2017, pp. 2584--2592.

\bibitem{chow1984portable}
F.~C.-T. Chow, \emph{A portable machine-independent global optimizer--Design and measurements}.\hskip 1em plus 0.5em minus 0.4em\relax Stanford University, 1984.

\bibitem{fang2020optimizing}
J.~Fang, Y.~Shen, Y.~Wang, and L.~Chen, ``Optimizing dnn computation graph using graph substitutions,'' \emph{Proceedings of the VLDB Endowment}, vol.~13, no.~12, pp. 2734--2746, 2020.

\bibitem{paszke2019pytorch}
A.~Paszke, S.~Gross, F.~Massa, A.~Lerer, J.~Bradbury, G.~Chanan, T.~Killeen, Z.~Lin, N.~Gimelshein, L.~Antiga \emph{et~al.}, ``Pytorch: An imperative style, high-performance deep learning library,'' \emph{Advances in neural information processing systems}, vol.~32, 2019.

\bibitem{wei2019overcoming}
X.~Wei, Y.~Liang, and J.~Cong, ``Overcoming data transfer bottlenecks in fpga-based dnn accelerators via layer conscious memory management,'' in \emph{Proceedings of the 56th Annual Design Automation Conference 2019}, 2019, pp. 1--6.

\bibitem{symons2021loma}
A.~Symons, L.~Mei, and M.~Verhelst, ``Loma: Fast auto-scheduling on dnn accelerators through loop-order-based memory allocation,'' in \emph{2021 IEEE 3rd International Conference on Artificial Intelligence Circuits and Systems (AICAS)}.\hskip 1em plus 0.5em minus 0.4em\relax IEEE, 2021, pp. 1--4.

\bibitem{ji2022task}
C.~Ji, Z.~Zhu, X.~Wang, W.~Zhai, X.~Zong, A.~Chen, and M.~Zhou, ``Task-aware swapping for efficient dnn inference on dram-constrained edge systems,'' \emph{International Journal of Intelligent Systems}, vol.~37, no.~10, pp. 8155--8169, 2022.

\bibitem{murthy2010optimal}
G.~S. Murthy, M.~Ravishankar, M.~M. Baskaran, and P.~Sadayappan, ``Optimal loop unrolling for gpgpu programs,'' in \emph{2010 IEEE International Symposium on Parallel \& Distributed Processing (IPDPS)}.\hskip 1em plus 0.5em minus 0.4em\relax IEEE, 2010, pp. 1--11.

\bibitem{vanholder2016efficient}
H.~Vanholder, ``Efficient inference with tensorrt,'' in \emph{GPU Technology Conference}, vol.~1, 2016, p.~2.

\bibitem{jiang2020mnn}
X.~Jiang, H.~Wang, Y.~Chen, Z.~Wu, L.~Wang, B.~Zou, Y.~Yang, Z.~Cui, Y.~Cai, T.~Yu \emph{et~al.}, ``Mnn: A universal and efficient inference engine,'' \emph{Proceedings of Machine Learning and Systems}, vol.~2, pp. 1--13, 2020.

\bibitem{jia2019taso}
Z.~Jia, O.~Padon, J.~Thomas, T.~Warszawski, M.~Zaharia, and A.~Aiken, ``Taso: optimizing deep learning computation with automatic generation of graph substitutions,'' in \emph{Proceedings of the 27th ACM Symposium on Operating Systems Principles}, 2019, pp. 47--62.

\bibitem{sekiyama2018profile}
T.~Sekiyama, T.~Imamichi, H.~Imai, and R.~Raymond, ``Profile-guided memory optimization for deep neural networks,'' \emph{arXiv preprint arXiv:1804.10001}, 2018.

\bibitem{huang2020swapadvisor}
C.-C. Huang, G.~Jin, and J.~Li, ``Swapadvisor: Pushing deep learning beyond the gpu memory limit via smart swapping,'' in \emph{Proceedings of the Twenty-Fifth International Conference on Architectural Support for Programming Languages and Operating Systems}, 2020, pp. 1341--1355.

\bibitem{rawat2018associative}
P.~S. Rawat, A.~Sukumaran-Rajam, A.~Rountev, F.~Rastello, L.-N. Pouchet, and P.~Sadayappan, ``Associative instruction reordering to alleviate register pressure,'' in \emph{SC18: International Conference for High Performance Computing, Networking, Storage and Analysis}.\hskip 1em plus 0.5em minus 0.4em\relax IEEE, 2018, pp. 590--602.

\bibitem{venkatesan2019magnet}
R.~Venkatesan, Y.~S. Shao, M.~Wang, J.~Clemons, S.~Dai, M.~Fojtik, B.~Keller, A.~Klinefelter, N.~Pinckney, P.~Raina \emph{et~al.}, ``Magnet: A modular accelerator generator for neural networks,'' in \emph{2019 IEEE/ACM International Conference on Computer-Aided Design (ICCAD)}.\hskip 1em plus 0.5em minus 0.4em\relax IEEE, 2019, pp. 1--8.

\bibitem{shen2017maximizing}
Y.~Shen, M.~Ferdman, and P.~Milder, ``Maximizing cnn accelerator efficiency through resource partitioning,'' \emph{ACM SIGARCH Computer Architecture News}, vol.~45, no.~2, pp. 535--547, 2017.

\bibitem{parashar2019timeloop}
A.~Parashar, P.~Raina, Y.~S. Shao, Y.-H. Chen, V.~A. Ying, A.~Mukkara, R.~Venkatesan, B.~Khailany, S.~W. Keckler, and J.~Emer, ``Timeloop: A systematic approach to dnn accelerator evaluation,'' in \emph{2019 IEEE international symposium on performance analysis of systems and software (ISPASS)}.\hskip 1em plus 0.5em minus 0.4em\relax IEEE, 2019, pp. 304--315.

\bibitem{stoutchinin2019optimally}
A.~Stoutchinin, F.~Conti, and L.~Benini, ``Optimally scheduling cnn convolutions for efficient memory access,'' \emph{arXiv preprint arXiv:1902.01492}, 2019.

\bibitem{peng2020capuchin}
X.~Peng, X.~Shi, H.~Dai, H.~Jin, W.~Ma, Q.~Xiong, F.~Yang, and X.~Qian, ``Capuchin: Tensor-based gpu memory management for deep learning,'' in \emph{Proceedings of the Twenty-Fifth International Conference on Architectural Support for Programming Languages and Operating Systems}, 2020, pp. 891--905.

\bibitem{yang2017designing}
T.-J. Yang, Y.-H. Chen, and V.~Sze, ``Designing energy-efficient convolutional neural networks using energy-aware pruning,'' in \emph{Proceedings of the IEEE conference on computer vision and pattern recognition}, 2017, pp. 5687--5695.

\bibitem{farley2021memory}
J.~Farley and A.~Gerstlauer, ``Memory-aware fusing and tiling of neural networks for accelerated edge inference,'' \emph{arXiv preprint arXiv:2107.06960}, 2021.

\bibitem{stahl2019fully}
R.~Stahl, Z.~Zhao, D.~Mueller-Gritschneder, A.~Gerstlauer, and U.~Schlichtmann, ``Fully distributed deep learning inference on resource-constrained edge devices,'' in \emph{International Conference on Embedded Computer Systems}.\hskip 1em plus 0.5em minus 0.4em\relax Springer, 2019, pp. 77--90.

\bibitem{naveen2022memory}
S.~Naveen and M.~R. Kounte, ``Memory optimization at edge for distributed convolution neural network,'' \emph{Transactions on Emerging Telecommunications Technologies}, vol.~33, no.~12, p. e4648, 2022.

\bibitem{zeng2019boomerang}
L.~Zeng, E.~Li, Z.~Zhou, and X.~Chen, ``Boomerang: On-demand cooperative deep neural network inference for edge intelligence on the industrial internet of things,'' \emph{IEEE Network}, vol.~33, no.~5, pp. 96--103, 2019.

\bibitem{chen2019iraf}
J.~Chen, S.~Chen, Q.~Wang, B.~Cao, G.~Feng, and J.~Hu, ``iraf: A deep reinforcement learning approach for collaborative mobile edge computing iot networks,'' \emph{IEEE Internet of Things Journal}, vol.~6, no.~4, pp. 7011--7024, 2019.

\bibitem{zhang2020towards}
S.~Zhang, Y.~Li, X.~Liu, S.~Guo, W.~Wang, J.~Wang, B.~Ding, and D.~Wu, ``Towards real-time cooperative deep inference over the cloud and edge end devices,'' \emph{Proceedings of the ACM on Interactive, Mobile, Wearable and Ubiquitous Technologies}, vol.~4, no.~2, pp. 1--24, 2020.

\bibitem{pan2022joint}
G.~Pan, H.~Zhang, S.~Xu, S.~Zhang, and X.~Chen, ``Joint optimization of dnn inference delay and energy under accuracy constraints for ar applications,'' in \emph{GLOBECOM 2022-2022 IEEE Global Communications Conference}.\hskip 1em plus 0.5em minus 0.4em\relax IEEE, 2022, pp. 2230--2235.

\bibitem{li2021throughput}
J.~Li, W.~Liang, Y.~Li, Z.~Xu, X.~Jia, and S.~Guo, ``Throughput maximization of delay-aware dnn inference in edge computing by exploring dnn model partitioning and inference parallelism,'' \emph{IEEE Transactions on Mobile Computing}, 2021.

\bibitem{jeong2018ionn}
H.-J. Jeong, H.-J. Lee, C.~H. Shin, and S.-M. Moon, ``Ionn: Incremental offloading of neural network computations from mobile devices to edge servers,'' in \emph{Proceedings of the ACM symposium on cloud computing}, 2018, pp. 401--411.

\bibitem{eshratifar2019jointdnn}
A.~E. Eshratifar, M.~S. Abrishami, and M.~Pedram, ``Jointdnn: An efficient training and inference engine for intelligent mobile cloud computing services,'' \emph{IEEE Transactions on Mobile Computing}, vol.~20, no.~2, pp. 565--576, 2019.

\bibitem{ma2018optimizing}
Y.~Ma, Y.~Cao, S.~Vrudhula, and J.-s. Seo, ``Optimizing the convolution operation to accelerate deep neural networks on fpga,'' \emph{IEEE Transactions on Very Large Scale Integration (VLSI) Systems}, vol.~26, no.~7, pp. 1354--1367, 2018.

\bibitem{laskaridis2020spinn}
S.~Laskaridis, S.~I. Venieris, M.~Almeida, I.~Leontiadis, and N.~D. Lane, ``Spinn: synergistic progressive inference of neural networks over device and cloud,'' in \emph{Proceedings of the 26th annual international conference on mobile computing and networking}, 2020, pp. 1--15.

\bibitem{zhang2020deep}
F.~Zhang, J.~Fang, B.~Wah, and P.~Torr, ``Deep fusionnet for point cloud semantic segmentation,'' in \emph{European Conference on Computer Vision}.\hskip 1em plus 0.5em minus 0.4em\relax Springer, 2020, pp. 644--663.

\bibitem{li2018edge}
E.~Li, Z.~Zhou, and X.~Chen, ``Edge intelligence: On-demand deep learning model co-inference with device-edge synergy,'' in \emph{Proceedings of the 2018 Workshop on Mobile Edge Communications}, 2018, pp. 31--36.

\bibitem{li2019edge}
E.~Li, L.~Zeng, Z.~Zhou, and X.~Chen, ``Edge ai: On-demand accelerating deep neural network inference via edge computing,'' \emph{IEEE Transactions on Wireless Communications}, vol.~19, no.~1, pp. 447--457, 2019.

\bibitem{ioffe2015batch}
S.~Ioffe and C.~Szegedy, ``Batch normalization: Accelerating deep network training by reducing internal covariate shift,'' in \emph{International conference on machine learning}.\hskip 1em plus 0.5em minus 0.4em\relax PMLR, 2015, pp. 448--456.

\bibitem{yang2022rep}
L.~Yang, A.~S. Rakin, and D.~Fan, ``Rep-net: Efficient on-device learning via feature reprogramming,'' in \emph{Proceedings of the IEEE/CVF Conference on Computer Vision and Pattern Recognition}, 2022, pp. 12\,277--12\,286.

\bibitem{zhou2016dorefa}
S.~Zhou, Y.~Wu, Z.~Ni, X.~Zhou, H.~Wen, and Y.~Zou, ``Dorefa-net: Training low bitwidth convolutional neural networks with low bitwidth gradients,'' \emph{arXiv preprint arXiv:1606.06160}, 2016.

\bibitem{friesen2017deep}
A.~L. Friesen and P.~Domingos, ``Deep learning as a mixed convex-combinatorial optimization problem,'' \emph{arXiv preprint arXiv:1710.11573}, 2017.

\bibitem{zhuang2018towards}
B.~Zhuang, C.~Shen, M.~Tan, L.~Liu, and I.~Reid, ``Towards effective low-bitwidth convolutional neural networks,'' in \emph{Proceedings of the IEEE conference on computer vision and pattern recognition}, 2018, pp. 7920--7928.

\bibitem{finkelstein2019fighting}
A.~Finkelstein, U.~Almog, and M.~Grobman, ``Fighting quantization bias with bias,'' \emph{arXiv preprint arXiv:1906.03193}, 2019.

\bibitem{micikevicius2017mixed}
P.~Micikevicius, S.~Narang, J.~Alben, G.~Diamos, E.~Elsen, D.~Garcia, B.~Ginsburg, M.~Houston, O.~Kuchaiev, G.~Venkatesh \emph{et~al.}, ``Mixed precision training,'' \emph{arXiv preprint arXiv:1710.03740}, 2017.

\bibitem{gupta2015deep}
S.~Gupta, A.~Agrawal, K.~Gopalakrishnan, and P.~Narayanan, ``Deep learning with limited numerical precision,'' in \emph{International conference on machine learning}.\hskip 1em plus 0.5em minus 0.4em\relax PMLR, 2015, pp. 1737--1746.

\bibitem{wang2022towards}
G.~Wang, Z.~Liu, Z.~Jiang, N.~Liu, N.~Zou, and X.~Hu, ``Towards memory efficient training via dual activation precision,'' \emph{arXiv preprint arXiv:2208.04187}, 2022.

\bibitem{yu2019tf}
J.~Yu, A.~Lukefahr, R.~Das, and S.~Mahlke, ``Tf-net: Deploying sub-byte deep neural networks on microcontrollers,'' \emph{ACM Transactions on Embedded Computing Systems (TECS)}, vol.~18, no.~5s, pp. 1--21, 2019.

\bibitem{lu2023quantization}
Q.~Lu, W.~Jiang, X.~Xu, J.~Hu, and Y.~Shi, ``Quantization through search: A novel scheme to quantize convolutional neural networks in finite weight space,'' in \emph{Proceedings of the 28th Asia and South Pacific Design Automation Conference}, 2023, pp. 378--383.

\bibitem{he2016deep}
K.~He, X.~Zhang, S.~Ren, and J.~Sun, ``Deep residual learning for image recognition,'' in \emph{Proceedings of the IEEE conference on computer vision and pattern recognition}, 2016, pp. 770--778.

\bibitem{huang2017densely}
G.~Huang, Z.~Liu, L.~Van Der~Maaten, and K.~Q. Weinberger, ``Densely connected convolutional networks,'' in \emph{Proceedings of the IEEE conference on computer vision and pattern recognition}, 2017, pp. 4700--4708.

\bibitem{kaplun2023subtuning}
G.~Kaplun, A.~Gurevich, T.~Swisa, M.~David, S.~Shalev-Shwartz, and E.~Malach, ``Subtuning: Efficient finetuning for multi-task learning,'' \emph{arXiv preprint arXiv:2302.06354}, 2023.

\bibitem{gruslys2016memory}
A.~Gruslys, R.~Munos, I.~Danihelka, M.~Lanctot, and A.~Graves, ``Memory-efficient backpropagation through time,'' \emph{Advances in neural information processing systems}, vol.~29, 2016.

\bibitem{kirisame2020dynamic}
M.~Kirisame, S.~Lyubomirsky, A.~Haan, J.~Brennan, M.~He, J.~Roesch, T.~Chen, and Z.~Tatlock, ``Dynamic tensor rematerialization,'' \emph{arXiv preprint arXiv:2006.09616}, 2020.

\bibitem{gim2022memory}
I.~Gim and J.~Ko, ``Memory-efficient dnn training on mobile devices,'' in \emph{Proceedings of the 20th Annual International Conference on Mobile Systems, Applications and Services}, 2022, pp. 464--476.

\bibitem{evans2020jpeg}
R.~D. Evans, L.~Liu, and T.~M. Aamodt, ``Jpeg-act: accelerating deep learning via transform-based lossy compression,'' in \emph{2020 ACM/IEEE 47th Annual International Symposium on Computer Architecture (ISCA)}.\hskip 1em plus 0.5em minus 0.4em\relax IEEE, 2020, pp. 860--873.

\bibitem{hosny2021sparse}
A.~Hosny, M.~Neseem, and S.~Reda, ``Sparse bitmap compression for memory-efficient training on the edge,'' in \emph{2021 IEEE/ACM Symposium on Edge Computing (SEC)}.\hskip 1em plus 0.5em minus 0.4em\relax IEEE, 2021, pp. 14--25.

\bibitem{unger2022unity}
C.~Unger, Z.~Jia, W.~Wu, S.~Lin, M.~Baines, C.~E.~Q. Narvaez, V.~Ramakrishnaiah, N.~Prajapati, P.~McCormick, J.~Mohd-Yusof \emph{et~al.}, ``Unity: Accelerating $\{$DNN$\}$ training through joint optimization of algebraic transformations and parallelization,'' in \emph{16th USENIX Symposium on Operating Systems Design and Implementation (OSDI 22)}, 2022, pp. 267--284.

\bibitem{hu2020jittor}
S.-M. Hu, D.~Liang, G.-Y. Yang, G.-W. Yang, and W.-Y. Zhou, ``Jittor: a novel deep learning framework with meta-operators and unified graph execution,'' \emph{Science China Information Sciences}, vol.~63, pp. 1--21, 2020.

\bibitem{zheng2020fusionstitching}
Z.~Zheng, P.~Zhao, G.~Long, F.~Zhu, K.~Zhu, W.~Zhao, L.~Diao, J.~Yang, and W.~Lin, ``Fusionstitching: boosting memory intensive computations for deep learning workloads,'' \emph{arXiv preprint arXiv:2009.10924}, 2020.

\bibitem{ivanov2021data}
A.~Ivanov, N.~Dryden, T.~Ben-Nun, S.~Li, and T.~Hoefler, ``Data movement is all you need: A case study on optimizing transformers,'' \emph{Proceedings of Machine Learning and Systems}, vol.~3, pp. 711--732, 2021.

\bibitem{zheng2022astitch}
Z.~Zheng, X.~Yang, P.~Zhao, G.~Long, K.~Zhu, F.~Zhu, W.~Zhao, X.~Liu, J.~Yang, J.~Zhai \emph{et~al.}, ``Astitch: enabling a new multi-dimensional optimization space for memory-intensive ml training and inference on modern simt architectures,'' in \emph{Proceedings of the 27th ACM International Conference on Architectural Support for Programming Languages and Operating Systems}, 2022, pp. 359--373.

\bibitem{nie2022tsplit}
X.~Nie, X.~Miao, Z.~Yang, and B.~Cui, ``Tsplit: Fine-grained gpu memory management for efficient dnn training via tensor splitting,'' in \emph{2022 IEEE 38th International Conference on Data Engineering (ICDE)}, 2022, pp. 2615--2628.

\bibitem{le2019automatic}
T.~D. Le, H.~Imai, Y.~Negishi, and K.~Kawachiya, ``Automatic gpu memory management for large neural models in tensorflow,'' in \emph{Proceedings of the 2019 ACM SIGPLAN International Symposium on Memory Management}, 2019, pp. 1--13.

\bibitem{shriram2019dynamic}
S.~Shriram, A.~Garg, and P.~Kulkarni, ``Dynamic memory management for gpu-based training of deep neural networks,'' in \emph{2019 IEEE International Parallel and Distributed Processing Symposium (IPDPS)}.\hskip 1em plus 0.5em minus 0.4em\relax IEEE, 2019, pp. 200--209.

\bibitem{ren2021sentinel}
J.~Ren, J.~Luo, K.~Wu, M.~Zhang, H.~Jeon, and D.~Li, ``Sentinel: Efficient tensor migration and allocation on heterogeneous memory systems for deep learning,'' in \emph{2021 IEEE International Symposium on High-Performance Computer Architecture (HPCA)}.\hskip 1em plus 0.5em minus 0.4em\relax IEEE, 2021, pp. 598--611.

\bibitem{chen2021cswap}
P.~Chen, S.~He, X.~Zhang, S.~Chen, P.~Hong, Y.~Yin, X.-H. Sun, and G.~Chen, ``Cswap: A self-tuning compression framework for accelerating tensor swapping in gpus,'' in \emph{2021 IEEE International Conference on Cluster Computing (CLUSTER)}.\hskip 1em plus 0.5em minus 0.4em\relax IEEE, 2021, pp. 271--282.

\bibitem{li2022application}
J.~Li, X.~Wang, X.~Chen, G.~Li, X.~Dong, P.~Zhao, X.~Yu, Y.~Yang, W.~Cao, L.~Liu \emph{et~al.}, ``An application-oblivious memory scheduling system for dnn accelerators,'' \emph{ACM Transactions on Architecture and Code Optimization (TACO)}, vol.~19, no.~4, pp. 1--26, 2022.

\bibitem{deepum2023jaehoon}
\BIBentryALTinterwordspacing
J.~Jung, J.~Kim, and J.~Lee, ``Deepum: Tensor migration and prefetching in unified memory,'' in \emph{Proceedings of the 28th ACM International Conference on Architectural Support for Programming Languages and Operating Systems, Volume 2}, ser. ASPLOS 2023.\hskip 1em plus 0.5em minus 0.4em\relax New York, NY, USA: Association for Computing Machinery, 2023, p. 207–221. [Online]. Available: \url{https://doi.org/10.1145/3575693.3575736}
\BIBentrySTDinterwordspacing

\bibitem{roesch2019relay}
J.~Roesch, S.~Lyubomirsky, M.~Kirisame, L.~Weber, J.~Pollock, L.~Vega, Z.~Jiang, T.~Chen, T.~Moreau, and Z.~Tatlock, ``Relay: A high-level compiler for deep learning,'' \emph{arXiv preprint arXiv:1904.08368}, 2019.

\bibitem{ncnn}
Tencent, ``Ncnn,'' \url{https://github.com/Tencent/ncnn/}.

\bibitem{mindspore}
Huawei, ``Mindspore,'' \url{https://github.com/mindspore-ai/mindspore}.

\bibitem{mirhoseini2018hierarchical}
A.~Mirhoseini, A.~Goldie, H.~Pham, B.~Steiner, Q.~V. Le, and J.~Dean, ``Hierarchical planning for device placement,'' 2018.

\bibitem{gao2018spotlight}
Y.~Gao, L.~Chen, and B.~Li, ``Spotlight: Optimizing device placement for training deep neural networks,'' in \emph{International Conference on Machine Learning}.\hskip 1em plus 0.5em minus 0.4em\relax PMLR, 2018, pp. 1676--1684.

\bibitem{addanki2019placeto}
R.~Addanki, S.~B. Venkatakrishnan, S.~Gupta, H.~Mao, and M.~Alizadeh, ``Placeto: Learning generalizable device placement algorithms for distributed machine learning,'' \emph{arXiv preprint arXiv:1906.08879}, 2019.

\bibitem{bortfeldt2006genetic}
A.~Bortfeldt, ``A genetic algorithm for the two-dimensional strip packing problem with rectangular pieces,'' \emph{European Journal of Operational Research}, vol. 172, no.~3, pp. 814--837, 2006.

\bibitem{MSmemoryswapping}
``Windows virtual address spaces,'' \url{https://learn.microsoft.com/en-us/windows-hardware/drivers/gettingstarted/virtual-address-spaces}.

\bibitem{Linuxmemoryswapping}
``The linux kernel documentation,'' \url{https://www.kernel.org/doc/html/latest/index.html#}.

\bibitem{zoph2018learning}
B.~Zoph, V.~Vasudevan, J.~Shlens, and Q.~V. Le, ``Learning transferable architectures for scalable image recognition,'' in \emph{Proceedings of the IEEE conference on computer vision and pattern recognition}, 2018, pp. 8697--8710.

\bibitem{schuster1997bidirectional}
M.~Schuster and K.~K. Paliwal, ``Bidirectional recurrent neural networks,'' \emph{IEEE transactions on Signal Processing}, vol.~45, no.~11, pp. 2673--2681, 1997.

\bibitem{narayanan2021memory}
D.~Narayanan, A.~Phanishayee, K.~Shi, X.~Chen, and M.~Zaharia, ``Memory-efficient pipeline-parallel dnn training,'' in \emph{Proceedings of ICML}.\hskip 1em plus 0.5em minus 0.4em\relax PMLR, 2021, pp. 7937--7947.

\bibitem{liu2023map}
Y.~Liu, S.~Li, J.~Fang, Y.~Shao, B.~Yao, and Y.~You, ``Map: Memory-aware automated intra-op parallel training for foundation models,'' \emph{arXiv preprint arXiv:2302.02599}, 2023.

\bibitem{mcdonald2009efficient}
R.~Mcdonald, M.~Mohri, N.~Silberman, D.~Walker, and G.~Mann, ``Efficient large-scale distributed training of conditional maximum entropy models,'' \emph{Advances in neural information processing systems}, vol.~22, 2009.

\bibitem{agarwal2014reliable}
A.~Agarwal, O.~Chapelle, M.~Dud{\'\i}k, and J.~Langford, ``A reliable effective terascale linear learning system,'' \emph{The Journal of Machine Learning Research}, vol.~15, no.~1, pp. 1111--1133, 2014.

\bibitem{wei2015managed}
J.~Wei, W.~Dai, A.~Qiao, Q.~Ho, H.~Cui, G.~R. Ganger, P.~B. Gibbons, G.~A. Gibson, and E.~P. Xing, ``Managed communication and consistency for fast data-parallel iterative analytics,'' in \emph{Proceedings of the Sixth ACM Symposium on Cloud Computing}, 2015, pp. 381--394.

\bibitem{li2013parameter}
M.~Li, L.~Zhou, Z.~Yang, A.~Li, F.~Xia, D.~G. Andersen, and A.~Smola, ``Parameter server for distributed machine learning,'' in \emph{Big learning NIPS workshop}, vol.~6, no.~2, 2013.

\bibitem{beaumont2022madpipe}
O.~Beaumont, L.~Eyraud-Dubois, and A.~Shilova, ``Madpipe: Memory aware dynamic programming algorithm for pipelined model parallelism,'' in \emph{2022 IEEE International Parallel and Distributed Processing Symposium Workshops (IPDPSW)}.\hskip 1em plus 0.5em minus 0.4em\relax IEEE, 2022, pp. 1063--1073.

\bibitem{xie2019asynchronous}
C.~Xie, S.~Koyejo, and I.~Gupta, ``Asynchronous federated optimization,'' \emph{arXiv preprint arXiv:1903.03934}, 2019.

\bibitem{li2020federated}
T.~Li, A.~K. Sahu, M.~Zaheer, M.~Sanjabi, A.~Talwalkar, and V.~Smith, ``Federated optimization in heterogeneous networks,'' \emph{Proceedings of Machine learning and systems}, vol.~2, pp. 429--450, 2020.

\bibitem{ouyang2021clusterfl}
X.~Ouyang, Z.~Xie, J.~Zhou, J.~Huang, and G.~Xing, ``Clusterfl: a similarity-aware federated learning system for human activity recognition,'' in \emph{Proceedings of the 19th Annual International Conference on Mobile Systems, Applications, and Services}, 2021, pp. 54--66.

\bibitem{rajbhandari2020zero}
S.~Rajbhandari, J.~Rasley, O.~Ruwase, and Y.~He, ``Zero: Memory optimizations toward training trillion parameter models,'' in \emph{SC20: International Conference for High Performance Computing, Networking, Storage and Analysis}.\hskip 1em plus 0.5em minus 0.4em\relax IEEE, 2020, pp. 1--16.

\bibitem{md2021distgnn}
V.~Md, S.~Misra, G.~Ma, R.~Mohanty, E.~Georganas, A.~Heinecke, D.~Kalamkar, N.~K. Ahmed, and S.~Avancha, ``Distgnn: Scalable distributed training for large-scale graph neural networks,'' in \emph{Proceedings of the International Conference for High Performance Computing, Networking, Storage and Analysis}, 2021, pp. 1--14.

\bibitem{shoeybi2019megatron}
M.~Shoeybi, M.~Patwary, R.~Puri, P.~LeGresley, J.~Casper, and B.~Catanzaro, ``Megatron-lm: Training multi-billion parameter language models using model parallelism,'' \emph{arXiv preprint arXiv:1909.08053}, 2019.

\bibitem{rosset2020turing}
C.~Rosset, ``Turing-nlg: A 17-billion-parameter language model by microsoft,'' \emph{Microsoft Blog}, vol.~1, no.~2, 2020.

\bibitem{heinecke2016libxsmm}
A.~Heinecke, G.~Henry, M.~Hutchinson, and H.~Pabst, ``Libxsmm: accelerating small matrix multiplications by runtime code generation,'' in \emph{SC'16: Proceedings of the International Conference for High Performance Computing, Networking, Storage and Analysis}.\hskip 1em plus 0.5em minus 0.4em\relax IEEE, 2016, pp. 981--991.

\bibitem{lai2021oort}
F.~Lai, X.~Zhu, H.~V. Madhyastha, and M.~Chowdhury, ``Oort: Efficient federated learning via guided participant selection.'' in \emph{OSDI}, 2021, pp. 19--35.

\bibitem{sun2022fedsea}
J.~Sun, A.~Li, L.~Duan, S.~Alam, X.~Deng, X.~Guo, H.~Wang, M.~Gorlatova, M.~Zhang, H.~Li \emph{et~al.}, ``Fedsea: A semi-asynchronous federated learning framework for extremely heterogeneous devices,'' 2022.

\bibitem{band2020memflow}
N.~Band, ``Memflow: Memory-aware distributed deep learning,'' in \emph{Proceedings of ACM SIGMOD}, 2020, pp. 2883--2885.

\bibitem{beaumont2020optimal}
O.~Beaumont, J.~Herrmann, G.~Pallez, and A.~Shilova, ``Optimal memory-aware backpropagation of deep join networks,'' \emph{Philosophical Transactions of the Royal Society A}, vol. 378, no. 2166, p. 20190049, 2020.

\bibitem{lai2018cmsis}
L.~Lai, N.~Suda, and V.~Chandra, ``Cmsis-nn: Efficient neural network kernels for arm cortex-m cpus,'' \emph{arXiv preprint arXiv:1801.06601}, 2018.

\bibitem{paoletti2020flop}
M.~E. Paoletti, J.~M. Haut, X.~Tao, J.~Plaza, and A.~Plaza, ``Flop-reduction through memory allocations within cnn for hyperspectral image classification,'' \emph{IEEE Transactions on Geoscience and Remote Sensing}, vol.~59, no.~7, pp. 5938--5952, 2020.

\bibitem{saha2020rnnpool}
O.~Saha, A.~Kusupati, H.~V. Simhadri, M.~Varma, and P.~Jain, ``Rnnpool: efficient non-linear pooling for ram constrained inference,'' \emph{Advances in Neural Information Processing Systems}, vol.~33, pp. 20\,473--20\,484, 2020.

\bibitem{oh2021quantum}
S.~Oh, J.~Choi, J.-K. Kim, and J.~Kim, ``Quantum convolutional neural network for resource-efficient image classification: A quantum random access memory (qram) approach,'' in \emph{2021 International Conference on Information Networking (ICOIN)}.\hskip 1em plus 0.5em minus 0.4em\relax IEEE, 2021, pp. 50--52.

\bibitem{liang2022distrihd}
D.~Liang, J.~Shiomi, N.~Miura, and H.~Awano, ``Distrihd: A memory efficient distributed binary hyperdimensional computing architecture for image classification,'' in \emph{2022 27th Asia and South Pacific Design Automation Conference (ASP-DAC)}.\hskip 1em plus 0.5em minus 0.4em\relax IEEE, 2022, pp. 43--49.

\bibitem{emara2019liteseg}
T.~Emara, H.~E. Abd El~Munim, and H.~M. Abbas, ``Liteseg: A novel lightweight convnet for semantic segmentation,'' in \emph{2019 Digital Image Computing: Techniques and Applications (DICTA)}.\hskip 1em plus 0.5em minus 0.4em\relax IEEE, 2019, pp. 1--7.

\bibitem{wang2019lednet}
Y.~Wang, Q.~Zhou, J.~Liu, J.~Xiong, G.~Gao, X.~Wu, and L.~J. Latecki, ``Lednet: A lightweight encoder-decoder network for real-time semantic segmentation,'' in \emph{2019 IEEE International Conference on Image Processing (ICIP)}.\hskip 1em plus 0.5em minus 0.4em\relax IEEE, 2019, pp. 1860--1864.

\bibitem{krevso2020efficient}
I.~Kre{\v{s}}o, J.~Krapac, and S.~{\v{S}}egvi{\'c}, ``Efficient ladder-style densenets for semantic segmentation of large images,'' \emph{IEEE Transactions on Intelligent Transportation Systems}, vol.~22, no.~8, pp. 4951--4961, 2020.

\bibitem{jin2021memory}
Y.~Jin, D.~Han, and H.~Ko, ``Memory-based semantic segmentation for off-road unstructured natural environments,'' in \emph{2021 IEEE/RSJ International Conference on Intelligent Robots and Systems (IROS)}.\hskip 1em plus 0.5em minus 0.4em\relax IEEE, 2021, pp. 24--31.

\bibitem{shangguan2019optimizing}
Y.~Shangguan, J.~Li, Q.~Liang, R.~Alvarez, and I.~McGraw, ``Optimizing speech recognition for the edge,'' \emph{arXiv preprint arXiv:1909.12408}, 2019.

\bibitem{guo2020efficient}
J.~Guo, G.~Tiwari, J.~Droppo, M.~Van~Segbroeck, C.-W. Huang, A.~Stolcke, and R.~Maas, ``Efficient minimum word error rate training of rnn-transducer for end-to-end speech recognition,'' \emph{arXiv preprint arXiv:2007.13802}, 2020.

\bibitem{winata2020lightweight}
G.~I. Winata, S.~Cahyawijaya, Z.~Lin, Z.~Liu, and P.~Fung, ``Lightweight and efficient end-to-end speech recognition using low-rank transformer,'' in \emph{ICASSP 2020-2020 IEEE International Conference on Acoustics, Speech and Signal Processing (ICASSP)}.\hskip 1em plus 0.5em minus 0.4em\relax IEEE, 2020, pp. 6144--6148.

\bibitem{wang2020voicefilter}
Q.~Wang, I.~L. Moreno, M.~Saglam, K.~Wilson, A.~Chiao, R.~Liu, Y.~He, W.~Li, J.~Pelecanos, M.~Nika \emph{et~al.}, ``Voicefilter-lite: Streaming targeted voice separation for on-device speech recognition,'' \emph{arXiv preprint arXiv:2009.04323}, 2020.

\bibitem{dosovitskiy2020image}
A.~Dosovitskiy, L.~Beyer, A.~Kolesnikov, D.~Weissenborn, X.~Zhai, T.~Unterthiner, M.~Dehghani, M.~Minderer, G.~Heigold, S.~Gelly \emph{et~al.}, ``An image is worth 16x16 words: Transformers for image recognition at scale,'' \emph{arXiv preprint arXiv:2010.11929}, 2020.

\bibitem{li2019deep}
S.~Li, W.~Song, L.~Fang, Y.~Chen, P.~Ghamisi, and J.~A. Benediktsson, ``Deep learning for hyperspectral image classification: An overview,'' \emph{IEEE Transactions on Geoscience and Remote Sensing}, vol.~57, no.~9, pp. 6690--6709, 2019.

\bibitem{weng2021semantic}
Z.~Weng and Z.~Qin, ``Semantic communication systems for speech transmission,'' \emph{IEEE Journal on Selected Areas in Communications}, vol.~39, no.~8, pp. 2434--2444, 2021.

\bibitem{das2021fundamentals}
N.~Das, S.~Chakraborty, J.~Chaki, N.~Padhy, and N.~Dey, ``Fundamentals, present and future perspectives of speech enhancement,'' \emph{International Journal of Speech Technology}, vol.~24, pp. 883--901, 2021.

\bibitem{zhao2017survey}
B.~Zhao, J.~Feng, X.~Wu, and S.~Yan, ``A survey on deep learning-based fine-grained object classification and semantic segmentation,'' \emph{International Journal of Automation and Computing}, vol.~14, no.~2, pp. 119--135, 2017.

\bibitem{gupta2018deep}
P.~Gupta, N.~Saxena, M.~Sharma, and J.~Tripathi, ``Deep neural network for human face recognition,'' \emph{International Journal of Engineering and Manufacturing (IJEM)}, vol.~8, no.~1, pp. 63--71, 2018.

\bibitem{you2019pixel}
H.~You, S.~Tian, L.~Yu, and Y.~Lv, ``Pixel-level remote sensing image recognition based on bidirectional word vectors,'' \emph{IEEE Transactions on Geoscience and Remote Sensing}, vol.~58, no.~2, pp. 1281--1293, 2019.

\bibitem{protasov2016fracture}
M.~Protasov, G.~Reshetova, and V.~Tcheverda, ``Fracture detection by gaussian beam imaging of seismic data and image spectrum analysis,'' \emph{Geophysical prospecting}, vol.~64, no.~1, pp. 68--82, 2016.

\bibitem{qian2022rex}
X.~Qian, R.~Hang, and Q.~Liu, ``Rex: an efficient approach to reducing memory cost in image classification,'' 2022.

\bibitem{sun2018fully}
W.~Sun and R.~Wang, ``Fully convolutional networks for semantic segmentation of very high resolution remotely sensed images combined with dsm,'' \emph{IEEE Geoscience and Remote Sensing Letters}, vol.~15, no.~3, pp. 474--478, 2018.

\bibitem{stojkoska2017review}
B.~L.~R. Stojkoska and K.~V. Trivodaliev, ``A review of internet of things for smart home: Challenges and solutions,'' \emph{Journal of cleaner production}, vol. 140, pp. 1454--1464, 2017.

\bibitem{alaa2017review}
M.~Alaa, A.~A. Zaidan, B.~B. Zaidan, M.~Talal, and M.~L.~M. Kiah, ``A review of smart home applications based on internet of things,'' \emph{Journal of Network and Computer Applications}, vol.~97, pp. 48--65, 2017.

\bibitem{feng2021intelligent}
S.~Feng, X.~Yan, H.~Sun, Y.~Feng, and H.~X. Liu, ``Intelligent driving intelligence test for autonomous vehicles with naturalistic and adversarial environment,'' \emph{Nature communications}, vol.~12, no.~1, p. 748, 2021.

\bibitem{khan2021modelling}
Q.-T.-A. Khan, S.~Abbas, M.~A. Khan, A.~Fatima, S.~Alanazi, and N.~S. Elmitwally, ``Modelling intelligent driving behaviour using machine learning,'' 2021.

\bibitem{yoshioka2018multi}
T.~Yoshioka, H.~Erdogan, Z.~Chen, and F.~Alleva, ``Multi-microphone neural speech separation for far-field multi-talker speech recognition,'' in \emph{2018 IEEE International Conference on Acoustics, Speech and Signal Processing (ICASSP)}.\hskip 1em plus 0.5em minus 0.4em\relax IEEE, 2018, pp. 5739--5743.

\bibitem{pratap2020scaling}
V.~Pratap, Q.~Xu, J.~Kahn, G.~Avidov, T.~Likhomanenko, A.~Hannun, V.~Liptchinsky, G.~Synnaeve, and R.~Collobert, ``Scaling up online speech recognition using convnets,'' \emph{arXiv preprint arXiv:2001.09727}, 2020.

\bibitem{han2017ese}
S.~Han, J.~Kang, H.~Mao, Y.~Hu, X.~Li, Y.~Li, D.~Xie, H.~Luo, S.~Yao, Y.~Wang \emph{et~al.}, ``Ese: Efficient speech recognition engine with sparse lstm on fpga,'' in \emph{Proceedings of the ACM/SIGDA International Symposium on Field-Programmable Gate Arrays}, 2017, pp. 75--84.

\bibitem{liciotti2020sequential}
D.~Liciotti, M.~Bernardini, L.~Romeo, and E.~Frontoni, ``A sequential deep learning application for recognising human activities in smart homes,'' \emph{Neurocomputing}, vol. 396, pp. 501--513, 2020.

\bibitem{mekruksavanich2021lstm}
S.~Mekruksavanich and A.~Jitpattanakul, ``Lstm networks using smartphone data for sensor-based human activity recognition in smart homes,'' \emph{Sensors}, vol.~21, no.~5, p. 1636, 2021.

\bibitem{schneider2019wav2vec}
S.~Schneider, A.~Baevski, R.~Collobert, and M.~Auli, ``wav2vec: Unsupervised pre-training for speech recognition,'' \emph{arXiv preprint arXiv:1904.05862}, 2019.

\bibitem{radford2022robust}
A.~Radford, J.~W. Kim, T.~Xu, G.~Brockman, C.~McLeavey, and I.~Sutskever, ``Robust speech recognition via large-scale weak supervision,'' \emph{arXiv preprint arXiv:2212.04356}, 2022.

\bibitem{liu2020privacy}
Y.~Liu, J.~James, J.~Kang, D.~Niyato, and S.~Zhang, ``Privacy-preserving traffic flow prediction: A federated learning approach,'' \emph{IEEE Internet of Things Journal}, vol.~7, no.~8, pp. 7751--7763, 2020.

\bibitem{hoffmann2019model}
E.~J. Hoffmann, Y.~Wang, M.~Werner, J.~Kang, and X.~X. Zhu, ``Model fusion for building type classification from aerial and street view images,'' \emph{Remote Sensing}, vol.~11, no.~11, p. 1259, 2019.

\bibitem{yi2018deep}
X.~Yi, J.~Zhang, Z.~Wang, T.~Li, and Y.~Zheng, ``Deep distributed fusion network for air quality prediction,'' in \emph{Proceedings of the 24th ACM SIGKDD international conference on knowledge discovery \& data mining}, 2018, pp. 965--973.

\bibitem{manogaran2021blockchain}
G.~Manogaran, M.~Alazab, P.~M. Shakeel, and C.-H. Hsu, ``Blockchain assisted secure data sharing model for internet of things based smart industries,'' \emph{IEEE Transactions on Reliability}, vol.~71, no.~1, pp. 348--358, 2021.

\bibitem{shahzad2020internet}
Y.~Shahzad, H.~Javed, H.~Farman, J.~Ahmad, B.~Jan, and M.~Zubair, ``Internet of energy: Opportunities, applications, architectures and challenges in smart industries,'' \emph{Computers \& Electrical Engineering}, vol.~86, p. 106739, 2020.

\bibitem{andronie2021artificial}
M.~Andronie, G.~L{\u{a}}z{\u{a}}roiu, M.~Iatagan, C.~Uț{\u{a}}, R.~Ștef{\u{a}}nescu, and M.~Cocoșatu, ``Artificial intelligence-based decision-making algorithms, internet of things sensing networks, and deep learning-assisted smart process management in cyber-physical production systems,'' \emph{Electronics}, vol.~10, no.~20, p. 2497, 2021.

\bibitem{liebig2017dynamic}
T.~Liebig, N.~Piatkowski, C.~Bockermann, and K.~Morik, ``Dynamic route planning with real-time traffic predictions,'' \emph{Information Systems}, vol.~64, pp. 258--265, 2017.

\bibitem{polson2017deep}
N.~G. Polson and V.~O. Sokolov, ``Deep learning for short-term traffic flow prediction,'' \emph{Transportation Research Part C: Emerging Technologies}, vol.~79, pp. 1--17, 2017.

\bibitem{ren2022manage}
H.~Ren, D.~Anicic, and T.~Runkler, ``How to manage tiny machine learning at scale: An industrial perspective,'' \emph{arXiv preprint arXiv:2202.09113}, 2022.

\bibitem{yang2020analysis}
K.~Yang, Z.~Yu, and Y.~Luo, ``Analysis on driving factors of lake surface water temperature for major lakes in yunnan-guizhou plateau,'' \emph{Water Research}, vol. 184, p. 116018, 2020.

\bibitem{banbury2021micronets}
C.~Banbury, C.~Zhou, I.~Fedorov, R.~Matas, U.~Thakker, D.~Gope, V.~Janapa~Reddi, M.~Mattina, and P.~Whatmough, ``Micronets: Neural network architectures for deploying tinyml applications on commodity microcontrollers,'' \emph{Proceedings of Machine Learning and Systems}, vol.~3, pp. 517--532, 2021.

\bibitem{sudharsan2020rce}
B.~Sudharsan, J.~G. Breslin, and M.~I. Ali, ``Rce-nn: a five-stage pipeline to execute neural networks (cnns) on resource-constrained iot edge devices,'' in \emph{Proceedings of the 10th International Conference on the Internet of Things}, 2020, pp. 1--8.

\bibitem{maskey2020cubesatnet}
A.~Maskey and M.~Cho, ``Cubesatnet: Ultralight convolutional neural network designed for on-orbit binary image classification on a 1u cubesat,'' \emph{Engineering Applications of Artificial Intelligence}, vol.~96, p. 103952, 2020.

\bibitem{hung2021end}
C.-W. Hung, S.-X. Zeng, C.-H. Lee, and W.-T. Li, ``End-to-end deep learning by mcu implementation: an intelligent gripper for shape identification,'' \emph{Sensors}, vol.~21, no.~3, p. 891, 2021.

\bibitem{wu2020enabling}
Y.~Wu, Z.~Wang, Y.~Shi, and J.~Hu, ``Enabling on-device cnn training by self-supervised instance filtering and error map pruning,'' \emph{IEEE Transactions on Computer-Aided Design of Integrated Circuits and Systems}, vol.~39, no.~11, pp. 3445--3457, 2020.

\bibitem{fedorov2019sparse}
I.~Fedorov, R.~P. Adams, M.~Mattina, and P.~Whatmough, ``Sparse: Sparse architecture search for cnns on resource-constrained microcontrollers,'' \emph{Advances in Neural Information Processing Systems}, vol.~32, 2019.

\bibitem{wang2020optimizing}
Q.~Wang, D.~Li, X.~Huang, S.~Shen, S.~Mei, and J.~Liu, ``Optimizing fft-based convolution on armv8 multi-core cpus,'' in \emph{European Conference on Parallel Processing}.\hskip 1em plus 0.5em minus 0.4em\relax Springer, 2020, pp. 248--262.

\bibitem{meng2022automatic}
J.~Meng, C.~Zhuang, P.~Chen, M.~Wahib, B.~Schmidt, X.~Wang, H.~Lan, D.~Wu, M.~Deng, Y.~Wei \emph{et~al.}, ``Automatic generation of high-performance convolution kernels on arm cpus for deep learning,'' \emph{IEEE Transactions on Parallel and Distributed Systems}, vol.~33, no.~11, pp. 2885--2899, 2022.

\bibitem{li2021optimizing}
D.~Li, D.~Huang, Z.~Chen, and Y.~Lu, ``Optimizing massively parallel winograd convolution on arm processor,'' in \emph{50th International Conference on Parallel Processing}, 2021, pp. 1--12.

\bibitem{huang2021numa}
X.~Huang, Q.~Wang, S.~Lu, R.~Hao, S.~Mei, and J.~Liu, ``Numa-aware fft-based convolution on armv8 many-core cpus,'' in \emph{2021 IEEE Intl Conf on Parallel \& Distributed Processing with Applications, Big Data \& Cloud Computing, Sustainable Computing \& Communications, Social Computing \& Networking (ISPA/BDCloud/SocialCom/SustainCom)}.\hskip 1em plus 0.5em minus 0.4em\relax IEEE, 2021, pp. 1019--1026.

\bibitem{zhou2022pipelining}
X.~Zhou, Y.~Dou, R.~Li, P.~Zhang, and Y.~Liu, ``A pipelining strategy for accelerating convolution neural networks on arm cpus,'' \emph{Concurrency and Computation: Practice and Experience}, vol.~34, no.~2, p. e6102, 2022.

\bibitem{wang2016dlau}
C.~Wang, L.~Gong, Q.~Yu, X.~Li, Y.~Xie, and X.~Zhou, ``Dlau: A scalable deep learning accelerator unit on fpga,'' \emph{IEEE Transactions on Computer-Aided Design of Integrated Circuits and Systems}, vol.~36, no.~3, pp. 513--517, 2016.

\bibitem{xie2021wpu}
X.~Xie and C.~Wu, ``Wpu: A fpga-based scalable, efficient and software/hardware co-design deep neural network inference acceleration processor,'' in \emph{2021 International Conference on High Performance Big Data and Intelligent Systems (HPBD\&IS)}.\hskip 1em plus 0.5em minus 0.4em\relax IEEE, 2021, pp. 1--5.

\bibitem{meng2021fixyfpga}
J.~Meng, S.~K. Venkataramanaiah, C.~Zhou, P.~Hansen, P.~Whatmough, and J.-s. Seo, ``Fixyfpga: Efficient fpga accelerator for deep neural networks with high element-wise sparsity and without external memory access,'' in \emph{2021 31st International Conference on Field-Programmable Logic and Applications (FPL)}.\hskip 1em plus 0.5em minus 0.4em\relax IEEE, 2021, pp. 9--16.

\bibitem{chang2019memory}
X.~Chang, H.~Pan, D.~Zhang, Q.~Sun, and W.~Lin, ``A memory-optimized and energy-efficient cnn acceleration architecture based on fpga,'' in \emph{2019 IEEE 28th International Symposium on Industrial Electronics (ISIE)}.\hskip 1em plus 0.5em minus 0.4em\relax IEEE, 2019, pp. 2137--2141.

\bibitem{rios2018performance}
A.~Rios-Navarro, R.~Tapiador-Morales, A.~Jimenez-Fernandez, C.~Amaya, M.~Dominguez-Morales, T.~Delbruck, and A.~Linares-Barranco, ``Performance evaluation over hw/sw co-design soc memory transfers for a cnn accelerator,'' in \emph{2018 IEEE 18th International Conference on Nanotechnology (IEEE-NANO)}.\hskip 1em plus 0.5em minus 0.4em\relax IEEE, 2018, pp. 1--4.

\bibitem{fu2016visual}
H.~Fu, Z.~Niu, C.~Zhang, J.~Ma, and J.~Chen, ``Visual cortex inspired cnn model for feature construction in text analysis,'' \emph{Frontiers in computational neuroscience}, vol.~10, p.~64, 2016.

\bibitem{wang2019high}
S.~Wang, G.~Ananthanarayanan, Y.~Zeng, N.~Goel, A.~Pathania, and T.~Mitra, ``High-throughput cnn inference on embedded arm big. little multicore processors,'' \emph{IEEE Transactions on Computer-Aided Design of Integrated Circuits and Systems}, vol.~39, no.~10, pp. 2254--2267, 2019.

\bibitem{2014Going}
C.~Szegedy, W.~Liu, Y.~Jia, P.~Sermanet, and A.~Rabinovich, ``Going deeper with convolutions,'' \emph{IEEE Computer Society}, 2014.

\bibitem{liu2021enterprise}
P.~Liu, W.~Qingqing, and W.~Liu, ``Enterprise human resource management platform based on fpga and data mining,'' \emph{Microprocessors and Microsystems}, vol.~80, p. 103330, 2021.

\bibitem{peng2021binary}
H.~Peng, S.~Zhou, S.~Weitze, J.~Li, S.~Islam, T.~Geng, A.~Li, W.~Zhang, M.~Song, M.~Xie \emph{et~al.}, ``Binary complex neural network acceleration on fpga,'' in \emph{2021 IEEE 32nd International Conference on Application-specific Systems, Architectures and Processors (ASAP)}.\hskip 1em plus 0.5em minus 0.4em\relax IEEE, 2021, pp. 85--92.

\bibitem{korol2019fpga}
G.~Korol and F.~G. Moraes, ``A fpga parameterizable multi-layer architecture for cnns,'' in \emph{Proceedings of the 32nd Symposium on Integrated Circuits and Systems Design}, 2019, pp. 1--6.

\bibitem{zainab2019fpga}
M.~Zainab, A.~R. Usmani, S.~Mehrban, and M.~Hussain, ``Fpga based implementations of rnn and cnn: A brief analysis,'' in \emph{2019 International Conference on Innovative Computing (ICIC)}.\hskip 1em plus 0.5em minus 0.4em\relax IEEE, 2019, pp. 1--8.

\bibitem{petrica2020memory}
L.~Petrica, T.~Alonso, M.~Kroes, N.~Fraser, S.~Cotofana, and M.~Blott, ``Memory-efficient dataflow inference for deep cnns on fpga,'' in \emph{2020 International Conference on Field-Programmable Technology (ICFPT)}.\hskip 1em plus 0.5em minus 0.4em\relax IEEE, 2020, pp. 48--55.

\bibitem{yang2020interstellar}
X.~Yang, M.~Gao, Q.~Liu, J.~Setter, J.~Pu, A.~Nayak, S.~Bell, K.~Cao, H.~Ha, P.~Raina \emph{et~al.}, ``Interstellar: Using halide's scheduling language to analyze dnn accelerators,'' in \emph{Proceedings of the Twenty-Fifth International Conference on Architectural Support for Programming Languages and Operating Systems}, 2020, pp. 369--383.

\bibitem{udagawa2022human}
K.~Udagawa, Y.~Saito, and H.~Saruwatari, ``Human-in-the-loop speaker adaptation for dnn-based multi-speaker tts,'' \emph{arXiv preprint arXiv:2206.10256}, 2022.

\bibitem{hao2019fpga}
C.~Hao, X.~Zhang, Y.~Li, S.~Huang, J.~Xiong, K.~Rupnow, W.-m. Hwu, and D.~Chen, ``Fpga/dnn co-design: An efficient design methodology for iot intelligence on the edge,'' in \emph{Proceedings of the 56th Annual Design Automation Conference 2019}, 2019, pp. 1--6.

\bibitem{li2020edd}
Y.~Li, C.~Hao, X.~Zhang, X.~Liu, Y.~Chen, J.~Xiong, W.-m. Hwu, and D.~Chen, ``Edd: Efficient differentiable dnn architecture and implementation co-search for embedded ai solutions,'' in \emph{2020 57th ACM/IEEE Design Automation Conference (DAC)}.\hskip 1em plus 0.5em minus 0.4em\relax IEEE, 2020, pp. 1--6.

\bibitem{jiang2019accuracy}
W.~Jiang, X.~Zhang, E.~H.-M. Sha, L.~Yang, Q.~Zhuge, Y.~Shi, and J.~Hu, ``Accuracy vs. efficiency: Achieving both through fpga-implementation aware neural architecture search,'' in \emph{Proceedings of the 56th Annual Design Automation Conference 2019}, 2019, pp. 1--6.

\bibitem{wang2020fann}
X.~Wang, M.~Magno, L.~Cavigelli, and L.~Benini, ``Fann-on-mcu: An open-source toolkit for energy-efficient neural network inference at the edge of the internet of things,'' \emph{IEEE Internet of Things Journal}, vol.~7, no.~5, pp. 4403--4417, 2020.

\bibitem{garofalo2020pulp}
A.~Garofalo, M.~Rusci, F.~Conti, D.~Rossi, and L.~Benini, ``Pulp-nn: accelerating quantized neural networks on parallel ultra-low-power risc-v processors,'' \emph{Philosophical Transactions of the Royal Society A}, vol. 378, no. 2164, p. 20190155, 2020.

\bibitem{niu2020patdnn}
W.~Niu, X.~Ma, S.~Lin, S.~Wang, X.~Qian, X.~Lin, Y.~Wang, and B.~Ren, ``Patdnn: Achieving real-time dnn execution on mobile devices with pattern-based weight pruning,'' in \emph{Proceedings of the Twenty-Fifth International Conference on Architectural Support for Programming Languages and Operating Systems}, 2020, pp. 907--922.

\bibitem{liu2020cocopie}
S.~Liu, B.~Ren, X.~Shen, and Y.~Wang, ``Cocopie: Making mobile ai sweet as pie--compression-compilation co-design goes a long way,'' \emph{arXiv preprint arXiv:2003.06700}, 2020.

\bibitem{ma2020pconv}
X.~Ma, F.-M. Guo, W.~Niu, X.~Lin, J.~Tang, K.~Ma, B.~Ren, and Y.~Wang, ``Pconv: The missing but desirable sparsity in dnn weight pruning for real-time execution on mobile devices,'' in \emph{Proceedings of the AAAI Conference on Artificial Intelligence}, vol.~34, no.~04, 2020, pp. 5117--5124.

\bibitem{jetson}
``Nvidia jetson embedded systems,'' \url{https://www.nvidia.com/en-us/autonomous-machines/embedded-systems/}.

\bibitem{xilinx}
``Xilinx zynq-7000 fpgas,'' \url{https://www.xilinx.com/products/silicon-devices/soc/zynq-7000.html}.

\bibitem{baymurzina2022review}
D.~Baymurzina, E.~Golikov, and M.~Burtsev, ``A review of neural architecture search,'' \emph{Neurocomputing}, vol. 474, pp. 82--93, 2022.

\bibitem{kim2022neural}
Y.~Kim, Y.~Li, H.~Park, Y.~Venkatesha, and P.~Panda, ``Neural architecture search for spiking neural networks,'' in \emph{Computer Vision--ECCV 2022: 17th European Conference, Tel Aviv, Israel, October 23--27, 2022, Proceedings, Part XXIV}.\hskip 1em plus 0.5em minus 0.4em\relax Springer, 2022, pp. 36--56.

\bibitem{liu2022federated}
X.~Liu, J.~Zhao, J.~Li, B.~Cao, and Z.~Lv, ``Federated neural architecture search for medical data security,'' \emph{IEEE transactions on industrial informatics}, vol.~18, no.~8, pp. 5628--5636, 2022.

\bibitem{yu2019evaluating}
K.~Yu, C.~Sciuto, M.~Jaggi, C.~Musat, and M.~Salzmann, ``Evaluating the search phase of neural architecture search,'' \emph{arXiv preprint arXiv:1902.08142}, 2019.

\bibitem{li2020random}
L.~Li and A.~Talwalkar, ``Random search and reproducibility for neural architecture search,'' in \emph{Uncertainty in artificial intelligence}.\hskip 1em plus 0.5em minus 0.4em\relax PMLR, 2020, pp. 367--377.

\bibitem{bib:hua2023edge}
H.~Hua, Y.~Li, T.~Wang, N.~Dong, W.~Li, and J.~Cao, ``Edge computing with artificial intelligence: A machine learning perspective,'' \emph{ACM Computing Surveys}, vol.~55, no.~9, pp. 1--35, 2023.

\bibitem{chen2019communication}
Y.~Chen, X.~Sun, and Y.~Jin, ``Communication-efficient federated deep learning with layerwise asynchronous model update and temporally weighted aggregation,'' \emph{IEEE transactions on neural networks and learning systems}, vol.~31, no.~10, pp. 4229--4238, 2019.

\bibitem{huang2021personalized}
Y.~Huang, L.~Chu, Z.~Zhou, L.~Wang, J.~Liu, J.~Pei, and Y.~Zhang, ``Personalized cross-silo federated learning on non-iid data,'' in \emph{Proceedings of the AAAI Conference on Artificial Intelligence}, vol.~35, no.~9, 2021, pp. 7865--7873.

\bibitem{bhardwaj2022ekya}
R.~Bhardwaj, Z.~Xia, G.~Ananthanarayanan, J.~Jiang, N.~Karianakis, Y.~Shu, K.~Hsieh, V.~Bahl, and I.~Stoica, ``Ekya: Continuous learning of video analytics models on edge compute servers,'' in \emph{USENIX NSDI}, 2022.

\bibitem{mullapudi2019online}
R.~T. Mullapudi, S.~Chen, K.~Zhang, D.~Ramanan, and K.~Fatahalian, ``Online model distillation for efficient video inference,'' in \emph{Proceedings of the IEEE/CVF International conference on computer vision}, 2019, pp. 3573--3582.

\bibitem{khani2021real}
M.~Khani, P.~Hamadanian, A.~Nasr-Esfahany, and M.~Alizadeh, ``Real-time video inference on edge devices via adaptive model streaming,'' in \emph{Proceedings of the IEEE/CVF International Conference on Computer Vision}, 2021, pp. 4572--4582.

\bibitem{wulf1995hitting}
W.~A. Wulf and S.~A. McKee, ``Hitting the memory wall: Implications of the obvious,'' \emph{ACM SIGARCH computer architecture news}, vol.~23, no.~1, pp. 20--24, 1995.

\bibitem{wordeman20123d}
M.~Wordeman, J.~Silberman, G.~Maier, and M.~Scheuermann, ``A 3d system prototype of an edram cache stacked over processor-like logic using through-silicon vias,'' in \emph{2012 IEEE International Solid-State Circuits Conference}.\hskip 1em plus 0.5em minus 0.4em\relax IEEE, 2012, pp. 186--187.

\bibitem{zhu20133d}
Q.~Zhu, B.~Akin, H.~E. Sumbul, F.~Sadi, J.~C. Hoe, L.~Pileggi, and F.~Franchetti, ``A 3d-stacked logic-in-memory accelerator for application-specific data intensive computing,'' in \emph{2013 IEEE international 3D systems integration conference (3DIC)}.\hskip 1em plus 0.5em minus 0.4em\relax IEEE, 2013, pp. 1--7.

\bibitem{zhu2013accelerating}
Q.~Zhu, T.~Graf, H.~E. Sumbul, L.~Pileggi, and F.~Franchetti, ``Accelerating sparse matrix-matrix multiplication with 3d-stacked logic-in-memory hardware,'' in \emph{2013 IEEE High Performance Extreme Computing Conference (HPEC)}.\hskip 1em plus 0.5em minus 0.4em\relax IEEE, 2013, pp. 1--6.

\bibitem{seshadri2015fast}
V.~Seshadri, K.~Hsieh, A.~Boroum, D.~Lee, M.~A. Kozuch, O.~Mutlu, P.~B. Gibbons, and T.~C. Mowry, ``Fast bulk bitwise and and or in dram,'' \emph{IEEE Computer Architecture Letters}, vol.~14, no.~2, pp. 127--131, 2015.

\bibitem{seshadri2013rowclone}
V.~Seshadri, Y.~Kim, C.~Fallin, D.~Lee, R.~Ausavarungnirun, G.~Pekhimenko, Y.~Luo, O.~Mutlu, P.~B. Gibbons, M.~A. Kozuch \emph{et~al.}, ``Rowclone: Fast and energy-efficient in-dram bulk data copy and initialization,'' in \emph{Proceedings of the 46th Annual IEEE/ACM International Symposium on Microarchitecture}, 2013, pp. 185--197.

\bibitem{jerger2008virtual}
N.~E. Jerger, L.-S. Peh, and M.~Lipasti, ``Virtual circuit tree multicasting: A case for on-chip hardware multicast support,'' \emph{ACM SIGARCH Computer Architecture News}, vol.~36, no.~3, pp. 229--240, 2008.

\bibitem{kang2012flexram}
Y.~Kang, W.~Huang, S.-M. Yoo, D.~Keen, Z.~Ge, V.~Lam, P.~Pattnaik, and J.~Torrellas, ``Flexram: Toward an advanced intelligent memory system,'' in \emph{2012 IEEE 30th International Conference on Computer Design (ICCD)}.\hskip 1em plus 0.5em minus 0.4em\relax IEEE, 2012, pp. 5--14.

\bibitem{wang2015energy}
Y.~Wang, T.~Tang, L.~Xia, B.~Li, P.~Gu, H.~Yang, H.~Li, and Y.~Xie, ``Energy efficient rram spiking neural network for real time classification,'' in \emph{Proceedings of the 25th edition on Great Lakes Symposium on VLSI}, 2015, pp. 189--194.

\bibitem{liu2015spiking}
C.~Liu, B.~Yan, C.~Yang, L.~Song, Z.~Li, B.~Liu, Y.~Chen, H.~Li, Q.~Wu, and H.~Jiang, ``A spiking neuromorphic design with resistive crossbar,'' in \emph{Proceedings of the 52nd Annual Design Automation Conference}, 2015, pp. 1--6.

\end{thebibliography}
\bibliographystyle{IEEEtran}

\end{document}